%% file: main.tex
\pdfoutput=1

\PassOptionsToPackage{prologue,dvipsnames}{xcolor} 

\documentclass[11pt]{article}

\usepackage[]{acl}

\usepackage{times}
\usepackage{latexsym}

\usepackage[T1]{fontenc}
\usepackage{CJKutf8}

\usepackage[utf8]{inputenc}

\usepackage{microtype}

\usepackage{inconsolata}

\usepackage{multirow}

\usepackage{listings}
\usepackage{tabularx}
\usepackage{ltablex}
\usepackage{longtable}
\usepackage{graphicx}
\usepackage{colortbl}
\usepackage[dvipsnames]{xcolor}
\usepackage[table]{xcolor} 
\usepackage{fontawesome5}
\usepackage{pythonhighlight}

\usepackage{supertabular}
\usepackage{xltabular}
\usepackage{amsmath}
\usepackage{amssymb}
\usepackage{mathrsfs}
\usepackage{bm,dutchcal}
\usepackage{float}
\usepackage{tcolorbox}
\usepackage{tipx}
\usepackage{marvosym}

\usepackage{adjustbox}
\usepackage{lineno}
\usepackage{array}
\usepackage{booktabs} 

\DeclareRobustCommand{\mail}{%
  \begingroup\normalfont
  \vspace{0em}%
  \raisebox{-0.2em}{%
  \includegraphics[height=1.em]{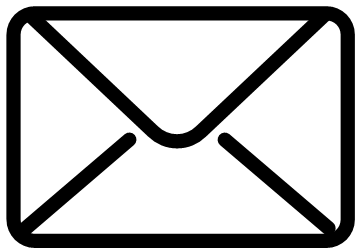}%
  }%
  \kern 0.4em%
  \endgroup
}

\DeclareRobustCommand{\github}{%
  \begingroup\normalfont
  \vspace{0.5em}%
  \raisebox{-0.2em}{%
  \includegraphics[height=1.2em]{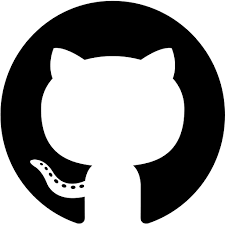}%
  }%
  \kern 0.4em%
  \endgroup
}

\DeclareRobustCommand{\MultiMedSTtitle}{%
  \begingroup\normalfont
  \raisebox{-0.2em}{%
  \includegraphics[height=2.2em]{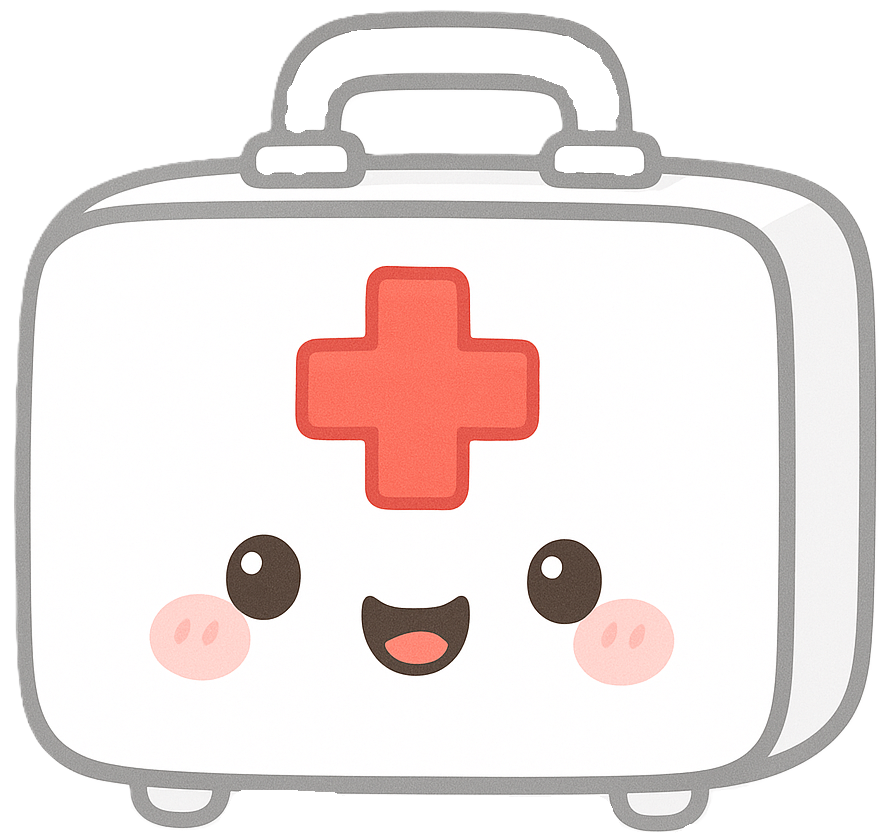}%
  }
  \kern 0.0em%
  \textbf{\textcolor{black}{MultiMed-ST: }}%
  \endgroup
}

\DeclareRobustCommand{\MultiMedST}{%
  \begingroup\normalfont
  \raisebox{-0.2em}{%
  \includegraphics[height=1.2em]{figures/MultiMedST_icon.png}%
  }
  \kern 0.0em%
  \textbf{\textcolor{red}{MultiMed-ST }}%
  \endgroup
}

\usepackage[acronym]{glossaries}
\makeglossaries

\newglossarystyle{acronym4col}{%
  \setglossarystyle{long}%
  {%
    \setlength{\arrayrulewidth}{0.4pt}%
    \renewcommand{\arraystretch}{1.15}%
    \begin{longtable}{p{1cm}|p{4.5cm}|%
      p{\dimexpr\linewidth-(1cm+4.5cm+1.5cm)-6\tabcolsep\relax}|%
      p{1.5cm}}%
  }%
  {\end{longtable}}%
}

\newacronym[description={General term for systems that perform tasks requiring human-like intelligence}]{AI}{AI}{Artificial Intelligence}
\newacronym[description={Speech translation is the process of automatically converting spoken language in one tongue into spoken or written language in another, combining speech recognition, machine translation, and speech synthesis in real time.}]{ST}{ST}{Speech Translation}
\newacronym[description={Machine translation is the automatic process of converting text or speech from one language into another using computational models.}]{MT}{MT}{Machine Translation}
\newacronym[description={It is a deep learning approach that translates text directly from one language to another using end-to-end neural networks.}]{NMT}{NMT}{Neural Machine Translation}
\newacronym[description={It is the technology that converts spoken language into written text by analyzing and transcribing audio signals.}]{ASR}{ASR}{Automatic Speech Recognition}
\newacronym[description={A Large Language Model is an AI system trained on vast text data to understand and generate human-like language.}]{LLM}{LLM}{Large Language Model}
\newacronym[description={seq2seq is a neural network architecture that transforms one sequence (like a sentence) into another (like a translation) using an encoder-decoder structure.}]{seq2seq}{seq2seq}{sequence-to-sequence}
\newacronym[description={SOTA means "state of the art," the best or most advanced performance achieved in a given field at the present time.}]{SOTA}{SOTA}{state-of-the-art}
\newacronym[description={In automatic speech recognition (ASR), AED (Attention-based Encoder Decoder) is a neural model that maps speech features to text by using an encoder to process the audio sequence and an attention-guided decoder to generate the transcription step by step.}]{AED}{AED}{Attention Encoder Decoder}
\newacronym[description={WER (Word Error Rate) is a common metric that measures how many words a speech recognition system got wrong compared to the reference transcript.}]{WER}{WER}{Word Error Rate}
\newacronym[description={CER (Character Error Rate) is a metric that measures the accuracy of text recognition systems by calculating the ratio of character insertions, deletions, and substitutions to the total number of characters in the reference text.}]{CER}{CER}{Character Error Rate}
\newacronym[description={The Recurrent Neural Network Transducer (RNN-T) is an end-to-end sequence model that jointly learns acoustic and language modeling to directly map input audio to text without needing frame-level alignment.}]{RNN-T}{RNN-T}{Recurrent Neural Network Transducer}
\newacronym[description={In automatic speech recognition (ASR), a CNN extracts local acoustic patterns from spectrograms by learning time-frequency features that help capture phonetic and speaker-invariant information.}]{CNN}{CNN}{Convolution Neural Network}
\newacronym[description={A Recurrent Neural Network (RNN) in Automatic Speech Recognition (ASR) models sequential dependencies by processing speech frames one at a time, using past context to better predict phonemes or words.}]{RNN}{RNN}{Recurrent Neural Network}
\newacronym[description={RBMT (Rule-Based Machine Translation) is a translation approach that relies on linguistic rules and bilingual dictionaries to convert text from a source language into a target language.}]{RBMT}{RBMT}{Rule-Based Machine Translation}
\newacronym[description={In machine translation, Statistical Machine Translation (SMT) is a method that generates translations by learning statistical patterns from large bilingual text corpora.
}]{SMT}{SMT}{Statistical Machine Translation}
\newacronym[description={MFCC (Mel-Frequency Cepstral Coefficients) are compact representations of speech audio that capture perceptually relevant frequency features, widely used in Automatic Speech Recognition (ASR) and as input features for speech-to-speech or speech-to-text Machine Translation (MT).}]{MFCC}{MFCC}{Mel-Frequency Cepstral Coefficients}
\newacronym[description={In automatic speech recognition (ASR), the Fast Fourier Transform (FFT) quickly converts audio signals from the time domain into frequency components, enabling models to analyze phonetic information.}]{FFT}{FFT}{Fast Fourier Transform}
\newacronym[description={In automatic speech recognition (ASR), the Short-Time Fourier Transform (STFT) converts the raw audio waveform into a time-frequency representation by analyzing short overlapping windows, making speech features easier for models to process.}]{STFT}{STFT}{Short-Time Fourier Transform}
\newacronym[description={ITN (Inverse Text Normalization) is the process of converting normalized text (like "twenty twenty-five") back into its spoken-style or symbolic form (e.g., "2025").}]{ITN}{ITN}{Inverse Text Normalization}
\newacronym[description={SFT (Supervised Fine-Tuning) means training a language model on parallel source-target sentence pairs so it learns to generate accurate translations in a supervised way.}]{SFT}{SFT}{Supervised Fine-tuning}
\newacronym[description={GQA (Grouped Query Attention) is a Transformer variant where multiple attention heads share key-value projections but keep separate query projections, reducing computation while preserving expressiveness.}]{GQA}{GQA}{Grouped Query Attention}
\newacronym[description={In machine learning, FFW (Feed-Forward Network) refers to a neural network where information moves only in one direction-from inputs through hidden layers to outputs-without feedback loops or recurrence.}]{FFW}{FFW}{Feed-Forward Network}

\definecolor{custom_light_blue}{rgb}{0.85, 0.95, 1}
\definecolor{custom_light_pink}{rgb}{1, 0.85, 0.85}

\definecolor{gtred}{RGB}{220,20,60} 
\definecolor{m2mblue}{RGB}{0,102,204}

\usepackage[table]{xcolor}
\definecolor{softgrey}{gray}{0.85}

\newcommand{\highlight}[2][custom_light_pink]{%
  \begingroup
  \setlength{\fboxsep}{1pt}
  \colorbox{#1}{#2}%
  \endgroup
}

\title{\MultiMedSTtitle Large-scale Many-to-many Multilingual \\Medical Speech Translation}

\author{Khai Le-Duc$^{*1,2,3}$, Tuyen Tran$^{*3,4}$, 
\\ {\bf Bach Phan Tat$^{5}$, Nguyen Kim Hai Bui$^{6}$, Quan Dang$^{4}$, Hung-Phong Tran$^{4}$,}
\\ {\bf Thanh-Thuy Nguyen$^{7}$, Ly Nguyen$^{8}$, Tuan-Minh Phan$^{9}$, Thi Thu Phuong Tran$^{10}$,}
\\ {\bf Chris Ngo$^{3}$, Nguyen X. Khanh$^{\heartsuit 11}$, Thanh Nguyen-Tang$^{\heartsuit \dagger 12}$}\\
$^1$University of Toronto, Canada
$^2$University Health Network, Canada\\
$^3$Knovel Engineering Lab, Singapore
$^4$Hanoi University of Science and Technology, Vietnam\\
$^5$KU Leuven, Belgium 
$^6$E\"{o}tv\"{o}s Lor\'{a}nd University, Hungary\\
$^7$HCMC Open University, Vietnam
$^8$I\'{E}SEG School of Management, France\\
$^9$Technische Universit\"{a}t Dortmund, Germany
$^{10}$University of Hertfordshire, United Kingdom\\
$^{11}$UC Berkeley, United States
$^{12}$New Jersey Institute of Technology, United States 
\\\mail\texttt{duckhai.le@mail.utoronto.ca}\\
\github \href{https://github.com/leduckhai/MultiMed-ST}{\textbf{\color{red!50!pink}{https://github.com/leduckhai/MultiMed-ST}}}\\}

\begin{document}
\maketitle
\begin{abstract}
Multilingual speech translation (ST) and machine translation (MT) in the medical domain enhances patient care by enabling efficient communication across language barriers, alleviating specialized workforce shortages, and facilitating improved diagnosis and treatment, particularly during pandemics. In this work, we present the \textit{first} systematic study on medical ST, to our best knowledge, by releasing \MultiMedST, a large-scale ST dataset for the medical domain, spanning \textit{all} translation directions in five languages: Vietnamese, English, German, French, and  Simplified/Traditional Chinese, together with the models. With 290,000 samples, this is \textbf{the largest medical MT dataset} and \textbf{the largest many-to-many multilingual ST among all domains}. Secondly, we present \textbf{the most comprehensive ST analysis in the field's history}, to our best knowledge, including: empirical baselines, bilingual-multilingual comparative study, end-to-end vs. cascaded comparative study, task-specific vs. multi-task sequence-to-sequence comparative study, code-switch analysis, and quantitative-qualitative error analysis. All code, data, and models are available online: \\\href{https://github.com/leduckhai/MultiMed-ST}{https://github.com/leduckhai/MultiMed-ST}.

\end{abstract}

\def\thefootnote{(*)}\footnotetext{Equal contribution}\def\thefootnote{\arabic{footnote}}
\def\thefootnote{($\heartsuit$)}\footnotetext{Equal advising}\def\thefootnote{\arabic{footnote}}
\def\thefootnote{($\dagger$)}\footnotetext{Done partly while at Johns Hopkins University}\def\thefootnote{\arabic{footnote}}


\begin{figure*}[t]
    \centering
    \includegraphics[width=0.8\linewidth]{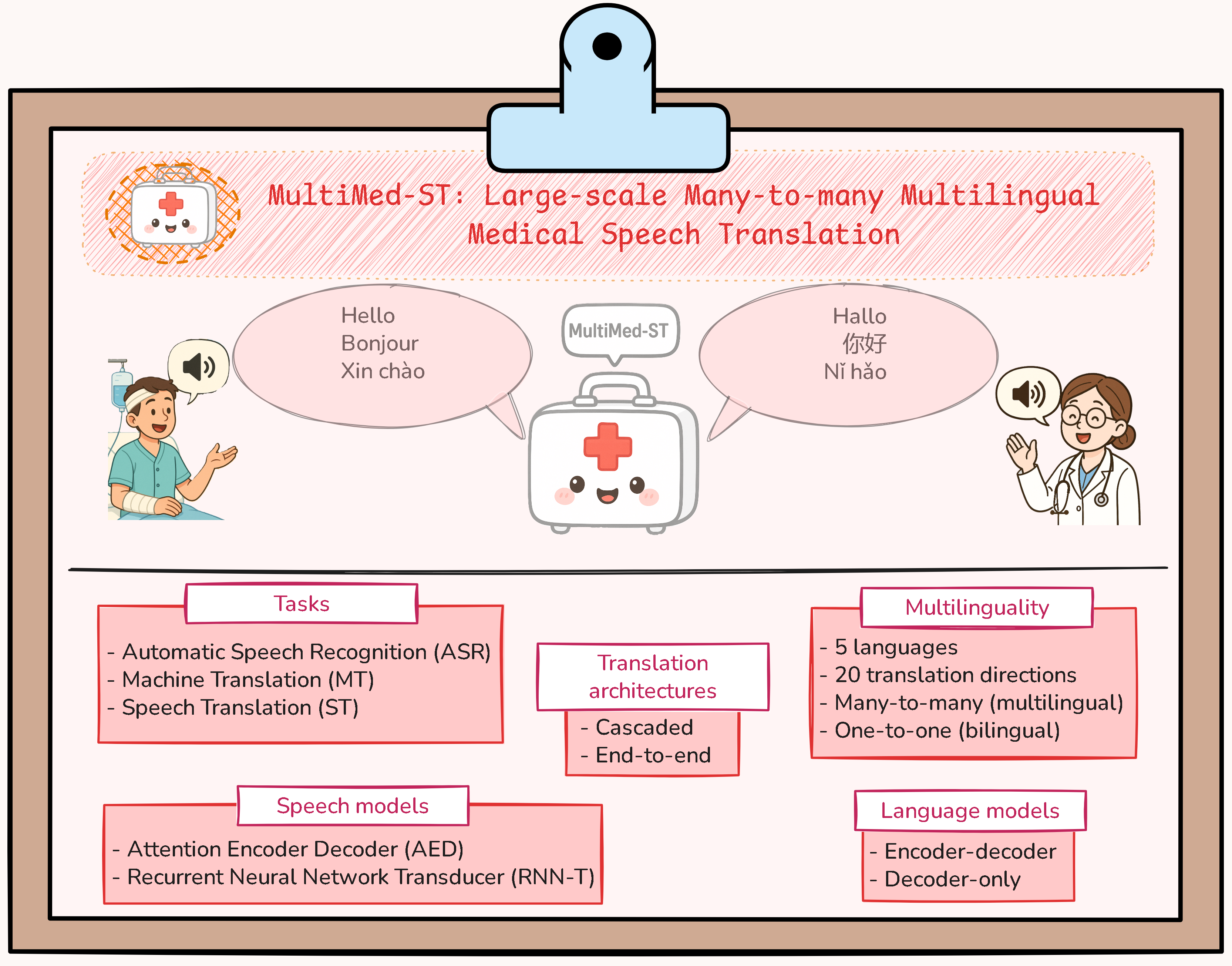}
    \caption{\textbf{An overview of \MultiMedST} -- A large-scale, many-to-many multilingual medical speech translation framework and dataset for facilitating cross-lingual communication in healthcare settings.}
    \label{fig:MultiMedST_pipeline}
\end{figure*}

\section{Introduction}
Effective communication between healthcare providers and patients is a foundation of quality medical care. However, linguistic barriers often hinder this communication, especially in multicultural and multilingual settings. These barriers can lead to misdiagnoses, improper treatment, and diminished patient satisfaction, ultimately compromising the overall quality of care \cite{al2020implications, woloshin1995language, cohen2005language, zhang2024assisting}. 

Medical \gls{ST}, also known as \gls{ST} in the medical domain, is a solution aimed at bridging these linguistic divides, by enabling (near) real-time communication between speakers of different languages. The demand for medical \gls{ST} has grown significantly with the increasing globalization of healthcare \cite{karwacka2015medical, khoong2022research}. Whether addressing the needs of immigrant populations, international patients seeking specialized treatments, or global health crises requiring cross-border collaboration, these technologies have the potential to transform how medical professionals deliver care \cite{dempere2023deploying, swaminathan2023natural, zhang2021leveraging}. Additionally, medical \gls{ST} aligns with broader efforts to promote health equity and accessibility, ensuring that language differences do not impede the right to quality healthcare \cite{nurminen2020machine, dahal2023exploring}.

Since the advent of large-scale pre-trained models adaptable to domain-specific tasks \cite{radford2022robustspeechrecognitionlargescale, chu2023qwen, touvron2023llama2openfoundation}, medical \gls{ST} research has gained attention. However, the scarcity of such publicly available datasets and models, driven by privacy concerns, hinders real-world deployment. Existing publicly available medical \gls{MT}\footnote{By definition, \gls{MT} encompasses both text-to-text translation (text-only \gls{MT}) and speech-to-text translation (\gls{ST}). As such, \gls{ST} is considered a subset of \gls{MT}.} datasets are text-only, small, and crawled from the Internet (\textcolor{pink}{\raisebox{-0.2ex}{\faInfoCircle}} see Appendix Table \ref{tab.data_stats_comparison_medical}). For medical \gls{ST}, previous works simply introduced the development of translation software without publishing datasets, models, or key findings, lacking a systematic and rigorous scientific approach \cite{bouillon2008many, marais2020awezamed, xu2024multilingual}.

To address the aforementioned issues, we introduce a large-scale high-quality, diverse dataset for many-to-many multilingual medical \gls{ST}, supporting 5 languages: Vietnamese, English, German, French, and Mandarin Chinese.  
Our key contributions are:
\begin{itemize}
    \item We present the \textit{first} systematic medical \gls{ST} study, to the best of our knowledge, by releasing \MultiMedST - a large-scale many-to-many multilingual medical \gls{ST} dataset for 5 languages, along with fine-tuned models. Built upon \textit{MultiMed} \gls{ASR} dataset, our translation annotation is the largest medical \gls{MT} dataset and the largest many-to-many multilingual \gls{ST} among all domains (\textcolor{pink}{\raisebox{-0.2ex}{\faInfoCircle}} see Section \ref{sec:data_stats}).
    \item We present the most extensive analysis ever conducted in \gls{ST} research to date, only enabled by the large-scale, many-to-many nature of \MultiMedST fine-tuning. It includes: \highlight{(i)} empirical baselines, \highlight{(ii)} task-specific vs. multi-task \gls{seq2seq} comparative study, \highlight{(iii)} end-to-end vs. cascaded comparative study, \highlight{(iv)} bilingual-multilingual comparative study, \highlight{(v)} code-switch analysis, and \highlight{(vi)} quantitative-qualitative error analysis. Our comprehensive analysis reveals guideline on how to build an effective many-to-many multilingual medical \gls{ST} model from a task-centric, model-centric, data-centric, and linguistic-centric perspective (\textcolor{pink}{\raisebox{-0.2ex}{\faInfoCircle}} see Section \ref{sec:conclusion} for the \highlight{five key findings} \textcolor{yellow}{\raisebox{-0.27ex}{\rotatebox{30}{\faLightbulb}}}). 
\end{itemize}

All code, data and models are published online.

\section{Data}

\subsection{Data Collection}
\label{sec:Data_Collection}
Speech data were sourced from the medical \gls{ASR} dataset provided by \citet{le2024multimed}, under the scientific research license. This dataset comprises manually transcribed recordings of real-world multi-speaker medical conversations across five languages: Vietnamese, English, German, French, and Mandarin Chinese. As pointed out by the authors, it represents the largest and most diverse medical \gls{ASR} resource, based on total duration (150 hours), number of recording conditions (10), number of accents (16), number of speaking roles (6), number of unique medical terms, and inclusion of all ICD-10 codes (\textcolor{pink}{\raisebox{-0.2ex}{\faInfoCircle}} see Table \ref{table:data_stats_ASR_literaturecompare} in Appendix Section \ref{sec.dataset_comparison_with_literature}).

\subsection{Annotation Process and Data Quality Control}
The data were initially translated from the source language into target languages (many-to-many) using the Gemini \gls{LLM}. Following the annotation process by \citet{zheng2023judging}, the \gls{LLM}-generated translated transcripts were treated as outputs from a \textit{real} human annotator. In the data quality process of the \textbf{test set}, five human annotators manually corrected and then cross-verified \textit{all} these translations based on the context of the whole conversation. To remove bias from \gls{LLM}-generated translations, only transcripts that received consensus approval from multiple annotators were retained, resulting in an \textbf{inter-annotator agreement of 100\%}. Roughly 90\% of \gls{LLM} translations need correction by our annotators. The estimated\footnote{Based on the publicly available price provided by professional translation services like VerboLabs or GTE Localize. We are not permitted to provide the true amount.} labor cost for the entire data quality process is 29k $\sim$ 58k USD.

All human annotators possessed a professional language proficiency of C1 or higher (or HSK5 for Chinese) in their respective working languages. Additionally, each annotator had completed basic medical training and demonstrated substantial knowledge of medical terminology in their selected language. Furthermore, they were either currently pursuing or had completed undergraduate or graduate studies in countries where their chosen language is predominantly spoken.

The dataset was subsequently uploaded to the Hugging Face platform.

\subsection{Data Statistics}
\label{sec:data_stats}
\input{tables/data_stats_main}
\input{tables/data_stats_comparison_medical_CutVersion}
The statistics of our data are described in Table \ref{tab.data_stats_main}. Our dataset has a total number of 290k samples for \textit{all} directions.

To the best of our knowledge, \textbf{\MultiMedST is the largest medical \gls{MT} dataset} when compared to existing medical \gls{MT} datasets, as shown in Table \ref{tab.data_stats_comparison_medical_CutVersion}, although speech data is much more difficult to collect and annotate.

Besides, in comparison with other large-scale \gls{ST} datasets reported in the literature, the size of \MultiMedST is comparable (\textcolor{pink}{\raisebox{-0.2ex}{\faInfoCircle}} see Table \ref{tab.data_stats_comparison_ST} in Appendix Section \ref{sec.dataset_comparison_with_literature}). However, \textbf{\MultiMedST is the largest many-to-many multilingual \gls{ST} among all domains}.

\section{Problem Formulation}
\highlight{\textbf{Informal definition:}} An \gls{ST} model aims to convert an audio signal to a translated language sequence. A \textbf{cascaded} \gls{ST} approach first transcribes speech to text (\gls{ASR}) and then translates it using a separate \gls{MT} model, while \textbf{end-to-end} \gls{ST} directly converts speech in one language to text in another without intermediate transcription.

\noindent\highlight{\textbf{Formal definition:}} Given an audio signal $x^{T}_{1} := x_{1}, x_{2}, ..., x_{T}$ of $T$ audio frames, a source language sequence $f^{J}_{1}$ of $J$ words, and a target language sequence $e^{I}_{1}$ of $I$ words, the maximization of the posterior probability $p$ of the target language sequence given the speech input is described as:

\highlight{\textbf{End-to-end approach:}}
\begin{equation}
x^{T}_{1} \rightarrow \hat{e}^{\hat{I}}_{1}(x^{T}_{1}) = \operatorname{arg}\max_{I, e^{I}_{1}}p(e^{I}_{1}|x^{T}_{1})
\end{equation}
where $\hat{e}^{\hat{I}}_{1}$ of length $\hat{I}$ words is the best target language sequence, and $\rightarrow$ is a mapping.

\highlight{\textbf{Cascaded approach:}}
\begin{equation}
x^{T}_{1} \rightarrow \hat{f}^{\hat{J}}_{1}(x^{T}_{1}) = \operatorname{arg}\max_{J, f^{J}_{1}}p(f^{J}_{1}|x^{T}_{1})
\label{eq:ASR_module}
\end{equation}

\begin{equation}
\hat{f}^{\hat{J}}_{1} \rightarrow \hat{e}^{\hat{I}}_{1}(\hat{f}^{\hat{J}}_{1}) = \operatorname{arg}\max_{I, e^{I}_{1}}p(e^{I}_{1}|\hat{f}^{\hat{J}}_{1})
\label{eq:MT_module}
\end{equation}
where Equation \ref{eq:ASR_module} is an \gls{ASR} model that transcribes the speech signal into the best source language sequence $\hat{f}^{\hat{J}}_{1}$, while Equation \ref{eq:MT_module} is an \gls{MT} model that generates the best target language sequence given the predicted source language sequence.

\textcolor{pink}{\raisebox{-0.2ex}{\faInfoCircle}} Further details of problem formulation are shown in Appendix Section \ref{sec:details_problem_formulation}.

\section{Experimental Setup}
We first establish empirical baselines, then derive key insights from task (task-specific vs. multi-task), model (end-to-end vs. cascaded), data (bilingual vs. multilingual training), and linguistic (code-switching analysis) perspectives.

\subsection{Training Setup}
\highlight{\textbf{Training system:}} We employed two standard training systems, \textbf{cascaded} (\gls{ASR}$\rightarrow$MT) and \textbf{end-to-end}.

\highlight{\textbf{\gls{ASR} models:}} We employed the 2 most \gls{SOTA} \gls{ASR} architectures with varying model sizes.
\begin{itemize}
    \item \gls{AED}: 
    \begin{itemize}
        \item Whisper models \cite{radford2023robust}: Whisper-small\footnote{https://huggingface.co/openai/whisper-small}, Whisper-large-v2\footnote{https://huggingface.co/openai/whisper-large-v2} 
        \item Deepgram\footnote{https://deepgram.com/}
    \end{itemize} 
    \item \gls{RNN-T}: AssemblyAI\footnote{https://www.assemblyai.com/}
\end{itemize}

\highlight{\textbf{\gls{MT} models:}} We employed various \gls{SOTA} open-source/closed-source, task-specific/multitask \gls{seq2seq} architectures and data representations.
\begin{itemize}
    \item Multilingual pre-trained models: 
    \begin{itemize}
        \item Encoder-decoder: mBART-large-50\footnote{https://huggingface.co/facebook/mbart-large-50} \cite{tang2020multilingual}, M2M100-418M\footnote{https://huggingface.co/facebook/m2m100\_418M} \cite{fan2020englishcentricmultilingualmachinetranslation}, 
        Marian\footnote{\url{https://huggingface.co/docs/transformers/model_doc/marian}} \cite{tiedemann2020opus}
        \item Decoder: Llama-3.1-8B\footnote{https://huggingface.co/meta-llama/Llama-3.1-8B} \cite{dubey2024llama3herdmodels}, Qwen-2.5-7B\footnote{https://huggingface.co/Qwen/Qwen2.5-7B} \cite{qwen2}, Mistral-v0.3-7B\footnote{https://huggingface.co/mistralai/Mistral-7B-v0.3} \cite{Jiang2023Mistral7}
        \item Commercial tool: Google Translate\footnote{https://cloud.google.com/translate/docs}
    \end{itemize}
    \item Bilingual pre-trained models: VinAI Translate\footnote{https://huggingface.co/vinai/vinai-translate-vi2en\\https://huggingface.co/vinai/vinai-translate-en2vi} \cite{vinaitranslate}, EnViT5\footnote{https://huggingface.co/VietAI/envit5-base} \cite{mtet}
\end{itemize}

\highlight{\textbf{End-to-end \gls{ST} models:}} For direct speech-to-text translation, we employed Whisper, SeamlessM4T-large-v2\footnote{https://huggingface.co/facebook/seamless-m4t-v2-large} \cite{Communication2023SeamlessME, communication2023seamlessm4tmassivelymultilingual}, Qwen2-Audio-7B-Instruct\footnote{https://huggingface.co/Qwen/Qwen2-Audio-7B-Instruct} \cite{Qwen2-Audio, chu2023qwen}.

All \gls{ASR} and \gls{MT} models are general-domain since \MultiMedST is the first attempt to fine-tune medical domain \gls{ST}. \textcolor{pink}{\raisebox{-0.2ex}{\faInfoCircle}} Full details of the training setup are shown in Appendix Section \ref{sec:details_experimental_setup}.

\input{tables/nmt_results_gt}

\subsection{Evaluation Metrics}
\noindent\textcolor{pink}{\raisebox{-0.2ex}{\faInfoCircle}} Advantage/disadvantage discussion of automatic metrics is described in Appendix Section \ref{sec.discussion_eval_metrics}.

\highlight{\textbf{Automatic \gls{MT} metrics:}} To evaluate \gls{MT} quality, two standard categories of evaluation metrics were utilized: \textbf{n-gram overlap metrics} (e.g., BLEU \cite{10.3115/1073083.1073135}, TER \cite{TER_metric}, METEOR \cite{banarjee2005}, ChrF \cite{popovic-2015-chrf}, ROUGE \cite{rouge}) and \textbf{embedding-based metrics} (e.g., BERTScore \cite{zhangbertscore}).

\highlight{\textbf{\gls{ASR} metrics:}} In the context of \gls{ST}, \gls{ASR} performance influences translation quality; therefore, \gls{ASR} was additionally assessed using \gls{WER} and \gls{CER}.

\highlight{\textbf{Human evaluation:}} Human evaluators directly assess \gls{MT} outputs by grading scores (0 to 10) based on three key criteria: \textit{adequacy}, \textit{fluency}, and \textit{comprehensibility} (\textcolor{pink}{\raisebox{-0.2ex}{\faInfoCircle}} see Appendix Section \ref{sec:details_human_eval}).

\highlight{\textbf{\gls{LLM}-as-a-judge:}} Unlike automated metrics, which rely on surface-level matching of n-grams, \gls{LLM}-as-a-judge \cite{zheng2023judging} can assess translations based on deeper semantic understanding, contextual appropriateness, and syntactic correctness (\textcolor{pink}{\raisebox{-0.2ex}{\faInfoCircle}} see Appendix Section \ref{sec:details_llm_as_judge} and Figure \ref{fig:llm_as_a_judge_prompt}).

\section{Experimental Results}
\subsection{Automatic Speech Recognition Baselines}
\input{tables/asr_results}
\textcolor{yellow}{\raisebox{-0.27ex}{\rotatebox{30}{\faLightbulb}}}  \textbf{What are trade-offs among model sizes, fine-tuning strategies, and performance of \gls{ASR} models?} As shown in Table \ref{tab:model_comparison}, the fine-tuned Whisper-small model achieved superior performance to larger pre-trained models, consistently outperforming all models across languages on the dev set. On test set, Whisper-small achieved the best \gls{WER} for Vietnamese (29.60\%) and \gls{CER} for Chinese (31.3\%), while Whisper-large-v2 excelled in English (\gls{WER} 25.5\%) and Chinese (\gls{CER} 37.3\%), and Deepgram outperformed others in French with a \gls{WER} of 40.3\%, highlighting the advantage of larger models for high-resource languages.

Also, results showed that monolingual fine-tuning consistently outperforms multilingual fine-tuning on both dev and test sets. Besides, SpecAugment \cite{park2019specaugment} does not help accuracy improvement.

\input{tables/asr_nmt_result}

\subsection{Ground-truth Translation Baselines}

\textcolor{yellow}{\raisebox{-0.27ex}{\rotatebox{30}{\faLightbulb}}} \textbf{Task-specific models outperform multi-task models on ground-truth transcript}: The experimental results for \gls{MT} on ground-truth transcript are presented in Table \ref{tab:translation-groundtruth}. Overall, translations from Google Translate achieved the highest results and outperformed other models across most language pairs in both settings. Encoder-decoder models, particularly those with English as the source language, generally outperformed the decoder models (\glspl{LLM}). Notably, the M2M100-418M model recorded higher BLEU scores than the \glspl{LLM} on many language pairs. This demonstrates the effectiveness of models trained for specific \gls{MT} tasks compared to multi-task models like \glspl{LLM}.

\subsection{Cascaded Speech Translation Baselines}
\textcolor{yellow}{\raisebox{-0.27ex}{\rotatebox{30}{\faLightbulb}}} \textbf{Multi-task models are on par with task-specific models in the \gls{ST} setting.} We evaluated the impact of \gls{ASR} models on text-to-text \gls{MT} models, as shown in Table \ref{tab:asr-translation}. 

Specifically, Whisper-large-v2 - M2M100-418M achieved the highest performance on most language pairs (16/20), except for the language pair with Vietnamese as the source language, where Whisper-small-mono - M2M100-418M achieved the best performance. This outcome stems from two factors: Whisper-large-v2's size and generalization enable more accurate transcripts for most languages, aiding \gls{MT} model, while Whisper-small-mono outperforms it for Vietnamese.

\gls{ASR} model performance differences reveal how \gls{ASR} transcript quality impacts \gls{MT}, with minor errors notably affecting complex languages like Vietnamese. Despite M2M100-418M's robustness on ground-truth text, it is sensitive to \gls{ASR} transcript quality. Also, M2M100-418M and mBart-large-50 do not significantly outperform \glspl{LLM} in the cascaded \gls{ST} setting, as shown in Table \ref{tab:translation-asr}. Therefore, multi-task models (\glspl{LLM}) still perform as well as task-specific models trained for \gls{MT} task.

\input{tables/nmt_results_asr}

\subsection{End-to-end and Cascaded Comparison}

\input{tables/nmt_multilingual_asr}
\input{tables/bilingual_translation}

\textcolor{yellow}{\raisebox{-0.27ex}{\rotatebox{30}{\faLightbulb}}} \textbf{\gls{MT} accuracy is dropped on speech}: Table \ref{tab:translation-groundtruth} and Table \ref{tab:translation-asr} show a significant decline in both BLEU and BERT scores due to the non-standard input text across all models, with the largest drop observed in the French-to-English from 50.18 to 30.15 with the LLama-3.1-8B model. This indicates that the \gls{ASR} model's poor performance for French significantly reduced translation accuracy. A similar trend was also observed in in-context learning experiments (\textcolor{pink}{\raisebox{-0.2ex}{\faInfoCircle}} see Appendix Section \ref{sec:incontext_learning_results}). 

\noindent\textcolor{yellow}{\raisebox{-0.27ex}{\rotatebox{30}{\faLightbulb}}} \textbf{Cascaded models significantly outperform end-to-end models}: Table \ref{tab:translation-asr} compares cascaded models with end-to-end  models. The results show a significant performance gap, with most cascaded models significantly outperforming end-to-end models. For a fair comparison with general-domain \gls{ST} in the literature, our findings align with prior insights that end-to-end models require extensive data (probably thousands of hours) and numerous parameters to match the accuracy of cascaded models \cite{sperber2020speech, sperber2019attention, xue2022large}.

\subsection{Bilingual-Multilingual Fine-tuning Comparison}

\textcolor{yellow}{\raisebox{-0.27ex}{\rotatebox{30}{\faLightbulb}}} \textbf{Bilingual fine-tuning outperforms multilingual \gls{MT} fine-tuning}: As shown in Table \ref{tab:translation-multilingual}, fine-tuning \gls{MT} models on all language pairs simultaneously resulted in a degradation of performance for most language pairs compared to fine-tuning on each language pair separately. When fine-tuning on multiple language pairs simultaneously, the shared parameters of the model must allocate their representational capacity across all pairs. This leads to interference between language pairs, especially when their linguistic structures or vocabularies differ significantly, as also observed in general-domain \gls{MT} \cite{dabre2020survey, blackwood2018multilingual}.

\subsection{Bilingual-Multilingual Pre-training Comparison}
\textcolor{yellow}{\raisebox{-0.27ex}{\rotatebox{30}{\faLightbulb}}} \textbf{Multilingual pre-trained \gls{MT} models match bilingual accuracy without needing multiple language-pair variants}: As shown in Table \ref{tab:bilingual-result}, the VinAI model achieved the highest BLEU score (50.79) for English-to-Vietnamese, while the M2M100-418M model excelled in BERTScore (0.95 vs. 0.88 for VinAI). For Vietnamese-to-English, M2M100-418M slightly outperformed VinAI with BLEU scores of 15.64 and 15.46, respectively. The EnViT5 model performed poorly for both translation directions.

These results show that bilingual pre-trained \gls{MT} models do not consistently outperform multilingual ones. This findings underscores the advantage of multilingual ones in leveraging diverse language pairs to achieve acceptable overall performance across metrics without requiring multiple variants for each language pair, as also observed in general-domain \gls{MT} \cite{dabre2020survey, nllb2024scaling, maillard2023small}. 

\subsection{Code-Switch Analysis}
\input{tables/code-switch}
In the medical domain, it is common for English terms or keywords to be retained in their original form when translated into other languages, a phenomenon referred to as code-switching. In Table \ref{tab:codeswitch-result}, this study filtered code-switched sentences for Vietnamese, German, French, and Chinese, evaluating model performance with BLEU and BERTScore metrics for each language pair.

\textcolor{yellow}{\raisebox{-0.27ex}{\rotatebox{30}{\faLightbulb}}} \textbf{Multilingual pre-trained \gls{MT} models could handle orthographic differences in code-switch  \gls{ST}}: Generally, results from code-switching in Table \ref{tab:codeswitch-result} are not consistently lower or higher than ground-truth baselines (Table \ref{tab:translation-groundtruth}) and cascaded monolingual fine-tuning \gls{ST} baselines (Table \ref{tab:asr-translation}). The results show that multilingual pre-trained \gls{MT} models can process multiple languages simultaneously within a single context, even with large orthographic differences like English-Chinese or smaller orthographic differences like English-Vietnamese/German.

\input{tables/human_and_auto_score}

\section{Error Analysis}
\subsection{Quantitative Error Analysis}
\textcolor{yellow}{\raisebox{-0.27ex}{\rotatebox{30}{\faLightbulb}}} \textbf{Strong correlation between n-gram overlap, contextual-embedding and subjective evaluation}: As shown in Table \ref{tab:human_auto_score}, for most language pairs and \gls{MT} models, there was a strong correlation between n-gram overlap metric and embedding-based metric and the evaluation outcomes obtained from both subjective \gls{LLM}-as-a-judge and subjective human evaluations in \gls{ST} quality. This alignment suggests that traditional automatic metrics remain reliable indicators of \gls{ST} quality, even as evaluation methodologies evolve. The consistency across these metrics reinforces their validity in assessing \textit{adequacy}, \textit{fluency} and \textit{comprehensibility} of medical \gls{ST} - the phenomenon is sometimes seen in general-domain \gls{MT} \cite{zheng2023judging, zhangbertscore, bavaresco2024llms}. \gls{LLM}-as-a-judge is a newly explored research trend, thus we found no reference for \gls{ST}, to our best knowledge.

\subsection{Qualitative Error Analysis}
We analyzed recurring translation errors in medical content across English, Vietnamese, German, Chinese, and French, identifying key areas for improvement.  

With English as the source, common issues included sentence fragmentation (notably in Chinese and Vietnamese), literal idiom translation, inconsistent medical terminology, and errors in proper noun handling. Vietnamese source texts led to grammatical errors in word order, verb tense, and articles, along with imprecise word choice, omissions, and register inconsistencies. German sources showed frequent word order errors, literal idiom translations, and issues with case, gender, and verb conjugation, especially in French and Vietnamese. Chinese texts often resulted in unnatural word-for-word translations, tense inaccuracies, missing grammatical elements, and misused measure words. French exhibited similar challenges to English, including sentence fragmentation, literal idiom translation, inconsistent terminology, and Vietnamese grammar errors in word order and verb conjugation.

\textcolor{pink}{\raisebox{-0.2ex}{\faInfoCircle}} More qualitative results are shown in Appendix Section \ref{sec:qualitative_results}.

\section{Conclusion}
\label{sec:conclusion}
In this work, we aim to remove language barriers in healthcare by  presenting the first systematic study on medical \gls{ST}, to our best knowledge. Specifically, we release \MultiMedST, a large-scale \gls{ST} dataset in the medical domain, covering \textit{all} translation directions in five languages: Vietnamese, English, German, French,  Simplified/Traditional Chinese, together with the models. With 290,000 samples, our dataset is the world's largest medical \gls{MT} dataset and the largest many-to-many multilingual \gls{ST} among all domains.

\textcolor{yellow}{\raisebox{-0.27ex}{\rotatebox{30}{\faLightbulb}}} Our key findings are: \highlight{(1)} Although task-specific models surpass multi-task models when evaluated on ground-truth transcripts, both exhibit comparable performance in the medical \gls{ST} setting. \highlight{(2)} Cascaded models still significantly outperform end-to-end models. \highlight{(3)} In the medical cascaded \gls{ST}, multilingual pre-trained \gls{MT} models should be selected for bilingual fine-tuning on each language pair for two primary reasons: first, multilingual pre-trained \gls{MT} models achieve bilingual accuracy without the need for multiple separate language-pair variants; second, bilingual fine-tuning has been shown to outperform multilingual \gls{MT} fine-tuning. \highlight{(4)} Multilingual pre-trained \gls{MT} models are capable of handling orthographic differences in code-switching with comparable effectiveness to non-code-switching in medical \gls{ST}. \highlight{(5)} In medical \gls{ST}, n-gram overlap evaluation exhibits a strong correlation with both contextual embedding-based evaluation and subjective assessment.

\section{Limitations}
Science and religion always go hand in hand. Carelessness in science can lead to serious consequences - not to mention the karmic repercussions researchers may face under the law of karma in Buddhism. Despite our best efforts to minimize human errors, mistakes in data, experiments, and processes are inevitable and often beyond our understanding or control.  

Medical research is a matter of great importance, as it can have direct negative impacts on human health. Given the critical nature of medical transcription (see Appendix Section \ref{sec:annotation_problem_longform_speech}), errors in \gls{ASR} and \gls{ST} outputs and annotation can lead to serious implications, potentially affecting patient diagnoses and treatment decisions \cite{adane2019role}. Therefore, \textbf{we earnestly urge readers to independently verify our hypotheses and experimental results using their own medical data}. We also strongly recommend conducting pilot tests in a simulated doctor-patient environment before full-scale deploying them in real-world applications.

\textcolor{pink}{\raisebox{-0.2ex}{\faInfoCircle}} Further limitations are extensively discussed in each Appendix Section.

\section*{Acknowledgement}
This work was initiated as part of a bachelor thesis by Khai Le-Duc at RWTH Aachen University under the supervision of Prof. Hermann Ney and PD. Ralf Schl\"uter.

Most of the theoretical formulations in this work were borrowed from lectures by Hermann Ney, Ralf Schl\"uter, and Albert Zeyer, as well as from PhD dissertations at the Machine Learning and Human Language Technology Group at RWTH Aachen University. 

We would like to thank other contributors, Long Vo-Dang, Nhut Huy Pham, and Viet Thanh Duy Nguyen for their precious initial efforts in this work.

\bibliography{custom}

\clearpage 


\appendix

\onecolumn
\tableofcontents
\newpage

\twocolumn
\section{Related Works}
\label{related_works}
\subsection{Neural Machine Translation}
\gls{NMT} has experienced substantial advancements with the development of Transformer-based models, such as the Transformer architecture \cite{vaswani2017attention}. The Transformer represents the first \gls{seq2seq} model solely reliant on the attention mechanism, wherein the recurrent layers of traditional models are replaced by multi-headed self-attention within the encoder-decoder framework. This architectural innovation has significantly accelerated training speeds in comparison to \gls{RNN} and \gls{CNN}, resulting in superior performance. BERT \cite{devlin-etal-2019-bert}, a pre-trained model designed to address the unidirectional constraints of earlier language models (such as the left-to-right processing in Transformers), incorporates a masked language model (MLM) to enable bidirectional representation, thereby enhancing machine translation tasks. Building on BERT and other pre-training paradigms, BART \cite{lewis2019bartdenoisingsequencetosequencepretraining} generalizes these techniques, achieving competitive results in various \gls{NMT} applications.

The GPT series, which demonstrates the efficacy of generative pre-training followed by fine-tuning for the \gls{MT} task in GPT-1 \cite{gpt1}, exhibits remarkable performance in text generation and zero-shot tasks. GPT-2 \cite{gpt2} and GPT-3 \cite{gpt3} further scale the model's size and training data, facilitating state-of-the-art performance in few-shot and zero-shot tasks, including translation. GPT-4 \cite{openai2024gpt4technicalreport} further improves capabilities in multilingual and domain-specific \gls{MT} tasks.

Several \gls{NMT} frameworks, such as OpenNMT \cite{klein-etal-2017-opennmt}, have been developed to facilitate the integration of custom deep learning models for translation tasks. These frameworks provide tools that optimize the efficiency of training, inference, and deployment in \gls{NMT} systems. MarianNMT \cite{mariannmt} emphasizes speed and scalability, enabling the implementation of state-of-the-art \gls{NMT} models with minimal computational overhead. OpenSeq2Seq \cite{kuchaiev-etal-2018-openseq2seq} offers reference implementations designed for efficient distributed and mixed-precision training. Tensor2Tensor \cite{vaswani2018tensor2tensorneuralmachinetranslation} and Sockeye \cite{hieber2018sockeyetoolkitneuralmachine} prioritize the security, reliability, and production-level performance of their software components. Fairseq \cite{ott-etal-2019-fairseq} is a fast, extensible toolkit for sequence modeling that offers scalability and is versatile across numerous applications.

\subsection{Cascaded Speech Translation}
\gls{ST} traditionally contains two components: \gls{ASR} (to convert audio into text) and \gls{NMT} (to translate text-to-text). The success in the \gls{ASR} technology starts with HTK \cite{htk} - a toolkit for manipulating Hidden Markov Models (HMM) provides comprehensive facilities for speech analysis, training, and recognition. Later success includes Julius \cite{DBLP:conf/interspeech/LeeKS01} - an open-source, high-performance, two-pass large vocabulary continuous speech recognition (LVCSR) decoder; Sphinx-4 \cite{10.5555/1698193} - a flexible, modular, and pluggable framework for \gls{ASR} written entirely in Java; RWTH ASR \cite{rwth_asr} - an open-source \gls{ASR} decoding system which includes state-of-the-art \gls{ASR} capabilities. Furthermore, Kaldi model \cite{Povey2011TheKS} provides a hybrid \gls{ASR} system based on finite-state transducers. Recent state-of-the-art framework was wav2vec 2.0 \cite{baevski2020wav2vec20frameworkselfsupervised} - a framework for self-supervised learning of speech representations which masks latent representations of the raw waveform and solves a contrastive task over quantized speech representations; and Whisper model \cite{radford2022robustspeechrecognitionlargescale} - which suggests that scaling weakly supervised pre-training has been underestimated in \gls{ASR} research. Other novel frameworks are from Facebook AI's end-to-end \gls{ASR} research, including wave2letter++ \cite{Pratap_2019} - the fastest open-source deep learning \gls{ASR}  framework, and Fairseq S2T \cite{wang2022fairseqs2tfastspeechtotext} - which bypassed traditional transcription steps, improving both latency and accuracy.

\subsection{End-to-end Speech Translation}
The development of end-to-end \gls{ST} models, which eliminate intermediary stages like \gls{ASR} outputs and lattices, has significantly reduced error propagation \cite{chen2024llast}. Research shows end-to-end \gls{ST} models achieve performance comparable to cascaded models \cite{sperber2019attention, ansari2020findings, bentivogli2021cascade}. Moreover, these models offer benefits like reduced latency and applicability to unwritten languages. \cite{berard2016listen}.

Some researchers have modified the multi-task encoder-decoder architecture \cite{weiss2017sequence} by splitting the decoder into two components \cite{liu2020synchronous, anastasopoulos2018tied}: one used to transcribe and the other one used to translate. Parallel research initiatives have likewise separated the encoder \cite{liu2020bridging, cheng2023m}, with subsequent studies demonstrating that a shared encoder can be independently segmented to optimize the utilization of \gls{ASR} data \cite{tang2021general, xu2023recent}. Furthermore, non-autoregressive (NAR) modeling has been investigated as a method to reduce latency \cite{inaguma2021orthros, chuang2021investigating}.

Recent advancements have notably explored multitasking within the framework of large-scale training, yielding remarkable performance on \gls{ST} benchmarks, like Whisper \cite{radford2022robustspeechrecognitionlargescale}, SeamlessM4T \cite{communication2023seamlessm4tmassivelymultilingual}. Another predominant approach involves the integration of an \gls{LLM} at the backend with a speech encoder at the frontend, like LauraGPT \cite{chen2024llast}, Qwen-Audio \cite{chu2023qwen}.

\subsection{Medical Machine Translation}
The translation of medical texts poses distinct challenges owing to the use of specialized terminology, frequent abbreviations, and the imperative requirement for precision \cite{neergard2003hospitals, flores2003errors}. Early methodologies predominantly utilized \gls{RBMT} and \gls{SMT}, both of which were tailored to medical language corpora \cite{10.3115/1220355.1220469}. \gls{RBMT} utilizes predefined rules and lexical databases to translate texts by analyzing their grammatical and lexical structures. It is particularly adept at managing medical terminology, provided that the dictionaries are up-to-date and comprehensive. However, \gls{RBMT} has limitations, including an inability to resolve ambiguity, interpret idiomatic expressions, and account for variations in language use. Additionally, \gls{RBMT} requires substantial human effort for the creation and ongoing maintenance of the rules and dictionaries specific to each language pair\cite{s2017statisticalvsrulebased}. \gls{SMT}, in contrast, depends on large parallel corpora-collections of aligned texts in two languages - to estimate the probability of translation equivalents \cite{10.5555/972470.972474}. In contrast to rule-based or dictionary-based systems, \gls{SMT} relies on data-driven algorithms to produce translations. This characteristic enables \gls{SMT} to be highly adaptable across different domains and genres, including specialized fields such as medical texts, by utilizing domain-specific corpora customized for both the source and target languages, as well as their respective contexts. However, \gls{SMT} is not without limitations. It often encounters challenges in generating fluent or grammatically accurate translations, particularly when dealing with low-resource languages or rare terminology, resulting in outputs that may be unnatural or imprecise.\cite{koehn2017challengesneuralmachinetranslation}. The occurrence of \gls{NMT} allowed for vast improvements, particularly with encoder-decoder architectures enhanced by attention mechanisms \cite{bahdanau2016neuralmachinetranslationjointly}. Recent studies have demonstrated that domain adaptation techniques, such as fine-tuning \glspl{LLM} on domain-specific datasets, can enhance the performance of medical tasks, including translation. \cite{bao2023discmedllmbridginggenerallarge, yang2024finetuningmedicallanguagemodels}.

\subsection{Domain Adaptation for Machine Translation}
Medical \gls{MT} for low-resource languages continues to present a significant challenge, primarily due to the absence of multilingual medical databases. Strategies such as data augmentation, which involves generating synthetic data to expand existing datasets \cite{Fadaee_2017, xia2019generalizeddataaugmentationlowresource}, back-translation, where target-to-source translations are utilized to create additional source-to-target pairs \cite{sennrich-etal-2016-improving}, and transfer learning \cite{zoph-etal-2016-transfer, nguyen2017transferlearninglowresourcerelated, gu2018universalneuralmachinetranslation}, which capitalizes on knowledge from high-resource languages to enhance performance in low-resource languages, have been proposed to address this issue.

Multilingual \gls{NMT} models such as mBART \cite{liu2020multilingualdenoisingpretrainingneural}, XLM-R \cite{conneau-etal-2020-unsupervised}, M2M-100 \cite{fan2020englishcentricmultilingualmachinetranslation}, and mT5 \cite{xue2021mt5massivelymultilingualpretrained} have demonstrated significant potential in overcoming the challenges associated with low-resource or domain-specific settings. This is achieved through the use of cross-lingual transfer learning, which allows the model to leverage shared linguistic representations across multiple languages. Consequently, this approach markedly improves the model's ability to generalize, even in the presence of limited training data in the target language.

Ethical considerations are also an essential problem in the context of medical \gls{MT}, given its potential implications for patient care \cite{Harishbhai2024}.

Future research is further centered on the integration of multimodal data, such as the combination of textual and audio-visual inputs, to improve translation accuracy within medical contexts \cite{huh2023improvingmedicalspeechtotextaccuracy, li2023llavamedtraininglargelanguageandvision}. Furthermore, fine-tuning pre-trained models on multilingual medical datasets, such as the Unified Medical Language System (UMLS), has shown promise in enhancing model performance while addressing the unique challenges associated with medical domains. However, these research directions lie beyond the scope of the our present study.

\subsection{Multilingual Machine Translation}
Recent research has increasingly focused on multilingual translation. For instance, studies by \citet{luong-manning-2015-stanford} and \citet{freitag2016fastdomainadaptationneural} have demonstrated that pre-training models on a diverse dataset, followed by fine-tuning on a smaller  target dataset, yields effective results. \citet{liu2020multilingualdenoisingpretrainingneural} extended the BART model with mBART and showed that multilingual denoising pre-training leads to significant performance improvements across a variety of \gls{MT} benchmarks. Additionally, \citet{verma-etal-2022-strategies} highlighted the effectiveness of multilingual pre-training in domain adaptation scenarios. Research by \citet{johnson-etal-2017-googles} further indicated that a trained multilingual \gls{NMT} system could perform zero-shot translation between previously unseen language pairs without direct supervision, provided that both source and target languages were included in the training process. \cite{arivazhagan2019missingingredientzeroshotneural} observed that the cosine similarity between the pooled encoder outputs of sentence pairs decreased during multilingual training. Meanwhile, \citet{sun2022zeroshotdomainadaptationneural} addressed domain adaptation by constructing bilingual phrase-level databases and retrieving contextually relevant prompts, which improved translation quality in unseen domains. On a different note, \cite{wu2024far100samplesgo} proposed an approach that fine-tuned models with a minimal amount of multi-parallel data, finding that a small, randomly sampled set of fine-tuning directions was sufficient for achieving comparable improvements.

\twocolumn
\section{Theoretical Formulation}
\label{sec:details_problem_formulation}
\subsection{Mel-Frequency Cepstral Coefficients (MFCCs)}
\begin{figure*}[t]
    \centering
    \includegraphics[width=1\linewidth]{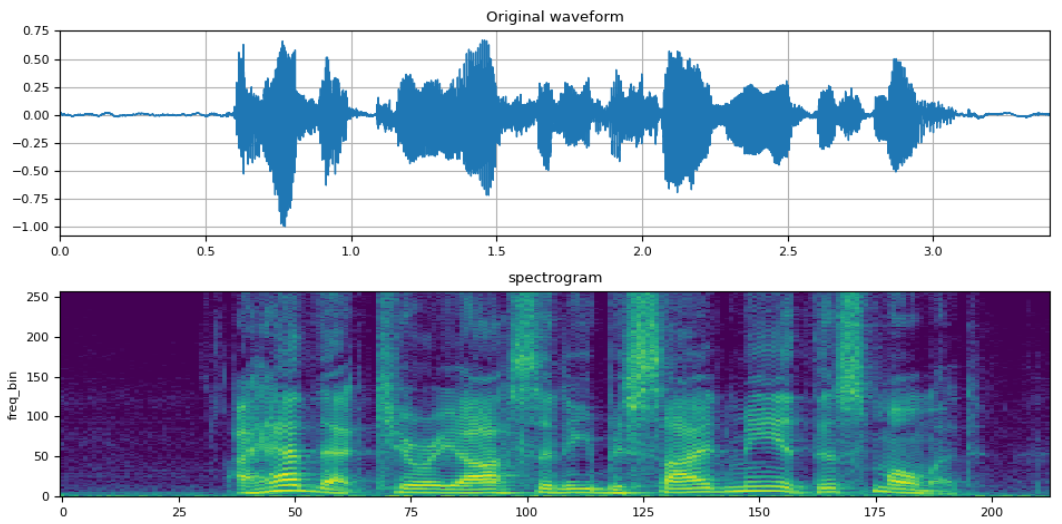}
    \caption{\textbf{\gls{MFCC} visualization}. The computation of \glspl{MFCC} begins by dividing the original waveform into overlapping 20ms frames.}
    \label{fig:MFCC_visualization}
\end{figure*}

\gls{MFCC} serves as a compact representation of the audio signal's spectral properties. The computation of \glspl{MFCC} begins by dividing the input signal $x^{T}_{1} := x_{1}, x_{2}, ..., x_{T}$ into overlapping frames, as visualized in Figure \ref{fig:MFCC_visualization}\footnote{\citet{golik2020data}'s Dissertation at RWTH Aachen University described \gls{MFCC} more comprehensively.\\
\gls{MFCC} visualization image is retrieved from Pytorch library.}.

\textbf{Pre-emphasis}: The audio signal, sampled at 16 kHz with a step size of 10 ms, is processed by extracting 160 consecutive samples from the Pulse Code Modulation (PCM) waveform for each frame. These 10 ms frames are non-overlapping, ensuring that stacking adjacent vectors avoids discontinuities. The 16-bit quantized samples, which span the integer range from $-2^{15}$ to $+2^{15}$, must be normalized to a numerically stable range. This normalization is achieved by applying mean and variance normalization, either globally across the entire training dataset or on a per-utterance basis. A commonly employed processing technique, known as high-frequency pre-emphasis, can be implemented by computing the differences between adjacent samples, as illustrated below:

\begin{equation}
\mathbcal{x}'_t = \mathbcal{x}_t - \mathbcal{x}_{t-1} \in \mathbb{R}
\end{equation}

A sequence of $16 \, \text{kHz} \times 10 \, \text{ms} = 160$ pre-emphasized waveform samples can then be considered a feature vector:

\begin{equation}
\hat{\mathbcal{x}}_t = {\mathbcal{x}'}^{t}_{t-160+1} \in \mathbb{R}^{160}
\end{equation}

\textbf{Amplitude spectrum - \gls{FFT}}: The \gls{STFT} is applied to overlapping windows with a duration of \( 25 \, \text{ms} \). Given a sampling rate of \( 16 \, \text{kHz} \), this window length corresponds to $25 \, \text{ms} \times 16 \, \text{kHz} = 400 \, \text{samples}$. To facilitate computation using the \gls{FFT}, the sample count is zero-padded to the next power of two, resulting in \( 2^9 = 512 \).
\begin{equation}
\begin{split}
&\mathbcal{z}_t \in \mathbb{R}^{512} \\
&=\begin{bmatrix}
\mathbcal{x}^{t'}_{t-400+1} & \mathbcal{x}^{t'}_{t-400+2} & \dots & \mathbcal{x}^{t'}
\underbrace{0 \dots 0}_{\text{zero-padding}}
\end{bmatrix}
\end{split}
\end{equation}

The extended sample vector is weighted using a Hann window, which exhibits smaller side lobes in the amplitude spectrum compared to a rectangular window:
\begin{equation}
\begin{split}
\mathbcal{w}^{(n)} &= 0.5 - 0.5 \cos\left(\frac{2\pi (n - 1)}{512 - 1}\right), \\
&\quad 1 \leq n \leq 512
\end{split}    
\end{equation}

\begin{equation}
    \mathbcal{s}_t^{(n)} = \mathbcal{z}_t^{(n)} \cdot \mathbcal{w}^{(n)}
\end{equation}

While the discrete \gls{STFT} could be done directly by evaluating the sum 
\begin{equation}
\begin{split}
\mathbcal{S}_t^{(\mathbb{F})} &= \sum_{n=0}^{512-1} \mathbcal{s}_t^{(n)} \cdot \exp\left(-j \frac{2\pi}{512}\mathbb{F}n\right),\\
&\quad 1 \leq \mathbb{F} \leq 512
\end{split}
\end{equation}
the complexity can be reduced from \(\mathcal{O}(N^2)\) to \(\mathcal{O}(N \log N)\) by applying the fast Fourier transform. 

The 512-\gls{FFT} results in a 257-dimensional vector because of the symmetry of the amplitude spectrum of a real-valued signal. The phase spectrum is removed. 
\begin{equation}
\begin{split}
\hat{\mathbcal{x}}_t &= 
\begin{bmatrix}
|\mathbcal{S}_t^{(0)}| & |\mathbcal{S}_t^{(1)}| & \dots & |\mathbcal{S}_t^{(512/2)}| 
\end{bmatrix} \\
&\in \mathbb{R}^{512/2+1}
\end{split}
\end{equation}

\textbf{\gls{MFCC}}: The \gls{MFCC} feature extraction is based on the \gls{STFT} of the pre-emphasized speech signal \cite{davis1980comparison}. It considers the nonlinear sensitivity of human auditory perception to variations in frequency. This is evidenced that the filter bank used to integrate the magnitude spectrum $|\mathbcal{S}^{(\mathbb{F})}_t|$ consists of $\mathbb{I}$ filters equidistantly spaced on the mel scale. The mel scale is a logarithmically scaled frequency axis. The $k$-th frequency bin of the \gls{FFT} centered around $\mathbb{F}_k$ Hz is then mapped to \( \tilde{\mathbb{F}}_k \) on the mel scale:
\begin{equation}
\mathbb{F}_k = \frac{k}{512} \cdot \mathbb{F}_\mathbcal{s}    
\end{equation}
\begin{equation}
\tilde{\mathbb{F}}_k = 2595 \cdot \log_{10}\left(1 + \frac{\mathbb{F}_k}{700 \text{ Hz}} \right)     
\end{equation} 

The filter center \( \tilde{\mathbb{F}}^{(i)}_c \) of the \( i \)-th triangular filter is then placed at \( i \cdot \tilde{\mathbb{F}}_b \), where the bandwidth \( \tilde{\mathbb{F}}_b \) corresponds to \( \tilde{\mathbb{F}}_{512} / \mathbb{I} \). With these parameters, the coefficients of the \( i \)-th triangular filter can be calculated explicitly as a piecewise linear function and stored in a weight vector \( \mathbcal{v}_i \in \mathbb{R}^{N/2+1} \). 

By applying discrete cosine transform (DCT), the \gls{MFCC} features are extracted from the logarithm filter outputs:
\begin{equation}
\mathbcal{X}^{(i)}_t = \log_{10}\left( \sum_{\mathbb{F}=0}^{512} |\mathbcal{S}^{(\mathbb{F})}_t| \mathbcal{v}^{(\mathbb{F})}_i \right)    
\end{equation}
\begin{equation}
\mathbcal{c}_{m,i} = \cos \left( \frac{\pi m (i + 0.5)}{\mathbb{I}} \right)   
\end{equation}
\begin{equation}
\mathbcal{C}^{(m)}_t = \sum_{i=0}^{\mathbb{I}-1} \mathbcal{c}_{m,i} \mathbcal{X}^{(i)}_t    
\end{equation}
\begin{equation}
\hat{\mathbcal{x}}_t = \left[ \mathbcal{C}^{(0)}_t \mathbcal{C}^{(1)}_t \dots \mathbcal{C}^{(\mathbb{I}-1)}_t \right] \in \mathbb{R}^\mathbb{I}    
\end{equation}

\subsection{Attention Encoder Decoder (AED)}
\label{sec:AED}
\begin{figure*}[h]
    \centering
    \includegraphics[width=1.0\linewidth]{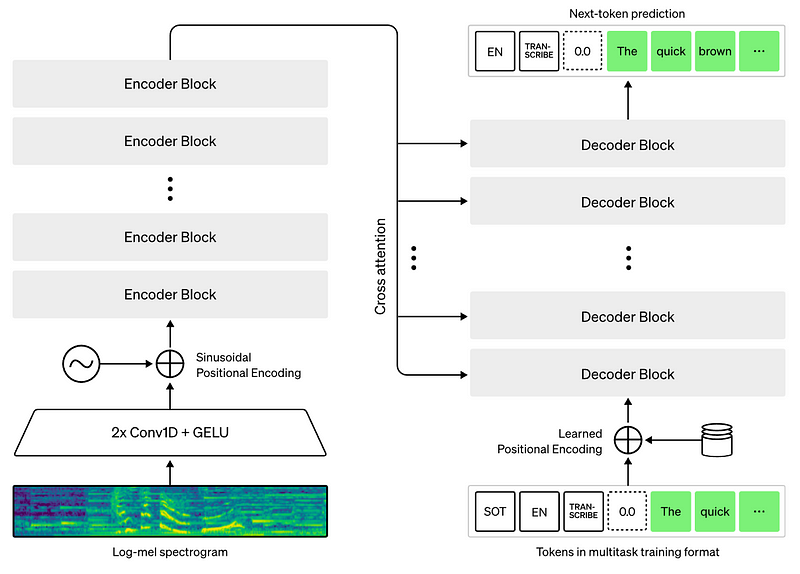}
    \caption{\textbf{OpenAI's Whisper architecture}. Whisper is a Transformer-based \gls{AED} architecture, using \gls{MFCC} features as input.}
    \label{fig:whisper_architecture}
\end{figure*}

\begin{figure*}[h]
    \centering
    \includegraphics[width=1.0\linewidth]{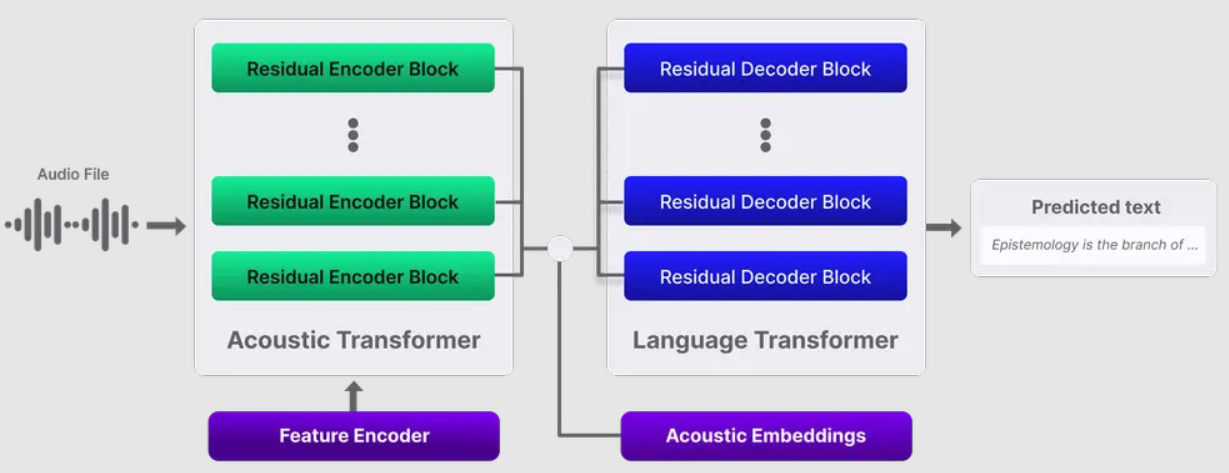}
    \caption{\textbf{Deepgram's Nova-2 architecture}. To our best understanding of Deepgram's documentation, Deepgram's Nova-2 is a Transformer-based \gls{AED} architecture, using raw waveform as input instead of \gls{MFCC} like Whisper. Feature extraction from raw waveform is probably  conducted by a learnable feature encoder, e.g. a block of \glspl{CNN} like wav2vec 2.0. Between encoder-decoder space, (unknown) acoustic embeddings are probably added as cross-attention.}
    \label{fig:deepgram_architecture}
\end{figure*}

As for \gls{AED} models, Whisper architecture is shown in Figure \ref{fig:whisper_architecture}, and Deepgram architecture is shown in Figure \ref{fig:deepgram_architecture}.

An \gls{ASR} model is used to transcribe speech into text by mapping an audio signal $x^{T}_{1} := x_{1}, x_{2}, ..., x_{T}$ of length $T$ to the most likely word sequence $w^{N}_{1}$ of length $N$. The word sequence probability is described as:

\begin{equation}
\label{eq:word_sequence_general}
p(w_{1}^{N}|x_{1}^{T}) = \prod_{n=1}^{N} p(w_n|w_{1}^{n-1},x_{1}^{T}).    
\end{equation}

In the \gls{ASR} encoder-decoder architecture, given $D$ as the feature dimension size, the input audio signal matrix could be described as $x^{T}_{1} \in \mathbb{R}^{T \times D_{input}}$. When simplified, downsampling before or inside the encoder - conducted by a fixed factor, such as striding in a \gls{CNN} - is removed. Thus, the encoder output sequence is as follows:

\begin{equation}
h_{1}^{T} = Encoder(x_{1}^{T}) \in \mathbb{R}^{T \times D_{encoder}}. 
\end{equation}

Using a stack of Transformer (\scalebox{1.44}{$\tau$}) blocks \cite{vaswani2017attention}, the encoder output sequence is described as function composition:
\begin{equation}
h_{1}^{T} = {\scalebox{1.44}{$\tau$}}_{0} \circ ... \circ {\scalebox{1.44}{$\tau$}}_{N_{EncLayers}}(x_{1}^{T}).
\end{equation}

In the decoder, the probability for each single word is defined as:

\begin{equation}
\begin{split}
p(w_n|w_{1}^{n-1},x_{1}^{T}) &= p(w_n|w_{1}^{n-1},h_{1}^{T}(x_{1}^{T}))\\
&= p(w_n|w_{1}^{n-1},h_{1}^{T}). 
\end{split}   
\end{equation}

Based on Equation \ref{eq:word_sequence_general}, the word sequence probability given the output of encoder is described as:
\begin{equation}
p(w_{1}^{N}|x_{1}^{T}) = \prod_{n=1}^{N} p(w_n|w_{1}^{n-1},h_{1}^{T}).    
\end{equation}

Then, decoder hidden state is formulated as:
\begin{equation}
g_n = \mathcal{F}(g_{n-1},w_{n-1}, c_n) \in \mathbb{R}^{D_{g}},
\end{equation}
where $\mathcal{F}$ is neural network; $D_{g}$ is hidden state dimension; and $c_n$ is context vector, e.g. weighted sum of encoder outputs via attention mechanism.

The attention mechanism in the decoder is described via 3 components: context vector $c_n$, attention weights $\alpha_{n,t}$, and attention energy $e_{n,t}$:

\begin{equation}
\begin{split}
c_n &= \sum_{t=1}^{T} \alpha_{n,t} {h}_{t} \in \mathbb{R}^{D_{encoder}},\\
\alpha_{n,t} &= \frac{\exp(e_{n,t})}{\sum_{t'=1}^{T}\exp(e_{n,t'})} \\
&= Softmax_{T}(\exp(e_{n,t})) \in \mathbb{R},\\
e_{n,t} &= Align(g_{n-1}, h_t) \in \mathbb{R} \\
&= W_{2} \cdot \tanh(W_{1} \cdot [g_{n-1}, h_t]),
\end{split}
\end{equation}
where $n$ is decoder step; $t$ is encoder frame; $\alpha \in \mathbb{R}^{T \times N}$ is attention weight matrix; $\alpha_n \in \mathbb{R}^{T}$ is normalized probability distribution over $t$; $Softmax_{T}$ is Softmax function over spatial dimension $T$, not feature dimension; $W_{1} \in \mathbb{R}^{(D_{g}+D_{encoder}) \times D_{key}}$; $W_{2} \in \mathbb{R}^{D_{key}}$.

In the decoding, the output probability distribution over vocabulary is defined as:
\begin{equation}
\begin{split}
&p(w_{n} = *|w_{1}^{n-1}, h_{1}^{T})\\
&= Softmax(MLP(w_{n-1}, g_n, c_n)) \in \mathbb{R}^{N}, 
\end{split}
\end{equation}
where $MLP$ is Multi-layer Perceptron.

To train an \gls{AED} model,  sequence-level frame-wise cross-entropy loss is employed:
\begin{equation}
\begin{split}
\mathscr{L}_{AED} &= - \sum_{(x_{1}^{T}, w_{1}^{N})} \log p(w_{1}^{N}|x_{1}^{T})\\
&= - \sum_{(x_{1}^{T}, w_{1}^{N})} \sum_{n=1}^{N} \log p(w_n|w_{1}^{n-1},x_{1}^{T}).
\end{split}
\end{equation}

During beam search, the auxilary quantity for each unknown partial string (tree of partial hypotheses) $w_{1}^{n}$ is defined as:
\begin{equation}
\begin{split}
Q(n; w_{1}^{n}) :&= \prod_{n'=1}^{n} p(w_{n'}|w_{0}^{n'-1},x_{1}^{T})\\
&= p(w_{n}|w_{0}^{n-1},x_{1}^{T}) \cdot Q(n-1, w_{1}^{n-1}).
\end{split}
\end{equation}

After discarding the less likely hypotheses in the beam search, the word sequence probability is calculated by the best hypothesis:
\begin{equation}
p(w_{1}^{N}|x_{1}^{T}) = Q(N; w_{1}^{N}).    
\end{equation}

\subsection{SpecAugment}
SpecAugment \cite{park2019specaugment} is a data augmentation technique for \gls{ASR} that manipulates spectrograms to improve model robustness by randomly applying masking in consecutive frames in the time axis as well as consecutive dimensions in the feature axis. It performs three main transformations\footnote{\citet{bahar2019using} analyzed deeply in end-to-end \gls{ST}. \citet{park2019specaugment} stated that time warping is the most expensive
and the least influential, we do not include it here}: time warping, frequency masking, and time masking.

\begin{figure}[h]
    \centering
    \includegraphics[width=1.0\linewidth]{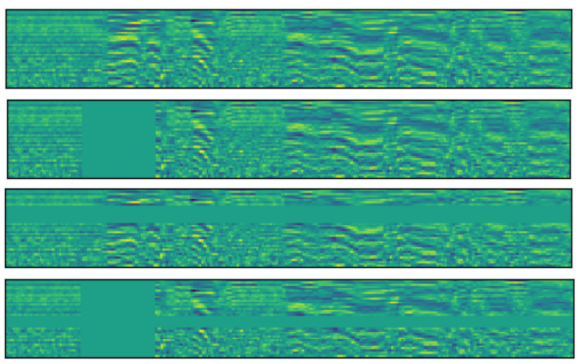}
    \caption{\textbf{SpecAugment visualization}. From top to bottom, the figures show the spectrogram of the input audio with no data augmentation, time masking, frequency masking and both masking applied.}
    \label{fig:specaug}
\end{figure}

Figure \ref{fig:specaug} shows examples of the individual augmentations applied to a single input. 

\textbf{Time Masking}: Given an audio signal $x^{T}_{1} := x_{1}, x_{2}, ..., x_{T}$ of length $T$. Time masking is masking of $\text{\textturntwo}$ successive time steps $[t, t + \tau)$, where we set:
\begin{equation}
(x_t, \dots, x_{t +\text{\textturntwo}}) := 0    
\end{equation}
where $\text{\textturntwo}$ is the masking window selected from a uniform distribution from $0$ to the maximum time mask parameter $\mathbb{TM}$. The time position $t$ is picked from another uniform distribution over $[0, T)$ such that the maximum sequence length $T$ is not exceeded (i.e. if $t +\text{\textturntwo} > T$, we set it to $T$).

\textbf{Frequency Masking}: Frequency masking is applied such that $\phi$ consecutive frequency channels $[f, f +\phi)$ are masked, where $\phi$ is selected from a uniform distribution from 0 to the frequency mask parameter $\mathbb{FM}$, and $f$ is chosen from $[0, \nu)$, where $\nu$ is the input feature dimension, e.g. the number of \gls{MFCC} channels. For raw waveform as input, $\nu=1$. Similar to time masking, if $f +\phi > \nu$, we set it to $f=\nu$.

\onecolumn
\section{Dataset Comparison with Literature}
\label{sec.dataset_comparison_with_literature}

\input{tables/data_stats_ASR_literaturecompare}

\onecolumn
\input{tables/data_stats_comparison_medical}

\onecolumn
\input{tables/data_stats_comparison_ST}

\onecolumn

\twocolumn
\section{Details of Experimental Setup}
\label{sec:details_experimental_setup}
\subsection{Training Setup: Whisper}
Whisper, a Transformer-based \gls{AED} (see Appendix Section \ref{sec:AED}), is an end-to-end multitask \gls{ASR} and \gls{ST} model pre-trained on 680k hours of labeled data. Approximately 65\% of the data (equivalent to 438,000 hours) consists of English-language audio paired with English transcripts. Around 18\% (or 126,000 hours) comprises non-English audio with English transcripts, while the remaining 17\% (or 117,000 hours) includes non-English audio along with their corresponding transcripts. The non-English data encompasses 98 distinct languages.

For \gls{ASR}, we performed a full fine-tuning (both encoder and decoder) monolingually (each language separately) and multilingually (all languages simultaneously). For \gls{ST}, we performed a full fine-tuning bilingually (each language pair separately) and multilingually (all language pairs simultaneously).

\begin{figure}[h]
\centering
\includegraphics[width=\columnwidth]{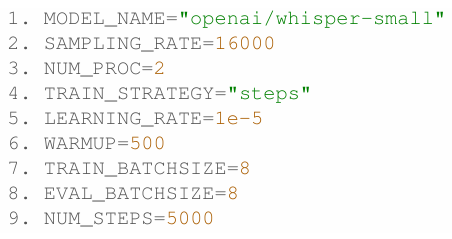}
\caption{Fine-tuning configuration of \textbf{Whisper-small} model}
\label{fig:config_whisper-ASR-small}
\end{figure}

\begin{figure}[h]
\centering
\includegraphics[width=\columnwidth]{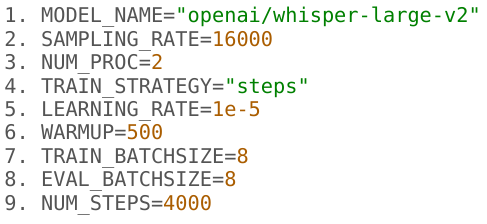}
\caption{Fine-tuning configuration of \textbf{Whisper-large-v2} model}
\label{fig:config_whisper-ASR-large-v2}
\end{figure}

\textbf{Whisper variants}: We employed 2 variants of Whisper models: Whisper-small\footnote{https://huggingface.co/openai/whisper-small} (244M parameters) and Whisper-large-v2\footnote{https://huggingface.co/openai/whisper-large-v2} (1550M parameters). Figure \ref{fig:config_whisper-ASR-small} and Figure \ref{fig:config_whisper-ASR-large-v2} show the fine-tuning configuration of Whisper-small model and Whisper-large-v2 model respectively.

\textbf{Pre-processing setup}: To preprocess data for the Whisper models, you must prepare audio files and their corresponding text transcriptions in a format suitable for training or fine-tuning. Begin by converting audio files to a consistent format (e.g., 16 kHz, mono-channel WAV files) to ensure compatibility. Use libraries like ffmpeg\footnote{https://www.ffmpeg.org/} or librosa\footnote{https://librosa.org/} for this purpose. Normalize and clean the transcriptions by removing extraneous characters, correcting spelling, and aligning timestamps with the audio. Tokenize the text using Whisper's tokenizer, ensuring it matches the pre-trained model's vocabulary. Additionally, segment long audio files into smaller chunks with overlapping windows to fit the model's input length constraints while preserving context. Finally, package the processed audio-text pairs into a dataset format such as JSON, which includes metadata like file paths, transcription text, and optional timestamps for alignment.

\begin{figure*}[t]
    \centering
    \includegraphics[width=\linewidth]{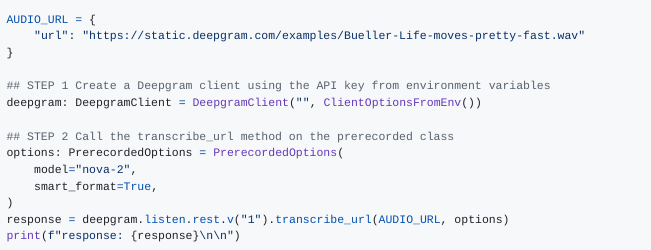}
    \caption{\textbf{Deepgram API call}. The model employed in our experiments is Nova-2 version. We used the API to directly recognize the audio files instead of fine-tuning.}
    \label{fig:deepgram_config_API}
\end{figure*}

\textbf{Training setup}: 
\begin{itemize}
    \item Whisper-small: As shown in Figure \ref{fig:config_whisper-ASR-small}, the training configuration for the \texttt{openai/whisper-small} model is detailed as follows. The model was trained using a sampling rate of 16,000 Hz (\texttt{SAMPLING\_RATE}) to process audio data effectively. The training utilized 2 processors (\texttt{NUM\_PROC}) to parallelize computations. A step-based training strategy (\texttt{TRAIN\_STRATEGY="steps"}) was adopted, where the model was trained for 5,000 steps (\texttt{NUM\_STEPS}). The learning rate was set to $1 \times 10^{-5}$ (\texttt{LEARNING\_RATE}), with a warmup period of 500 steps (\texttt{WARMUP}) to stabilize training. Both the training and evaluation batch sizes were configured as 8 (\texttt{TRAIN\_BATCHSIZE} and \texttt{EVAL\_BATCHSIZE}, respectively) to ensure efficient memory usage while maintaining model performance.
    \item Whisper-large-v2: As shown in Figure \ref{fig:config_whisper-ASR-large-v2}, the training setup for fine-tuning the Whisper model leverages the \texttt{openai/whisper-large-v2} architecture. The audio inputs are resampled to a sampling rate of 16,000 Hz (\texttt{SAMPLING\_RATE=16000}) for consistency with the model's requirements. Training is distributed across two processing units (\texttt{NUM\_PROC=2}) using a step-based training strategy (\texttt{TRAIN\_STRATEGY="steps"}). The learning rate is set to a modest value of $1 \times 10^{-5}$ (\texttt{LEARNING\_RATE=1e-5}) with a warm-up phase spanning 500 steps (\texttt{WARMUP=500}) to stabilize the optimization process. The training batch size and evaluation batch size are both configured to 8 (\texttt{TRAIN\_BATCHSIZE=8} and \texttt{EVAL\_BATCHSIZE=8}, respectively), balancing computational efficiency with memory constraints. The total number of training steps is capped at 4,000 (\texttt{NUM\_STEPS=4000}), ensuring effective model convergence without overfitting.
\end{itemize}

\subsection{Training Setup: Deepgram}
Deepgram\footnote{https://deepgram.com/} is also a Transformer-based \gls{AED} architecture. We employed the Nova-2 version, which supports over 30 languages. Since this is a commercial API, we could only employ direct recognition on audio files instead of fine-tuning, as shown in Figure \ref{fig:deepgram_config_API}. 

\subsection{Training Setup: AssemblyAI}
AssemblyAI\footnote{https://www.assemblyai.com/} is a Conformer-based \gls{RNN-T} architecture. We used Universal-2 with 600M parameters, pre-trained on from 150,000 hours to 300,000 hours of supervised multilingual data.

\textbf{Special Tokenization}\footnote{https://www.assemblyai.com/research/universal-2}: \gls{RNN-T} demonstrates a constrained capacity to generate consecutive identical tokens. Previous studies \cite{ghodsi2020rnn, xu2023efficient} have indicated that \gls{RNN-T} possesses a pronounced inductive bias that inhibits the prediction of identical tokens in succession. Therefore, Universal-2 incorporates a unique \textcolor{blue}{<repeat\_token>} within its tokenization scheme, which is inserted between repeated tokens in the target sequences of the training data. This modification eliminates the need for the \gls{RNN-T} model to predict the same token multiple times consecutively. Consequently, it enables accurate recognition of repeated tokens without deletions, addressing a limitation of the \gls{RNN-T} architecture. During inference, the \textcolor{blue}{<repeat\_token>} is removed from the final \gls{ASR} output.

\textbf{\gls{RNN-T}}: The \gls{RNN-T} encoder was pre-trained on 12.5 million hours of diverse, multilingual audio data. Following the pre-training phase, the encoder was integrated with a randomly initialized decoder, and the complete model underwent fine-tuning utilizing a combination of the aforementioned supervised dataset and a pseudo-labeled dataset.

\begin{figure*}[h]
    \centering
    \includegraphics[width=\linewidth]{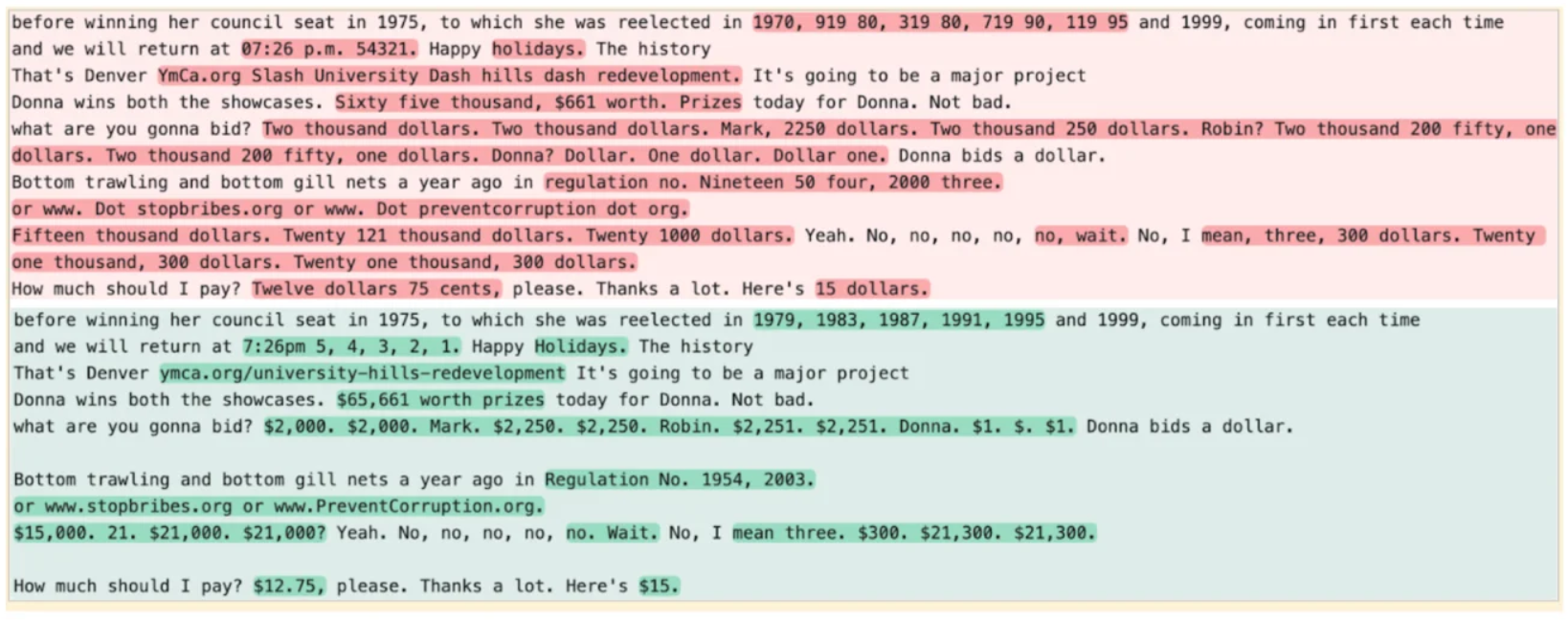}
    \caption{\textbf{An example of text formating in AssemblyAI's Universal-2}. Green text is the final \gls{ASR} output and red text is the \gls{ASR} output before it is processed by text formatting module.}
    \label{fig:assemblyAI_textformating_example}
\end{figure*}

\textbf{Text Formating}: The Text Formatting module processes raw transcripts into well-structured text by incorporating Punctuation Restoration, Truecasing, and \gls{ITN}, ensuring the final output is both highly readable and adaptable for diverse applications, as shown in Figure \ref{fig:assemblyAI_textformating_example}.

\begin{figure*}[h]
    \centering
    \includegraphics[width=\linewidth]{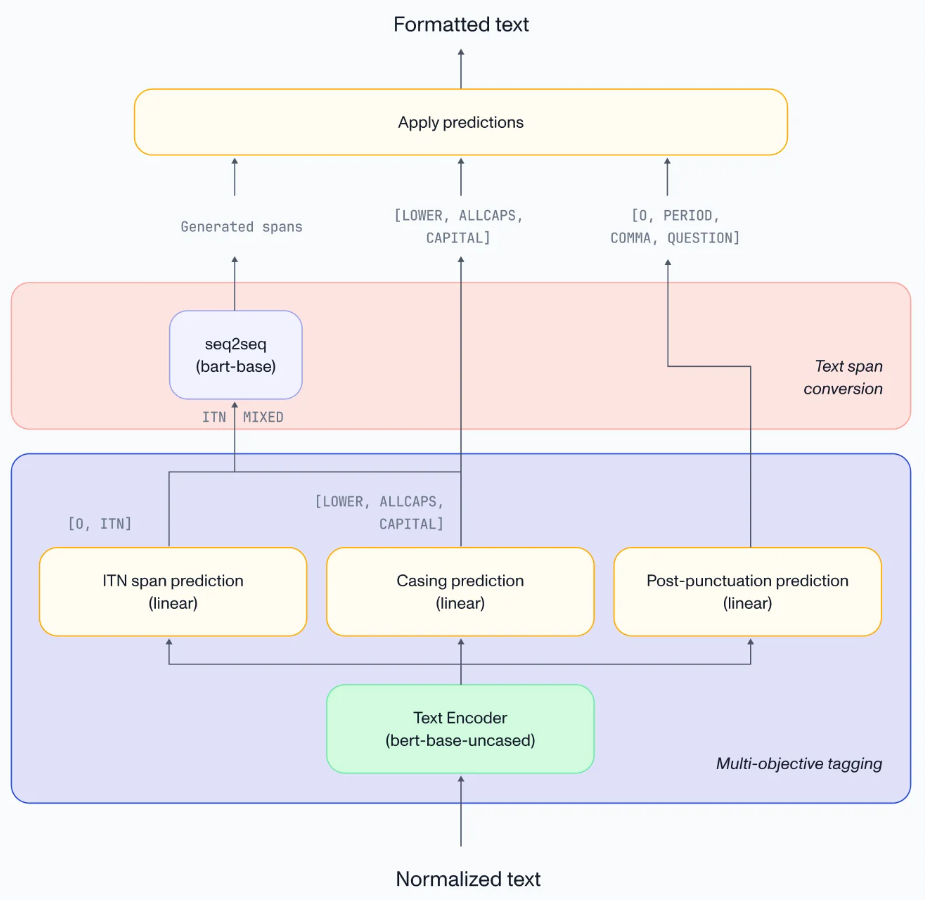}
    \caption{\textbf{AssemblyAI's Universal-2 Text Formatting Architecture}.}
    \label{fig:assemblyAI_textformating_architecture}
\end{figure*}

\textbf{Text Formatting Architecture}: Figure \ref{fig:assemblyAI_textformating_architecture} shows the visualization of Universal-2 Text Formatting Architecture. The architecture is described as below:
\begin{itemize}
    \item Token-based Truecasing: Universal-1 employed a character-based model for Truecasing, which exhibited susceptibility to hallucination errors and incurred increased computational overhead. Universal-2 switched to token-based modeling resulting in more accurate Truecasing with reduced computational demands.
    \item \gls{seq2seq} Modeling for \gls{ITN}: Universal-2 employs a \gls{seq2seq} model, which more effectively captures contextual information for \gls{ITN} compared to a rule-based approach.
    \item Multi-objective tagging model: The model comprises a shared Transformer encoder, followed by three distinct classification heads designed to perform specific tasks: (1) post-punctuation prediction, (2) token-level truecasing to address all-uppercase, all-lowercase, word capitalization, and mixed-case word identification, and (3) textual span detection for \gls{ITN} processing.
    \item Text span conversion model: The \gls{seq2seq} model employs a Transformer-based encoder-decoder architecture and is utilized to process normalized mixed-case and \gls{ITN} spans identified by the multi-objective tagging model, generating their corresponding formatted representations.
\end{itemize}

\subsection{Training Setup: mBART}
mBART-50 \cite{tang2020multilingual} is a multilingual \gls{seq2seq} model designed to demonstrate the feasibility of creating multilingual \gls{MT} models through multilingual fine-tuning. Rather than fine-tuning the model for a single translation direction, it is fine-tuned across multiple translation directions simultaneously. The mBART-50 model extends the original mBART framework by incorporating 25 additional languages, enabling support for multilingual \gls{MT} across 50 languages. The pre-trained model, mBART-large-50, is primarily optimized for fine-tuning on \gls{MT} but can also be adapted for other multilingual \gls{seq2seq} applications.

\begin{figure}
    \centering
    \includegraphics[width=\linewidth]{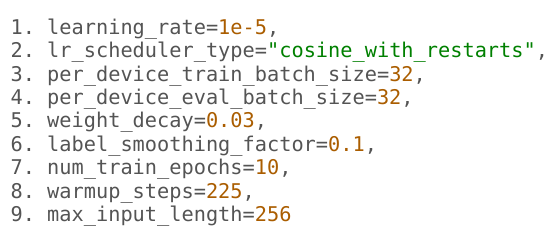}
    \caption{Fine-tuning configuration of \textbf{mBART-large-50} model}
    \label{fig:config_mbart}
\end{figure}

\textbf{Pre-processing setup}: Due to the multilingual nature of the model, it requires input sequences to adhere to a specific format. A unique language identifier token is employed as a prefix in both the source and target texts. The format for the text is \textcolor{blue}{$[lang\_code] X [eos]$}, where \textcolor{blue}{$X$} represents the source or target text, and \textcolor{blue}{$lang\_code$} corresponds to the source language code for the source text and the target language code for the target text. The beginning-of-sequence (\textcolor{blue}{$bos$}) token is not utilized. Once the examples are formatted in this manner, the model can be trained as a standard \gls{seq2seq} model. Pre-processing might also involve cleaning data (removing noise, handling encoding issues), truncating or padding sentences to the maximum sequence length supported by the model, and batching data for efficient processing.

\textbf{Training setup}: As shown Figure \ref{fig:config_mbart}, the training setup for the mBART-large-50 model is designed to optimize performance with a range of hyperparameters. The learning rate is set to a relatively low value of 1e-5, ensuring fine-tuning without overshooting optimal solutions. The learning rate schedule follows a cosine annealing with restarts strategy, allowing for periodic adjustments to prevent overfitting as training progresses. A batch size of 32 for both training and evaluation is chosen to balance computational efficiency and model convergence. Regularization is applied with a weight decay of 0.03, and label smoothing of 0.1 is used to help the model generalize better by softening the target labels. Training runs for 10 epochs, providing ample opportunity for the model to adapt to the dataset. A warmup phase with 225 steps is included to gradually ramp up the learning rate and avoid instability at the start of training. The maximum input sequence length is capped at 256 tokens, optimizing memory usage while accommodating most text sequences.

\subsection{Training Setup: M2M100}
M2M100-418M is a multilingual encoder-decoder model designed for many-to-many multilingual \gls{MT}. This model is capable of directly translating across 9,900 translation directions involving 100 languages.

\textbf{Preprocessing setup}: To translate into a target language, the target language identifier (id) is designated as the first generated token. This can be achieved by specifying the \textcolor{blue}{$forced\_bos\_token\_id$} parameter in the $generate$ method. The M2M100Tokenizer relies on SentencePiece\footnote{https://pypi.org/project/sentencepiece/} (SPM). All datasets must undergo detokenization prior to the application of SPM during the data pre-processing phase. Following the download of raw data, it is necessary to post-process the data, apply SPM, and then binarize the dataset.

\begin{figure}[t]
    \centering
    \includegraphics[width=\linewidth]{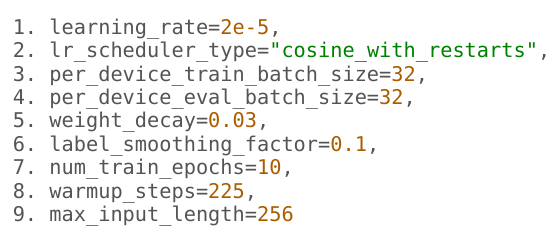}
    \caption{Fine-tuning configuration of \textbf{M2M100-148M} model}
    \label{fig:config_m2m100}
\end{figure}

\textbf{Training setup}: As shown in Figure \ref{fig:config_m2m100}, the learning rate was set to $2 \times 10^{-5}$, with a \texttt{cosine\_with\_restarts} learning rate scheduler employed to dynamically adjust the learning rate during training. A per-device batch size of 32 was used for both training and evaluation to balance computational efficiency and memory usage. Weight decay was applied with a factor of 0.03 to regularize the model and prevent overfitting. To further enhance generalization, a label smoothing factor of 0.1 was introduced. The model was trained for 10 epochs, with the first 225 steps dedicated to warmup to allow a gradual ramp-up of the learning rate. Additionally, the maximum input length for sequences was capped at 256 tokens to ensure efficient processing of data. This training configuration was chosen to achieve optimal performance on multilingual \gls{MT} tasks.

\subsection{Training Setup: Marian}
Marian is an encoder-decoder fine-tuned on one-to-one translation task, which is built upon BART architecture. The original Marian is a highly efficient and open-source \gls{NMT} framework, implemented in pure C++ with minimal external dependencies. Its development is primarily led by the Microsoft Translator team. All models are transformer encoder-decoders with 6 layers in each component. 

\begin{figure}[h]
    \centering
    \includegraphics[width=\linewidth]{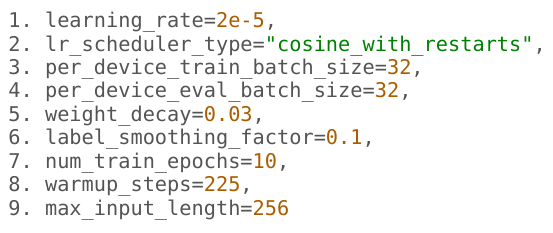}
    \caption{Fine-tuning configuration of \textbf{Marian} model}
    \label{fig:config_marian}
\end{figure}

\begin{figure*}[t]
    \centering
    \includegraphics[width=\linewidth]{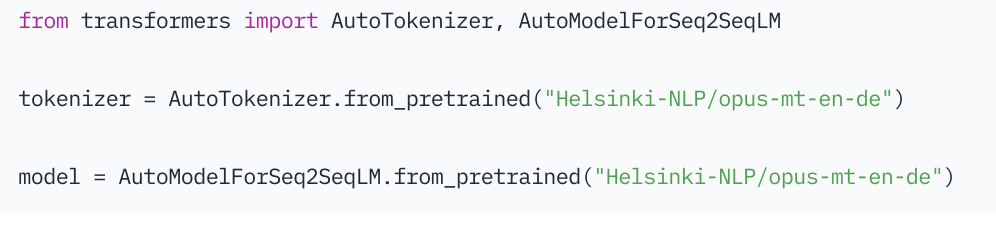}
    \caption{\textbf{Python Hugging Face version}. Marian C++ version is much more efficient for training but it is more difficult to train and deploy. Thus in the scope of our experiments, we only used Hugging Face implementation.}
    \label{fig:marian_HFcall}
\end{figure*}

We employed the Python Hugging Face version for training instead of C++ version, as shown in Figure \ref{fig:marian_HFcall}.

\textbf{Training setup}: As shown in Figure \ref{fig:config_marian}, the learning rate was set to $2 \times 10^{-5}$, and a cosine learning rate scheduler with restarts (\texttt{cosine\_with\_restarts}) was utilized to adjust the learning rate dynamically during training. The batch size for both training and evaluation was fixed at 32 samples per device (\texttt{per\_device\_train\_batch\_size} and \texttt{per\_device\_eval\_batch\_size}). A weight decay value of 0.03 was applied to mitigate overfitting, and label smoothing with a factor of 0.1 was incorporated to improve model generalization. The model was trained for a total of 10 epochs (\texttt{num\_train\_epochs}), with 225 warmup steps (\texttt{warmup\_steps}) to stabilize the optimization process. Additionally, the maximum input sequence length was restricted to 256 tokens (\texttt{max\_input\_length}) to efficiently handle the computational requirements.

\subsection{Training Setup: Llama}
The Meta Llama 3.1 series comprises a collection of multilingual \glspl{LLM} designed for text-based input and output. These pre-trained and instruction-tuned generative models, specifically the 8B parameter variant, are optimized for multilingual dialogue applications. The Llama 3.1 instruction-tuned models demonstrate superior performance compared to numerous open-source and proprietary conversational models across standard industry benchmarks. In our experiments, we employed Llama-3.1-8B model.

\textbf{Model architecture}: Llama-3.1-8B is an autoregressive language model (decoder-only model) designed with an optimized transformer architecture. The fine-tuned variants employ \gls{SFT} and reinforcement learning with human feedback (RLHF) to enhance alignment with human preferences for both helpfulness and safety.

\begin{figure*}[t]
    \centering
    \includegraphics[width=\linewidth]{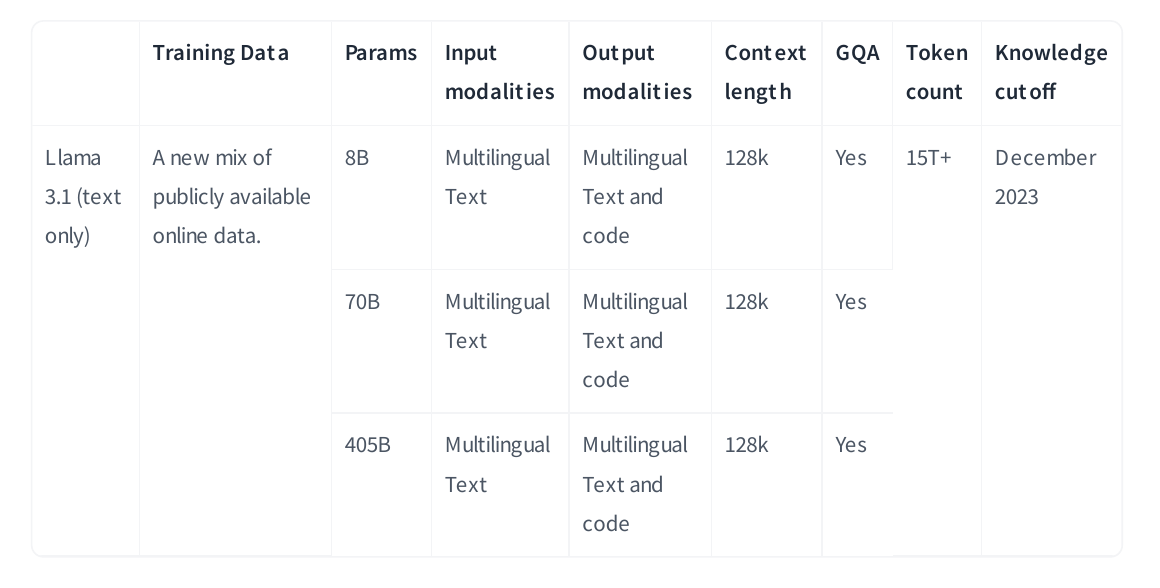}
    \caption{\textbf{Model card of Llama 3.1 family}. The Llama 3.1 family of models was pre-trained on approximately 15 trillion tokens sourced from publicly available datasets, with token counts reflecting pre-training data exclusively. All versions of Llama 3.1 utilize \gls{GQA} to enhance inference scalability. Fine-tuning was conducted using a combination of publicly available instruction datasets and over 25 million synthetically generated examples. The pre-training dataset has a cutoff date of December 2023.}
    \label{fig:llama3_modelcard}
\end{figure*}

\begin{figure}[h]
    \centering
    \includegraphics[width=\linewidth]{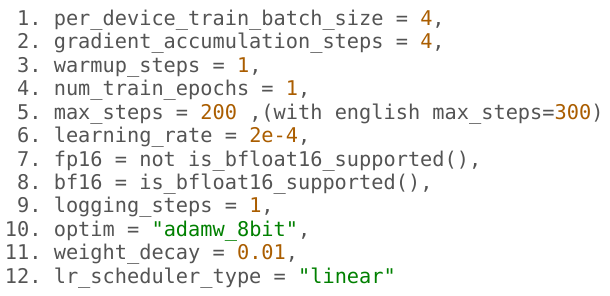}
    \caption{Fine-tuning configuration of \textbf{\gls{LLM}} model}
    \label{fig:config_LLM}
\end{figure}

\textbf{Training setup}: As shown in Figure \ref{fig:config_LLM}, the training setup for the Llama-3.1-8B model was configured with a per-device training batch size of 4 and a gradient accumulation of 4 steps to effectively utilize hardware resources. The training process involved a warmup phase consisting of 1 step, followed by a single training epoch. The maximum number of training steps was set to 200, with an extended English configuration allowing up to 300 steps. A learning rate of $2 \times 10^{-4}$ was used, alongside mixed-precision training, where FP16 was enabled if BF16 support was unavailable. Conversely, BF16 was activated on supported devices. The AdamW optimizer in its 8-bit variant was employed, with a weight decay of 0.01 to mitigate overfitting. A linear learning rate scheduler was adopted, and logging was performed at every step to ensure detailed progress tracking throughout the training process.

\subsection{Training Setup: Qwen}
The Qwen2.5 \glspl{LLM} have been pre-trained on a newly developed large-scale dataset comprising up to 18 trillion tokens, representing a substantial expansion compared to Qwen2. This enhanced pre-training has endowed Qwen2.5 with significantly improved capabilities, including advanced instruction-following, the ability to generate extended texts exceeding 8,000 tokens, improved comprehension of structured data (e.g., tables), and enhanced generation of structured outputs, particularly in JSON format. Qwen2.5 supports a context length of up to 128,000 tokens and can produce outputs of up to 8,000 tokens. Additionally, these models maintain multilingual functionality, encompassing more than 29 languages, such as Chinese, English, French, Spanish, Portuguese, German, Italian, Russian, Japanese, Korean, and Vietnamese.

\begin{figure*}[t]
    \centering
    \includegraphics[width=\linewidth]{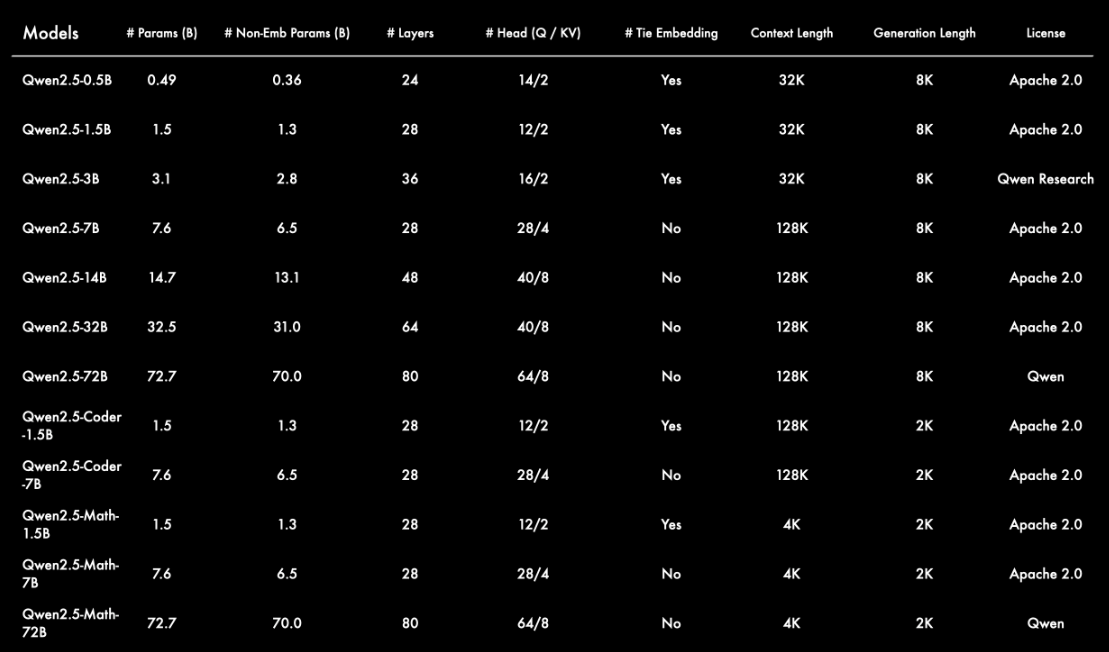}
    \caption{\textbf{Model card of Qwen2.5 family}.The training setup for Qwen-2.5-7B features a causal language model architecture. This model employs transformers with various advanced components, including RoPE (Rotary Positional Embeddings), SwiGLU activation functions, RMSNorm normalization, and Attention QKV bias. With a total of 7.61 billion parameters, the Qwen2.5 model has 6.53 billion parameters dedicated to non-embedding components. The model consists of 28 layers, with attention heads configured as 28 for the Query (Q) and 4 for Key-Value (KV). The model is designed for a context length of 131,072 tokens, allowing for processing of long-range dependencies in text sequences during pre-training.}
    \label{fig:qwen_modelcard}
\end{figure*}

\begin{figure*}[t]
    \centering
    \includegraphics[width=\linewidth]{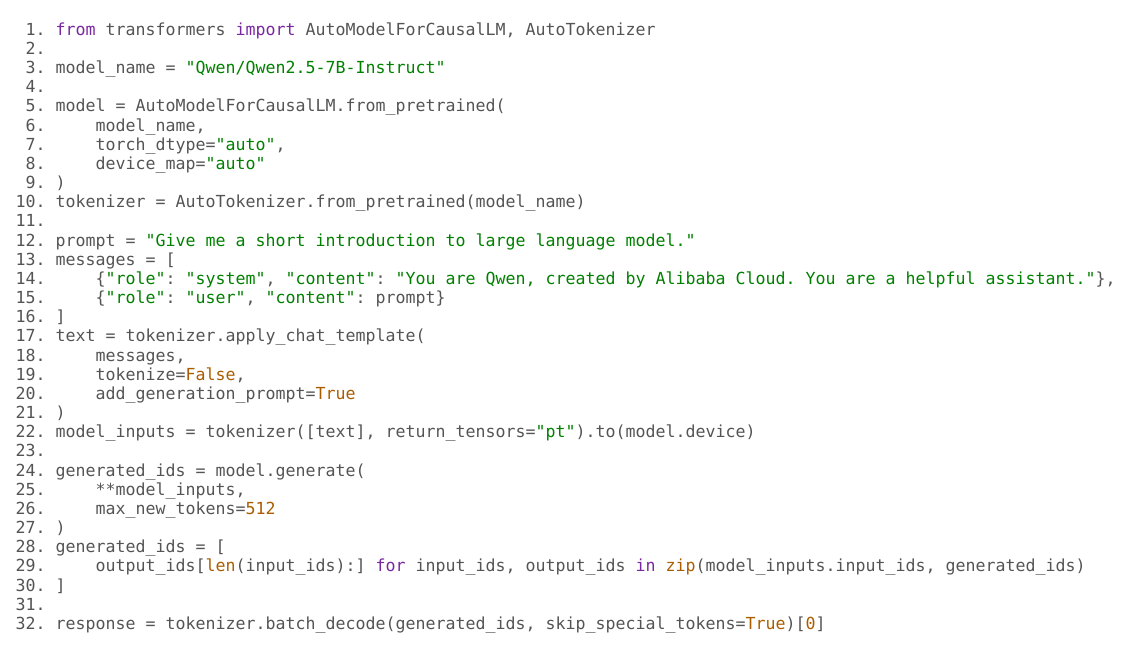}
    \caption{\textbf{Hugging Face implementation of Qwen-2.5-7B model}.The implementation for Qwen-2.5-7B is also conducted via Ollama to run locally. We can also access the Ollama service via its OpenAI-compatible API. However, in the scope of our experiments, we only used Hugging Face for training.}
    \label{fig:qwen_HFcall}
\end{figure*}

We employed Qwen-2.5-7B version. Its model card is shown in Figure \ref{fig:qwen_modelcard} and its Hugging Face implementation is shown in Figure \ref{fig:qwen_HFcall}. Qwen-2.5-7B could also be run locally via Ollama\footnote{https://github.com/ollama/ollama} service. However, in the scope of our experiments, we only used Hugging Face for training.

\textbf{Training setup}: As shown in Figure \ref{fig:config_LLM}, the training setup for the Qwen-2.5-7B model was configured with a per-device training batch size of 4 and a gradient accumulation of 4 steps to effectively utilize hardware resources. The training process involved a warmup phase consisting of 1 step, followed by a single training epoch. The maximum number of training steps was set to 200, with an extended English configuration allowing up to 300 steps. A learning rate of $2 \times 10^{-4}$ was used, alongside mixed-precision training, where FP16 was enabled if BF16 support was unavailable. Conversely, BF16 was activated on supported devices. The AdamW optimizer in its 8-bit variant was employed, with a weight decay of 0.01 to mitigate overfitting. A linear learning rate scheduler was adopted, and logging was performed at every step to ensure detailed progress tracking throughout the training process.

\subsection{Training Setup: Mistral}
Mistral 7B is an \gls{LLM} consisting of 7 billion parameters, developed and released by Mistral AI. This model has been meticulously engineered to offer a balance of computational efficiency and high performance, making it suitable for practical applications. Upon its release, Mistral 7B demonstrated superior performance across all evaluated benchmarks, surpassing the leading open-source 13B-parameter model, Llama 2. We employed Mistral-v0.3-7B\footnote{https://huggingface.co/mistralai/Mistral-7B-v0.3} version.

\textbf{Model architecture}: The model incorporates attention mechanisms such as
\begin{itemize}
    \item \gls{GQA}: which enhances inference speed and reduces memory usage during decoding
    \item Sliding Window Attention (SWA) \cite{child2019generating, beltagy2020longformer}: enabling the processing of sequences of arbitrary length while minimizing inference cost, in which each layer attends to the previous 4,096 hidden states. The primary advancement, and the primary motivation for the initial investigation, is the linear computational cost of \( O(\text{sliding\_window.seq\_len}) \).\\
    Sliding window attention leverages the hierarchical structure of transformer layers to extend the receptive field beyond the fixed window size. Specifically, a token \(i\) at layer \(k\) attends to the tokens in the range \([i - \text{sliding\_window}, i]\) at layer \(k-1\). These attended tokens, in turn, have attended to tokens in the range \([i - 2 \times \text{sliding\_window}, i]\) at layer \(k-2\). As a result, higher layers are able to access information from tokens further in the past than what the local attention pattern of the current layer suggests.\\
    Finally, a constrained attention span allows for the limitation of the cache size to that of a sliding window of tokens, facilitated by the use of rotating buffers. This approach reduces the cache memory requirement by 50\% for inference on sequences of length 8192, without compromising model performance.
    \item FlashAttention \cite{dao2022flashattention, dao2023flashattention2} and xFormers \cite{xFormers2022}: In practice, changes made to FlashAttention and xFormers yield a 2x speed improvement for sequence length of 16k with a window of 4k.
\end{itemize}

\textbf{Training setup}: As shown in Figure \ref{fig:config_LLM}, the training setup for the Mistral-v0.3-7B model was configured with a per-device training batch size of 4 and a gradient accumulation of 4 steps to effectively utilize hardware resources. The training process involved a warmup phase consisting of 1 step, followed by a single training epoch. The maximum number of training steps was set to 200, with an extended English configuration allowing up to 300 steps. A learning rate of $2 \times 10^{-4}$ was used, alongside mixed-precision training, where FP16 was enabled if BF16 support was unavailable. Conversely, BF16 was activated on supported devices. The AdamW optimizer in its 8-bit variant was employed, with a weight decay of 0.01 to mitigate overfitting. A linear learning rate scheduler was adopted, and logging was performed at every step to ensure detailed progress tracking throughout the training process.

\subsection{Training Setup: Google Translate}
Using Google Translate as a \gls{MT} model in a cascaded \gls{ST} system can provide a powerful and scalable solution for real-time multilingual communication. In a cascaded \gls{ST} setup, the process typically involves two stages: first, \gls{ASR} module converts audio into text, and then an \gls{MT} model like Google Translate is used to render that text into the desired language. 

By leveraging Google Translate, which is backed by advanced \gls{NMT} techniques, the system can provide high-quality, context-aware translations. The integration of Google Translate into the \gls{ST} system offers several benefits, including the ability to handle a wide range of language pairs, rapid updates, and continuous improvements due to the vast data the system processes. Additionally, Google Translate has been trained on massive multilingual corpora, which helps it deal with diverse linguistic nuances and idiomatic expressions. 

However, this approach also comes with challenges. One potential issue is that the quality of the \gls{ASR} output plays a critical role in the overall effectiveness of the \gls{MT}. If the \gls{ASR} system produces too many errors or misinterprets the audio, Google Translate will likely propagate these errors, leading to inaccuracies in the final translated output. Furthermore, Google Translate may struggle with medical-domain language or highly medical content, which may require fine-tuning or customization to ensure higher \gls{MT} accuracy. 

Despite these challenges, using Google Translate in a cascaded \gls{ST} system remains a viable and practical solution for multilingual communication, especially when quick deployment and ease of integration are paramount. It is also an ideal solution when working with a wide array of languages, as Google Translate supports over 100 languages, making it adaptable to diverse linguistic needs.

\subsection{Training Setup: VinAI Translate}
\begin{figure}[h]
    \centering
    \includegraphics[width=\linewidth]{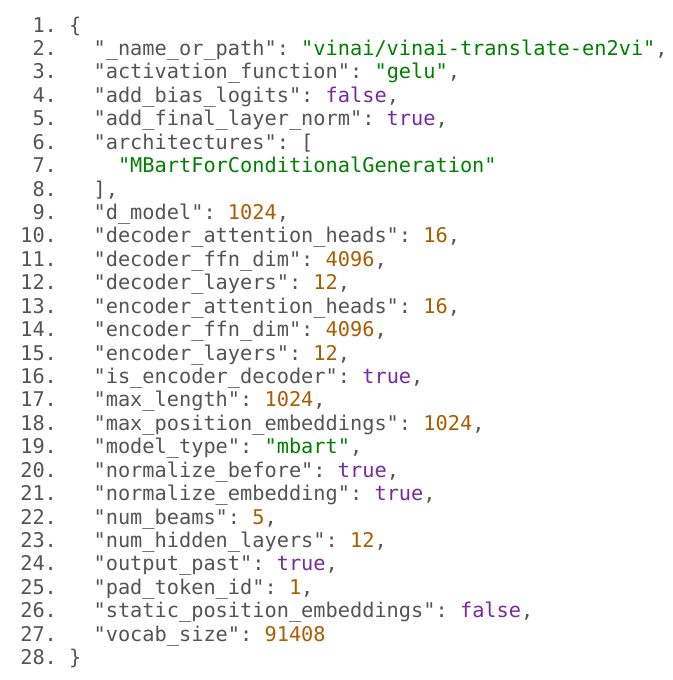}
    \caption{Fine-tuning configuration of \textbf{VinAI Translate} model}
    \label{fig:config_vinAItranslate}
\end{figure}

The pre-trained VinAI Translate models represent state-of-the-art systems for Vietnamese-to-English and English-to-Vietnamese text translation. The platform features a user-friendly, interactive interface and incorporates advanced models for \gls{ASR}, \gls{MT}, and text-to-speech (TTS). Experimental results demonstrate that the system achieves state-of-the-art performance, surpassing Google Translate in both automated and human evaluations on publicly available Vietnamese-English translation benchmarks.

In our experiments, we only leveraged the \gls{MT} module of VinAI Translate for our bilingual Vietnamese-English cascaded \gls{ST} systems.

\textbf{The pre-trained \gls{MT} component}: The pre-trained \gls{seq2seq} model mBART is first fine-tuned using 3M high-quality English-Vietnamese sentence pairs from the PhoMT dataset \cite{doan2021phomt} for English-to-Vietnamese translation. Subsequently, the fine-tuned model is employed to translate English sentences from "noisy" datasets into Vietnamese. Sentence pairs with BLEU scores in the range of 0.15 to 0.95 are selected, resulting in an additional 6M pairs. Combining these with the initial 3M pairs yields a total of 9M high-quality sentence pairs. To simulate \gls{ASR} output, this dataset is augmented for each translation direction (English-to-Vietnamese and Vietnamese-to-English) by applying lowercase conversion and punctuation removal to the source sentences while keeping the target sentences unchanged. This augmentation adds another 9M sentence pairs for each direction, resulting in 18M sentence pairs per direction. The mBART model is then fine-tuned for each translation direction using the full 18M sentence pairs to develop the machine translation \gls{MT} component.

\textbf{Training setup}: As shown in Figure \ref{fig:config_vinAItranslate}, we adopted pre-trained \gls{MT} model from VinAI Translate to fine-tune on our own dataset. The model utilizes the \texttt{MBartForConditionalGeneration} framework with a transformer-based encoder-decoder structure. The encoder and decoder each consist of 12 layers, with 16 attention heads per layer, and a \gls{FFW} dimension of 4096. The hidden layer size is set to 1024, with a maximum sequence length and position embeddings capped at 1024. The activation function employed is \texttt{gelu}, and both embedding and layer normalization are applied before each layer. The model includes a vocabulary size of 91,408 tokens and does not use static position embeddings. During generation, the beam search decoding strategy is employed with 5 beams. Additional features include the use of a pad token ID of 1, final layer normalization, and bias-free logits.

\subsection{Training Setup: EnViT5}
\begin{figure}[h]
    \centering
    \includegraphics[width=\linewidth]{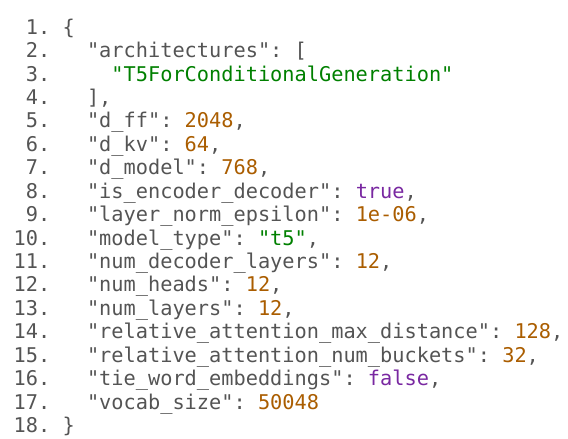}
    \caption{Fine-tuning configuration of \textbf{EnViT5} model}
    \label{fig:config_EnViT5}
\end{figure}

The EnViT5 model is a Text-to-Text Transformer based on the encoder-decoder architecture introduced within the T5 framework proposed by \citet{raffel2020exploring}. For pre-training, the model utilizes the CC100 dataset, a monolingual dataset derived from web crawl data, as described by \citet{wenzek2020ccnet}. This corpus comprises monolingual data for over 100 languages. Subsequently, the model is fine-tuned using MTet, the largest publicly available parallel corpus for English-Vietnamese translation. MTet, as published along with EnViT5 model, contains 4.2 million high-quality training sentence pairs and includes a multi-domain test set curated by the Vietnamese research community.

\textbf{Training setup}: As shown in Figure \ref{fig:config_EnViT5}, the training setup utilizes a model architecture based on the \texttt{T5ForConditionalGeneration} class, designed for tasks requiring a Transformer-based encoder-decoder structure. The model configuration includes a hidden dimensionality (\texttt{d\_model}) of 768, with \gls{FFW} sublayers of size 2048 (\texttt{d\_ff}) and key-value dimensionality (\texttt{d\_kv}) of 64. It consists of 12 encoder layers and 12 decoder layers (\texttt{num\_layers} and \texttt{num\_decoder\_layers}), each employing 12 attention heads (\texttt{num\_heads}). Relative position embeddings are implemented with a maximum attention distance of 128 (\texttt{relative\_attention\_max\_distance}) and 32 buckets (\texttt{relative\_attention\_num\_buckets}). Layer normalization is applied with an epsilon value of $10^{-6}$ (\texttt{layer\_norm\_epsilon}). The model does not tie word embeddings (\texttt{tie\_word\_embeddings} = \texttt{false}) and supports a vocabulary size of 50,048 tokens (\texttt{vocab\_size}).

\subsection{Training Setup: SeamlessM4T}
\begin{figure}[h]
    \centering
    \includegraphics[width=\linewidth]{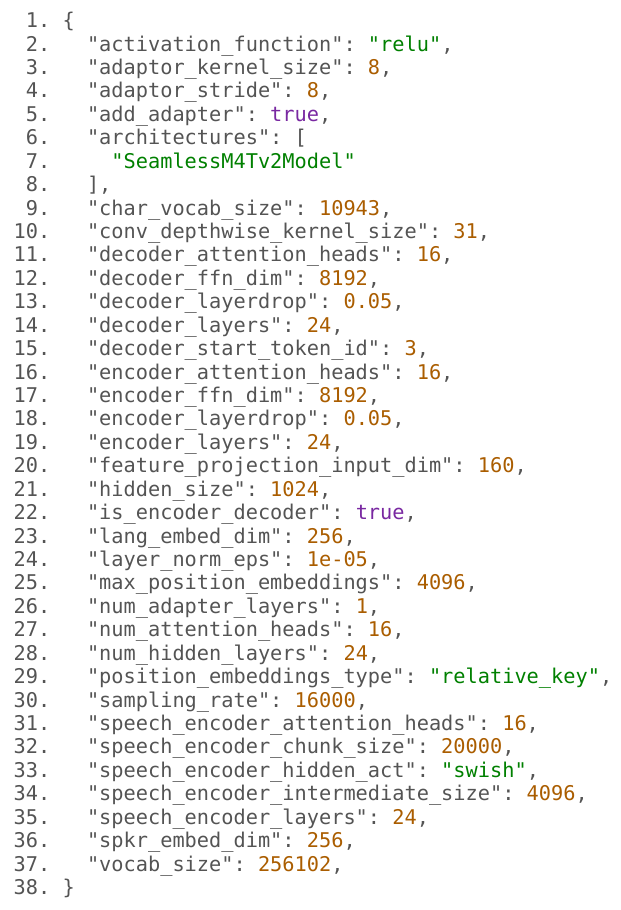}
    \caption{Fine-tuning configuration of \textbf{SeamlessM4T-large-v2} model}
    \label{fig:config_seamlessm4t}
\end{figure}

SeamlessM4T is a foundational, all-in-one massively multilingual and multimodal \gls{MT} model designed to provide high-quality translations for both speech and text across nearly 100 languages. The SeamlessM4T framework supports the following tasks: Speech-to-speech translation (S2ST), Speech-to-text translation (S2TT), Text-to-speech translation (T2ST), Text-to-text translation (T2TT), \gls{ASR}.

Key language support capabilities include: 101 languages for speech input, 96 languages for text input and output, 35 languages for speech output. As the first model of its kind, SeamlessM4T enables simultaneous S2ST and S2ST for multiple source and target languages.  

The latest version of SeamlessM4T incorporates multitask-UnitY2, featuring a non-autoregressive unit decoder and hierarchical upsampling to enhance data efficiency in predicting translation units. Additionally, the model includes the w2v-BERT 2.0 speech encoder, pre-trained on 4.5 million hours of unlabeled audio data. The multitask model has been fine-tuned with increased supervision using automatically aligned data pairs to improve performance, particularly for low-resource languages.  

SeamlessM4T leverages the Efficient Monotonic Multihead Attention (EMMA) mechanism \cite{ma2023efficient}, allowing for low-latency generation of target translations without requiring complete source utterances, thereby enabling real-time \gls{MT} capabilities.

\textbf{Training setup}: As shown in Figure \ref{fig:config_seamlessm4t}, the SeamlessM4T-large-v2 model is designed to leverage a robust architecture with an encoder-decoder framework, incorporating 24 encoder and 24 decoder layers. The encoder and decoder use 16 attention heads, a hidden size of 1024, and \gls{FFW} network dimensions of 8192. The activation function is set to \texttt{ReLU}, with \texttt{layer\_norm\_eps} configured at $10^{-5}$. The maximum position embeddings extend up to 4096, utilizing a relative key-based position embedding type. The system integrates a speech encoder with 24 layers, 16 attention heads, an intermediate size of 4096, and a \texttt{swish} activation function, operating on a sampling rate of 16 kHz and a chunk size of 20,000. Adapter layers are added with one adapter per layer, featuring a kernel size of 8, a stride of 8, and a depthwise convolution kernel size of 31. The model, based on the \texttt{SeamlessM4Tv2Model} architecture, supports a vocabulary size of 256,102 for text and 10,943 for characters. Additional features include speaker embeddings and language embeddings, both of dimension 256. To ensure stability, a dropout rate of 0.05 is applied to both encoder and decoder layers. 

\subsection{Training Setup: Qwen-Audio}
\begin{figure}[h]
    \centering
    \includegraphics[width=\linewidth]{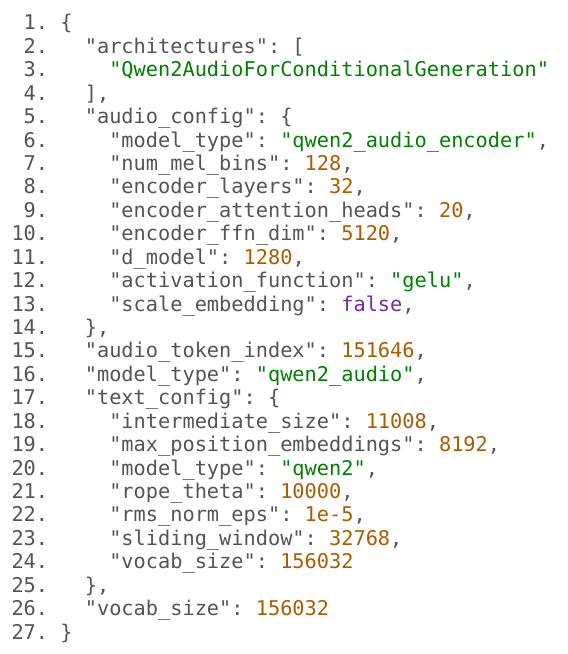}
    \caption{Fine-tuning configuration of \textbf{Qwen2-Audio-7B-Instruct} model}
    \label{fig:config_qwen2audio}
\end{figure}
The Qwen2-Audio series represents the latest advancements in the Qwen large audio-language model framework. This model is designed to process diverse audio signal inputs, enabling comprehensive audio analysis and generating direct textual responses based on spoken instructions. Qwen2-Audio supports over eight languages and dialects, including but not limited to Chinese, English, Cantonese, French, Italian, Spanish, German, and Japanese.

The Qwen language model and an audio encoder serve as the foundational models. Multi-task pre-training is subsequently applied to achieve audio-language alignment, followed by supervised fine-tuning and direct preference optimization (DPO). These steps are designed to enhance the model's performance on downstream tasks and align with human preferences.

\textbf{Training setup}: As shown in Figure \ref{fig:config_qwen2audio}, we fine-tuned Qwen2-Audio-7B-Instruct in an end-to-end \gls{ST} manner. The training setup leverages the \texttt{Qwen2AudioForConditionalGeneration} architecture, which combines an advanced audio encoder and a text-based generative model. The audio encoder configuration (\texttt{qwen2\_audio\_encoder}) includes 128 mel bins for the \gls{MFCC}, 32 encoder layers, 20 attention heads per layer, a \gls{FFW} dimension of 5120, and a model dimension (\texttt{d\_model}) of 1280. The encoder employs the \texttt{GELU} activation function and does not scale embeddings. For token-level alignment, the audio token index is set to 151646. The text model (\texttt{qwen2}) features an intermediate size of 11008, supports a maximum of 8192 positional embeddings, and uses rotary position encoding with a scaling factor (\texttt{rope\_theta}) of 10000. Additionally, the sliding window size is 32768 to handle long-context text processing. Both the audio and text models share a vocabulary size of 156032.

\onecolumn
\subsection{In-context Learning Prompt}
We present our prompt templates used in in-context learning experiments.  

\begin{figure*}[h]
    \centering
    \includegraphics[width=\linewidth]{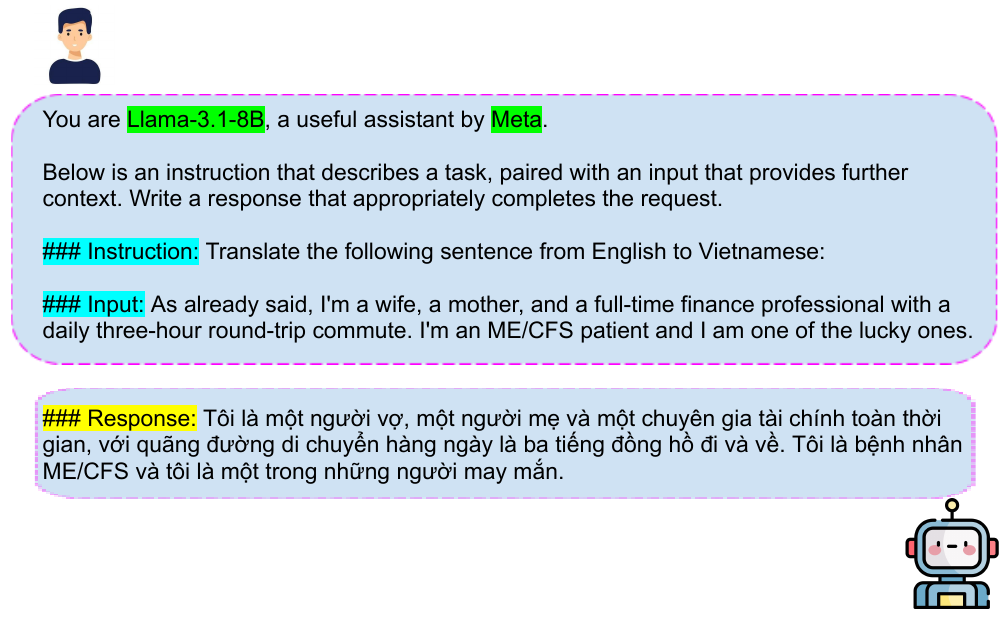}
    \caption{Prompt template for \textbf{\gls{SFT} on the entire dataset} using \textbf{Llama-3.1-8B} model. This prompt is used for cascaded \gls{ST} system.}
    \label{fig:prompt_llama_finetune}
\end{figure*}

\begin{figure*}[h]
    \centering
    \includegraphics[width=\linewidth]{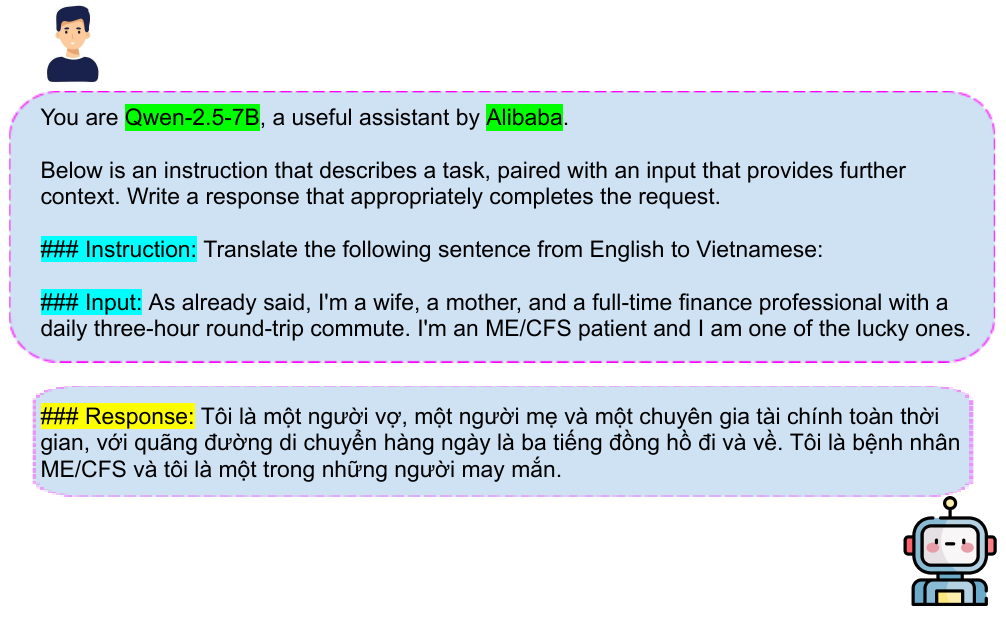}
    \caption{Prompt template for \textbf{\gls{SFT} on the entire dataset} using \textbf{Qwen-2.5-7B} model. This prompt is used for cascaded \gls{ST} system.}
    \label{fig:prompt_qwen_finetune}
\end{figure*}

\begin{figure*}[h]
    \centering
    \includegraphics[width=\linewidth]{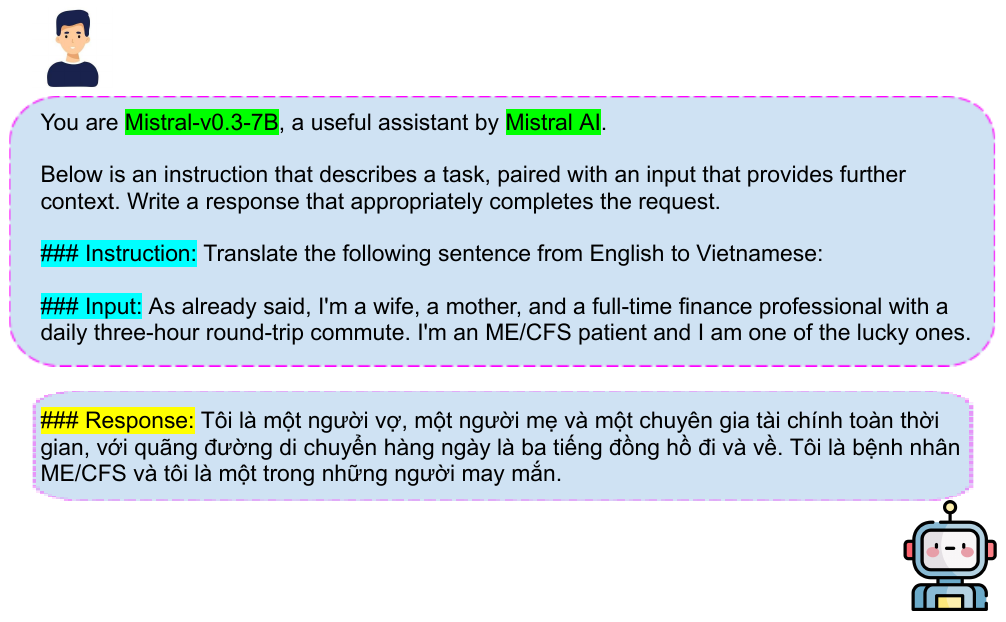}
    \caption{Prompt template for \textbf{\gls{SFT} on the entire dataset} using \textbf{Mistral-v0.3-7B} model. This prompt is used for cascaded \gls{ST} system.}
    \label{fig:prompt_mistral_finetune}
\end{figure*}

Prompt template for \textbf{\textit{\gls{SFT}}} on the entire dataset is shown in Figure \ref{fig:prompt_llama_finetune} for Llama-3.1-8B, Figure \ref{fig:prompt_qwen_finetune} for Qwen-2.5-7B, and Figure \ref{fig:prompt_mistral_finetune} for Mistral-v0.3-7B.  

\begin{figure*}[h]
    \centering
    \includegraphics[width=\linewidth]{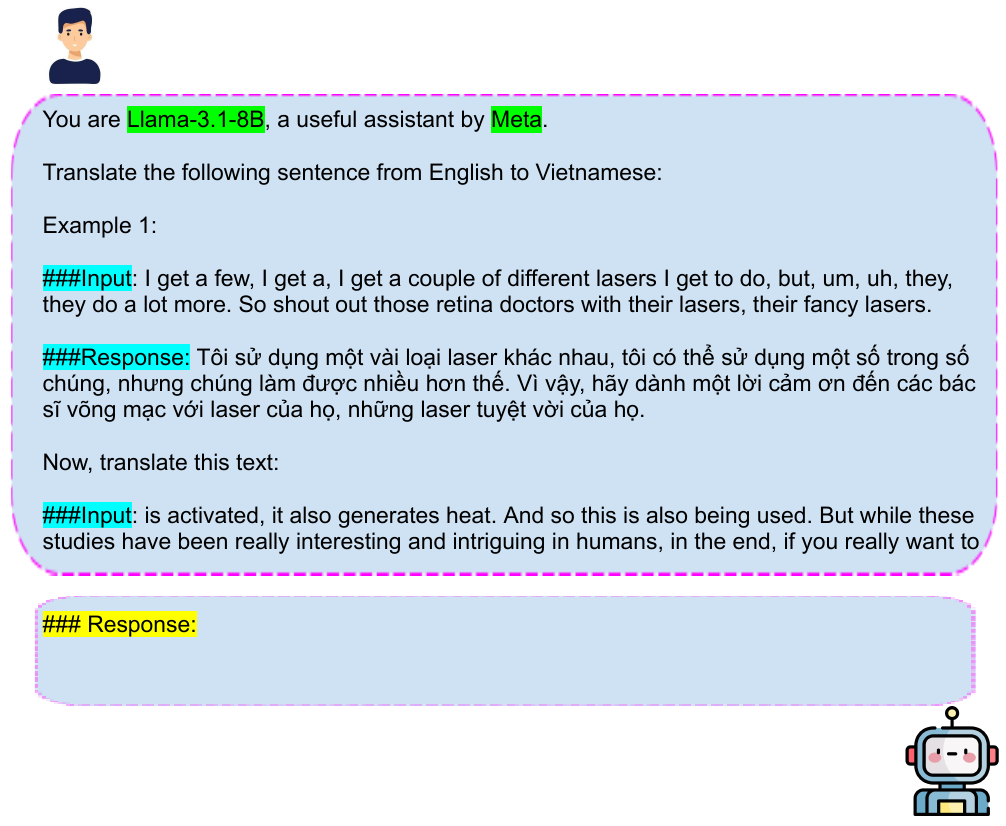}
    \caption{Prompt template for \textbf{few-shot learning} using \textbf{Llama-3.1-8B} model. This prompt is used for cascaded \gls{ST} system.}
    \label{fig:prompt_llama_fewshot}
\end{figure*}

\begin{figure*}[h]
    \centering
    \includegraphics[width=\linewidth]{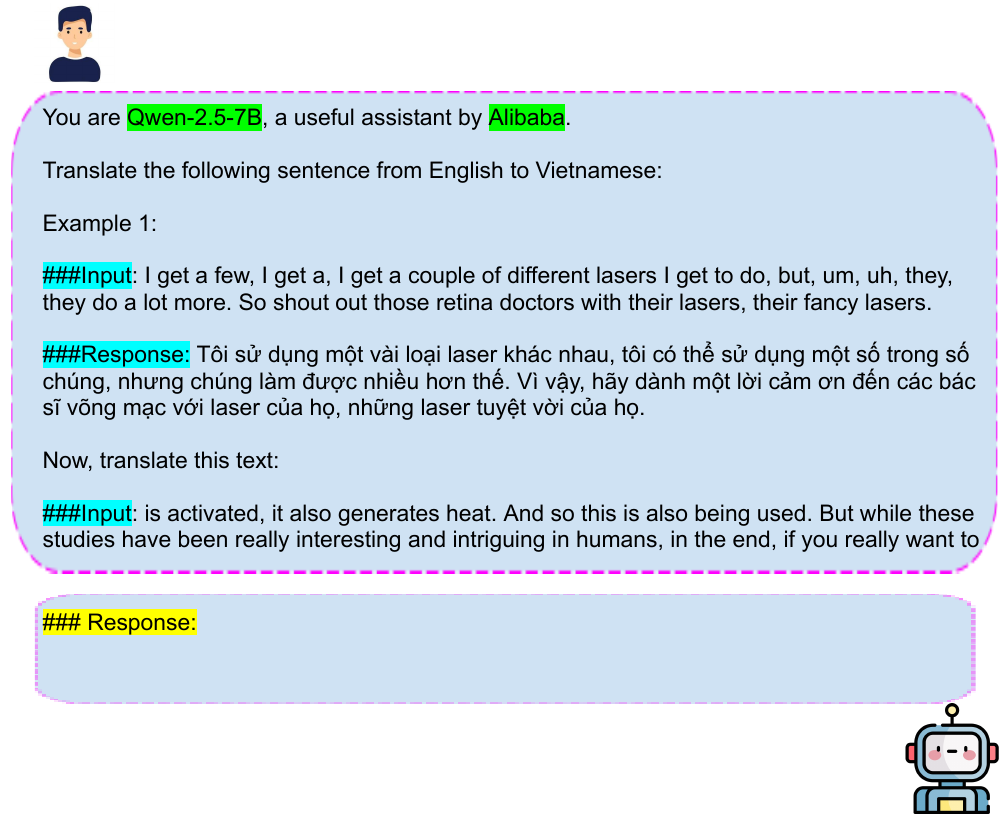}
    \caption{Prompt template for \textbf{few-shot learning} using \textbf{Qwen-2.5-7B} model. This prompt is used for cascaded \gls{ST} system.}
    \label{fig:prompt_qwen_fewshot}
\end{figure*}

\begin{figure*}[h]
    \centering
    \includegraphics[width=\linewidth]{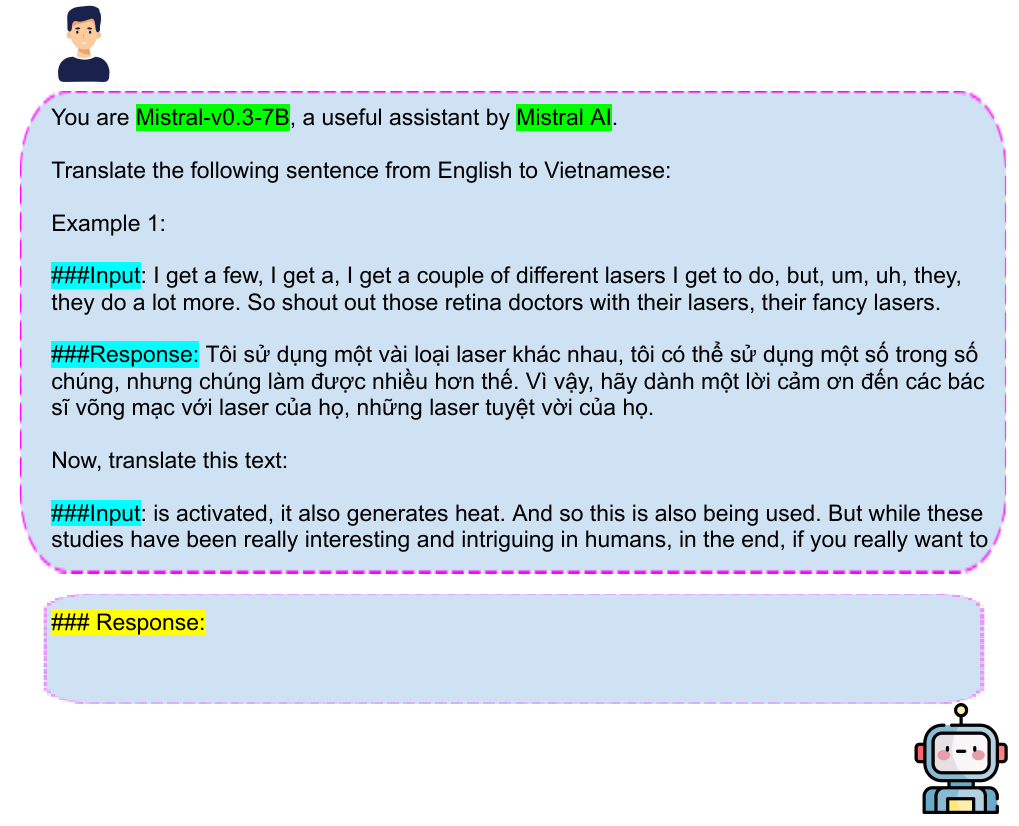}
    \caption{Prompt template for \textbf{few-shot learning} using \textbf{Mistral-v0.3-7B} model. This prompt is used for cascaded \gls{ST} system.}
    \label{fig:prompt_mistral_fewshot}
\end{figure*}

Prompt template for \textbf{\textit{few-shot learning}} is shown in Figure \ref{fig:prompt_llama_fewshot} for Llama-3.1-8B, Figure \ref{fig:prompt_qwen_fewshot} for Qwen-2.5-7B, and Figure \ref{fig:prompt_mistral_fewshot} for Mistral-v0.3-7B. 

\begin{figure*}[h]
    \centering
    \includegraphics[width=\linewidth]{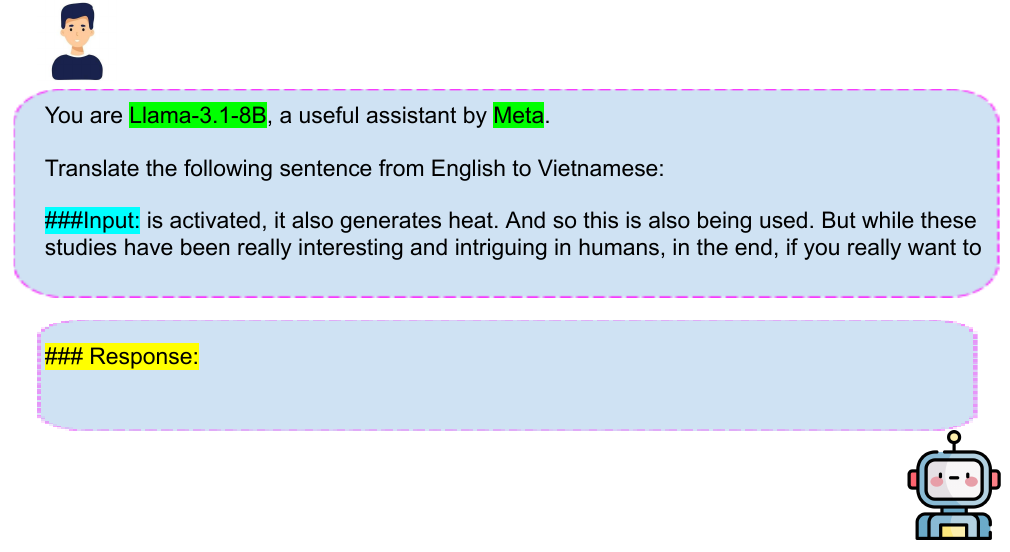}
    \caption{Prompt template for \textbf{zero-shot learning} using \textbf{Llama-3.1-8B} model. This prompt is used for cascaded \gls{ST} system.}
    \label{fig:prompt_llama_zeroshot}
\end{figure*}

\begin{figure*}[h]
    \centering
    \includegraphics[width=\linewidth]{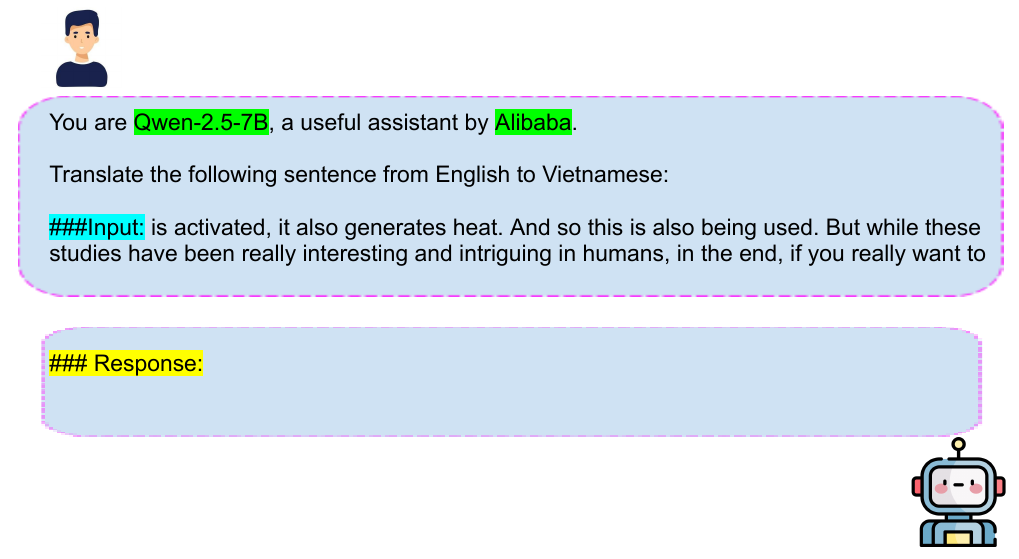}
    \caption{Prompt template for \textbf{zero-shot learning} using \textbf{Qwen-2.5-7B} model. This prompt is used for cascaded \gls{ST} system.}
    \label{fig:prompt_qwen_zeroshot}
\end{figure*}

\begin{figure*}[h]
    \centering
    \includegraphics[width=\linewidth]{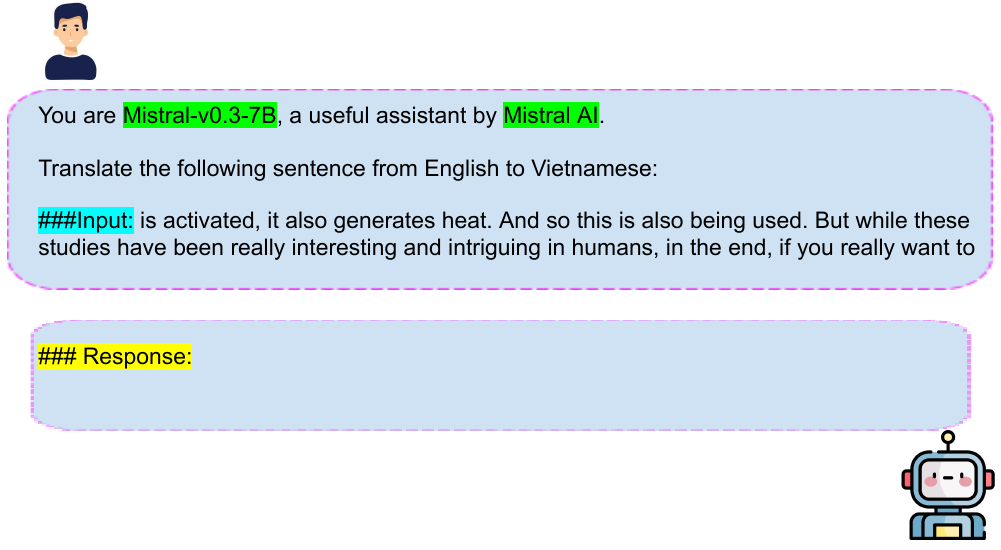}
    \caption{Prompt template for \textbf{zero-shot learning} using \textbf{Mistral-v0.3-7B} model. This prompt is used for cascaded \gls{ST} system.}
    \label{fig:prompt_mistral_zeroshot}
\end{figure*}

Prompt template for \textbf{\textit{zero-shot learning}} is shown in Figure \ref{fig:prompt_llama_zeroshot} for Llama-3.1-8B, Figure \ref{fig:prompt_qwen_zeroshot} for Qwen-2.5-7B, and Figure \ref{fig:prompt_mistral_zeroshot} for Mistral-v0.3-7B. 

\twocolumn

\twocolumn

\twocolumn

\twocolumn

\twocolumn

\twocolumn

\twocolumn

\twocolumn

\twocolumn
\section{Details of Evaluation Metrics}
\subsection{Discussion about Automatic Evaluation Metrics}
\label{sec.discussion_eval_metrics}
In this section, we discuss the advantages and disadvantages of two types of automatic evaluation metrics in \gls{MT}: n-gram overlap metrics (e.g. BLEU, METEOR, etc.) and embeddings-based metrics (e.g. BERTScore)

\textbf{N-gram overlap metrics}:
\begin{itemize}
\item Advantages:
\begin{itemize}
    \item Simplicity and widespread use: N-gram overlap metrics are widely used in \gls{MT}, especially BLEU, making them a standard for benchmarking models and enabling easy comparison across studies.  
    \item Efficient computation: N-gram overlap metrics are computationally efficient and works well for quick assessments of translation quality.  
    \item Word n-gram matching: By focusing on n-gram overlaps, these metrics capture the degree of lexical similarity between the hypothesis and reference translations.
\end{itemize}
\item Disadvantages:
\begin{itemize}
    \item Insensitive to semantics: N-gram overlap metrics rely solely on surface-level word matches, failing to account for semantic similarity or paraphrasing.  
    \item Context ignorance: N-gram overlap metrics do not account for context, which is crucial in capturing the nuances of \gls{MT}.  
    \item Reliance on references: The quality of n-gram overlap metrics heavily depends on the availability of high-quality reference translations, limiting its reliability in low-resource scenarios.  
    \item Bias towards short phrases: N-gram overlap metrics may over-penalize longer, valid translations due to brevity penalties or under-represented n-grams.
\end{itemize}
\end{itemize}

\textbf{Embeddings-based metrics}:
\begin{itemize}
\item Advantages:
\begin{itemize}
    \item Semantic sensitivity: By leveraging contextual embeddings from models like BERT \cite{devlin-etal-2019-bert} for BERTScore, embeddings-based metrics capture semantic similarity and accounts for paraphrasing better than n-gram overlap metrics.  
    \item Robust to variations: Embeddings-based metrics are more robust to word order and phrasing differences, making it suitable for languages with flexible syntactic structures. 
    \item Handles low-resource scenarios: Embeddings-based metrics perform well even with a limited number of reference translations by emphasizing meaning over exact matches.   
\end{itemize}
\item Disadvantages:
\begin{itemize}
    \item Higher computational cost: Calculating embeddings-based metrics requires the use of pre-trained transformer models, making it more resource-intensive.  
    \item Dependency on pre-trained models: The quality of embeddings-based metrics depends on the pre-trained embeddings, which might not always align well with the target language or domain.  
    \item Less established: Embeddings-based metrics are relatively newer and less standardized, which may hinder direct comparisons across different studies.  
    \item Overemphasis on semantic similarity: While beneficial, embeddings-based metrics may overlook syntactic errors or stylistic mismatches that are critical in \gls{ST}.  
\end{itemize}
\end{itemize}

Both n-gram overlap metrics and embeddings-based metrics have their merits and limitations, and their effectiveness often depends on the specific requirements of the \gls{ST} task. Combining them or using them in tandem with human evaluation can provide a more comprehensive assessment.

\textbf{BLEU (Bilingual Evaluation Understudy)}: The BLEU score measures the similarity between a candidate translation  $\texttt{Can}$ and one or more reference translations $\texttt{Ref}$ by calculating the precision of n-gram matches, penalized for brevity.

Modified n-gram precision: The modified precision $\texttt{Prec}_{n}$ is calculated for n-grams of size $n$ as below
{\small
\begin{equation}
\text{Prec}_{n} = 
\frac{\sum_{g \in \text{Can}} \min(\text{count}(g, \text{Can}), \text{count}(g, \text{Ref}))}
     {\sum_{g \in \text{Can}} \text{count}(g, \text{Can})}
\end{equation}
}
where: \begin{itemize}
    \item[-] \( g \) represents an n-gram
    \item[-] \( \text{count}(g, \texttt{Can}) \) is the frequency of \( g \) in the candidate
    \item[-] \( \text{count}(g, \texttt{Ref}) \) is the frequency of \( g \) in the reference
\end{itemize}

Brevity Penalty (BP): A brevity penalty accounts for the candidate translation being shorter than the reference:  
\begin{equation}
BP =
   \begin{cases}
   1, & \text{if } \text{len}(\texttt{Can}) > \text{len}(\texttt{Ref}), \\
   e^{1 - \frac{\text{len}(\texttt{Ref})}{\text{len}(\texttt{Can})}}, & \text{otherwise}.
   \end{cases}    
\end{equation}
  
BLEU Score: The BLEU score is computed as the geometric mean of the n-gram precisions, weighted by a constant \( w_n \):  
\begin{equation}
\text{BLEU} = BP \cdot \exp \left( \sum_{n=1}^N w_n \log \texttt{Prec}_{n} \right)    
\end{equation}
Typically, \( N = 4 \) (up to 4-grams), and \( w_n = \frac{1}{N} \).

\textbf{BERTScore}: BERTScore evaluates the semantic similarity between the candidate \( \texttt{Can} \) and reference \( \texttt{Ref} \) by computing cosine similarities of their token embeddings.

Token Embeddings: Let \( \textbf{E}(\texttt{Can}) = \{\textbf{\texttt{can}}_i\} \) and \( \textbf{E}(\texttt{Ref}) = \{\textbf{\texttt{ref}}_j\} \) be the token embeddings of \( \texttt{Can} \) and \( \texttt{Ref} \), obtained from a pre-trained model like BERT.

Cosine Similarity Matrix: Compute the cosine similarity between all pairs of token embeddings as below
\begin{equation}
   \texttt{Sim}_{ij} = \frac{\textbf{\texttt{can}}_{i} \cdot \textbf{\texttt{ref}}_{j}}{
   \lVert \textbf{\texttt{can}}_{i} \rVert \lVert \textbf{\texttt{ref}}_{j} \rVert}    
\end{equation}

Precision, Recall, and F1 Score:  

- Precision:  
\begin{equation}
\text{Prec} = \frac{1}{|\texttt{Can}|} \sum_{i=1}^{|\texttt{Can}|} \max_{j}  \texttt{Sim}_{ij}    
\end{equation}
     
- Recall:  
\begin{equation}
\text{Rec} = \frac{1}{|\texttt{Ref}|} \sum_{j=1}^{|\texttt{Ref}|} \max_{i}  \texttt{Sim}_{ij}    
\end{equation}
     
- F1 Score (BERTScore): 
\begin{equation}
\text{BERTScore} = 2 \cdot \frac{\text{Prec} \cdot \text{Rec}}{\text{Prec} + \text{Rec}}    
\end{equation}

In practice, BERTScore can be averaged across a dataset to produce a final evaluation score.

\textbf{\gls{WER}}: Both \gls{WER} and \gls{CER} are common metrics for evaluating the quality of \gls{ASR} for \gls{ST}. They compare the output sequence (hypothesis) with a reference sequence and compute the number of errors in terms of word or character differences. \gls{WER} measures the ratio of the total number of word-level errors (insertions, deletions, and substitutions) to the total number of words in the reference.

Definition:
\begin{equation}
\text{WER} = \frac{\texttt{Sub} + \texttt{Del} + \texttt{Ins}}{\texttt{Num}}
\end{equation}
where:
   
   - \( \texttt{Sub} \): Number of word substitutions
   
   - \( \texttt{Del} \): Number of word deletions
   
   - \( \texttt{Ins} \): Number of word insertions
   
   - \( \texttt{Num} \): Total number of words in the reference

Steps:
\begin{enumerate}
    \item Align the hypothesis and reference sequences using dynamic programming (e.g., Levenshtein distance)
    \item Count the number of substitutions, deletions, and insertions
\end{enumerate}

\textbf{\gls{CER}}: \gls{CER} operates similarly to \gls{WER} but at the character level, making it suitable for scripts where words are not clearly delineated, such as Chinese or languages with agglutination.

Definition:
\begin{equation}
\text{CER} = \frac{\texttt{SUB} + \texttt{DEL} + \texttt{INS}}{\texttt{NUM}}
\end{equation}
   Where:
   
   - \( \texttt{SUB} \): Number of character substitutions.
   
   - \( \texttt{DEL} \): Number of character deletions.
   
   - \( \texttt{INS} \): Number of character insertions.
   
   - \( \texttt{NUM} \): Total number of characters in the reference.

Steps:
\begin{enumerate}
    \item Align the hypothesis and reference sequences character by character
    \item Count substitutions, deletions, and insertions
\end{enumerate}
These formulations highlight the alignment-based approach to calculating \gls{WER} and \gls{CER}, which can be implemented using algorithms like dynamic programming to find the minimal edit distance between the hypothesis and the reference.

\subsection{Details of Human Evaluation}
\label{sec:details_human_eval}
Various approaches exist for eliciting judgments from informants regarding the quality of machine-translated sentences. Human evaluators may be tasked with directly assessing \gls{MT} outputs by assigning scores to specific indicators on a predefined scale (0 to 10) for the same source sentence. These evaluations are typically based on three key criteria: adequacy, fluency, and comprehensibility. 

\begin{itemize}
    \item \textbf{Adequacy}: Measures how well the meaning of the source text is conveyed in the translation
    \item \textbf{Fluency}: Evaluates the grammatical and stylistic quality of the translated text, irrespective of the source text
    \item \textbf{Comprehensibility}: Assesses how easily a human reader can understand the translated text without referring to the source.
\end{itemize}
Direct assessments of \gls{MT} ranking serve as the standard evaluation methods in recent biomedical \gls{MT} shared task campaigns conducted by WMT \cite{bojar2016findings, bojar2017findings}, as well as in \gls{MT} research from the 1990s led by the Advanced Research Projects Agency (ARPA) \cite{church1993good, white1994arpa}.

\subsection{Details of LLM-as-a-judge}
\label{sec:details_llm_as_judge}
The concept of using \glspl{LLM} as a "judge" in \gls{MT} has emerged as a promising method for evaluating translation quality. Unlike traditional evaluation metrics such as BLEU, ROUGE, or METEOR, which rely on n-gram overlap between machine-generated translations and reference texts, \gls{LLM}-based evaluators leverage their advanced understanding of language semantics, context, and grammar. This approach allows for a more nuanced assessment of translation fidelity, fluency, and adequacy.

\textbf{Advantages}:
\begin{itemize}
    \item Contextual understanding: \glspl{LLM} excel at evaluating translations by considering the broader context and subtle nuances in language use, which traditional metrics often overlook \cite{karpinska2023large, fernandes2023context_chat_translation}.
    \item Reference-free evaluation: \glspl{LLM} can perform evaluations without requiring reference translations, which can reduce biases introduced by specific linguistic choices in reference texts \cite{chen2024humans, stureborg2024large}.
    \item Scalability and automation: \glspl{LLM} enable scalable, automated evaluation pipelines, reducing the dependency on human annotators for large-scale \gls{MT} tasks \cite{yang2023human, he2024exploring}.
\end{itemize}

\textbf{Disadvantages}:
\begin{itemize}
    \item Biases in \glspl{LLM}: \glspl{LLM} may inherit biases from their training data, potentially influencing their judgment of translation quality \cite{behnke2022bias, huang2023improving}.
    \item Generalization: Ensuring that \glspl{LLM} perform reliably across languages, domains, and translation styles remains a significant challenge \cite{yan2024gpt, singh2024translating}.
\end{itemize}

\begin{figure}[h]
    \centering
    \includegraphics[width=\linewidth]{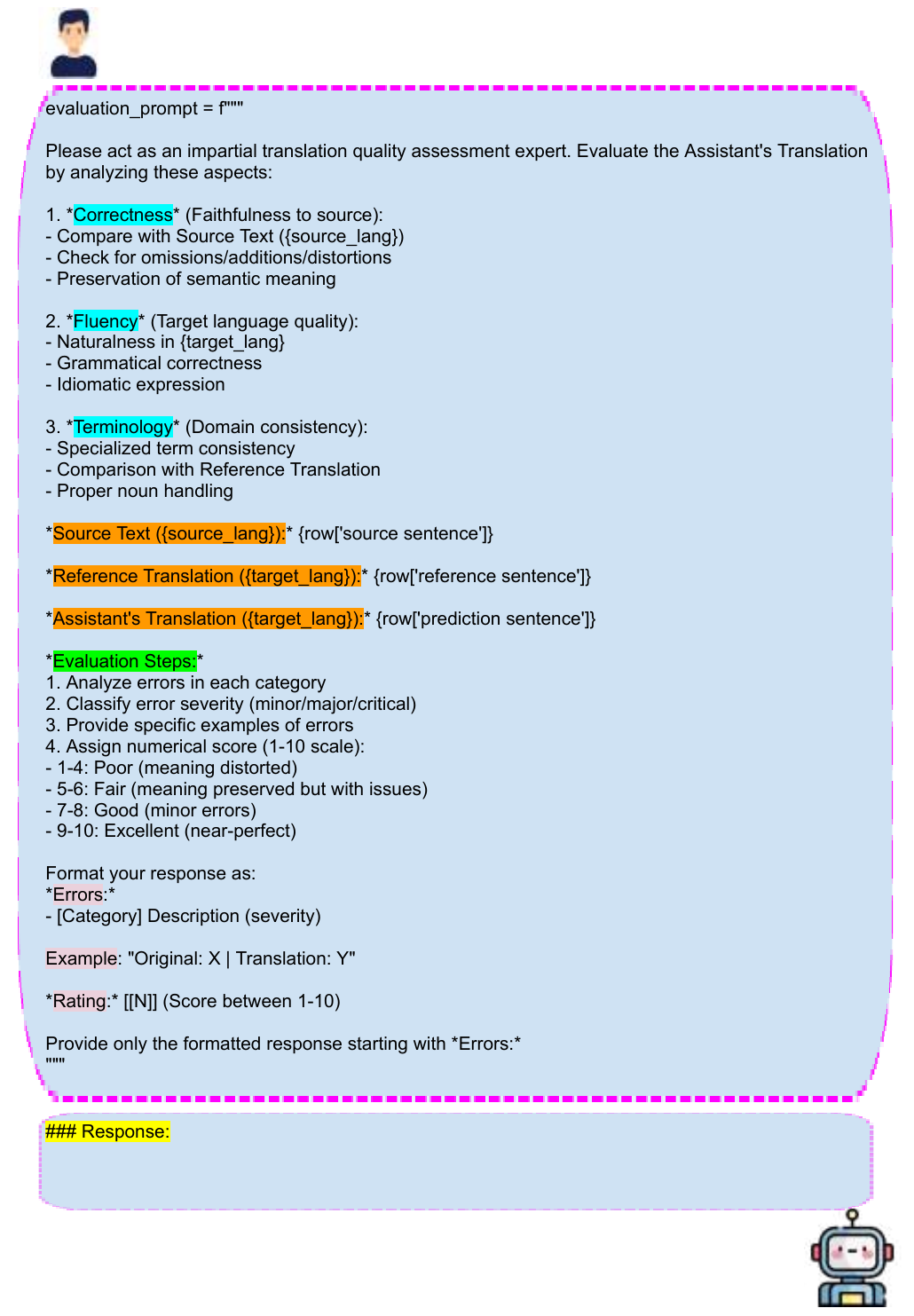}
    \caption{\gls{LLM}-as-a-judge prompt template we used for \gls{ST} transcript evaluation.}
    \label{fig:llm_as_a_judge_prompt}
\end{figure}

Figure \ref{fig:llm_as_a_judge_prompt} shows the \gls{LLM}-as-a-judge prompt template we used for \gls{ST} transcript evaluation.

\onecolumn

\onecolumn
\section{Extra Experimental Results}

\subsection{In-context Learning Results}
\label{sec:incontext_learning_results}
This section presents in-context learning results, comparing full fine-tuning, zero-shot, few-shot on both ground-truth transcript and \gls{ASR} transcript in the cascaded setting.

Full results are shown in Table \ref{tab:appx_nmt_fewshot-En-X} (English to X), Table \ref{tab:appx_nmt_fewshot-Vi-X} (Vietnamese to X), Table \ref{tab:appx_nmt_fewshot-Fr-X} (French to X), Table \ref{tab:appx_nmt_fewshot-De-X} (German to X), and Table \ref{tab:appx_nmt_fewshot-Zh-X} (Chinese to X) below.

For \glspl{LLM}, we conducted few-shot learning experiments to assess their strength in \gls{MT} tasks. On ground-truth transcripts, fine-tuned models significantly outperformed \glspl{LLM} on most language pairs for both Qwen-2.5-7B and Mistral-v0.3-7B. Notably, the Llama-3.1-8B model only showed better results than few-shot versions for source languages such as English, French, and German. Furthermore, as the number of few-shot examples increased, performance improved across all three \glspl{LLM}, as shown in the tables. For \gls{ASR} transcripts, a similar trend was observed, although there was a notable exception: when the source language was English, few-shot models maintained BLEU scores comparable to those of ground-truth text, despite a slight drop in BERTScores, which were still higher than those of fine-tuned \glspl{LLM}.

\input{tables/appx_nmt_fewshot-En-X}

\onecolumn \input{tables/appx_nmt_fewshot-Vi-X}

\onecolumn \input{tables/appx_nmt_fewshot-Fr-X}

\onecolumn \input{tables/appx_nmt_fewshot-De-X}

\onecolumn \input{tables/appx_nmt_fewshot-Zh-X}

\onecolumn
\subsection{Full Results: Ground-truth Translation Baselines}
\label{sec:full_results_groundtruth_MT_baselines}
This section presents full results of ground-truth \gls{MT} baselines for all evaluation metrics, which is an extension of Table \ref{tab:translation-groundtruth} in the main paper. Full results are shown in Table 
\ref{tab:appx_nmt_gt_allMetrics-En-X} (English to X), Table \ref{tab:appx_nmt_gt_allMetrics-Vi-X} (Vietnamese to X), Table \ref{tab:appx_nmt_gt_allMetrics-Fr-X} (French to X), Table \ref{tab:appx_nmt_gt_allMetrics-De-X} (German to X), and Table \ref{tab:appx_nmt_gt_allMetrics-Zh-X} (Chinese to X) below.
\clearpage
\input{tables/appx_nmt_gt_allMetrics-En-X}

\onecolumn \input{tables/appx_nmt_gt_allMetrics-Vi-X}

\onecolumn \input{tables/appx_nmt_gt_allMetrics-Fr-X}

\onecolumn \input{tables/appx_nmt_gt_allMetrics-De-X}

\onecolumn \input{tables/appx_nmt_gt_allMetrics-Zh-X}

\onecolumn
\subsection{Extra Results: Cascaded Speech Translation Baselines}
\label{sec:extra_results_cascaded_ST_baselines}
This section presents extra results of cascaded \gls{ST} baselines for all evaluation metrics, which is a supplement of Table \ref{tab:asr-translation} in the main paper. Extra results are shown in Table \ref{tab:appx_nmt_asr_allMetrics-En-X} (English to X), Table \ref{tab:appx_nmt_asr_allMetrics-Vi-X} (Vietnamese to X), Table \ref{tab:appx_nmt_asr_allMetrics-Fr-X} (French to X), Table \ref{tab:appx_nmt_asr_allMetrics-De-X} (German to X), and Table \ref{tab:appx_nmt_asr_allMetrics-Zh-X} (Chinese to X) below.
\clearpage
\input{tables/appx_nmt_asr_allMetrics-En-X}

\onecolumn \input{tables/appx_nmt_asr_allMetrics-Vi-X}

\onecolumn \input{tables/appx_nmt_asr_allMetrics-Fr-X}

\onecolumn \input{tables/appx_nmt_asr_allMetrics-De-X}

\onecolumn \input{tables/appx_nmt_asr_allMetrics-Zh-X}

\onecolumn
\subsection{Qualitative Results}
\label{sec:qualitative_results}
\subsubsection{Vietnamese to German Speech Translation}
\begin{figure}[h]
    \centering
    \includegraphics[width=0.8\linewidth]{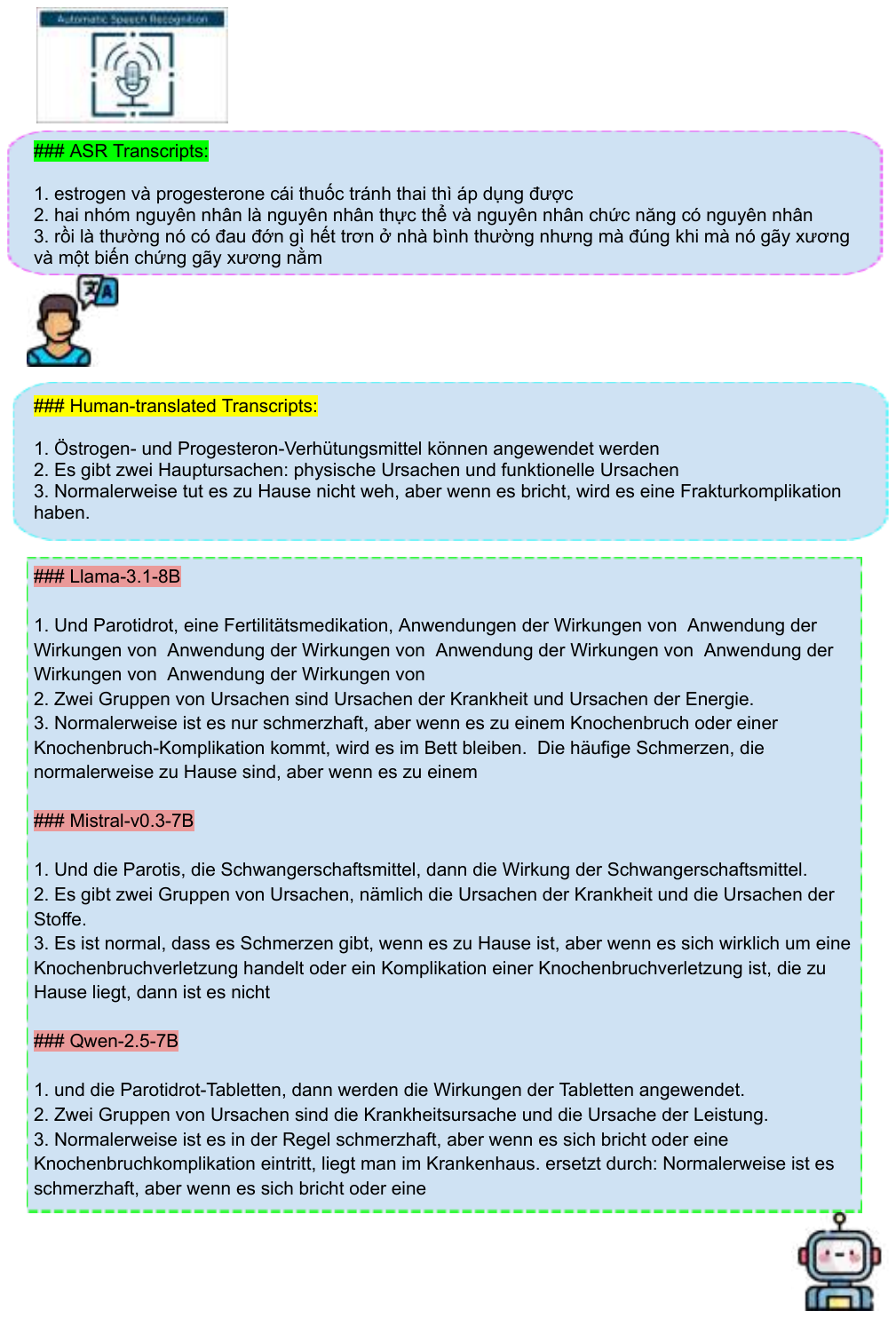}
    \caption{\textbf{Qualitative Results}. Vietnamese to German \gls{ST}}
\end{figure}

\onecolumn
\subsubsection{Vietnamese to English Speech Translation}
\begin{figure}[h]
    \centering
    \includegraphics[width=0.8\linewidth]{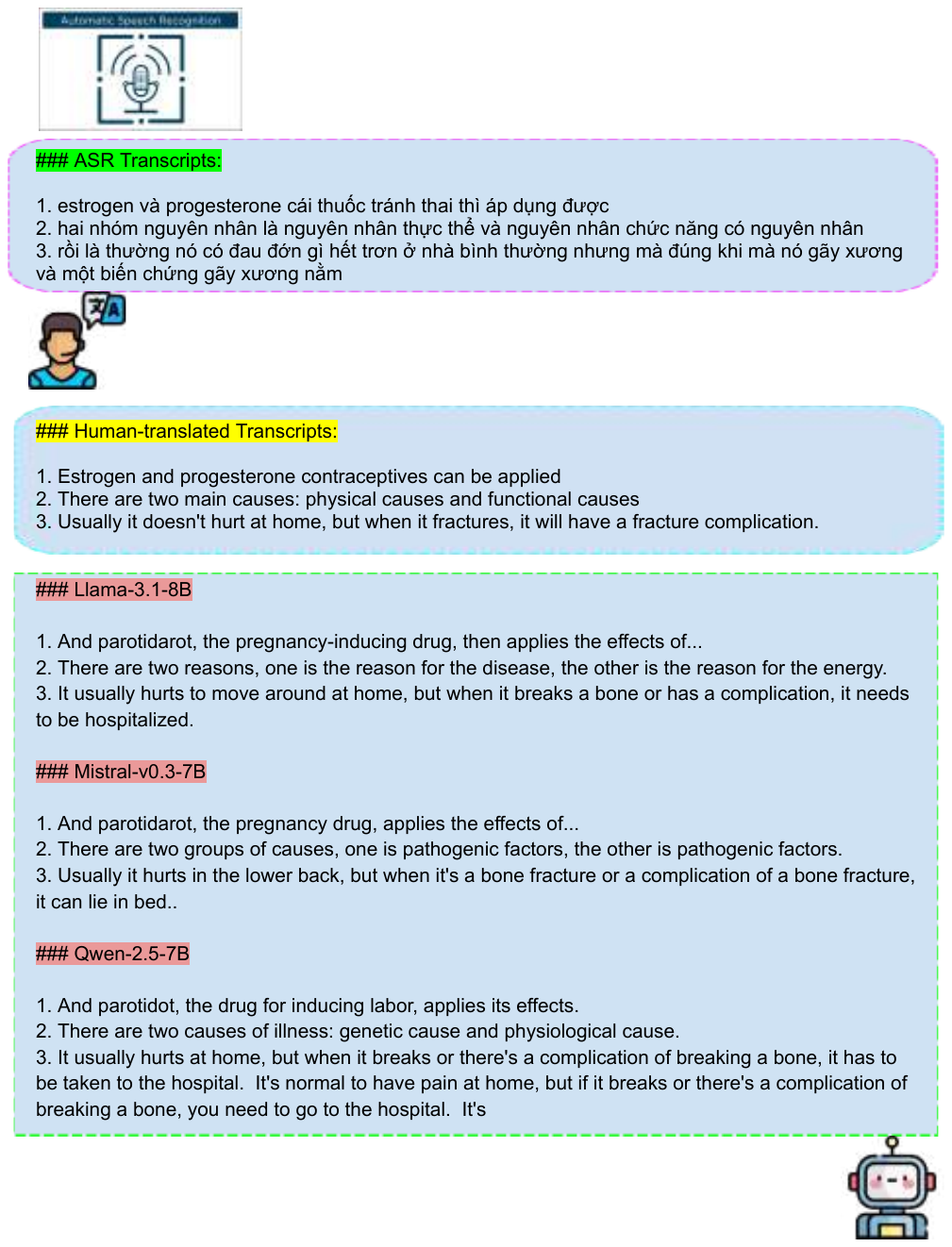}
    \caption{\textbf{Qualitative Results}. Vietnamese to English \gls{ST}}
\end{figure}

\onecolumn
\subsubsection{Vietnamese to French Speech Translation}
\begin{figure}[h]
    \centering
    \includegraphics[width=0.8\linewidth]{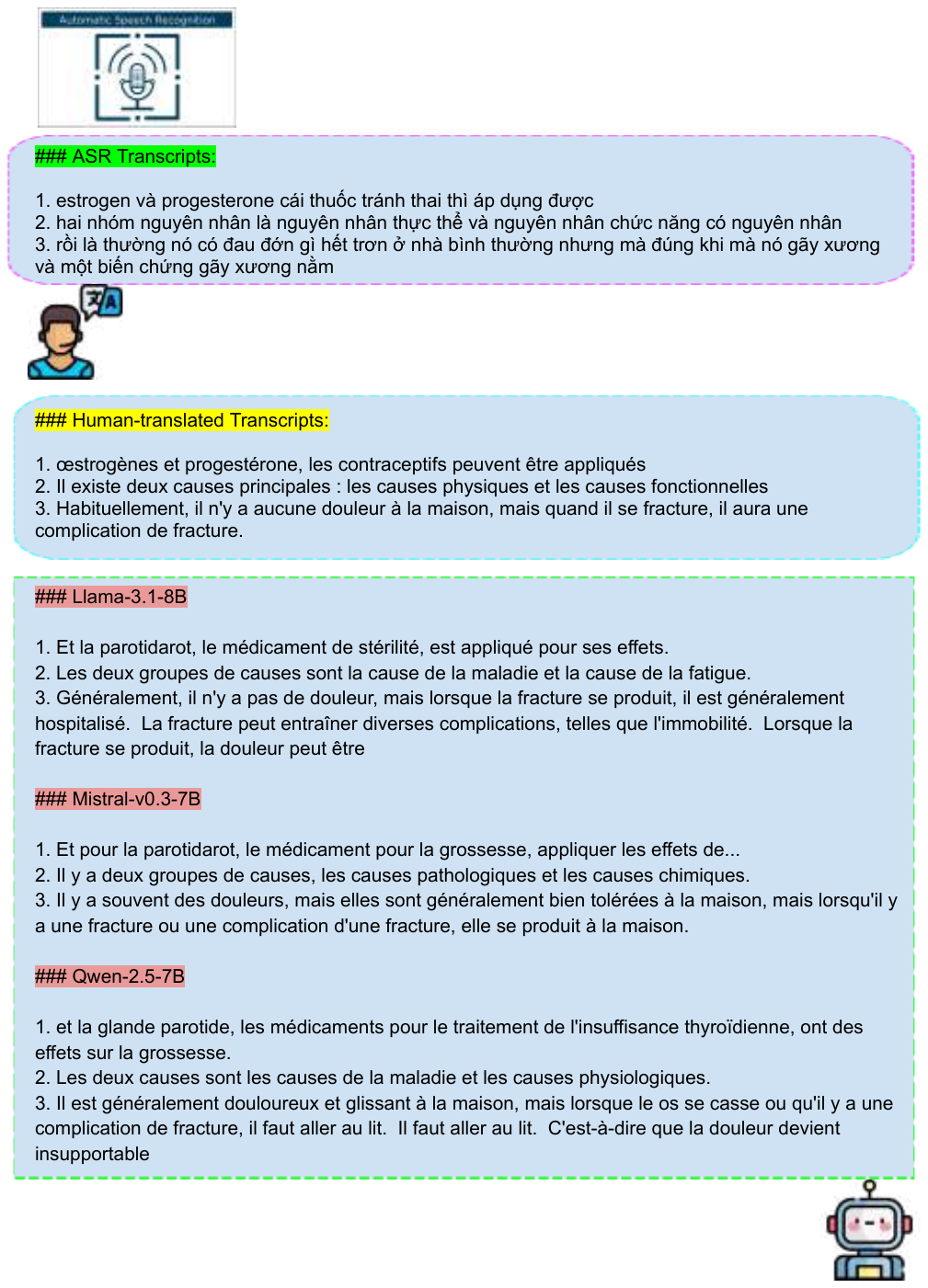}
    \caption{\textbf{Qualitative Results}. Vietnamese to French \gls{ST}}
\end{figure}

\onecolumn
\subsubsection{Vietnamese to Chinese Speech Translation}
\begin{figure}[h]
    \centering
    \includegraphics[width=0.8\linewidth]{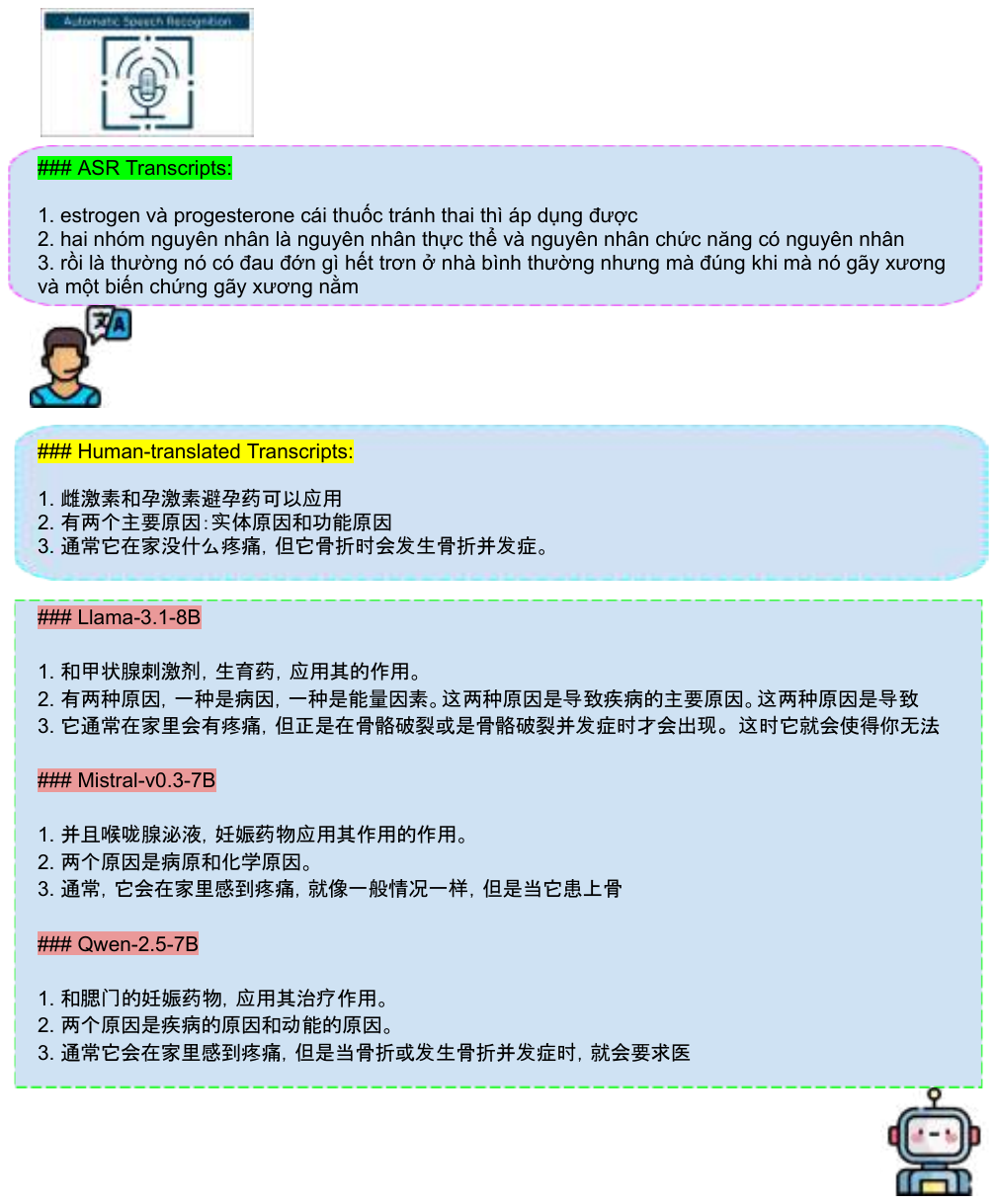}
    \caption{\textbf{Qualitative Results}. Vietnamese to Chinese \gls{ST}}
\end{figure}

\onecolumn
\subsubsection{English to Vietnamese Speech Translation}
\begin{figure}[h]
    \centering
    \includegraphics[width=0.8\linewidth]{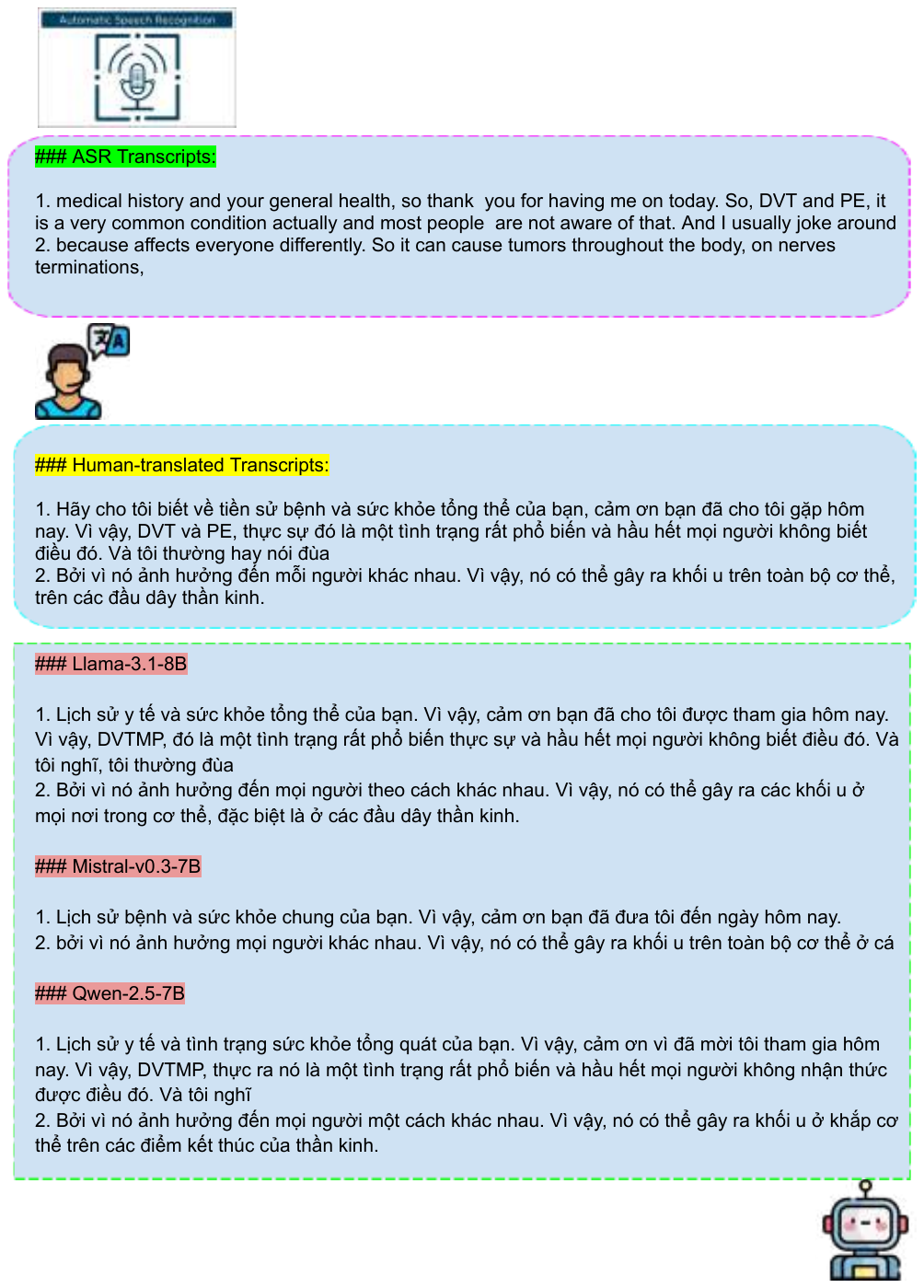}
    \caption{\textbf{Qualitative Results}. English to Vietnamese \gls{ST}}
\end{figure}

\onecolumn
\subsubsection{English to German Speech Translation}
\begin{figure}[h]
    \centering
    \includegraphics[width=0.8\linewidth]{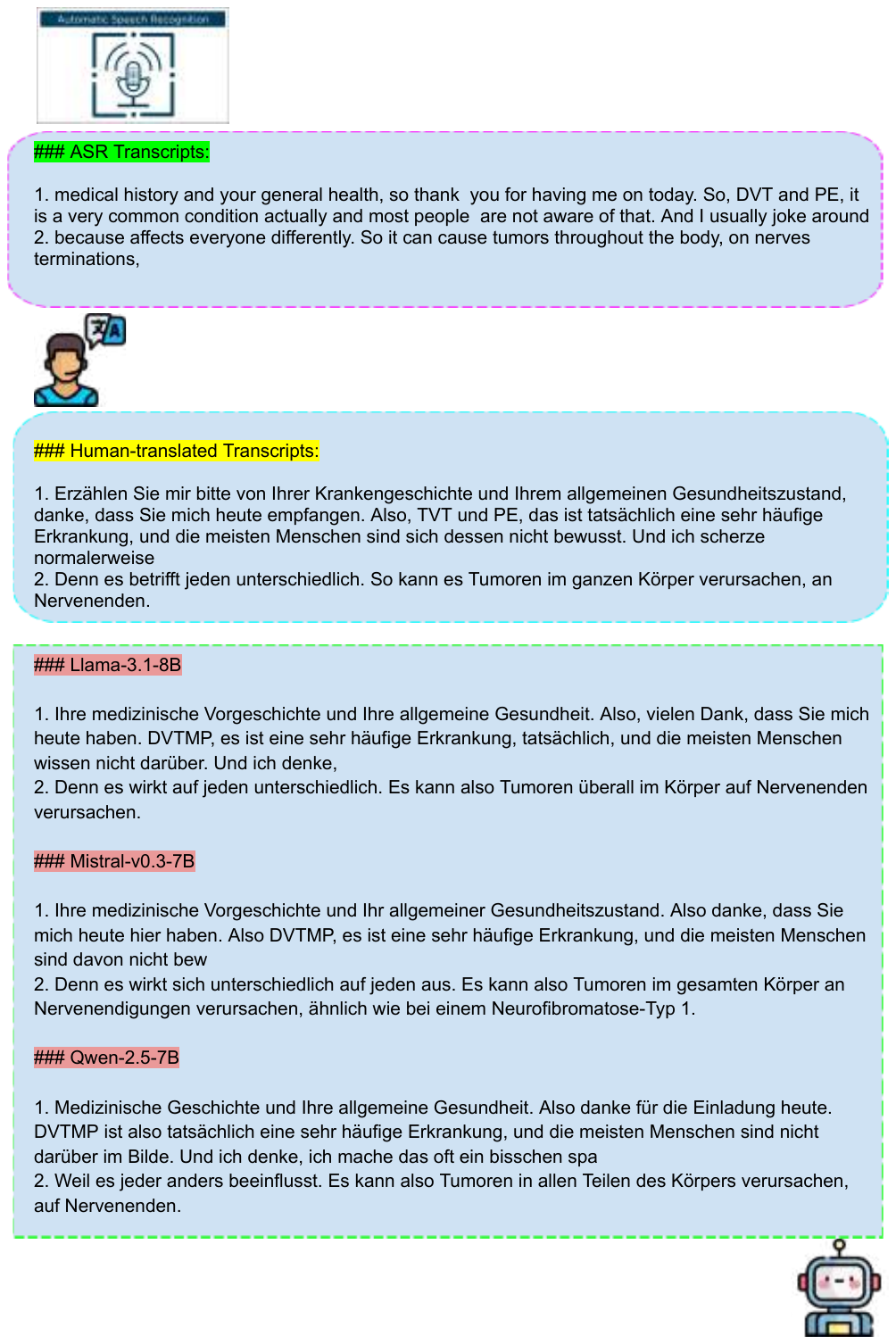}
    \caption{\textbf{Qualitative Results}. English to German \gls{ST}}
\end{figure}

\onecolumn
\subsubsection{English to French Speech Translation}
\begin{figure}[h]
    \centering
    \includegraphics[width=0.8\linewidth]{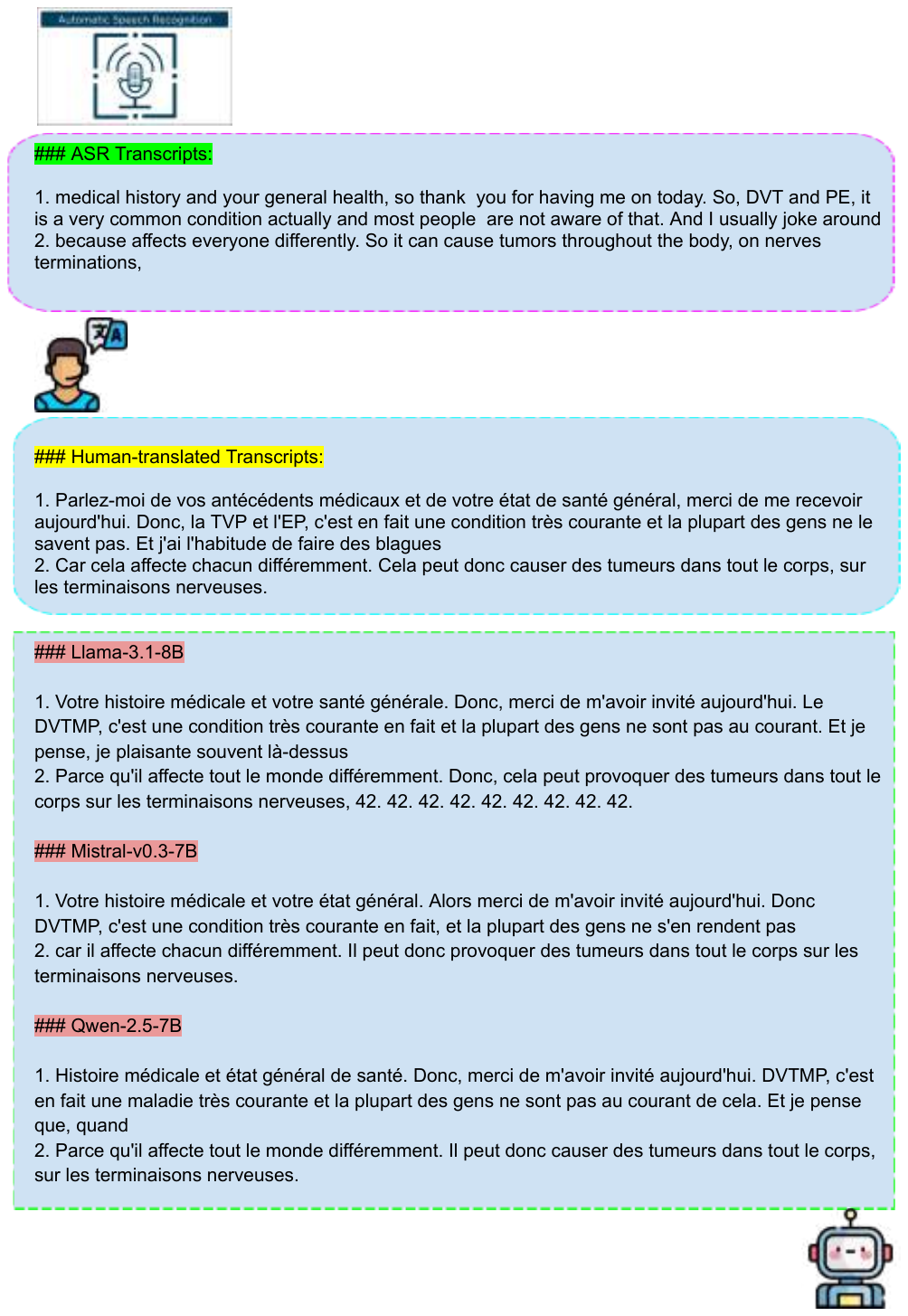}
    \caption{\textbf{Qualitative Results}. English to French \gls{ST}}
\end{figure}

\onecolumn
\subsubsection{English to Chinese Speech Translation}
\begin{figure}[h]
    \centering
    \includegraphics[width=0.8\linewidth]{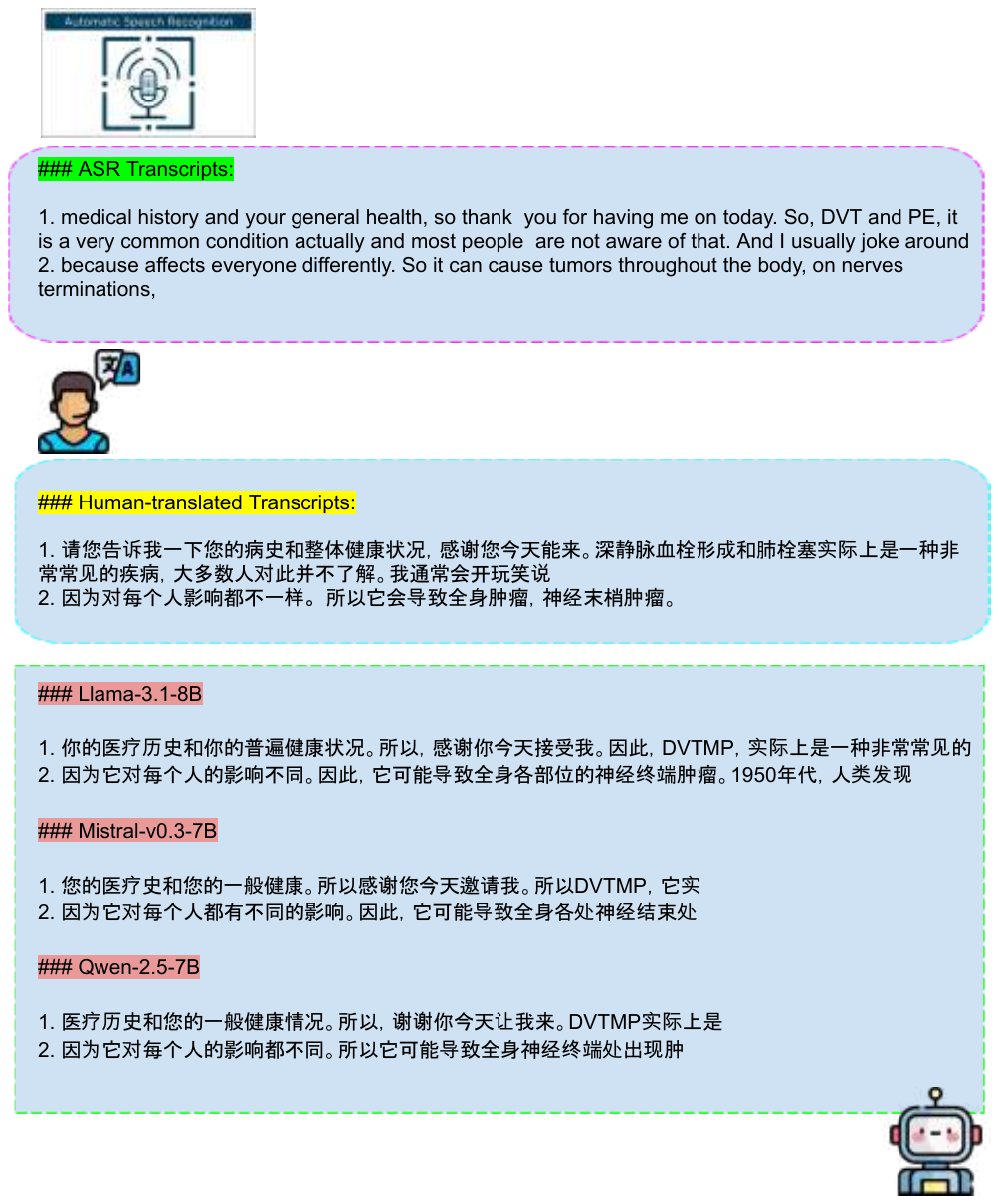}
    \caption{\textbf{Qualitative Results}. English to Chinese \gls{ST}}
\end{figure}

\onecolumn
\subsubsection{German to Vietnamese Speech Translation}
\begin{figure}[h]
    \centering
    \includegraphics[width=0.8\linewidth]{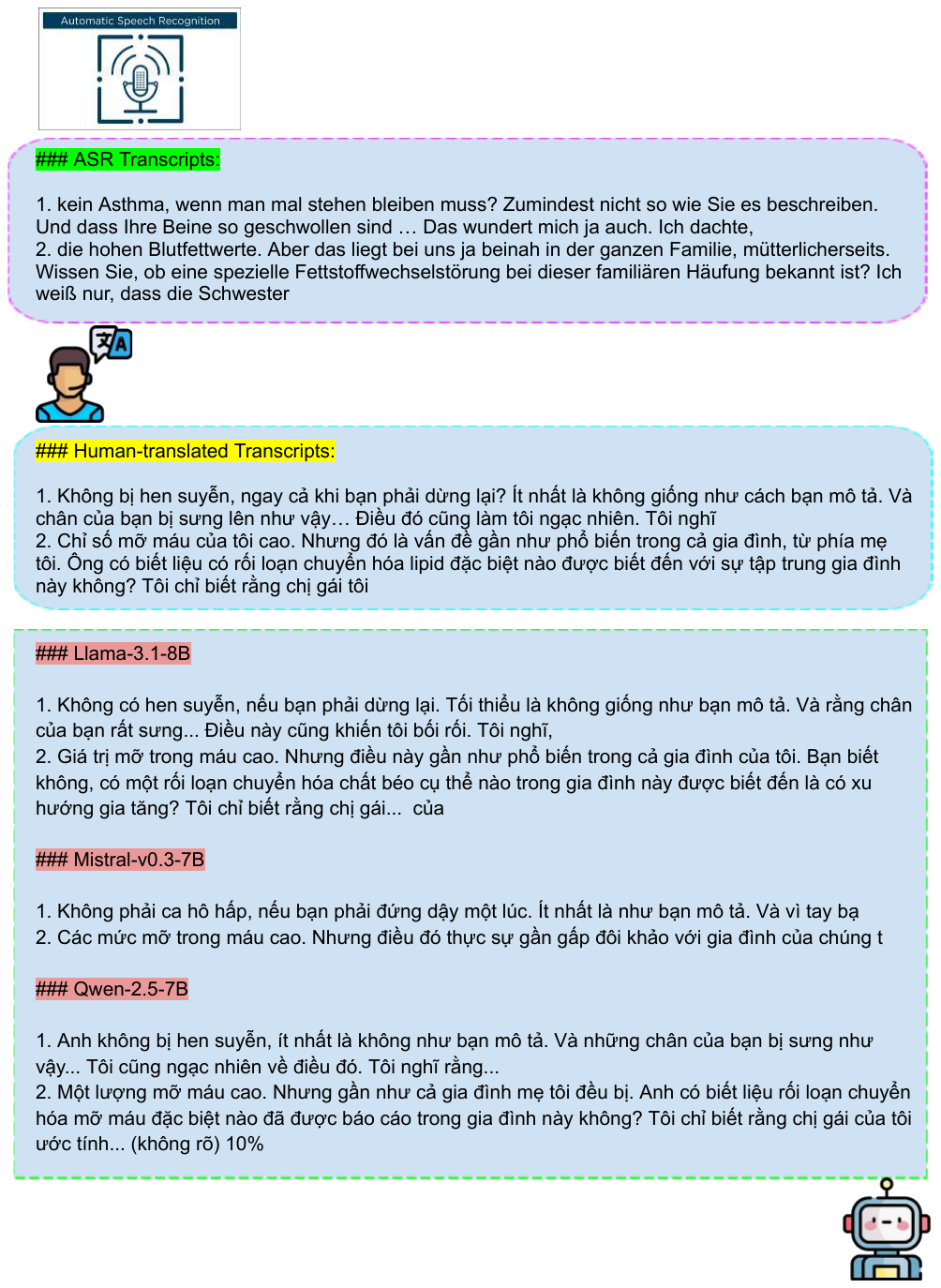}
    \caption{\textbf{Qualitative Results}. German to Vietnamese \gls{ST}}
\end{figure}

\onecolumn
\subsubsection{German to English Speech Translation}
\begin{figure}[h]
    \centering
    \includegraphics[width=0.8\linewidth]{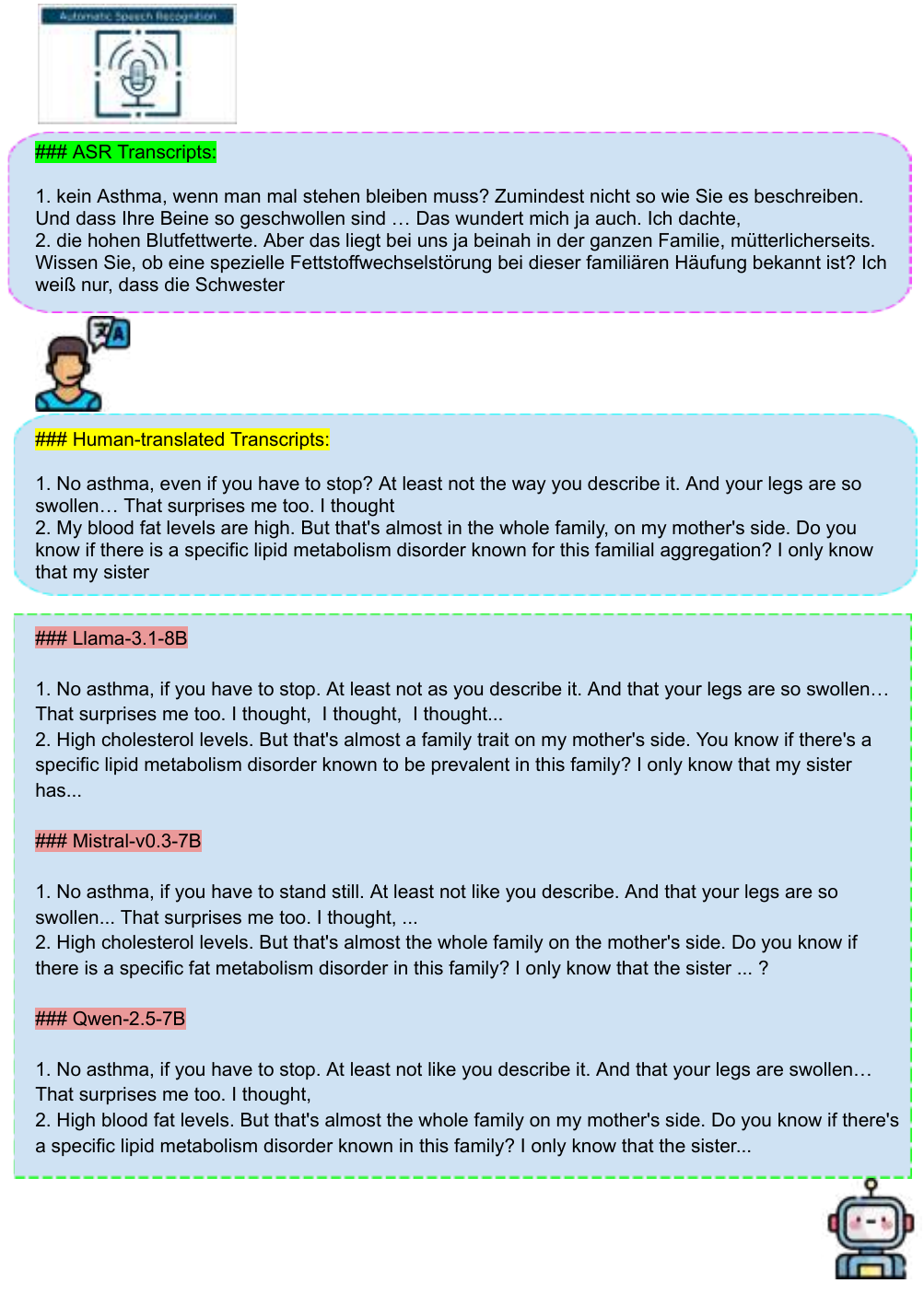}
    \caption{\textbf{Qualitative Results}. German to English \gls{ST}}
\end{figure}

\onecolumn
\subsubsection{German to French Speech Translation}
\begin{figure}[h]
    \centering
    \includegraphics[width=0.8\linewidth]{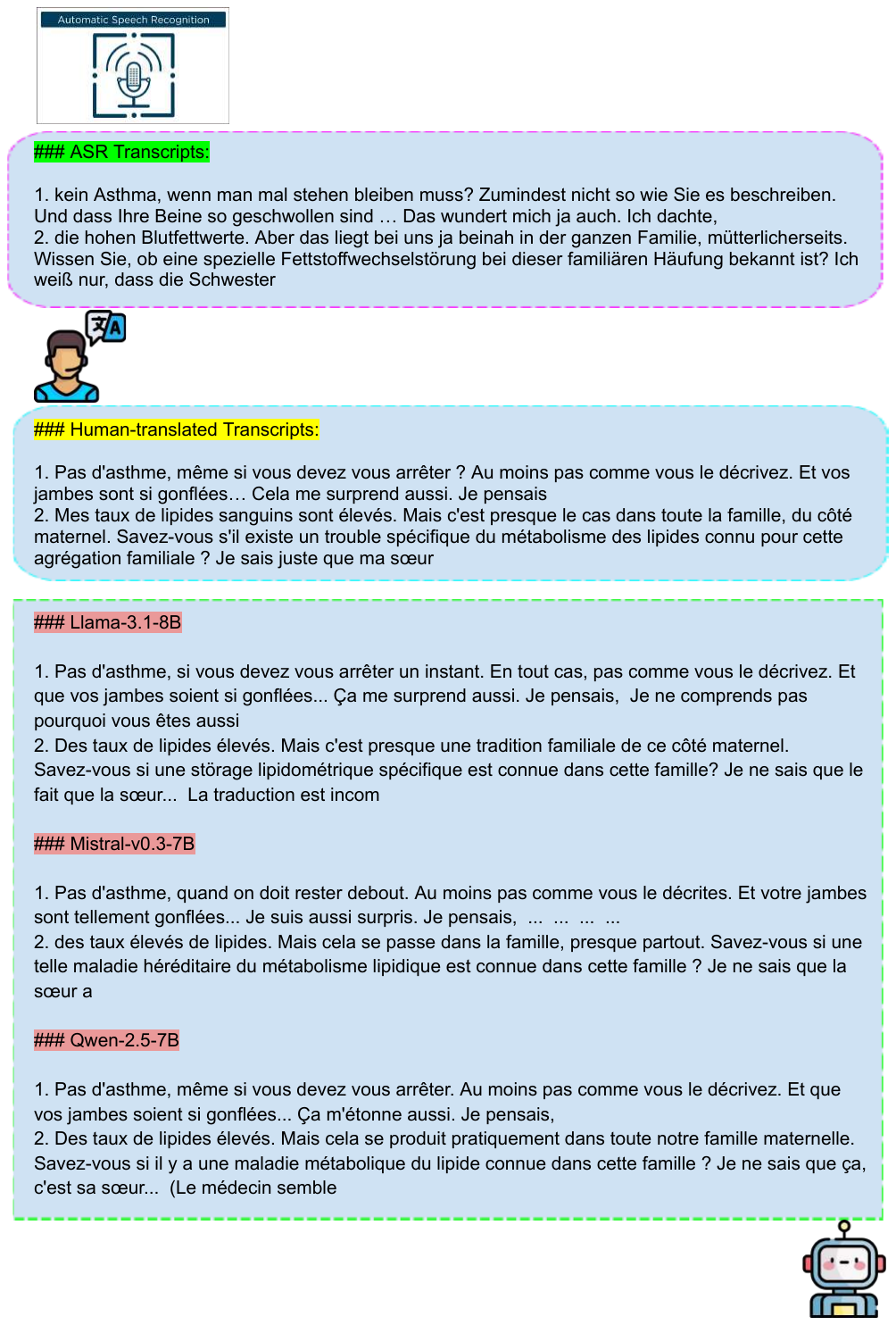}
    \caption{\textbf{Qualitative Results}. German to French \gls{ST}}
\end{figure}

\onecolumn
\subsubsection{German to Chinese Speech Translation}
\begin{figure}[h]
    \centering
    \includegraphics[width=0.8\linewidth]{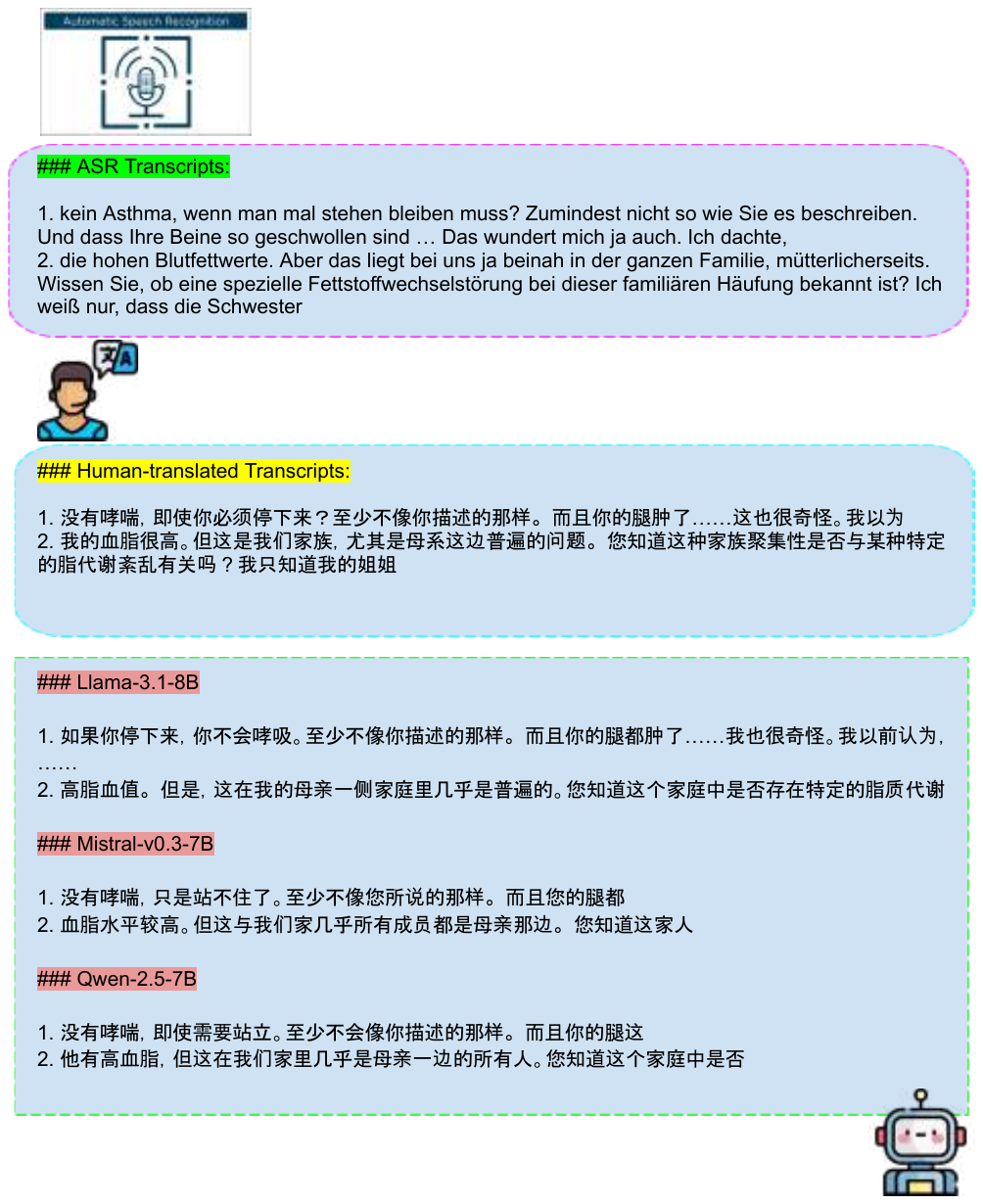}
    \caption{\textbf{Qualitative Results}. German to Chinese \gls{ST}}
\end{figure}

\onecolumn
\subsubsection{French to Vietnamese Speech Translation}
\begin{figure}[h]
    \centering
    \includegraphics[width=0.8\linewidth]{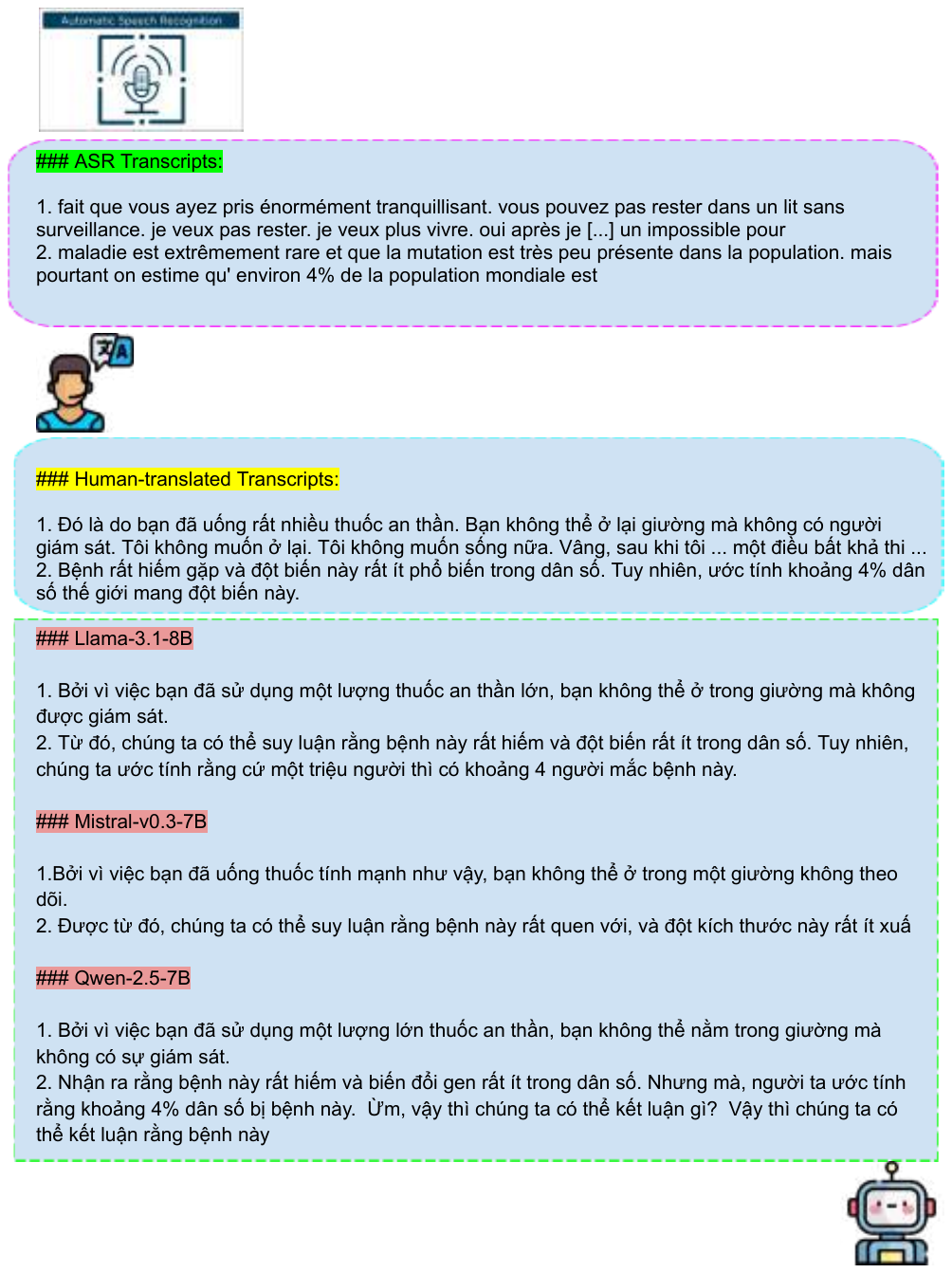}
    \caption{\textbf{Qualitative Results}. French to Vietnamese \gls{ST}}
\end{figure}

\onecolumn
\subsubsection{French to German Speech Translation}
\begin{figure}[h]
    \centering
    \includegraphics[width=0.8\linewidth]{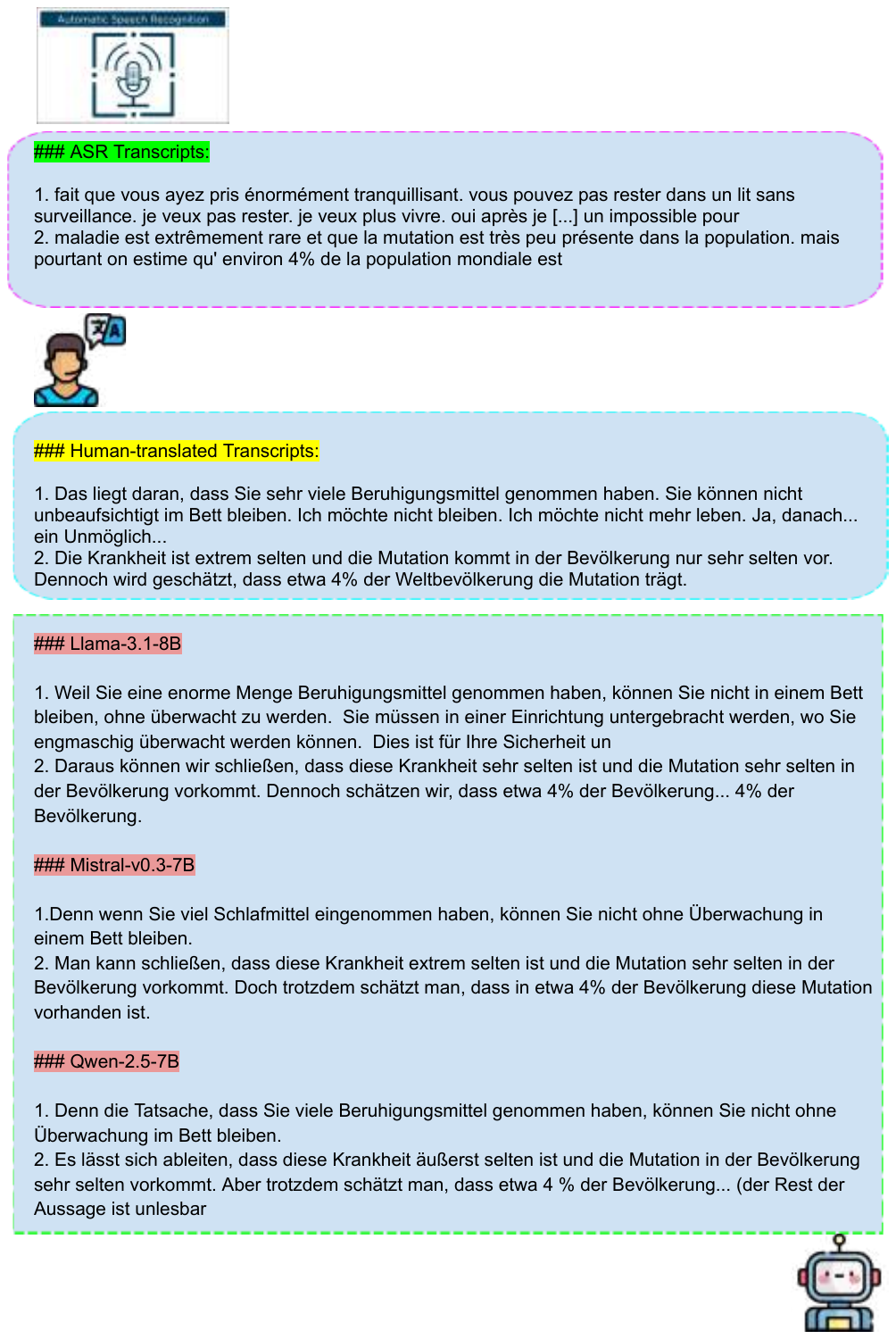}
    \caption{\textbf{Qualitative Results}. French to German \gls{ST}}
\end{figure}

\onecolumn
\subsubsection{French to English Speech Translation}
\begin{figure}[h]
    \centering
    \includegraphics[width=0.8\linewidth]{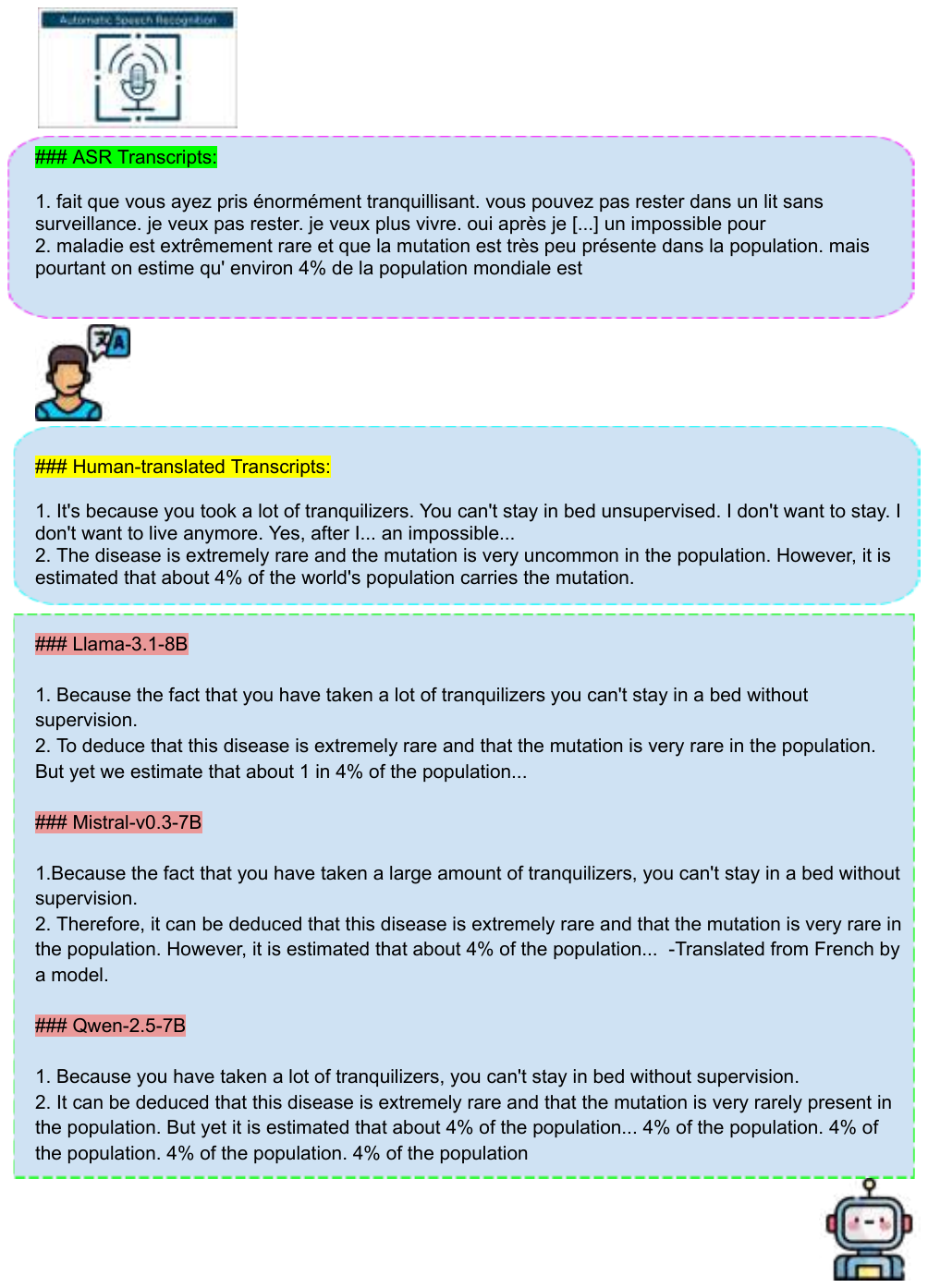}
    \caption{\textbf{Qualitative Results}. French to English \gls{ST}}
\end{figure}

\onecolumn
\subsubsection{French to Chinese Speech Translation}
\begin{figure}[h]
    \centering
    \includegraphics[width=0.8\linewidth]{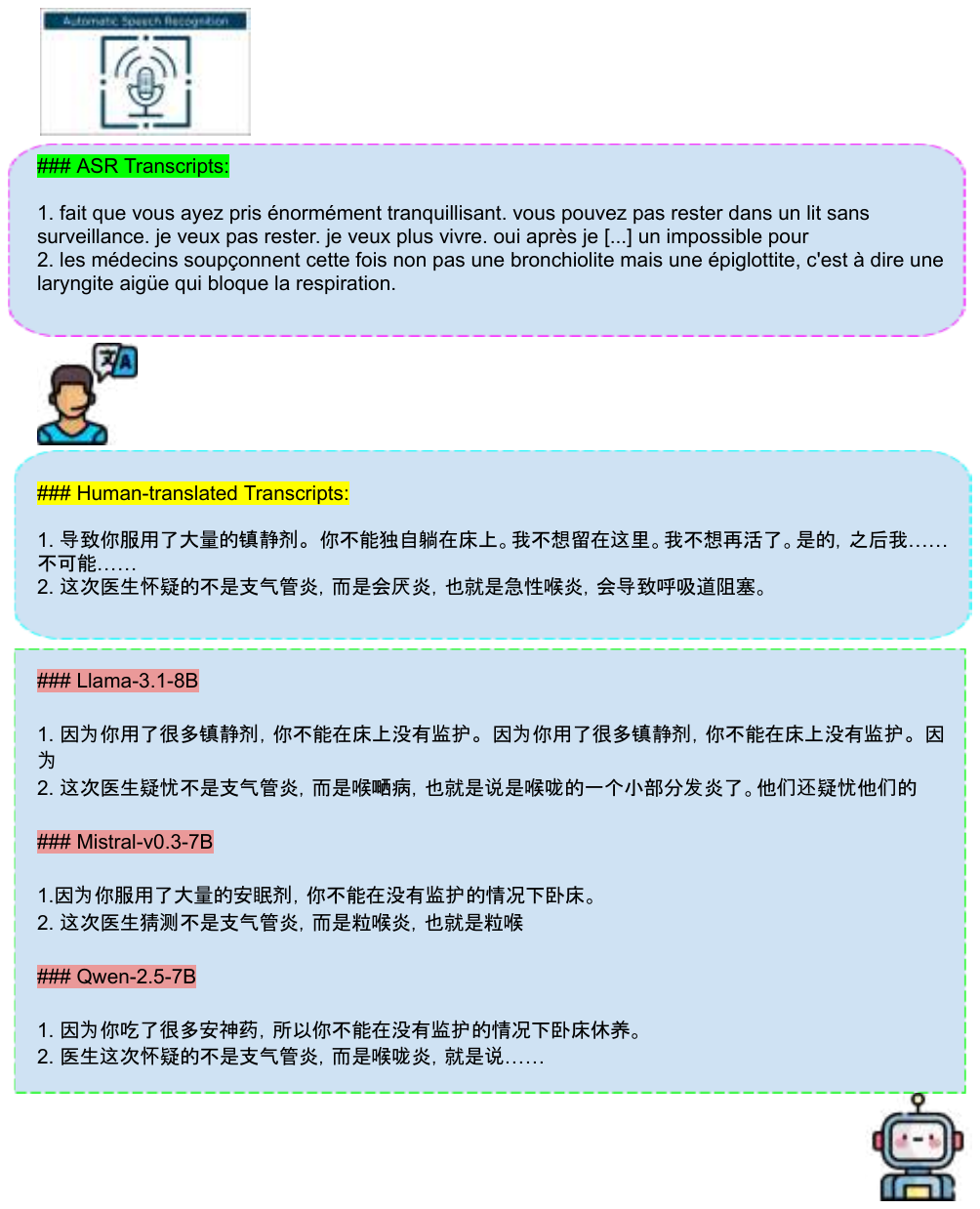}
    \caption{\textbf{Qualitative Results}. French to Chinese \gls{ST}}
\end{figure}

\onecolumn
\subsubsection{Chinese to Vietnamese Speech Translation}
\begin{figure}[h]
    \centering
    \includegraphics[width=0.8\linewidth]{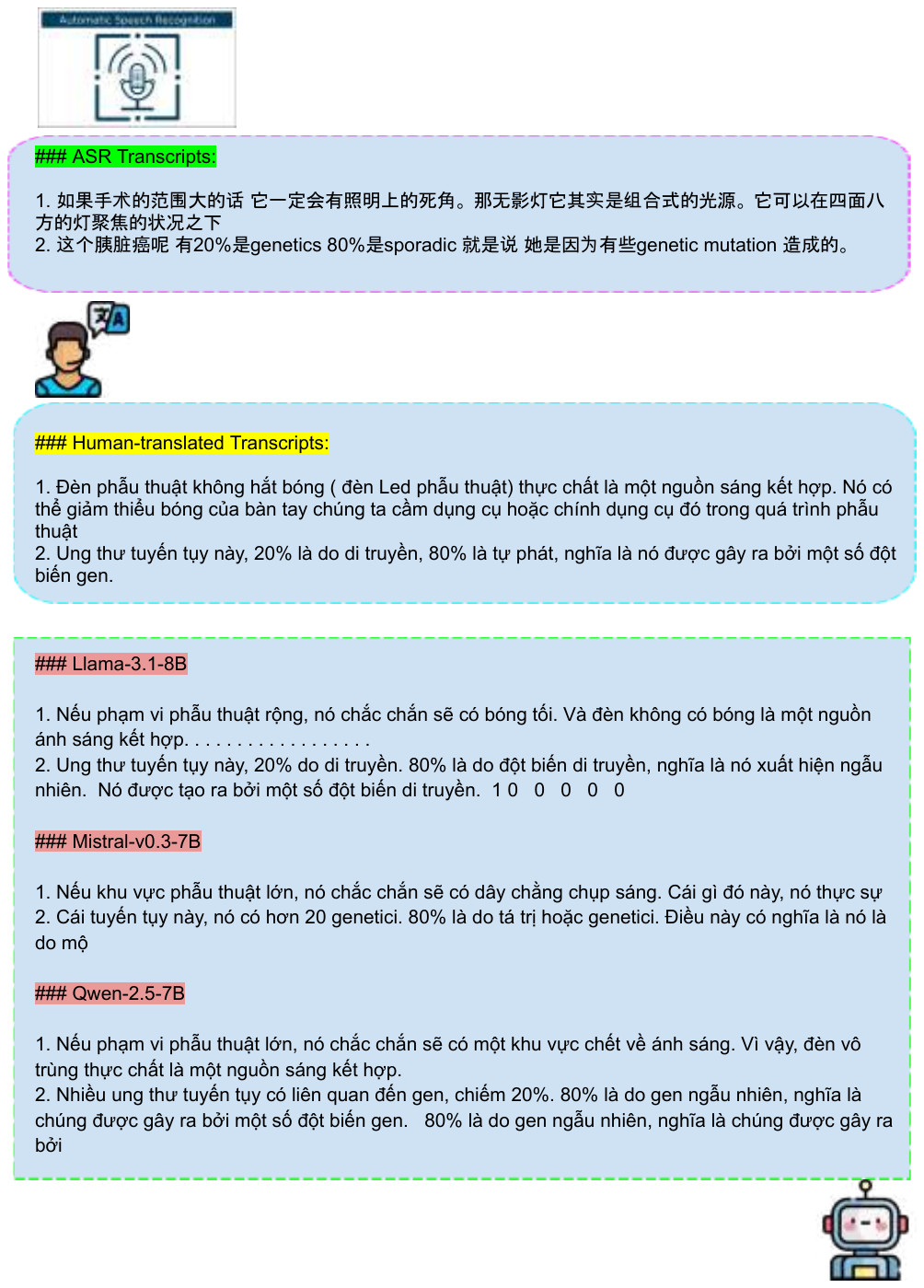}
    \caption{\textbf{Qualitative Results}. Chinese to Vietnamese \gls{ST}}
\end{figure}

\onecolumn
\subsubsection{Chinese to English Speech Translation}
\begin{figure}[h]
    \centering
    \includegraphics[width=0.8\linewidth]{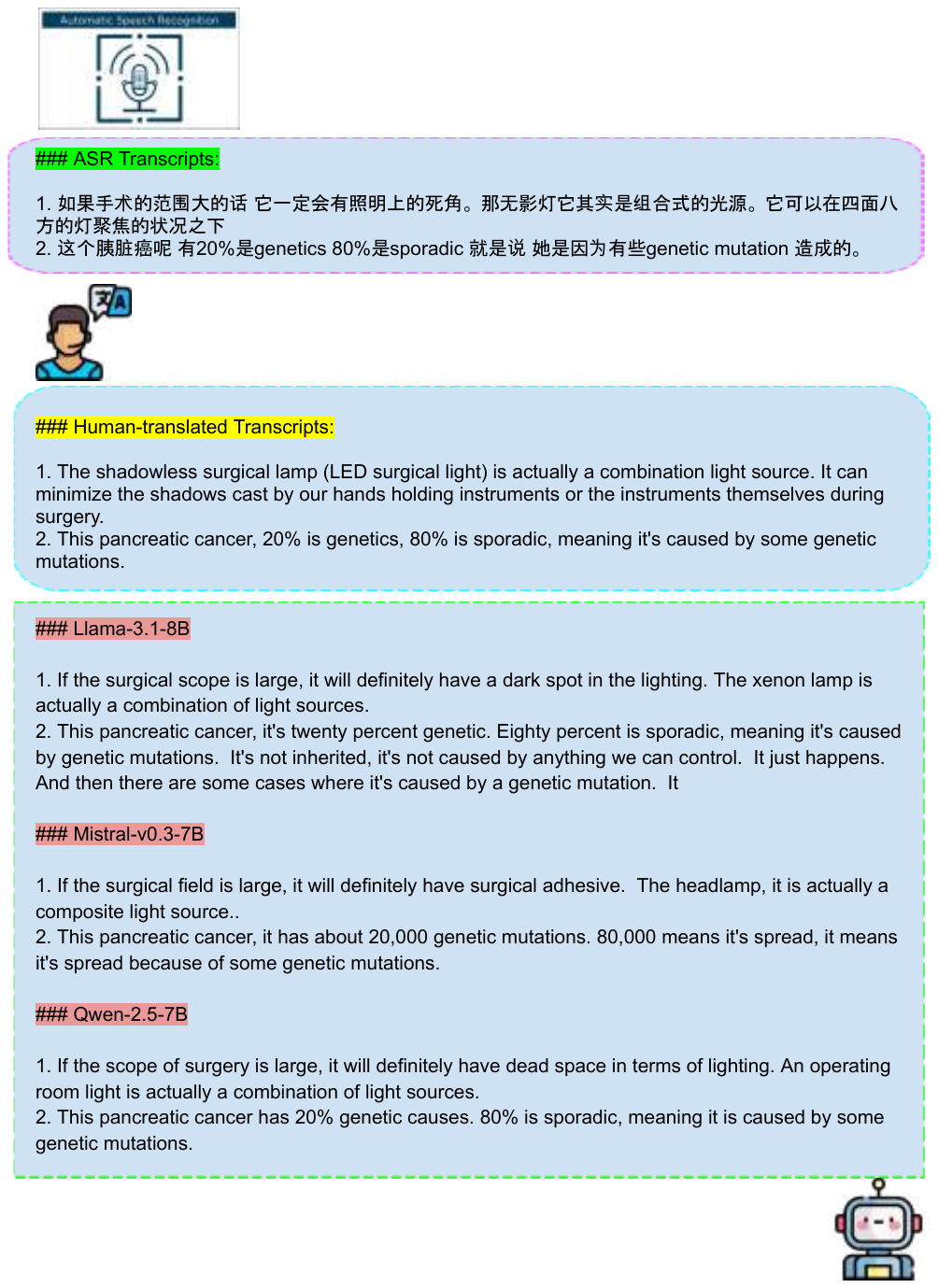}
    \caption{\textbf{Qualitative Results}. Chinese to English \gls{ST}}
\end{figure}

\onecolumn
\subsubsection{Chinese to French Speech Translation}
\begin{figure}[h]
    \centering
    \includegraphics[width=0.8\linewidth]{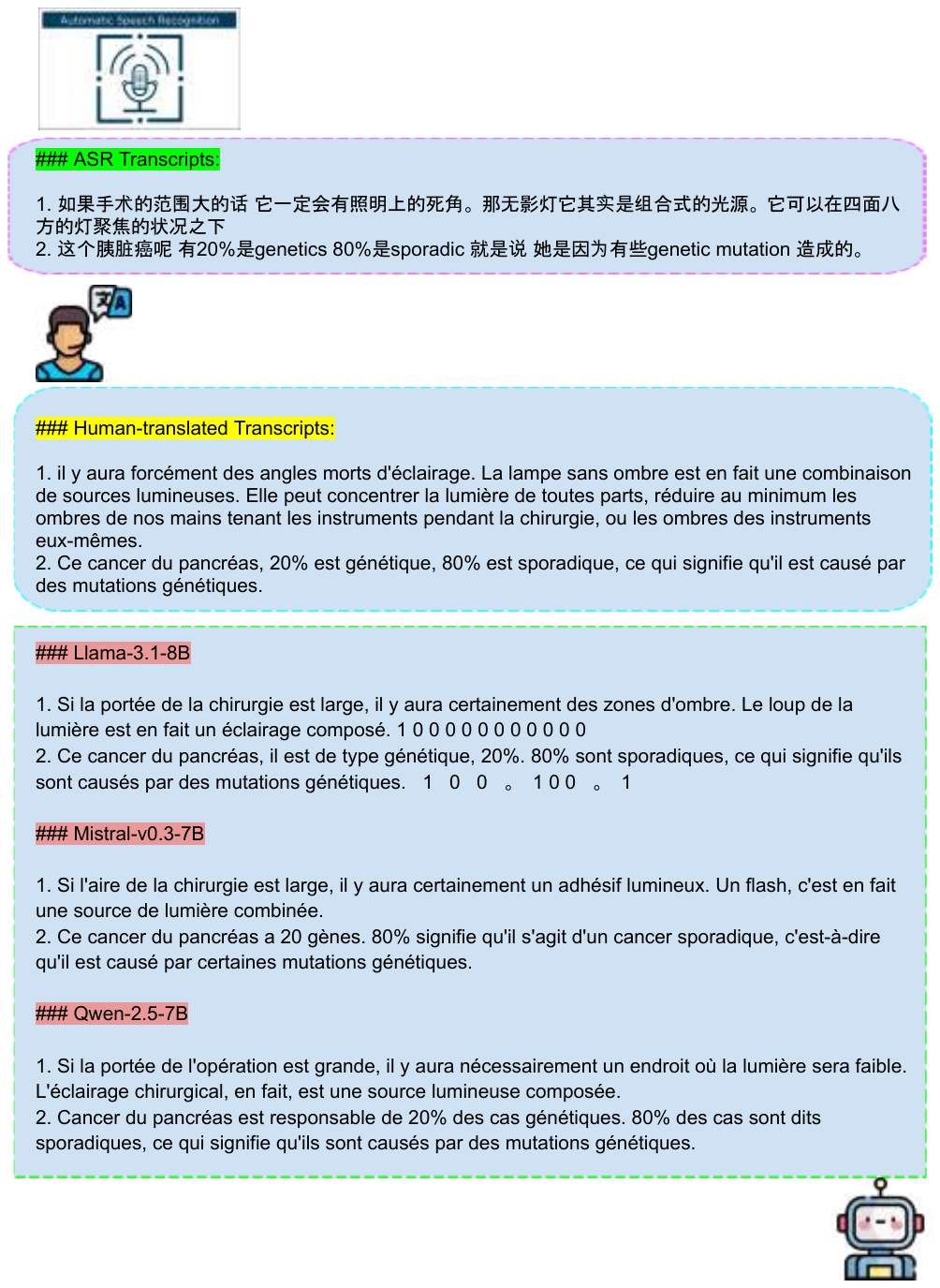}
    \caption{\textbf{Qualitative Results}. Chinese to French \gls{ST}}
\end{figure}

\onecolumn
\subsubsection{Chinese to German Speech Translation}
\begin{figure}[h]
    \centering
    \includegraphics[width=0.8\linewidth]{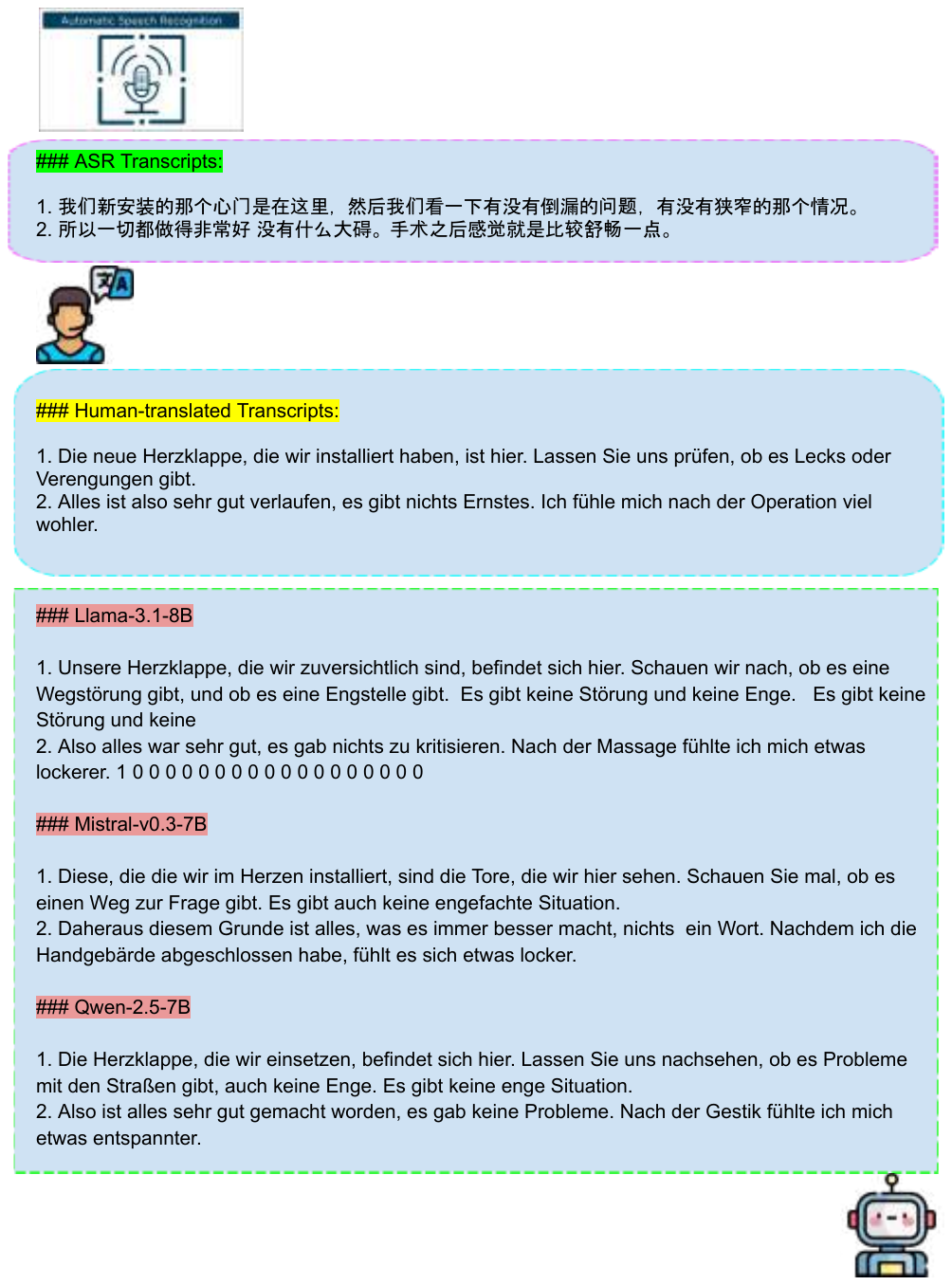}
    \caption{\textbf{Qualitative Results}. Chinese to German \gls{ST}}
\end{figure}

\twocolumn
\section{Ethical Statements}
\subsection{Fair Use}
\subsubsection{Fair Use Considerations}
The concept of fair use is critical when creating, curating, and utilizing medical \gls{ST} datasets \cite{sobel2017artificial}. Fair use provides a legal framework allowing limited use of copyrighted material without obtaining permission from the copyright holders \cite{yankwich1954fair}. However, its application is context-dependent and often requires careful analysis of specific factors to ensure compliance with the law, particularly when dealing with sensitive domains like healthcare and multilingual communication. Below is an in-depth discussion of how fair use applies in the context of medical \gls{ST} datasets.

\textbf{The Four Factors of Fair Use}:  Fair Use is governed by Section 107 of the Copyright Act, which provides a legal framework for evaluating whether a specific use of copyrighted material qualifies under this doctrine. The determination of whether the use of copyrighted material qualifies as fair use typically hinges on four key factors, as described in Fair Use defined by the U.S. Copyright Office\footnote{https://www.copyright.gov/fair-use/}. We elaborate as below:

\textbf{1. Purpose and character of the use}: 
\begin{itemize}
    \item The purpose and character of the use focus on whether the use is for nonprofit, educational, or research purposes, which are more likely to qualify as fair use, as opposed to commercial purposes. In the case of medical \gls{ST} datasets, fair use might apply if the data is used for research aimed at advancing public health, improving medical communication, or fostering innovations in machine learning for medical applications.
    \item Transformative use, where the material is repurposed or recontextualized for a different objective, also strengthens the argument for fair use. For example, using patient speech data to train \gls{AI} models for real-time \gls{MT} in healthcare settings can be considered transformative because the original purpose (e.g., doctor-patient communication) is being altered to advance language and accessibility technologies.
\end{itemize}

\textbf{2. Nature of the copyrighted work}:
\begin{itemize}
    \item The nature of the copyrighted material evaluates whether the original work is creative or factual. Works that are more factual, such as medical speech recordings or transcripts used for diagnosis and treatment, are generally more likely to fall under fair use than highly creative works like fictional narratives.
    \item Medical \gls{ST} datasets often consist of factual information, such as conversations regarding symptoms, diagnoses, or medical instructions. These factual elements weigh in favor of fair use when the data is used to support research, training, or public health interventions.
\end{itemize}

\textbf{3. Amount and substantiality of the portion used}:
\begin{itemize}
    \item This factor examines both the quantity and quality of the copyrighted material used. While fair use does not provide a specific threshold for the amount of material that can be used, using only what is necessary for the intended purpose is a key principle.
    \item In the context of medical \gls{ST} datasets, this means limiting the dataset to include only the audio or textual data required for training or evaluation. Anonymizing or redacting non-essential information, such as personally identifiable details, can further support the argument for fair use by demonstrating that the dataset minimizes unnecessary use of protected content.
\end{itemize}
   
\textbf{4. Effect of the use on the market for the original work}:
\begin{itemize}
    \item The potential impact on the market value of the original work is a crucial consideration. If the use of the copyrighted material negatively affects the market for the original work, it may weigh against fair use. For example, using proprietary medical transcription data for commercial purposes without authorization could harm the value of the original service or content.
    \item Conversely, using the material for nonprofit research or educational purposes, where there is no direct competition with the original work, is less likely to harm the market. In the case of medical speech datasets, it is important to consider whether the use might substitute for a commercial service or create a competitive disadvantage for the copyright holder.
\end{itemize}
   
\subsubsection{Ensuring Fair Use Compliance}
While the principles of fair use offer a framework for leveraging copyrighted material in medical \gls{ST} datasets, organizations and researchers should take proactive measures to minimize legal risks and maximize the ethical integrity of their projects. These measures include:

\textbf{1. Anonymization and de-identification}: Removing all personally identifiable information (PII) from the dataset is critical in the healthcare domain to comply with privacy laws such as HIPAA\footnote{https://www.hhs.gov/hipaa/for-professionals/index.html} (Health Insurance Portability and Accountability Act) and GDPR\footnote{https://gdpr-info.eu/} (General Data Protection Regulation). This step not only enhances patient confidentiality but also strengthens the argument for fair use by limiting the dataset to factual, non-identifiable content.

\textbf{2. Licensing and permissions}: Where possible, obtaining licenses or permissions to use copyrighted material ensures that the dataset is fully compliant with intellectual property laws. For medical \gls{ST} datasets, this might involve collaborating with healthcare providers, transcription services, or language experts who can provide content under appropriate agreements.

\textbf{3. Transparency and documentation}: Maintaining transparent records of how the data is sourced, processed, and used can demonstrate good-faith efforts to adhere to fair use principles, as we did here. Our documentation  includes information on the purpose of the dataset and the intended audience or beneficiaries of the research.

\textbf{4. Limiting commercialization}: Restricting the dataset's use to non-commercial purposes, such as academic research or public health initiatives, can further justify fair use, as we did here. If commercialization is pursued, ensuring that the product or service is sufficiently transformative and does not compete with the original work is critical.

\subsection{Data Consent}
It is essential to approach this topic with sensitivity and a clear understanding of ethical, legal, and scientific principles. Researchers working with medical data, including datasets for \gls{ST}, must carefully navigate the balance between advancing scientific progress and protecting patients' privacy and rights. Below is a discussion on why, in certain circumstances, researchers may not require patients' explicit consent to use medical \gls{ST} data, grounded in ethical reasoning and legal frameworks. 

\textbf{1. De-identification and anonymization of data}: One of our central arguments for not requiring explicit patient consent is that medical \gls{ST} data used for research purposes is de-identified or anonymized. This means that all PII is removed or masked in a way that makes it impossible to trace the data back to a specific individual. Under frameworks such as the HIPAA in the United States, once data is de-identified, it is no longer considered protected health information (PHI). In such cases, researchers are not legally obligated to obtain patient consent, as the data no longer poses a risk to the individual's privacy or confidentiality.

\textbf{2. Public benefit and the advancement of science}: Medical research, including the development of \gls{ST} systems, serves a broader public good by improving healthcare delivery, accessibility, and outcomes. For instance, creating accurate and effective \gls{ST} systems can help break language barriers in healthcare settings, enabling better communication between patients and providers. By allowing researchers to access de-identified data without requiring individual consent, delays in critical advancements can be avoided, ultimately benefiting society as a whole. The collective societal gain is often considered to outweigh the need for individual consent in such cases.

\textbf{3. Impracticality of obtaining consent}: In many cases, medical \gls{ST} datasets involve large-scale online collections of voice recordings or transcripts, often spanning years. Obtaining consent from every individual whose data is included in the dataset can be logistically impossible or financially prohibitive. This impracticality is especially pronounced when dealing with legacy data or when the individuals are no longer reachable. By allowing the use of such data without requiring explicit consent, researchers can ensure that valuable information is not lost to scientific progress.

\textbf{Precedents in data use for research}: The use of medical data for research purposes without explicit patient consent is not unprecedented. For example, population-level studies, epidemiological research, and biobank studies often use de-identified data without seeking individual consent. These practices are typically justified by their alignment with ethical guidelines, legal frameworks, and public health objectives. \gls{ST} datasets are no different in this regard, as long as they are managed under similar principles of de-identification.

\textbf{Transparency and accountability}: Although individual consent may not be required, transparency remains a cornerstone of ethical research. Researchers are encouraged to publicly disclose the purpose, methods, and intended outcomes of their studies, as we did here. This helps build trust with the public and ensures accountability in the use of sensitive medical data. 

In conclusion, the decision not to require explicit patient consent for medical \gls{ST} datasets is grounded in the ethical principles of beneficence, justice, and respect for privacy, as well as the legal standards governing de-identified data. While consent is a vital component of ethical research in many contexts, exceptions are made when data is anonymized, the research serves a compelling public interest, and robust safeguards are in place to protect individuals' privacy. By adhering to these principles, we can balance the need for scientific progress with the ethical imperative to protect patient rights.

\subsection{Annotation Problem for Long-form Speech}
\label{sec:annotation_problem_longform_speech}
Transcription annotation for long-form audio often suffers from timestamp mismatches, which can significantly impact the accuracy of transcriptions and the usability of the data. These mismatches arise from multiple factors, including technical limitations of \gls{ASR} models, human annotation inconsistencies, and the inherent challenges of handling long-form audio. Below are some of the key reasons why timestamp errors frequently occur in long-form \gls{ASR} annotation.

\textbf{1. \gls{ASR} model drift}: \gls{ASR} systems process audio sequentially, and small timing drifts accumulate over long durations. Many \gls{ASR} models generate timestamps by predicting words frame by frame based on phonetic models and language models \cite{lee2013joint}. However, minor deviations in phoneme alignment at the start of the audio can snowball, leading to timestamp mismatches by the middle or end of a long recording.

Frame-based decoding delays: \gls{ASR} models break audio into small frames (e.g., 10-20 ms each) \cite{tyagi2006variable}. Slight misalignments in early frames can lead to progressive timestamp shifts.

\textbf{2. Variable speech rates and pauses}: Speakers naturally change their speaking rate, pause for effect, or speed up at certain points \cite{mirghafori1996towards}. \gls{ASR} systems and humans rely on predefined acoustic models or raw audio files that may not always capture these variations accurately.
\begin{itemize}
    \item Fast speech compression: If a speaker speeds up, the \gls{ASR} systems and humans may drop or misalign words, causing an early drift in timestamps \cite{mirghafori1996towards}.
    \item Extended pauses misinterpretation: When a speaker pauses significantly, the \gls{ASR} system may either insert silence markers inaccurately or assume the next word starts too early \cite{chen2015pronunciation}, leading a wrong reference for human validation and later manual translation.
\end{itemize}

\textbf{3. Inconsistent segmentation strategies}: Long-form audio is typically segmented into smaller chunks for processing efficiency, and different segmentation strategies can cause timestamp mismatches \cite{chang2021hypothesis}.
\begin{itemize}
    \item Fixed-length segmentation: Some \gls{ASR} pipelines divide audio into fixed intervals (e.g., 30-second segments) \cite{radford2022robustspeechrecognitionlargescale}. If these segments do not align with sentence boundaries, words near the edges may be duplicated or omitted, leading to inaccurate timestamps.
    \item Overlap handling issues: Some \gls{ASR} systems introduce small overlaps to avoid word truncation at segment boundaries \cite{cetin2006speaker}. This overlap can lead to duplicate word recognition and timestamp mismatches \cite{flynn2023much}, also leading wrong reference for human annotators for transcription and translation.
\end{itemize}

\textbf{4. Human annotation variability}: Even in manually corrected \gls{ASR} transcriptions, human annotators introduce timestamp errors due to cognitive biases, differences in annotation tools, and subjective interpretations of timing.
\begin{itemize}
    \item Different perceptions of word onset and offset: Annotators may not consistently agree on where a word starts and ends, especially for words with soft or gradual onsets (e.g., "uhh", "well")
    \item Tool latency and interface limitations: Annotation software often has playback controls that introduce slight delays, affecting manual timestamp adjustments.
    \item Multispeaker confusion: When multiple speakers are involved, annotators may struggle to pinpoint precise timestamps for overlapping speech.
\end{itemize}

\textbf{5. Background noise and acoustic challenges}: Long-form audio often includes background noise, cross-talk, and varying microphone quality, which impact the accuracy of human validation and thus timestamp precision. \cite{maas2012recurrent}

\textbf{6. Post-processing and formatting issues}:
After \gls{ASR} transcription, further processing (such as punctuation insertion, casing normalization, or formatting corrections) can introduce wrong context for human validation.
\begin{itemize}
    \item Text normalization adjustments: Some \gls{ASR} systems correct text formatting after the initial transcription, potentially shifting timestamps \cite{manohar2024lost}.
    \item Forced alignment corrections: Post-processing often involves forced alignment techniques to match text to audio, but this can introduce additional errors when realigning sentences \cite{mathad2021impact}, making synthetic \gls{MT} transcripts unreliable for human validation.
\end{itemize}





\onecolumn
\section{Contribution Statements}
This list shows major contributions to this work:
\begin{enumerate}
    \item \textbf{Khai Le-Duc} led all aspects of this work, including ideation, experimental setup, paper writing and business development.
    \item \textbf{Tuyen Tran} conducted the experiments.
    \item \textbf{Bach Phan Tat} led the data annotation team.
    \item \textbf{Nguyen Kim Hai Bui} conducted all the data processing.
\end{enumerate}

\end{document}

%% file: tables/data_stats_main.tex
\begin{table}[h]
\resizebox{\columnwidth}{!}{%
\begin{tabular}{ll|ccccc}
\multicolumn{2}{l|}{\textbf{Language}} & \textbf{vi$\rightarrow$X} & \textbf{en$\rightarrow$X} & \textbf{de$\rightarrow$X} & \textbf{fr$\rightarrow$X} & \textbf{zh$\rightarrow$X} \\ \hline
\multicolumn{1}{l|}{\multirow{4}{*}{\textbf{\#Samples}}} & \textbf{Train} & 4k5 & 25k5 & 1k4 & 1k4 & 1k2 \\
\multicolumn{1}{l|}{} & \textbf{Dev} & 1k1 & 2k8 & 300 & 40 & 90 \\
\multicolumn{1}{l|}{} & \textbf{Test} & 3k4 & 4k8 & 1k1 & 300 & 200 \\
\multicolumn{1}{l|}{} & \textbf{All} & 9k1 & 33k1 & 2k8 & 1k8 & 1k6 \\ \hline
\multicolumn{1}{l|}{\multirow{5}{*}{\textbf{Med. length}}} & \textbf{$\rightarrow$vi} & 70 & 140 & 180 & 160 & 250 \\
\multicolumn{1}{l|}{} & \textbf{$\rightarrow$en} & 90 & 150 & 160 & 150 & 250 \\
\multicolumn{1}{l|}{} & \textbf{$\rightarrow$de} & 110 & 170 & 180 & 180 & 300 \\
\multicolumn{1}{l|}{} & \textbf{$\rightarrow$fr} & 100 & 160 & 200 & 140 & 290 \\
\multicolumn{1}{l|}{} & \textbf{$\rightarrow$zh} & 30 & 50 & 50 & 40 & 80
\\\hline
\end{tabular}%
}
\caption{\textbf{\highlight{Statistics of our \MultiMedST dataset.}} In total, our dataset has \textbf{290k samples (utterances) for all directions of 5 languages}: Vietnamese (vi), English (en), German (de), French (fr), and both traditional and simplified Chinese (zh). \\
Median text length is calculated based on the number of characters. 
}
\label{tab.data_stats_main}
\end{table}

%% file: tables/data_stats_comparison_medical_CutVersion.tex
\begin{table}[ht]
\resizebox{\columnwidth}{!}{%
\begin{tabular}{l|c|c|c}
\multicolumn{1}{c|}{\textbf{Dataset}} & \textbf{\#Samples} & \textbf{Lang.} & \textbf{Direction} \\ \hline
\citet{neves2017parallel} & 46k & 2 & one-to-one \\
ParaMed \cite{liu2021paramed} & 200k & 2 & one-to-one \\
Khresmoi \cite{Khresmoi_dataset} & 12k & 8 & many-to-many \\
WMT Biomed. \cite{bawden2020findings} & 160k & 9 & one-to-one \\
YuQ \cite{yu2020yuq} & 130k & 2 & one-to-one \\
\citet{berard2020multilingual} & 1k5 & 2 & one-to-one \\
MedEV \cite{vo2024improving} & 36k & 2 & one-to-one \\ \hline
\MultiMedST \textbf{(ours)} & \textbf{290k} & 5 & \textbf{many-to-many}
\\\hline
\end{tabular}%
}
\caption{\textbf{\highlight{Dataset comparison with literature.}} All publicly available datasets listed here are \textit{text-only} medical \gls{MT}. \textbf{Our \MultiMedST is the first medical \gls{ST} dataset, and is the largest medical \gls{MT} dataset}.\\ 
\textcolor{pink}{\faInfoCircle} Full details are shown in Table \ref{tab.data_stats_comparison_medical} in Appendix Section \ref{sec.dataset_comparison_with_literature}.}
\label{tab.data_stats_comparison_medical_CutVersion}
\end{table}

%% file: tables/nmt_results_gt.tex
\begin{table*}[t]
\centering
\tiny
\setlength{\tabcolsep}{2pt}
\renewcommand{\arraystretch}{1.1}

\resizebox{\textwidth}{!}{%
\begin{tabular}{llcccccccccccccccccccc}
\hline
\multicolumn{1}{l|}{\textbf{MT}} & \multicolumn{1}{l|}{\textbf{Metrics}} & \textbf{en-vi} & \textbf{en-fr} & \textbf{en-zh} & \multicolumn{1}{c|}{\textbf{en-de}} & \textbf{vi-en} & \textbf{vi-fr} & \textbf{vi-zh} & \multicolumn{1}{c|}{\textbf{vi-de}} & \textbf{fr-en} & \textbf{fr-vi} & \textbf{fr-zh} & \multicolumn{1}{c|}{\textbf{fr-de}} & \textbf{de-en} & \textbf{de-vi} & \textbf{de-fr} & \multicolumn{1}{c|}{\textbf{de-zh}} & \textbf{zh-en} & \textbf{zh-vi} & \textbf{zh-fr} & \textbf{zh-de} \\ \hline
\multicolumn{22}{c}{\colorbox{custom_light_blue}{Decoder}} \\ \hline
\multicolumn{1}{l|}{\multirow{2}{*}{\begin{tabular}[c]{@{}l@{}}Llama\\ -3.1-8B\end{tabular}}} & \multicolumn{1}{l|}{BLEU} & 53.44 & 48.24 & 37.50 & \multicolumn{1}{c|}{40.49} & 23.16 & 15.57 & 16.09 & \multicolumn{1}{c|}{11.61} & 50.18 & 39.63 & 29.25 & \multicolumn{1}{c|}{27.46} & 49.44 & 40.01 & 33.45 & \multicolumn{1}{c|}{31.16} & 28.21 & 23.49 & 18.87 & 13.07 \\
\multicolumn{1}{l|}{} & \multicolumn{1}{l|}{BERTSc} & 0.90 & 0.89 & 0.83 & \multicolumn{1}{c|}{0.87} & 0.92 & 0.79 & 0.74 & \multicolumn{1}{c|}{0.77} & 0.95 & 0.86 & 0.79 & \multicolumn{1}{c|}{0.81} & 0.95 & 0.87 & 0.84 & \multicolumn{1}{c|}{0.81} & 0.91 & 0.79 & 0.77 & 0.74 \\ \hline
\multicolumn{1}{l|}{\multirow{2}{*}{\begin{tabular}[c]{@{}l@{}}Qwen\\ -2.5-7B\end{tabular}}} & \multicolumn{1}{l|}{BLEU} & 54.50 & 49.63 & 28.61 & \multicolumn{1}{c|}{38.75} & 26.21 & 19.25 & 29.06 & \multicolumn{1}{c|}{14.44} & 49.69 & 40.67 & 20.97 & \multicolumn{1}{c|}{33.91} & 52.10 & 43.73 & 40.72 & \multicolumn{1}{c|}{23.26} & 35.63 & 32.95 & 24.05 & 16.95 \\
\multicolumn{1}{l|}{} & \multicolumn{1}{l|}{BERTSc} & 0.90 & 0.90 & 0.81 & \multicolumn{1}{c|}{0.87} & 0.93 & 0.81 & 0.81 & \multicolumn{1}{c|}{0.79} & 0.95 & 0.86 & 0.78 & \multicolumn{1}{c|}{0.84} & 0.96 & 0.88 & 0.88 & \multicolumn{1}{c|}{0.79} & 0.95 & 0.85 & 0.84 & 0.83 \\ \hline
\multicolumn{1}{l|}{\multirow{2}{*}{\begin{tabular}[c]{@{}l@{}}Mistral\\ -v0.3-7B\end{tabular}}} & \multicolumn{1}{l|}{BLEU} & 24.77 & 51.71 & 26.38 & \multicolumn{1}{c|}{43.99} & 24.56 & 16.00 & 25.04 & \multicolumn{1}{c|}{13.38} & 34.95 & 14.47 & 19.92 & \multicolumn{1}{c|}{33.73} & 36.39 & 15.68 & 40.77 & \multicolumn{1}{c|}{21.28} & 27.68 & 10.67 & 18.46 & 11.40 \\
\multicolumn{1}{l|}{} & \multicolumn{1}{l|}{BERTSc} & 0.82 & 0.89 & 0.81 & \multicolumn{1}{c|}{0.88} & 0.92 & 0.79 & 0.78 & \multicolumn{1}{c|}{0.78} & 0.91 & 0.79 & 0.78 & \multicolumn{1}{c|}{0.85} & 0.92 & 0.79 & 0.86 & \multicolumn{1}{c|}{0.78} & 0.93 & 0.75 & 0.80 & 0.76 \\ \hline
\multicolumn{22}{c}{\colorbox{custom_light_blue}{Encoder-decoder}} \\ \hline
\multicolumn{1}{l|}{\multirow{2}{*}{\begin{tabular}[c]{@{}l@{}}mBart\\ -large-50\end{tabular}}} & \multicolumn{1}{l|}{BLEU} & 59.73 & 56.23 & 44.77 & \multicolumn{1}{c|}{46.48} & 16.48 & 12.61 & 22.97 & \multicolumn{1}{c|}{10.43} & 39.58 & 36.17 & 24.63 & \multicolumn{1}{c|}{28.73} & 41.45 & 41.12 & 40.48 & \multicolumn{1}{c|}{30.43} & 15.03 & 14.26 & 15.70 & 10.67 \\
\multicolumn{1}{l|}{} & \multicolumn{1}{l|}{BERTSc} & 0.92 & 0.92 & 0.86 & \multicolumn{1}{c|}{0.89} & 0.89 & 0.80 & 0.78 & \multicolumn{1}{c|}{0.75} & 0.93 & 0.86 & 0.77 & \multicolumn{1}{c|}{0.83} & 0.94 & 0.87 & 0.87 & \multicolumn{1}{c|}{0.80} & 0.90 & 0.82 & 0.79 & 0.77 \\ \hline
\multicolumn{1}{l|}{\multirow{2}{*}{\begin{tabular}[c]{@{}l@{}}M2M100\\ -418M\end{tabular}}} 
& \multicolumn{1}{l|}{BLEU} 
& \textcolor{m2mblue}{62.31} & \textcolor{m2mblue}{57.49} & \textcolor{m2mblue}{46.38} & \multicolumn{1}{c|}{\textcolor{m2mblue}{49.36}} 
& 23.01 & \textcolor{m2mblue}{21.10} & 24.95 & \multicolumn{1}{c|}{\textcolor{m2mblue}{16.72}} 
& 43.73 & 35.04 & \textcolor{m2mblue}{29.41} & \multicolumn{1}{c|}{\textcolor{m2mblue}{34.72}} 
& 44.76 & \textcolor{m2mblue}{43.83} & \textcolor{m2mblue}{43.53} & \multicolumn{1}{c|}{30.42} 
& 21.65 & 27.69 & 21.88 & 15.17 \\
\multicolumn{1}{l|}{} & \multicolumn{1}{l|}{BERTSc} & \textcolor{m2mblue}{0.97} & \textcolor{m2mblue}{0.95} & \textcolor{m2mblue}{0.93} & \multicolumn{1}{c|}{\textcolor{m2mblue}{0.94}} & 0.82 & \textcolor{m2mblue}{0.81} & 0.80 & \multicolumn{1}{c|}{\textcolor{m2mblue}{0.79}} & 0.88 & 0.82 & \textcolor{m2mblue}{0.82} & \multicolumn{1}{c|}{0.83} & 0.83 & 0.85 & \textcolor{m2mblue}{0.88} & \multicolumn{1}{c|}{0.75} & 0.76 & \textcolor{m2mblue}{0.85} & 0.82 & 0.82 \\ \hline
\multicolumn{1}{l|}{\multirow{2}{*}{Marian}} & \multicolumn{1}{l|}{BLEU} & 58.22 & 53.84 & 38.67 & \multicolumn{1}{c|}{45.81} & 17.63 & 15.97 & 15.56 & \multicolumn{1}{c|}{12.84} & 39.97 & 33.41 & 17.13 & \multicolumn{1}{c|}{32.62} & 42.74 & 38.26 & 39.59 & \multicolumn{1}{c|}{18.11} & 11.44 & 16.14 & 11.33 & 6.24 \\
\multicolumn{1}{l|}{} & \multicolumn{1}{l|}{BERTSc} & 0.91 & 0.91 & 0.85 & \multicolumn{1}{c|}{0.89} & 0.80 & 0.79 & 0.78 & \multicolumn{1}{c|}{0.77} & 0.87 & 0.86 & 0.78 & \multicolumn{1}{c|}{\textcolor{m2mblue}{0.85}} & 0.88 & 0.87 & 0.87 & \multicolumn{1}{c|}{0.78} & 0.78 & 0.79 & 0.77 & 0.75 \\ \hline
\multicolumn{22}{c}{\colorbox{custom_light_blue}{Commercial tool}} \\ \hline
\multicolumn{1}{l|}{\multirow{2}{*}{\begin{tabular}[c]{@{}l@{}}Google\\ Translate\end{tabular}}} & \multicolumn{1}{l|}{BLEU} & 59.50 & \textcolor{gtred}{59.28} & \textcolor{gtred}{57.13} & \multicolumn{1}{c|}{\textcolor{gtred}{49.12}} & \textcolor{gtred}{28.62} & \textcolor{gtred}{25.25} & \textcolor{gtred}{31.24} & \multicolumn{1}{c|}{\textcolor{gtred}{19.00}} & 47.47 & \textcolor{gtred}{39.28} & \textcolor{gtred}{39.38} & \multicolumn{1}{c|}{\textcolor{gtred}{38.89}} & \textcolor{gtred}{53.35} & \textcolor{gtred}{42.47} & \textcolor{gtred}{43.67} & \multicolumn{1}{c|}{\textcolor{gtred}{40.54}} & \textcolor{gtred}{39.34} & \textcolor{gtred}{44.41} & \textcolor{gtred}{29.48} & \textcolor{gtred}{24.77} \\
\multicolumn{1}{l|}{} & \multicolumn{1}{l|}{BERTSc} & \textcolor{gtred}{0.91} & \textcolor{gtred}{0.91} & \textcolor{gtred}{0.90} & \multicolumn{1}{c|}{0.89} & 0.84 & \textcolor{gtred}{0.83} & \textcolor{gtred}{0.83} & \multicolumn{1}{c|}{\textcolor{gtred}{0.81}} & 0.88 & \textcolor{gtred}{0.86} & \textcolor{gtred}{0.85} & \multicolumn{1}{c|}{\textcolor{gtred}{0.86}} & 0.90 & \textcolor{gtred}{0.88} & \textcolor{gtred}{0.88} & \multicolumn{1}{c|}{\textcolor{gtred}{0.86}} & 0.88 & \textcolor{gtred}{0.87} & \textcolor{gtred}{0.85} & \textcolor{gtred}{0.85} \\ \hline
\end{tabular}%
}
\caption{\textbf{\highlight{Ground-truth MT baselines.}} All \gls{MT} models were fine-tuned monolingually (on each respective language pair separately) except Google Translate being recognized directly on test set. \textit{en-vi} denotes translation from \textit{en} to \textit{vi}. Only BLEU (n-gram overlap metric) and BERTScore (embedding-based metric) were reported in this table. \\
\textcolor{red}{\faChartLine} Google Translate leads overall, with Encoder-decoder \gls{MT} models often surpassing \glspl{LLM} on many language pairs.\\
\textcolor{pink}{\faInfoCircle} Full results for all evaluation metrics (including other n-gram overlap metrics) are shown in Table \ref{tab:appx_nmt_gt_allMetrics-En-X} (English to X), Table \ref{tab:appx_nmt_gt_allMetrics-Vi-X} (Vietnamese to X), Table \ref{tab:appx_nmt_gt_allMetrics-Fr-X} (French to X), Table \ref{tab:appx_nmt_gt_allMetrics-De-X} (German to X), and Table \ref{tab:appx_nmt_gt_allMetrics-Zh-X} (Chinese to X) in Appendix Section \ref{sec:full_results_groundtruth_MT_baselines}. \\ 
\textbf{\textcolor{gtred}{Red}} highlight: best result. \textbf{\textcolor{m2mblue}{Blue}} highlight: second-best result (Encoder-decoder models outperform Decoder-only)
}
\label{tab:translation-groundtruth}
\end{table*}

%% file: tables/asr_results.tex
\begin{table}[h]
\centering
\tiny
\setlength{\tabcolsep}{2pt}
\renewcommand{\arraystretch}{1.6}
\adjustbox{width=\columnwidth}{
\begin{tabular}{l|ccccc|ccccc}
\multirow{2}{*}{\textbf{ASR}} & \multicolumn{5}{c|}{\textbf{dev}} & \multicolumn{5}{c}{\textbf{test}} \\ \cline{2-11} 
 & \textbf{vi} & \textbf{en} & \textbf{zh} & \textbf{de} & \textbf{fr} & \textbf{vi} & \textbf{en} & \textbf{zh} & \textbf{de} & \textbf{fr} \\ \hline
Whisper-small-mono & 21.2 & 24.4 & \textcolor{gtred}{25.9} & \textcolor{gtred}{17.5} & \textcolor{gtred}{35.8} & \textcolor{gtred}{29.6} & 33.8 & 31.3 & 26.3 & 45.7 \\
+ SpecAugment & \textcolor{gtred}{19.8} & \textcolor{gtred}{23.5} & 43.3 & 17.9 & 44.1 & \textcolor{gtred}{31.7} & 36.9 & 46.9 & 24.1 & 45.6 \\
Whisper-small-multi & 25.7 & 46.1 & 73.9 & 22.2 & 50.6 & 33.4 & 40.9 & 89.8 & \textcolor{gtred}{19.6} & 55.3 \\
Whisper-large-v2-mono & 57.7 & 26.9 & 39.0 & 23.7 & 52.9 & 62.6 & \textcolor{gtred}{25.5} & 37.3 & 24.2 & \textcolor{gtred}{41.7} \\
Assembly & 51.9 & 31.7 & 49.8 & 27.9 & 49.4 & 65.5 & 30.6 & 45.2 & 28.9 & 42.1 \\
Deepgram & 35.8 & 33.9 & 40.4 & 27.8 & 50.7 & 40.0 & 32.1 & 46.7 & 28.4 & 40.3
\\\hline
\end{tabular}
}
\caption{\textbf{\highlight{ASR baseline results.}} Chinese (zh) is evaluated by \gls{CER} (\%), while other languages are evaluated by \gls{WER} (\%). Whisper is fine-tuned monolingually (each language separately) or multilingually (all languages simultaneously). SpecAugment \cite{park2019specaugment} is tested on Whisper-small-mono as data augmentation. Commercial models like Assembly and Deepgram only allows direct recognition. \\
\textcolor{red}{\faChartLine} Monolingual Whisper-small leads overall, while larger models excel in high-resource languages.}
\label{tab:model_comparison}
\end{table}

%% file: tables/asr_nmt_result.tex
\begin{table*}[h]
\centering
\setlength{\tabcolsep}{1pt}
\renewcommand{\arraystretch}{1.1}
\resizebox{\textwidth}{!}{%
\begin{tabular}{l|l|l|cccc|cccc|cccc|cccc|cccc}
\hline
\textbf{ASR} & \textbf{MT} & \textbf{Metrics} & \textbf{en-vi} & \textbf{en-fr} & \textbf{en-zh} & \textbf{en-de} & \textbf{vi-en} & \textbf{vi-fr} & \textbf{vi-zh} & \textbf{vi-de} & \textbf{fr-en} & \textbf{fr-vi} & \textbf{fr-zh} & \textbf{fr-de} & \textbf{de-en} & \textbf{de-vi} & \textbf{de-fr} & \textbf{de-zh} & \textbf{zh-en} & \textbf{zh-vi} & \textbf{zh-fr} & \textbf{zh-de} \\ \hline
\multirow{4}{*}{\begin{tabular}[c]{@{}l@{}}Ground\\ -truth\end{tabular}} & \multirow{2}{*}{\begin{tabular}[c]{@{}l@{}}mBart-\\ large-50\end{tabular}} & BLEU & 59.73 & 56.23 & 44.77 & 46.48 & 16.48 & 12.61 & 22.97 & 10.43 & 39.58 & 36.17 & 24.63 & 28.73 & 41.45 & 41.12 & 40.48 & 30.43 & 15.03 & 14.26 & 15.70 & 10.67 \\
 &  & BERTSc & 0.92 & 0.92 & 0.86 & 0.89 & 0.89 & 0.80 & 0.78 & 0.75 & 0.93 & 0.86 & 0.77 & 0.83 & 0.94 & 0.87 & 0.87 & 0.80 & 0.90 & 0.82 & 0.79 & 0.77 \\ \cline{2-23} 
& \multirow{2}{*}{\begin{tabular}[c]{@{}l@{}}M2M100\\ -418M\end{tabular}} & BLEU & \textcolor{gtred}{62.31} & \textcolor{gtred}{57.49} & \textcolor{gtred}{46.38} & \textcolor{gtred}{49.36} & \textcolor{gtred}{23.01} & \textcolor{gtred}{21.10} & \textcolor{gtred}{24.95} & \textcolor{gtred}{16.72} & \textcolor{gtred}{43.73} & \textcolor{gtred}{35.04} & \textcolor{gtred}{29.41} & \textcolor{gtred}{34.72} & \textcolor{gtred}{44.76} & \textcolor{gtred}{43.83} & \textcolor{gtred}{43.53} & \textcolor{gtred}{30.42} & \textcolor{gtred}{21.65} & \textcolor{gtred}{27.69} & \textcolor{gtred}{21.88} & \textcolor{gtred}{15.17} \\
 &  & BERTSc & 0.97 & 0.95 & 0.93 & 0.94 & 0.82 & 0.81 & 0.80 & 0.79 & 0.88 & 0.82 & 0.82 & 0.83 & 0.83 & 0.85 & 0.88 & 0.75 & 0.76 & 0.85 & 0.82 & 0.82 \\ \hline
\multirow{4}{*}{\begin{tabular}[c]{@{}l@{}}Whisper\\ small-\\ mono\end{tabular}} & \multirow{2}{*}{\begin{tabular}[c]{@{}l@{}}mBart-\\ large-50\end{tabular}} & BLEU & 48.00 & 43.20 & 35.70 & 35.07 & 10.17 & 12.80 & 16.77 & 7.23 & 23.82 & 22.86 & 16.46 & 17.39 & 31.95 & 32.62 & 31.96 & 25.07 & 11.88 & 18.40 & 12.30 & 9.64 \\
 &  & BERTSc & 0.87 & 0.86 & 0.81 & 0.84 & 0.88 & 0.76 & 0.73 & 0.72 & 0.90 & 0.80 & 0.72 & 0.77 & 0.92 & 0.84 & 0.84 & 0.77 & 0.89 & 0.79 & 0.76 & 0.75 \\ \cline{2-23} 
 & \multirow{2}{*}{\begin{tabular}[c]{@{}l@{}}M2M100\\ -418M\end{tabular}} & BLEU & 48.21 & 43.16 & 36.94 & 36.55 & \textcolor{m2mblue}{15.64} & 13.95 & \textcolor{m2mblue}{16.99} & \textcolor{m2mblue}{11.10} & 25.65 & 21.88 & 18.44 & 19.98 & 33.66 & 34.70 & 34.67 & 24.31 & 16.65 & 21.83 & 16.94 & 13.06 \\
 &  & BERTSc & 0.95 & 0.92 & 0.92 & 0.92 & 0.78 & 0.77 & 0.74 & 0.75 & 0.81 & 0.76 & 0.75 & 0.76 & 0.77 & 0.81 & 0.85 & 0.72 & 0.76 & 0.83 & 0.79 & 0.78 \\ \hline
\multirow{4}{*}{\begin{tabular}[c]{@{}l@{}}Whisper\\ small-\\ multi\end{tabular}} & \multirow{2}{*}{\begin{tabular}[c]{@{}l@{}}mBart-\\ large-50\end{tabular}} & BLEU & 47.99 & 43.22 & 36.02 & 34.93 & 10.16 & 12.18 & 16.95 & 6.88 & 25.53 & 25.34 & 19.15 & 19.26 & 34.6 & 35.2 & 34.45 & 27.19 & 12.39 & 18.4 & 11.5 & 9.31 \\
 &  & BERTSc & 0.87 & 0.86 & 0.81 & 0.84 & 0.88 & 0.76 & 0.73 & 0.72 & 0.91 & 0.81 & 0.73 & 0.78 & 0.93 & 0.85 & 0.85 & 0.78 & 0.89 & 0.78 & 0.75 & 0.74 \\ \cline{2-23} 
 & \multirow{2}{*}{\begin{tabular}[c]{@{}l@{}}M2M100\\ -418M\end{tabular}} & BLEU & 48.1 & 42.98 & 36.94 & 36.28 & 14.76 & 13.1 & 17.17 & 10.38 & 27.97 & 24.84 & 22.99 & 22.97 & 37.25 & 37.26 & 37.61 & 27.20 & 14.64 & 21.35 & 15.37 & 11.89 \\
 &  & BERTSc & 0.95 & 0.92 & 0.92 & 0.91 & 0.78 & 0.77 & 0.74 & 0.75 & 0.82 & 0.77 & 0.77 & 0.76 & 0.79 & 0.81 & 0.85 & 0.74 & 0.74 & 0.85 & 0.78 & 0,76 \\ \hline
\multirow{4}{*}{\begin{tabular}[c]{@{}l@{}}Whisper\\ large-\\ mono\end{tabular}} & \multirow{2}{*}{\begin{tabular}[c]{@{}l@{}}mBart-\\ large-50\end{tabular}} & BLEU & 53.43 & 47.69 & 40.82 & 39.19 & 6.71 & 8.67 & 11.30 & 4.19 & 29.47 & 28.01 & 20.63 & 21.39 & 35.29 & 35.96 & 34.56 & 28.81 & 7.41 & 12.39 & 9.51 & 5.90 \\
 &  & BERTSc & 0.89 & 0.88 & 0.84 & 0.86 & 0.86 & 0.72 & 0.65 & 0.68 & 0.92 & 0.82 & 0.74 & 0.79 & 0.93 & 0.85 & 0.85 & 0.79 & 0.81 & 0.61 & 0.74 & 0.71 \\ \cline{2-23} 
& \multirow{2}{*}{\begin{tabular}[c]{@{}l@{}}M2M100\\ -418M\end{tabular}} & BLEU & \textcolor{m2mblue}{53.42} & \textcolor{m2mblue}{47.96} & \textcolor{m2mblue}{42.05} & \textcolor{m2mblue}{40.52} & 10.85 & 9.68 & 11.45 & 7.76 & \textcolor{m2mblue}{32.19} & \textcolor{m2mblue}{29.84} & \textcolor{m2mblue}{25.52} & \textcolor{m2mblue}{25.25} & \textcolor{m2mblue}{37.90} & \textcolor{m2mblue}{38.51} & \textcolor{m2mblue}{37.72} & \textcolor{m2mblue}{28.69} & \textcolor{m2mblue}{18.71} & \textcolor{m2mblue}{24.20} & \textcolor{m2mblue}{16.83} & \textcolor{m2mblue}{13.66} \\
&  & BERTSc & 0.96 & 0.93 & 0.92 & 0.93 & 0.73 & 0.72 & 0.70 & 0.71 & 0.84 & 0.79 & 0.79 & 0.79 & 0.81 & 0.83 & 0.86 & 0.74 & 0.78 & 0.85 & 0.78 & 0.78 \\ \hline
\multirow{4}{*}{Assembly} & \multirow{2}{*}{\begin{tabular}[c]{@{}l@{}}mBart-\\ large-50\end{tabular}} & BLEU & 51.23 & 45.45 & 40.51 & 37.37 & 8.37 & 11.03 & 14.55 & 4.74 & 29.00 & 26.52 & 18.77 & 19.84 & 33.84 & 34.64 & 32.76 & 28.42 & 4.90 & 10.21 & 7.79 & 4.97 \\
 &  & BERTSc & 0.88 & 0.88 & 0.83 & 0.86 & 0.87 & 0.75 & 0.71 & 0.70 & 0.91 & 0.82 & 0.74 & 0.79 & 0.93 & 0.85 & 0.85 & 0.79 & 0.76 & 0.60 & 0.73 & 0.72 \\ \cline{2-23} 
 & \multirow{2}{*}{\begin{tabular}[c]{@{}l@{}}M2M100\\ -418M\end{tabular}} & BLEU & 51.30 & 45.85 & 41.91 & 38.38 & 13.60 & 12.56 & 13.95 & 9.71 & 31.20 & 27.12 & 22.84 & 22.93 & 35.89 & 37.15 & 35.83 & 30.17 & 14.90 & 19.79 & 13.11 & 10.23 \\
 &  & BERTSc & 0.95 & 0.93 & 0.92 & 0.92 & 0.77 & 0.76 & 0.69 & 0.75 & 0.83 & 0.78 & 0.77 & 0.76 & 0.77 & 0.81 & 0.86 & 0.77 & 0.77 & 0.83 & 0.78 & 0.77 \\ \hline
\multirow{4}{*}{Deepgram} & \multirow{2}{*}{\begin{tabular}[c]{@{}l@{}}mBart-\\ large-50\end{tabular}} & BLEU & 50.93 & 45.37 & 39.93 & 37.30 & 9.44 & 12.05 & 15.48 & 5.88 & 28.95 & 27.39 & 18.82 & 20.52 & 33.99 & 35.37 & 33.49 & 27.90 & 4.90 & 7.79 & 07.03 & 3.31 \\
 &  & BERTSc & 0.88 & 0.88 & 0.83 & 0.85 & 0.88 & 0.67 & 0.65 & 0.70 & 0.91 & 0.82 & 0.73 & 0.79 & 0.93 & 0.85 & 0.85 & 0.79 & 0.80 & 0.60 & 0.71 & 0.69 \\ \cline{2-23} 
 & \multirow{2}{*}{\begin{tabular}[c]{@{}l@{}}M2M100\\ -418M\end{tabular}} & BLEU & 51.01 & 45.34 & 41.18 & 38.42 & 15.60 & \textcolor{m2mblue}{14.20} & 16.24 & 11.10 & 31.02 & 28.38 & 24.04 & 22.75 & 36.02 & 37.55 & 36.45 & 28.78 & 13.47 & 16.50 & 12.40 & 9.57 \\
 &  & BERTSc & 0.95 & 0.93 & 0.92 & 0.92 & 0.76 & 0.76 & 0.73 & 0.74 & 0.82 & 0.77 & 0.78 & 0.74 & 0.79 & 0.82 & 0.86 & 0.75 & 0.76 & 0.82 & 0.76 & 0.76 \\ \hline
\end{tabular}%
}
\caption{\textbf{\highlight{Cascaded ST baseline results.}} The effect of ASR models on \gls{MT} quality is compared with \gls{MT} on ground-truth text. Monolingual translation fine-tuning refers to fine-tuning \gls{MT} models on each language pair separately, while multilingual translation fine-tuning refers to fine-tuning \gls{MT} models on all language pairs simutaneously. \\
\textcolor{red}{\faChartLine} Whisper-large-v2 with M2M100-418M achieved the best overall \gls{ST} performance, except for Vietnamese where Whisper-small-mono was superior.\\
\textcolor{pink}{\faInfoCircle} Extra results for all evaluation metrics and models are shown in Table \ref{tab:appx_nmt_asr_allMetrics-En-X} (English to X), Table \ref{tab:appx_nmt_asr_allMetrics-Vi-X} (Vietnamese to X), Table \ref{tab:appx_nmt_asr_allMetrics-Fr-X} (French to X), Table \ref{tab:appx_nmt_asr_allMetrics-De-X} (German to X), and Table \ref{tab:appx_nmt_asr_allMetrics-Zh-X} (Chinese to X) in Appendix Section \ref{sec:extra_results_cascaded_ST_baselines}.}
\label{tab:asr-translation}
\end{table*}

%% file: tables/nmt_results_asr.tex
\begin{table*}[t]
\centering
\tiny
\setlength{\tabcolsep}{2pt}
\renewcommand{\arraystretch}{1.}
\resizebox{\textwidth}{!}{%
\begin{tabular}{llcccccccccccccccccccc}
\hline
\multicolumn{1}{l|}{\textbf{Model}} & \multicolumn{1}{l|}{\textbf{Metrics}} & \textbf{en-vi} & \textbf{en-fr} & \textbf{en-zh} & \multicolumn{1}{c|}{\textbf{en-de}} & \textbf{vi-en} & \textbf{vi-fr} & \textbf{vi-zh} & \multicolumn{1}{c|}{\textbf{vi-de}} & \textbf{fr-en} & \textbf{fr-vi} & \textbf{fr-zh} & \multicolumn{1}{c|}{\textbf{fr-de}} & \textbf{de-en} & \textbf{de-vi} & \textbf{de-fr} & \multicolumn{1}{c|}{\textbf{de-zh}} & \textbf{zh-en} & \textbf{zh-vi} & \textbf{zh-fr} & \textbf{zh-de} \\ \hline
\multicolumn{22}{c}{\colorbox{custom_light_blue}{Cascaded}} \\ \hline
\multicolumn{1}{l|}{\multirow{2}{*}{\begin{tabular}[c]{@{}l@{}}Llama\\ -3.1-8B\end{tabular}}} & \multicolumn{1}{l|}{BLEU} & 43.32 & 37.92 & 30.78 & \multicolumn{1}{c|}{31.36} & 14.55 & 10.29 & 11.56 & \multicolumn{1}{c|}{7.71} & 30.15 & 25.36 & 20.28 & \multicolumn{1}{c|}{16.38} & 40.63 & 33.63 & 26.97 & \multicolumn{1}{c|}{26.31} & 19.01 & 17.65 & 13.84 & 11.13 \\
\multicolumn{1}{l|}{} & \multicolumn{1}{l|}{BERTSc} & 0.85 & 0.84 & 0.8 & \multicolumn{1}{c|}{0.83} & 0.78 & 0.75 & 0.73 & \multicolumn{1}{c|}{0.73} & 0.82 & 0.80 & 0.75 & \multicolumn{1}{c|}{0.74} & 0.86 & 0.84 & 0.80 & \multicolumn{1}{c|}{0.79} & 0.79 & 0.85 & 0.76 & 0.74 \\ \hline
\multicolumn{1}{l|}{\multirow{2}{*}{\begin{tabular}[c]{@{}l@{}}Qwen\\ -2.5-7B\end{tabular}}} & \multicolumn{1}{l|}{BLEU} & 43.37 & 37.34 & 23.46 & \multicolumn{1}{c|}{28.5} & 13.97 & 11.66 & 20.27 & \multicolumn{1}{c|}{8.75} & \textcolor{gtred}{30.35} & \textcolor{gtred}{25.59} & 15.33 & \multicolumn{1}{c|}{20.38} & 40.52 & 34.24 & 31.45 & \multicolumn{1}{c|}{19.87} & 25.36 & 26.31 & 17.84 & 12.61 \\
\multicolumn{1}{l|}{} & \multicolumn{1}{l|}{BERTSc} & 0.85 & 0.85 & 0.8 & \multicolumn{1}{c|}{0.82} & 0.78 & 0.76 & 0.78 & \multicolumn{1}{c|}{0.75} & 0.81 & 0.80 & 0.76 & \multicolumn{1}{c|}{0.78} & 0.86 & 0.84 & 0.84 & \multicolumn{1}{c|}{0.79} & 0.82 & 0.90 & 0.80 & 0.78 \\ \hline
\multicolumn{1}{l|}{\multirow{2}{*}{\begin{tabular}[c]{@{}l@{}}Mistral\\ -v0.3-7B\end{tabular}}} & \multicolumn{1}{l|}{BLEU} & 17.72 & 36.58 & 20.27 & \multicolumn{1}{c|}{29.9} & 15.86 & 10.92 & 17.92 & \multicolumn{1}{c|}{9.03} & 29.35 & 9.20 & 13.94 & \multicolumn{1}{c|}{18.65} & 28.33 & 12.38 & 31.15 & \multicolumn{1}{c|}{17.82} & 20.17 & 8.01 & 12.58 & 7.14 \\
\multicolumn{1}{l|}{} & \multicolumn{1}{l|}{BERTSc} & 0.77 & 0.83 & 0.77 & \multicolumn{1}{c|}{0.81} & 0.78 & 0.75 & 0.77 & \multicolumn{1}{c|}{0.75} & 0.79 & 0.74 & 0.76 & \multicolumn{1}{c|}{0.78} & 0.78 & 0.77 & 0.83 & \multicolumn{1}{c|}{0.78} & 0.81 & 0.86 & 0.78 & 0.72 \\ \hline
\multicolumn{1}{l|}{\multirow{2}{*}{\begin{tabular}[c]{@{}l@{}}mBart\\ -large-50\end{tabular}}} & \multicolumn{1}{l|}{BLEU} & 48.00 & 43.20 & 35.70 & \multicolumn{1}{c|}{35.07} & 10.17 & 12.80 & 16.77 & \multicolumn{1}{c|}{7.23} & 23.82 & 22.86 & 16.46 & \multicolumn{1}{c|}{17.39} & 31.95 & 32.62 & 31.96 & \multicolumn{1}{c|}{25.07} & 11.88 & 18.40 & 12.30 & 9.64 \\
\multicolumn{1}{l|}{} & \multicolumn{1}{l|}{BERTSc} & 0.87 & 0.86 & 0.81 & \multicolumn{1}{c|}{0.84} & 0.88 & 0.76 & 0.73 & \multicolumn{1}{c|}{0.72} & 0.90 & 0.80 & 0.72 & \multicolumn{1}{c|}{0.77} & 0.92 & 0.84 & 0.84 & \multicolumn{1}{c|}{0.77} & 0.89 & 0.79 & 0.76 & 0.75 \\ \hline
\multicolumn{1}{l|}{\multirow{2}{*}{\begin{tabular}[c]{@{}l@{}}M2M100\\ -418M\end{tabular}}} & \multicolumn{1}{l|}{BLEU} & \textcolor{gtred}{48.21} & 43.16 & 36.94 & \multicolumn{1}{c|}{\textcolor{gtred}{36.55}} & 15.64 & 13.95 & 16.99 & \multicolumn{1}{c|}{11.10} & 25.65 & 21.88 & 18.44 & \multicolumn{1}{c|}{19.98} & 33.66 & \textcolor{gtred}{34.70} & \textcolor{gtred}{34.67} & \multicolumn{1}{c|}{24.31} & 16.65 & 21.83 & 16.94 & 13.06 \\
\multicolumn{1}{l|}{} & \multicolumn{1}{l|}{BERTSc} & 0.95 & 0.92 & 0.92 & \multicolumn{1}{c|}{0.92} & 0.78 & 0.77 & 0.74 & \multicolumn{1}{c|}{0.75} & 0.81 & 0.76 & 0.75 & \multicolumn{1}{c|}{0.76} & 0.77 & 0.81 & 0.85 & \multicolumn{1}{c|}{0.72} & 0.76 & 0.83 & 0.79 & 0.78 \\ \hline
\multicolumn{1}{l|}{\multirow{2}{*}{Marian}} & \multicolumn{1}{l|}{BLEU} & 45.07 & 40.54 & 31.17 & \multicolumn{1}{c|}{33.90} & 12.95 & 11.23 & 12.09 & \multicolumn{1}{c|}{09.08} & 24.03 & 22.20 & 11.27 & \multicolumn{1}{c|}{19.14} & 34.09 & 29.72 & 30.48 & \multicolumn{1}{c|}{14.79} & 8.50 & 13.37 & 8.39 & 5.76 \\
\multicolumn{1}{l|}{} & \multicolumn{1}{l|}{BERTSc} & 0.87 & 0.86 & 0.82 & \multicolumn{1}{c|}{0.84} & 0.77 & 0.76 & 0.75 & \multicolumn{1}{c|}{0.74} & 0.81 & 0.80 & 0.74 & \multicolumn{1}{c|}{0.79} & 0.85 & 0.83 & 0.84 & \multicolumn{1}{c|}{0.76} & 0.75 & 0.77 & 0.74 & 0.73 \\ \hline
\multicolumn{1}{l|}{\multirow{2}{*}{\begin{tabular}[c]{@{}l@{}}Google\\ Translate\end{tabular}}} & \multicolumn{1}{l|}{BLEU} 
& 46.21
& \textcolor{gtred}{44.77} 
& \textcolor{gtred}{44.74} 
& \multicolumn{1}{c|}{36.29}
& \textcolor{gtred}{18.79} 
& \textcolor{gtred}{16.42} 
& \textcolor{gtred}{21.63} 
& \multicolumn{1}{c|}{\textcolor{gtred}{12.54}} 
& 27.82
& 24.18 
& \textcolor{gtred}{24.49} 
& \multicolumn{1}{c|}{\textcolor{gtred}{22.38}} 
& \textcolor{gtred}{40.74} 
& 32.69
& 33.15 
& \multicolumn{1}{c|}{\textcolor{gtred}{31.89}} 
& \textcolor{gtred}{27.74} 
& \textcolor{gtred}{30.70} 
& \textcolor{gtred}{20.71} 
& \textcolor{gtred}{19.11} \\
\multicolumn{1}{l|}{} & \multicolumn{1}{l|}{BERTSc} & 0.86 & 0.86 & 0.85 & \multicolumn{1}{c|}{0.84} & 0.78 & 0.78 & 0.78 & \multicolumn{1}{c|}{0.76} & 0.81 & 0.80 & 0.79 & \multicolumn{1}{c|}{0.79} & 0.86 & 0.85 & 0.84 & \multicolumn{1}{c|}{0.83} & 0.83 & 0.82 & 0.80 & 0.81 \\ \hline
\multicolumn{22}{c}{\colorbox{custom_light_blue}{End-to-end}} \\ \hline
\multicolumn{1}{l|}{\multirow{2}{*}{\begin{tabular}[c]{@{}l@{}}SeamlessM4T\\ -large-v2\end{tabular}}} & \multicolumn{1}{l|}{BLEU} & 24,59 & 25,68 & 20,43 & \multicolumn{1}{c|}{20,19} & 14,4 & 10,19 & 11,49 & \multicolumn{1}{c|}{7,4} & 29,23 & 17,49 & 11,37 & \multicolumn{1}{c|}{15,94} & 25,09 & 15,07 & 12,88 & \multicolumn{1}{c|}{11,45} & 14,22 & 11,39 & 6,83 & 4,16 \\
\multicolumn{1}{l|}{} & \multicolumn{1}{l|}{BERTSc} & 0,81 & 0,82 & 0,76 & \multicolumn{1}{c|}{0,8} & 0,77 & 0,75 & 0,74 & \multicolumn{1}{c|}{0,72} & 0,82 & 0,78 & 0,72 & \multicolumn{1}{c|}{0,77} & 0,82 & 0,77 & 0,75 & \multicolumn{1}{c|}{0,73} & 0.79 & 0,74 & 0,73 & 0,70 \\ \hline
\multicolumn{1}{l|}{\multirow{2}{*}{\begin{tabular}[c]{@{}l@{}}QwenAudio-2\\ -7B-Instruct\end{tabular}}} & \multicolumn{1}{l|}{BLEU} & 24,46 & 30,16 & 23,3 & \multicolumn{1}{c|}{22,69} & 1,66 & 1,17 & 2,36 & \multicolumn{1}{c|}{1,13} & 23,63 & 11,49 & 15,37 & \multicolumn{1}{c|}{14,51} & 23,29 & 11,07 & 14,88 & \multicolumn{1}{c|}{16,04} & 19,63 & 15,72 & 13,52 & 10.37 \\
\multicolumn{1}{l|}{} & \multicolumn{1}{l|}{BERTSc} & 0,8 & 0,82 & 0,76 & \multicolumn{1}{c|}{0,79} & 0,66 & 0,65 & 0,65 & \multicolumn{1}{c|}{0,66} & 0,79 & 0,74 & 0,71 & \multicolumn{1}{c|}{0,74} & 0,8 & 0,73 & 0,76 & \multicolumn{1}{c|}{0,72} & 0,8 & 0,78 & 0,77 & 0,77 \\ \hline
\multicolumn{1}{l|}{\multirow{2}{*}{Whisper}} & \multicolumn{1}{l|}{BLEU} &  &  &  & \multicolumn{1}{c|}{} & 8.18 &  &  & \multicolumn{1}{c|}{} & 26.06 &  &  & \multicolumn{1}{c|}{} & 37.32 &  &  & \multicolumn{1}{c|}{} & 16.54 &  &  &  \\
\multicolumn{1}{l|}{} & \multicolumn{1}{l|}{BERTSc} &  &  &  & \multicolumn{1}{c|}{} & 0.75 &  &  & \multicolumn{1}{c|}{} & 0.81 &  &  & \multicolumn{1}{c|}{} & 0.85 &  &  & \multicolumn{1}{c|}{} & 0.79 &  &  &  \\ \hline
\end{tabular}%
}
\caption{\textbf{\highlight{End-to-end and cascaded comparison.}} All cascaded models use Whisper$_{small-mono}$ as \gls{ASR} model (Whisper \gls{ASR} is fine-tuned monolingually - on each source language separately), then MT models translate into target languages. End-to-end Whisper for \gls{ST} is fine-tuned bilingually - on each language pair separately. End-to-end Whisper \gls{ST} only supports X to English, thus no results for other translation directions were reported. \\
\textcolor{red}{\faChartLine} Cascaded models significantly outperform end-
to-end models.}
\label{tab:translation-asr}
\end{table*}

%% file: tables/nmt_multilingual_asr.tex
\begin{table*}[t]
\centering
\tiny
\setlength{\tabcolsep}{2pt}
\renewcommand{\arraystretch}{1.3}
\resizebox{\textwidth}{!}{%
\begin{tabular}{llcccccccccccccccccccc}
\hline
\multicolumn{1}{l|}{\textbf{Model}} & \multicolumn{1}{l|}{\textbf{Metrics}} & \textbf{en-vi} & \textbf{en-fr} & \textbf{en-zh} & \multicolumn{1}{c|}{\textbf{en-de}} & \textbf{vi-en} & \textbf{vi-fr} & \textbf{vi-zh} & \multicolumn{1}{c|}{\textbf{vi-de}} & \textbf{fr-en} & \textbf{fr-vi} & \textbf{fr-zh} & \multicolumn{1}{c|}{\textbf{fr-de}} & \textbf{de-en} & \textbf{de-vi} & \textbf{de-fr} & \multicolumn{1}{c|}{\textbf{de-zh}} & \textbf{zh-en} & \textbf{zh-vi} & \textbf{zh-fr} & \textbf{zh-de} \\ \hline
\multicolumn{22}{c}{\colorbox{custom_light_blue}{Multilingual MT fine-tuning}} \\ \hline
\multicolumn{1}{l|}{\multirow{2}{*}{\begin{tabular}[c]{@{}l@{}}Llama\\ -3.1-8B\end{tabular}}} & \multicolumn{1}{l|}{BLEU} & 41.79 & 36.14 & 32.71 & \multicolumn{1}{c|}{28.19} & 15.41 & 10.71 & 19.55 & \multicolumn{1}{c|}{8.33} & 27.47 & 21.63 & 18.05 & \multicolumn{1}{c|}{17.40} & 36.47 & 27.5 & 27.06 & \multicolumn{1}{c|}{25.05} & 20.48 & 21.52 & 15.37 & 10.64 \\
\multicolumn{1}{l|}{} & \multicolumn{1}{l|}{BERTSc} & 0.85 & 0.84 & 0.82 & \multicolumn{1}{c|}{0.82} & 0.78 & 0.76 & 0.78 & \multicolumn{1}{c|}{0.74} & 0.81 & 0.79 & 0.77 & \multicolumn{1}{c|}{0.78} & 0.85 & 0.82 & 0.83 & \multicolumn{1}{c|}{0.80} & 0.80 & 0.79 & 0.77 & 0.77 \\ \hline
\multicolumn{1}{l|}{\multirow{2}{*}{\begin{tabular}[c]{@{}l@{}}Qwen\\ -2.5-7B\end{tabular}}} & \multicolumn{1}{l|}{BLEU} & 41.71 & 36.39 & \textcolor{gtred}{32.78} & \multicolumn{1}{c|}{27.89} & 15.11 & 10.55 & 19.58 & \multicolumn{1}{c|}{7.80} & 27.56 & 22.09 & 19.06 & \multicolumn{1}{c|}{17.69} & 36.05 & 26.27 & 27.36 & \multicolumn{1}{c|}{25.11} & 20.62 & 21.37 & 15.51 & 10.47 \\
\multicolumn{1}{l|}{} & \multicolumn{1}{l|}{BERTSc} & 0.85 & 0.84 & 0.82 & \multicolumn{1}{c|}{0.82} & 0.78 & 0.76 & 0.78 & \multicolumn{1}{c|}{0.74} & 0.81 & 0.79 & 0.77 & \multicolumn{1}{c|}{0.78} & 0.85 & 0.82 & 0.83 & \multicolumn{1}{c|}{0.80} & 0.80 & 0.79 & 0.78 & 0.76 \\ \hline
\multicolumn{1}{l|}{\multirow{2}{*}{\begin{tabular}[c]{@{}l@{}}Mistral\\ -v0.3-7B\end{tabular}}} & \multicolumn{1}{l|}{BLEU} & 19.09 & 35.89 & 20.22 & \multicolumn{1}{c|}{28.83} & 15.4 & 10.7 & 16.83 & \multicolumn{1}{c|}{8.61} & 27.95 & 9.83 & 13.59 & \multicolumn{1}{c|}{16.18} & 37.82 & 11.42 & 21.13 & \multicolumn{1}{c|}{15.37} & 21.07 & 9.21 & 13.02 & 9.14 \\
\multicolumn{1}{l|}{} & \multicolumn{1}{l|}{BERTSc} & 0.8 & 0.84 & 0.79 & \multicolumn{1}{c|}{0.83} & 0.78 & 0.75 & 0.77 & \multicolumn{1}{c|}{0.74} & 0.82 & 0.75 & 0.76 & \multicolumn{1}{c|}{0.78} & 0.86 & 0.77 & 0.81 & \multicolumn{1}{c|}{0.72} & 0.81 & 0.73 & 0.76 & 0.76 \\ \hline
\multicolumn{22}{c}{\colorbox{custom_light_blue}{Bilingual MT fine-tuning}} \\ \hline
\multicolumn{1}{l|}{\multirow{2}{*}{\begin{tabular}[c]{@{}l@{}}Llama\\ -3.1-8B\end{tabular}}} & \multicolumn{1}{l|}{BLEU} & 43.32 & \textcolor{gtred}{37.92} & 30.78 & \multicolumn{1}{c|}{\textcolor{gtred}{31.36}} & 14.55 & 10.29 & 11.56 & \multicolumn{1}{c|}{7.71} & 30.15 & 25.36 & \textcolor{gtred}{20.28} & \multicolumn{1}{c|}{16.38} & \textcolor{gtred}{40.63} & 33.63 & 26.97 & \multicolumn{1}{c|}{\textcolor{gtred}{26.31}} & 19.01 & 17.65 & 13.84 & 11.13 \\
\multicolumn{1}{l|}{} & \multicolumn{1}{l|}{BERTSc} & 0.85 & 0.84 & 0.8 & \multicolumn{1}{c|}{0.83} & 0.78 & 0.75 & 0.73 & \multicolumn{1}{c|}{0.73} & 0.82 & 0.80 & 0.75 & \multicolumn{1}{c|}{0.74} & 0.86 & 0.84 & 0.80 & \multicolumn{1}{c|}{0.79} & 0.79 & 0.85 & 0.76 & 0.74 \\ \hline
\multicolumn{1}{l|}{\multirow{2}{*}{\begin{tabular}[c]{@{}l@{}}Qwen\\ -2.5-7B\end{tabular}}} & \multicolumn{1}{l|}{BLEU} & \textcolor{gtred}{43.37} & 37.34 & 23.46 & \multicolumn{1}{c|}{28.5} & 13.97 & \textcolor{gtred}{11.66} & \textcolor{gtred}{20.27} & \multicolumn{1}{c|}{8.75} & \textcolor{gtred}{30.35} & \textcolor{gtred}{25.59} & 15.33 & \multicolumn{1}{c|}{\textcolor{gtred}{20.38}} & 40.52 & \textcolor{gtred}{34.24} & \textcolor{gtred}{31.45} & \multicolumn{1}{c|}{19.87} & \textcolor{gtred}{25.36} & \textcolor{gtred}{26.31} & \textcolor{gtred}{17.84} & \textcolor{gtred}{12.61} \\
\multicolumn{1}{l|}{} & \multicolumn{1}{l|}{BERTSc} & 0.85 & 0.85 & 0.8 & \multicolumn{1}{c|}{0.82} & 0.78 & 0.76 & 0.78 & \multicolumn{1}{c|}{0.75} & 0.81 & 0.80 & 0.76 & \multicolumn{1}{c|}{0.78} & 0.86 & 0.84 & 0.84 & \multicolumn{1}{c|}{0.79} & 0.82 & 0.90 & 0.80 & 0.78 \\ \hline
\multicolumn{1}{l|}{\multirow{2}{*}{\begin{tabular}[c]{@{}l@{}}Mistral\\ -v0.3-7B\end{tabular}}} & \multicolumn{1}{l|}{BLEU} & 17.72 & 36.58 & 20.27 & \multicolumn{1}{c|}{29.9} & \textcolor{gtred}{15.86} & 10.92 & 17.92 & \multicolumn{1}{c|}{\textcolor{gtred}{9.03}} & 29.35 & 9.20 & 13.94 & \multicolumn{1}{c|}{18.65} & 28.33 & 12.38 & 31.15 & \multicolumn{1}{c|}{17.82} & 20.17 & 8.01 & 12.58 & 7.14 \\
\multicolumn{1}{l|}{} & \multicolumn{1}{l|}{BERTSc} & 0.77 & 0.83 & 0.77 & \multicolumn{1}{c|}{0.81} & 0.78 & 0.75 & 0.77 & \multicolumn{1}{c|}{0.75} & 0.79 & 0.74 & 0.76 & \multicolumn{1}{c|}{0.78} & 0.78 & 0.77 & 0.83 & \multicolumn{1}{c|}{0.78} & 0.81 & 0.86 & 0.78 & 0.72 \\ \hline
\end{tabular}%
}
\caption{\textbf{\highlight{Bilingual-multilingual fine-tuning comparison.}} All \gls{ST} results are from cascaded \gls{ST} models with \gls{ASR} transcript generated by Whisper Small fine-tuned monolingually on source language. \\
\textcolor{red}{\faChartLine} Overall, Bilingual fine-tuning outperforms multilingual \gls{MT} fine-tuning.}
\label{tab:translation-multilingual}
\end{table*}

%% file: tables/bilingual_translation.tex
\begin{table}[t]
\centering
\tiny
\resizebox{\columnwidth}{!}{%
\begin{tabular}{llcccc}
\hline
\multicolumn{1}{l|}{\multirow{2}{*}{\textbf{MT}}} & \multicolumn{1}{l|}{\multirow{2}{*}{\textbf{Metrics}}} & \multicolumn{2}{c|}{\textbf{Ground-truth}} & \multicolumn{2}{c}{\textbf{ASR}} \\ \cline{3-6} 
\multicolumn{1}{l|}{} & \multicolumn{1}{l|}{} & \textbf{en-vi} & \multicolumn{1}{c|}{\textbf{vi-en}} & \textbf{en-vi} & \textbf{vi-en} \\ \hline
\multicolumn{6}{c}{\colorbox{custom_light_blue}{Bilingual pre-trained MT}} \\ \hline
\multicolumn{1}{l|}{\multirow{2}{*}{VinAI}} & \multicolumn{1}{l|}{BLEU} & \textcolor{gtred}{65.85} & \multicolumn{1}{c|}{\textcolor{gtred}{28.55}} & \textcolor{gtred}{50.79} & 15.46 \\
\multicolumn{1}{l|}{} & \multicolumn{1}{l|}{BERTSc} & 0.93 & \multicolumn{1}{c|}{0.84} & 0.88 & 0.77 \\ \hline
\multicolumn{1}{l|}{\multirow{2}{*}{EnViT5}} & \multicolumn{1}{l|}{BLEU} & 20.72 & \multicolumn{1}{c|}{23.46} & 17.26 & 15.16 \\
\multicolumn{1}{l|}{} & \multicolumn{1}{l|}{BERTSc} & 0.83 & \multicolumn{1}{c|}{0.82} & 0.80 & 0.78 \\ \hline
\multicolumn{6}{c}{\colorbox{custom_light_blue}{Multilingual pre-trained MT}} \\ \hline
\multicolumn{1}{l|}{\multirow{2}{*}{\begin{tabular}[c]{@{}l@{}}mBart\\ -large-50\end{tabular}}} & \multicolumn{1}{l|}{BLEU} & 59.73 & \multicolumn{1}{c|}{16.48} & 48.00 & 10.17 \\
\multicolumn{1}{l|}{} & \multicolumn{1}{l|}{BERTSc} & 0.92 & \multicolumn{1}{c|}{\textcolor{gtred}{0.89}} & 0.87 & \textcolor{gtred}{0.88} \\ \hline
\multicolumn{1}{l|}{\multirow{2}{*}{\begin{tabular}[c]{@{}l@{}}M2M100\\ -418M\end{tabular}}} & \multicolumn{1}{l|}{BLEU} & 62.31 & \multicolumn{1}{c|}{23.01} & 48.21 & \textcolor{gtred}{15.64} \\
\multicolumn{1}{l|}{} & \multicolumn{1}{l|}{BERTSc} & \textcolor{gtred}{0.97} & \multicolumn{1}{c|}{0.82} & \textcolor{gtred}{0.95} & 0.78 \\ \hline
\end{tabular}%
}
\caption{\textbf{\highlight{Bilingual-multilingual pre-training}} \textbf{\highlight{comparison.}} All \gls{ST} results are from cascaded \gls{ST} models with \gls{ASR} transcript generated by Whisper Small fine-tuned monolingually on source language. \\
\textcolor{red}{\faChartLine} Multilingual \gls{MT} models perform on par with bilingual ones.}
\label{tab:bilingual-result}
\end{table}

%% file: tables/code-switch.tex
\begin{table}[h]
\centering
\tiny
\setlength{\tabcolsep}{2pt}
\renewcommand{\arraystretch}{1.2}
\resizebox{\columnwidth}{!}{%
\begin{tabular}{llcccccccc}
\hline
\multicolumn{1}{l|}{\multirow{2}{*}{\textbf{MT}}} & \multicolumn{1}{l|}{\multirow{2}{*}{\textbf{Metrics}}} & \multicolumn{4}{c|}{\textbf{Ground-truth}} & \multicolumn{4}{c}{\textbf{ASR}} \\ \cline{3-10} 
\multicolumn{1}{l|}{} & \multicolumn{1}{l|}{} & \multicolumn{1}{r}{\textbf{en-vi}} & \multicolumn{1}{r}{\textbf{en-fr}} & \multicolumn{1}{r}{\textbf{en-zh}} & \multicolumn{1}{r|}{\textbf{en-de}} & \multicolumn{1}{r}{\textbf{en-vi}} & \multicolumn{1}{r}{\textbf{en-fr}} & \multicolumn{1}{r}{\textbf{en-zh}} & \multicolumn{1}{r}{\textbf{en-de}} \\ \hline
\multicolumn{10}{c}{\colorbox{custom_light_blue}{Decoder}} \\ \hline
\multicolumn{1}{l|}{\multirow{2}{*}{\begin{tabular}[c]{@{}l@{}}Llama\\ -3.1-8B\end{tabular}}} & \multicolumn{1}{l|}{BLEU} & 51.92 & 51.12 & 39.42 & \multicolumn{1}{c|}{39.96} & 41.68 & 38.21 & 33.02 & 30.49 \\
\multicolumn{1}{l|}{} & \multicolumn{1}{l|}{BERTSc} & 0.90 & 0.90 & 0.83 & \multicolumn{1}{c|}{0.87} & 0.85 & 0.85 & 0.80 & 0.82 \\ \hline
\multicolumn{1}{l|}{\multirow{2}{*}{\begin{tabular}[c]{@{}l@{}}Qwen\\ -2.5-7B\end{tabular}}} & \multicolumn{1}{l|}{BLEU} & 51.60 & 50.00 & 29.62 & \multicolumn{1}{c|}{37.39} & 41.81 & 36.13 & 24.18 & 27.18 \\
\multicolumn{1}{l|}{} & \multicolumn{1}{l|}{BERTSc} & 0.90 & 0.90 & 0.82 & \multicolumn{1}{c|}{0.87} & 0.85 & 0.85 & 0.80 & 0.82 \\ \hline
\multicolumn{1}{l|}{\multirow{2}{*}{\begin{tabular}[c]{@{}l@{}}Mistral\\ -v0.3-7B\end{tabular}}} & \multicolumn{1}{l|}{BLEU} & 26.31 & 52.74 & 25.48 & \multicolumn{1}{c|}{44.75} & 18.51 & 37.80 & 19.11 & 30.99 \\
\multicolumn{1}{l|}{} & \multicolumn{1}{l|}{BERTSc} & 0.83 & 0.90 & 0.81 & \multicolumn{1}{c|}{0.88} & 0.78 & 0.85 & 0.76 & 0.82 \\ \hline
\multicolumn{10}{c}{\colorbox{custom_light_blue}{Encoder-decoder}} \\ \hline
\multicolumn{1}{l|}{\multirow{2}{*}{\begin{tabular}[c]{@{}l@{}}mBart\\ -large-50\end{tabular}}} & \multicolumn{1}{l|}{BLEU} & 60.69 & 56.47 & 49.20 & \multicolumn{1}{c|}{45.67} & 46.20 & 40.91 & 38.02 & 33.78 \\
\multicolumn{1}{l|}{} & \multicolumn{1}{l|}{BERTSc} & 0.92 & 0.92 & 0.88 & \multicolumn{1}{c|}{0.89} & 0.87 & 0.86 & 0.84 & 0.84 \\ \hline
\multicolumn{1}{l|}{\multirow{2}{*}{\begin{tabular}[c]{@{}l@{}}M2M100\\ -418M\end{tabular}}} & \multicolumn{1}{l|}{BLEU} & 61.11 & 57.07 & 52.06 & \multicolumn{1}{c|}{48.75} & 46.26 & 41.91 & 39.70 & 35.95 \\
\multicolumn{1}{l|}{} & \multicolumn{1}{l|}{BERTSc} & 0.92 & 0.92 & 0.88 & \multicolumn{1}{c|}{0.90} & 0.87 & 0.86 & 0.84 & 0.85 \\ \hline
\multicolumn{1}{l|}{\multirow{2}{*}{Marian}} & \multicolumn{1}{l|}{BLEU} & 56.70 & 53.20 & 43.00 & \multicolumn{1}{c|}{43.86} & 43.54 & 38.59 & 33.42 & 32.48 \\
\multicolumn{1}{l|}{} & \multicolumn{1}{l|}{BERTSc} & 0.91 & 0.91 & 0.86 & \multicolumn{1}{c|}{0.89} & 0.87 & 0.86 & 0.82 & 0.84 \\ \hline
\end{tabular}%
}
\caption{\textbf{\highlight{Code-switch analysis.}} All \gls{ST} results are from cascaded \gls{ST} models with \gls{ASR} transcript generated by Whisper Small fine-tuned monolingually on source language. The original dataset shows code-switching percentages of 11.2\%, 7\%, 7.9\%, and 12.8\% for Vietnamese, French, Chinese, and German, respectively.}
\label{tab:codeswitch-result}
\end{table}

%% file: tables/human_and_auto_score.tex
\begin{table*}[t]
\centering
\tiny
\setlength{\tabcolsep}{1pt}
\renewcommand{\arraystretch}{1.1}
\resizebox{\textwidth}{!}{%
\begin{tabular}{l|l|cccc|cccc|cccc|cccc|cccc}
\hline
\textbf{Model} & \textbf{Metrics} & \textbf{en-vi} & \textbf{en-fr} & \textbf{en-zh} & \textbf{en-de} & \textbf{vi-en} & \textbf{vi-fr} & \textbf{vi-zh} & \textbf{vi-de} & \textbf{fr-en} & \textbf{fr-vi} & \textbf{fr-zh} & \textbf{fr-de} & \textbf{de-en} & \textbf{de-vi} & \textbf{de-fr} & \textbf{de-zh} & \textbf{zh-en} & \textbf{zh-vi} & \textbf{zh-fr} & \textbf{zh-de} \\ \hline
\multirow{4}{*}{\begin{tabular}[c]{@{}l@{}}Llama\\ -3.1-8B\end{tabular}} & BLEU & 41.79 & 36.14 & 32.71 & 28.19 & 15.41 & 10.71 & 19.55 & 8.33 & 27.47 & 21.63 & 18.05 & 17.40 & 36.47 & 27.50& 27.06 & 25.05 & 20.48 & 21.52 & 15.37 & 10.64 \\
 & BERTSc & 0.85 & 0.84 & 0.82 & 0.82 & 0.78 & 0.76 & 0.78 & 0.74 & 0.81 & 0.79 & 0.77 & 0.78 & 0.85 & 0.82 & 0.83 & 0.80 & 0.80 & 0.79 & 0.77 & 0.77 \\
 & LLM-judge & 5.14 & 4.64 & 4.45 & 4.63 & 3.88 & 3.49 & 3.15 & 3.41 & 4.38 & 4.01 & 3.43 & 3.44 & 5.81 & 5.39 & 4.52 & 4.06 & 3.88 & 3.61 & 3.78 & 3.69\\ 
 & Human & 6.85 & 6.47 & 4.31 & 8.53 & 6.54 & 5.64 & 4.12 & 7.24 & 5.19 & 5.45 & 4.04 & 6.42 & 6.15 & 8.05 & 6.64 & 4.14 & 4.08 & 3.58 & 5.64 & 6.54\\ \hline
\multirow{4}{*}{\begin{tabular}[c]{@{}l@{}}Qwen\\ -2.5-7B\end{tabular}} & BLEU & 41.71 & 36.39 & 32.78 & 27.89 & 15.11 & 10.55 & 19.58 & 7.80 & 27.56 & 22.09 & 19.06 & 17.69 & 36.05 & 26.27 & 27.36 & 25.11 & 20.62 & 21.37 & 15.51 & 10.47 \\
 & BERTSc & 0.85 & 0.84 & 0.82 & 0.82 & 0.78 & 0.76 & 0.78 & 0.74 & 0.81 & 0.79 & 0.77 & 0.78 & 0.85 & 0.82 & 0.83 & 0.80 & 0.80 & 0.79 & 0.78 & 0.76 \\
 & LLM-judge & 4.93 & 4.91 & 3.46 & 4.52 & 4.04 & 3.91 & 4.05 & 3.48 & 4.39 & 4.10& 3.62 & 3.96 & 6.17 & 5.51 & 5.52 & 3.99 & 4.97 & 4.50& 4.67 & 4.36 \\
 & Human & 7.97 & 6.56 & 4.41 & 8.55 & 6.72 & 5.73 & 4.17 & 7.42 & 7.50& 5.51 & 4.06 & 7.39 & 8.30& 7.71 & 6.57 & 4.22 & 7.69 & 5.09 & 5.86 & 7.58 \\ \hline
\multirow{4}{*}{\begin{tabular}[c]{@{}l@{}}Mistral\\ -v0.3-7B\end{tabular}} & BLEU & 19.09 & 35.89 & 20.22 & 28.83 & 15.40& 10.70& 16.83 & 8.61 & 27.95 & 9.83 & 13.59 & 16.18 & 37.82 & 11.42 & 21.13 & 15.37 & 21.07 & 9.21 & 13.02 & 9.14 \\
 & BERTSc & 0.80& 0.84 & 0.79 & 0.83 & 0.78 & 0.75 & 0.77 & 0.74 & 0.82 & 0.75 & 0.76 & 0.78 & 0.86 & 0.77 & 0.81 & 0.72 & 0.81 & 0.73 & 0.76 & 0.76 \\
 & LLM-judge & 2.40& 4.54 & 3.20& 4.50& 3.57 & 3.18 & 3.41 & 3.09 & 4.49 & 2.30& 3.56 & 3.86 & 4.91 & 2.73 & 5.02 & 3.53 & 4.46 & 2.0 & 3.97 & 2.94 \\
 & Human & 6.36 & 6.52 & 3.43 & 6.19 & 6.00 & 5.73 & 5.08 & 4.74 & 7.55 & 5.51 & 3.69 & 4.64 & 7.78 & 4.07 & 6.57 & 3.91 & 7.33 & 2.53 & 6.14 & 5.20\\ \hline
\end{tabular}%
}
\caption{\textbf{\highlight{LLM-as-a-judge and human evaluation results.}} All \gls{ST} results are from cascaded \gls{ST} models with \gls{ASR} transcript generated
by Whisper Small fine-tuned monolingually on source language. A BERTScore of $>0.8$ is often seen as good translation quality. while $>0.9$ is excellent translation quality. \\
\textcolor{red}{\faChartLine} Automatic metrics (BLEU, BERTScore) strongly correlate with both \gls{LLM}-as-a-judge and human evaluations across most language pairs.}
\label{tab:human_auto_score}
\end{table*}

%% file: tables/data_stats_ASR_literaturecompare.tex
\begin{table*}[h]
\centering
\begin{adjustbox}{width=\textwidth}
\begin{tabular}{l|ccccccc}
\hline
\textbf{Dataset} & \textbf{Dur.} & \textbf{Language} & \textbf{Nature} & \textbf{\#Rec. Cond.} & \textbf{\#Spk} & \textbf{\#Acc} & \textbf{\#Roles} \\
\hline
VietMed \cite{vietmed_dataset} & 16h & Vietnamese & Real-world & 8 & 61 & 6 & 6 \\
PriMock57$^2$ \cite{korfiatis2022primock57} & 9h & English & Simulated & 1 & 64 & 4 & 2 \\
\citet{fareez2022dataset}$^3$ & 55h & English & Simulated & 1 & N/A & 1 & 2 \\
AfriSpeech-200$^4$ \cite{olatunji2023afrispeech} & $\approx$123h & African English & Read speech & 1 & N/A & N/A & 1 \\
myMediCon$^5$ \cite{htun2024mymedicon} & 11h & Burmese & Read speech & 1 & 12 & 5 & 2 \\ \hline
\textbf{\MultiMedST$^1$ (ours)} & \textbf{150h} & \textbf{Multiling.} & \textbf{Real-world} & \textbf{10} & \textbf{198} & \textbf{16} & \textbf{6}\\ \hline
\end{tabular}%
\end{adjustbox}
\caption{\textbf{Dataset comparison with literature: A list of all publicly available medical \gls{ASR} datasets.} \\ \textbf{Our \MultiMedST is the largest and most diverse medical \gls{ASR} dataset.}\\ From left to right: Total duration in hours (h), language, nature of speech, number of recording conditions, number of speakers, number of accents, speaking roles.\\
$^1$In our dataset, only the number of recording conditions, speakers, accents and speaking roles for Vietnamese and English are identified because of technical and privacy issues. Therefore, the exact number of speakers and accents must be much larger than the currently reported number. 10 recording conditions include: Documentary, Interview, Lecture, News, Podcast, Webinar, Speech, Talk, Vlog, Workshop. 10 English accents include: Main US, Southern US, UK, Australian, Indian, Mexican, European, Japanese, Uzbekistan, Russian. 6 Vietnamese accents include: North, South Central Coast, South East, South West, Central Highland, North Central Coast.\\
$^2$Speech collected by simulated medical conversations between 2 speaking roles - clinicians and actors/actresses. 4 English accents include:  British English, European, other English, and other non-English.\\
$^3$Speech was recorded as patient-physician interviews (counted as 1 recording condition and 2 speaking roles) by West England speakers (counted as 1 accent)\\
$^4$AfriSpeech-200 dataset is a mix of general-domain and medical-domain speech. To our best understanding of the paper, we estimate the total duration of medical-domain speech to be around 123 hours. Recordings were collected by crowd-sourced workers to read aloud the medical transcripts (also known as read speech), thus both the number of recording conditions and speaking roles are counted as 1.\\
$^5$myMediCon dataset hired speakers to read aloud the translated medical transcripts from English corpus (thus known as read speech). 5 speakers' accents include: Native Burmese, Pa'O, Kachin, Dawei, and Mon. 2 speaking roles are patients and doctors.
}
\label{table:data_stats_ASR_literaturecompare}
\end{table*}

%% file: tables/data_stats_comparison_medical.tex
\begin{table}[h]
\resizebox{\textwidth}{!}{%
\begin{tabular}{l|c|c|c|c}
\hline
\multicolumn{1}{c|}{\textbf{Dataset}} & \textbf{Size} & \textbf{Domain} & \textbf{Language} & \textbf{Direction} \\ \hline
\citet{neves2017parallel}$^1$ & 23k pairs = 46k samples & clinical trials & en-pt & one-to-one \\
ParaMed$^2$ \cite{liu2021paramed} & 100k pairs = 200k samples & medical documents & en-zh & one-to-one \\
Khresmoi$^3$ \cite{Khresmoi_dataset} & 1k5 pairs = 12k samples & medical documents & 8 EU lang. & many-to-many \\
WMT Biomedical Task$^4$ \cite{bawden2020findings} & 160k samples & medical documents & 9 lang. & one-to-one \\
YuQ$^5$ \cite{yu2020yuq} & 65k pairs = 130k samples & medical articles & en-ug & one-to-one \\
\citet{berard2020multilingual}$^6$ & 1500 samples & COVID-19 & en-kr & one-to-one \\
MedEV$^7$ \cite{vo2024improving} & 18k pairs = 36k samples & medical documents & en-vi & one-to-one \\ \hline
\textbf{\MultiMedST (ours)} & 48k pairs = \textbf{290k samples} & medical conversations & 5 lang. & \textbf{many-to-many} \\\hline
\end{tabular}
}
\caption{\textbf{Dataset comparison with literature: A list of all publicly available medical \gls{MT} datasets.} \\
\textbf{Our \MultiMedST is the first medical \gls{ST} dataset, and is the largest medical \gls{MT} dataset}, to the best of our knowledge, given the fact that speech data is much more difficult to collect compared to medical text data. \\
$^1$ Text-only medical \gls{MT} dataset for English - Portuguese\\
$^2$ Text-only medical \gls{MT} dataset for English - Chinese crawled from the New England Journal of Medicine, thus leading to low diversity\\
$^3$ Text-only medical \gls{MT} dataset for 8 European languages: Czech, English, French, German, Hungarian, Polish, Spanish, and Swedish. The dataset requires users' costly payment.\\
$^4$ Text-only medical \gls{MT} dataset for 9 European languages: English,  Basque, Chinese, French, German, Italian, Portuguese, Spanish, Russian\\
$^5$ Text-only medical \gls{MT} dataset for Chinese-Uyghur, covering seven clinical disciplines and five sense organs science\\
$^6$ Text-only medical \gls{MT} test set for Korean-English, collected from official COVID-19 guidelines and recent papers\\
$^7$ Text-only medical \gls{MT} dataset for English - Vietnamese, containing 18k high-quality sentence pairs as dev and test set. The rest training data was not quality-controlled by human annotators.
}
\label{tab.data_stats_comparison_medical}
\end{table}

%% file: tables/data_stats_comparison_ST.tex
\begin{table*}[h]
\resizebox{\textwidth}{!}{%
\begin{tabular}{l|c|c|c|c}
\hline
\multicolumn{1}{c|}{\textbf{Dataset}} & \textbf{Size} & \textbf{Domain} & \textbf{Language} & \textbf{Direction} \\ \hline
BhasaAnuvaad$^1$ \cite{jain2024bhasaanuvaad} & 47k samples & general-domain, spontaneuous & en-13 Indic lang. & one-to-many \\
Europarl-ST$^2$ \cite{iranzo2020europarl} & 200k samples & parliamentary debates & 6 EU lang. & many-to-many \\
MaSS$^3$ \cite{boito2020mass}& 8k samples & bible & 8 lang. & many-to-many \\
Fisher \& Callhome$^4$ \cite{post2013improved} & 170k samples & telephone, spontaneuous & en-es & one-to-one \\
BSTC$^5$ \cite{zhang2021bstc} & 40k samples & various TED-like domains & en-zh & one-to-one \\ \hline
\textbf{\MultiMedST (ours)} & \textbf{290k samples} & medical conversations & 5 lang. & \textbf{many-to-many}
\\\hline
\end{tabular}
}
\caption{\textbf{Dataset comparison with literature: A list of some of the largest publicly available medical \gls{ST} datasets.} \\
Although medical \gls{ST} data is widely known to be very difficult to collect, \textbf{our \MultiMedST is as large as popular large-scale general-domain \gls{ST} datasets}. Although ours is not the largest among all existing \gls{ST} datasets, \textbf{our \MultiMedST is the largest many-to-many multilingual \gls{ST} datasets}.\\
$^1$ Bidirectional \gls{ST} dataset from English into 13 Indian languages, known as the largest Indic language \gls{ST} dataset\\
$^2$ Many-to-many multilingual \gls{ST} dataset, covering English, German, French, Spanish, Italian and Portuguese. The domain is about parliamentary debates,  thus leading to low diversity\\
$^3$ Clean \gls{ST} dataset extracted from the Bible, covering English, Spanish, Basque, Finnish, French, Hungarian, Romanian, and Russian\\
$^4$ Crowd-sourced Spanish-English \gls{ST} dataset derived from two costly \gls{ASR} datasets Fisher \cite{cieri2004fisher} and Callhome\\
$^5$ The first large-scale Chinese-English \gls{ST} dataset, containing 68 hours of mandarin speeches from three TED-like content producers
}
\label{tab.data_stats_comparison_ST}
\end{table*}

%% file: tables/appx_nmt_fewshot-En-X.tex
\setlength{\tabcolsep}{4pt}
\begin{longtable}{l|l|cccc|cccc}
 &  & \multicolumn{4}{c|}{\textbf{Ground-truth}} & \multicolumn{4}{c}{\textbf{ASR}} \\ \cline{3-10} 
\multirow{-2}{*}{\textbf{Model}} & \multirow{-2}{*}{\textbf{Metrics}} & \textbf{en-vi} & \textbf{en-fr} & \textbf{en-zh} & \textbf{en-de} & \textbf{en-vi} & \textbf{en-fr} & \textbf{en-zh} & \textbf{en-de} \\ \hline
\endfirsthead
\multicolumn{10}{c}%
{{\bfseries Table \thetable\ continued from previous page}} \\
 &  & \multicolumn{4}{c|}{\textbf{Ground-truth}} & \multicolumn{4}{c}{\textbf{ASR}} \\ \cline{3-10} 
\multirow{-2}{*}{\textbf{Model}} & \multirow{-2}{*}{\textbf{Metrics}} & \textbf{en-vi} & \textbf{en-fr} & \textbf{en-zh} & \textbf{en-de} & \textbf{en-vi} & \textbf{en-fr} & \textbf{en-zh} & \textbf{en-de} \\ \hline
\endhead
\hline
\endfoot
\endlastfoot
 & BLEU & 53.44 & 48.24 & 37.5 & 40.49 & 43.32 & 37.92 & 30.78 & 31.36 \\
 & BERTScore & 0.9 & 0.89 & 0.83 & 0.87 & 0.78 & 0.76 & 0.73 & 0.74 \\
 & TER & 39.42 & 46.9 & 58.37 & 53.12 & 53.71 & 60.93 & 66.54 & 67.05 \\
 & METEOR & 0.77 & 0.72 & 0.6 & 0.66 & 0.63 & 0.59 & 0.51 & 0.53 \\
 & ChrF & 67.74 & 70.97 & 32.39 & 66.08 & 57.23 & 60.63 & 26.34 & 56.58 \\
 & ROUGE-1 & 0.83 & 0.73 & 0.15 & 0.67 & 0.76 & 0.63 & 0.13 & 0.57 \\
 & ROUGE-2 & 0.68 & 0.57 & 0.13 & 0.47 & 0.58 & 0.47 & 0.11 & 0.37 \\
\multirow{-8}{*}{\textbf{\begin{tabular}[c]{@{}l@{}}Llama-3.1-8B \\ - Ft.\end{tabular}}} & ROUGE-L & 0.76 & 0.7 & 0.15 & 0.63 & 0.67 & 0.59 & 0.13 & 0.53 \\ \hline
 & BLEU & 42.24 & 41.25 & 24.22 & 33.89 & 14.59 & 11.01 & 10.01 & 8.84 \\
 & BERTScore & 0.94 & 0.92 & 0.88 & 0.91 & 0.75 & 0.75 & 0.63 & 0.74 \\
 & TER & 47.12 & 48.21 & 90.92 & 55.95 & 95.01 & 110.53 & 119.62 & 119.14 \\
 & METEOR & 0.66 & 0.65 & 0.51 & 0.59 & 0.32 & 0.32 & 0.25 & 0.28 \\
 & ChrF & 57.44 & 63.84 & 25.17 & 58.71 & 30.2 & 38.57 & 10.25 & 35.75 \\
 & ROUGE-1 & 0.77 & 0.68 & 0.17 & 0.61 & 0.5 & 0.34 & 0.09 & 0.29 \\
 & ROUGE-2 & 0.59 & 0.51 & 0.15 & 0.4 & 0.28 & 0.18 & 0.07 & 0.13 \\
\multirow{-8}{*}{\textbf{\begin{tabular}[c]{@{}l@{}}Llama-3.1-8B\\ - 0 Shot\end{tabular}}} & ROUGE-L & 0.69 & 0.66 & 0.16 & 0.58 & 0.39 & 0.31 & 0.09 & 0.26 \\ \hline
 & BLEU & 47.77 & 46.22 & 31.69 & 38.34 & 14.57 & 11.33 & 13.75 & 8.77 \\
 & BERTScore & 0.95 & 0.93 & 0.88 & 0.92 & 0.77 & 0.77 & 0.69 & 0.74 \\
 & TER & 40.66 & 42.69 & 59.98 & 50.31 & 101.77 & 115.8 & 83.28 & 129.49 \\
 & METEOR & 0.72 & 0.7 & 0.56 & 0.64 & 0.36 & 0.34 & 0.31 & 0.3 \\
 & ChrF & 62.4 & 68.12 & 27.1 & 63.19 & 33.13 & 40.18 & 12.97 & 37.59 \\
 & ROUGE-1 & 0.81 & 0.73 & 0.17 & 0.66 & 0.54 & 0.34 & 0.11 & 0.28 \\
 & ROUGE-2 & 0.64 & 0.56 & 0.16 & 0.44 & 0.3 & 0.19 & 0.09 & 0.13 \\
\multirow{-8}{*}{\textbf{\begin{tabular}[c]{@{}l@{}}Llama-3.1-8B\\ - 1 Shot\end{tabular}}} & ROUGE-L & 0.74 & 0.7 & 0.17 & 0.63 & 0.41 & 0.31 & 0.1 & 0.25 \\ \hline
 & BLEU & 46.7 & 45.58 & 31.28 & 38.0 & 16.14 & 12.38 & 13.97 & 9.38 \\
 & BERTScore & 0.95 & 0.93 & 0.88 & 0.92 & 0.77 & 0.77 & 0.69 & 0.75 \\
 & TER & 41.73 & 43.32 & 59.24 & 50.92 & 95.97 & 109.27 & 80.78 & 123.43 \\
 & METEOR & 0.7 & 0.7 & 0.56 & 0.64 & 0.37 & 0.34 & 0.32 & 0.3 \\
 & ChrF & 61.5 & 67.39 & 26.85 & 62.46 & 34.01 & 40.42 & 13.29 & 37.53 \\
 & ROUGE-1 & 0.81 & 0.73 & 0.17 & 0.66 & 0.55 & 0.35 & 0.11 & 0.29 \\
 & ROUGE-2 & 0.63 & 0.56 & 0.15 & 0.44 & 0.31 & 0.19 & 0.09 & 0.13 \\
\multirow{-8}{*}{\textbf{\begin{tabular}[c]{@{}l@{}}Llama-3.1-8B\\ - 8 Shot\end{tabular}}} & ROUGE-L & 0.73 & 0.7 & 0.17 & 0.62 & 0.43 & 0.32 & 0.1 & 0.26 \\ \hline
 & BLEU & 47.77 & 46.22 & 43.62 & 38.34 & 15.45 & 12.04 & 13.94 & 9.17 \\
 & BERTScore & 0.95 & 0.93 & 0.94 & 0.92 & 0.77 & 0.75 & 0.69 & 0.74 \\
 & TER & 40.66 & 42.69 & 46.76 & 50.31 & 101.83 & 109.92 & 81.05 & 124.29 \\
 & METEOR & 0.72 & 0.7 & 0.68 & 0.64 & 0.37 & 0.33 & 0.32 & 0.29 \\
 & ChrF & 62.4 & 68.12 & 37.57 & 63.19 & 34.13 & 39.64 & 13.27 & 36.84 \\
 & ROUGE-1 & 0.81 & 0.73 & 0.2 & 0.66 & 0.54 & 0.34 & 0.11 & 0.28 \\
 & ROUGE-2 & 0.64 & 0.56 & 0.18 & 0.44 & 0.31 & 0.19 & 0.09 & 0.13 \\
\multirow{-8}{*}{\textbf{\begin{tabular}[c]{@{}l@{}}Llama-3.1-8B\\ - 16 Shot\end{tabular}}} & ROUGE-L & 0.74 & 0.7 & 0.19 & 0.63 & 0.42 & 0.31 & 0.1 & 0.25 \\ \hline
 & BLEU & 48.27 & 46.92 & 30.56 & 38.27 & 15.21 & 15.29 & 13.97 & 12.82 \\
 & BERTScore & 0.95 & 0.93 & 0.87 & 0.92 & 0.77 & 0.76 & 0.69 & 0.74 \\
 & TER & 41.04 & 42.64 & 59.25 & 50.85 & 102.67 & 78.6 & 81.42 & 84.12 \\
 & METEOR & 0.72 & 0.7 & 0.55 & 0.64 & 0.37 & 0.33 & 0.31 & 0.3 \\
 & ChrF & 62.63 & 68.33 & 26.38 & 62.97 & 33.87 & 38.04 & 13.28 & 36.11 \\
 & ROUGE-1 & 0.81 & 0.73 & 0.17 & 0.65 & 0.53 & 0.39 & 0.1 & 0.35 \\
 & ROUGE-2 & 0.64 & 0.56 & 0.16 & 0.44 & 0.3 & 0.22 & 0.08 & 0.16 \\
\multirow{-8}{*}{\textbf{\begin{tabular}[c]{@{}l@{}}Llama-3.1-8B\\ - 32 Shot\end{tabular}}} & ROUGE-L & 0.74 & 0.7 & 0.17 & 0.62 & 0.41 & 0.35 & 0.1 & 0.31 \\ \hline
 & BLEU & 54.5 & 49.63 & 28.61 & 38.75 & 43.37 & 37.34 & 23.46 & 28.5 \\
 & BERTScore & 0.9 & 0.9 & 0.81 & 0.87 & 0.77 & 0.76 & 0.74 & 0.74 \\
 & TER & 38.21 & 42.42 & 59.1 & 51.55 & 53.52 & 57.76 & 64.75 & 66.7 \\
 & METEOR & 0.77 & 0.72 & 0.5 & 0.64 & 0.63 & 0.58 & 0.44 & 0.51 \\
 & ChrF & 67.82 & 70.5 & 27.81 & 63.39 & 57.34 & 60.21 & 23.19 & 54.19 \\
 & ROUGE-1 & 0.83 & 0.74 & 0.14 & 0.65 & 0.76 & 0.63 & 0.12 & 0.55 \\
 & ROUGE-2 & 0.69 & 0.57 & 0.13 & 0.43 & 0.59 & 0.47 & 0.1 & 0.34 \\
\multirow{-8}{*}{\textbf{\begin{tabular}[c]{@{}l@{}}Qwen-2.5-7B\\ - Ft.\end{tabular}}} & ROUGE-L & 0.77 & 0.71 & 0.14 & 0.62 & 0.67 & 0.6 & 0.12 & 0.51 \\ \hline
 & BLEU & 38.32 & 39.1 & 36.16 & 30.0 & 13.95 & 13.43 & 17.4 & 10.05 \\
 & BERTScore & 0.93 & 0.91 & 0.92 & 0.9 & 0.73 & 0.73 & 0.7 & 0.72 \\
 & TER & 53.71 & 52.77 & 63.44 & 62.2 & 82.95 & 84.55 & 95.85 & 89.8 \\
 & METEOR & 0.61 & 0.61 & 0.66 & 0.55 & 0.3 & 0.3 & 0.36 & 0.26 \\
 & ChrF & 53.19 & 60.89 & 34.98 & 55.39 & 28.72 & 35.86 & 16.47 & 33.06 \\
 & ROUGE-1 & 0.72 & 0.64 & 0.17 & 0.56 & 0.51 & 0.36 & 0.1 & 0.31 \\
 & ROUGE-2 & 0.54 & 0.47 & 0.15 & 0.34 & 0.28 & 0.19 & 0.08 & 0.13 \\
\multirow{-8}{*}{\textbf{\begin{tabular}[c]{@{}l@{}}Qwen-2.5-7B\\ - 0 Shot\end{tabular}}} & ROUGE-L & 0.64 & 0.61 & 0.17 & 0.53 & 0.4 & 0.32 & 0.1 & 0.27 \\ \hline
 & BLEU & 43.81 & 44.22 & 26.17 & 28.21 & 16.09 & 16.18 & 15.76 & 11.56 \\
 & BERTScore & 0.93 & 0.93 & 0.87 & 0.91 & 0.75 & 0.75 & 0.68 & 0.68 \\
 & TER & 45.94 & 44.7 & 62.27 & 62.57 & 79.98 & 81.36 & 85.38 & 84.33 \\
 & METEOR & 0.67 & 0.68 & 0.48 & 0.5 & 0.34 & 0.35 & 0.33 & 0.3 \\
 & ChrF & 58.22 & 66.19 & 25.15 & 51.07 & 31.66 & 39.96 & 14.43 & 35.66 \\
 & ROUGE-1 & 0.78 & 0.71 & 0.15 & 0.52 & 0.56 & 0.41 & 0.1 & 0.35 \\
 & ROUGE-2 & 0.6 & 0.53 & 0.13 & 0.32 & 0.31 & 0.22 & 0.08 & 0.15 \\
\multirow{-8}{*}{\textbf{\begin{tabular}[c]{@{}l@{}}Qwen-2.5-7B\\ - 1 Shot\end{tabular}}} & ROUGE-L & 0.7 & 0.68 & 0.15 & 0.49 & \cellcolor[HTML]{FFFFFF}0.43 & \cellcolor[HTML]{FFFFFF}0.37 & 0.1 & \cellcolor[HTML]{FFFFFF}0.31 \\ \hline
 & BLEU & 42.62 & 42.14 & 26.13 & 31.26 & 15.15 & 13.73 & 9.85 & 10.12 \\
 & BERTScore & 0.93 & 0.92 & 0.87 & 0.9 & 0.73 & 0.72 & 0.64 & 0.7 \\
 & TER & 50.59 & 48.99 & 62.56 & 62.16 & 84.56 & 82.99 & 86.47 & 92.32 \\
 & METEOR & 0.65 & 0.65 & 0.48 & 0.56 & 0.31 & 0.3 & 0.23 & 0.26 \\
 & ChrF & 56.84 & 63.67 & 24.91 & 55.45 & 29.24 & 34.64 & 10.07 & 31.7 \\
 & ROUGE-1 & 0.76 & 0.68 & 0.15 & 0.57 & 0.5 & 0.36 & 0.08 & 0.29 \\
 & ROUGE-2 & 0.58 & 0.5 & 0.14 & 0.36 & 0.28 & 0.19 & 0.06 & 0.13 \\
\multirow{-8}{*}{\textbf{\begin{tabular}[c]{@{}l@{}}Qwen-2.5-7B\\ - 8 Shot\end{tabular}}} & ROUGE-L & 0.68 & 0.65 & 0.15 & 0.53 & 0.39 & 0.32 & 0.08 & 0.26 \\ \hline
 & BLEU & 43.81 & 44.22 & 26.17 & 28.21 & 16.03 & 14.52 & 12.21 & 7.8 \\
 & BERTScore & 0.94 & 0.93 & 0.87 & 0.89 & 0.75 & 0.74 & 0.68 & 0.68 \\
 & TER & 45.94 & 44.7 & 62.27 & 62.57 & 82.91 & 84.47 & 86.09 & 96.92 \\
 & METEOR & 0.67 & 0.68 & 0.48 & 0.5 & 0.33 & 0.32 & 0.27 & 0.21 \\
 & ChrF & 58.22 & 66.19 & 25.15 & 51.07 & 31.11 & 36.74 & 11.81 & 27.56 \\
 & ROUGE-1 & 0.78 & 0.71 & 0.15 & 0.52 & 0.54 & 0.38 & 0.09 & 0.24 \\
 & ROUGE-2 & 0.6 & 0.53 & 0.13 & 0.32 & 0.3 & 0.2 & 0.07 & 0.09 \\
\multirow{-8}{*}{\textbf{\begin{tabular}[c]{@{}l@{}}Qwen-2.5-7B\\ - 16 Shot\end{tabular}}} & ROUGE-L & 0.7 & 0.68 & 0.15 & 0.49 & 0.42 & 0.34 & 0.09 & 0.21 \\ \hline
 & BLEU & 43.51 & 44.95 & 25.72 & 33.24 & 16.96 & 16.86 & 14.63 & 9.33 \\
 & BERTScore & 0.94 & 0.93 & 0.87 & 0.91 & 0.75 & 0.76 & 0.7 & 0.71 \\
 & TER & 45.74 & 44.76 & 62.47 & 56.02 & 82.13 & 80.46 & 83.36 & 97.53 \\
 & METEOR & 0.66 & 0.69 & 0.47 & 0.59 & 0.35 & 0.36 & 0.31 & 0.24 \\
 & ChrF & 58.02 & 66.94 & 24.8 & 58.63 & 32.33 & 40.69 & 13.9 & 31.57 \\
 & ROUGE-1 & 0.77 & 0.71 & 0.15 & 0.61 & 0.56 & 0.42 & 0.1 & 0.27 \\
 & ROUGE-2 & 0.6 & 0.54 & 0.13 & 0.38 & 0.31 & 0.23 & 0.08 & 0.11 \\
\multirow{-8}{*}{\textbf{\begin{tabular}[c]{@{}l@{}}Qwen-2.5-7B\\ - 32 Shot\end{tabular}}} & ROUGE-L & 0.7 & 0.68 & 0.15 & 0.57 & 0.44 & 0.38 & 0.1 & 0.24 \\ \hline
 & BLEU & 24.77 & 51.71 & 26.38 & 43.99 & 17.72 & 36.58 & 20.27 & 29.9 \\
 & BERTScore & 0.82 & 0.89 & 0.81 & 0.88 & 0.68 & 0.74 & 0.69 & 0.72 \\
 & TER & 63.33 & 43.0 & 60.43 & 48.54 & 71.27 & 59.78 & 70.94 & 65.81 \\
 & METEOR & 0.45 & 0.73 & 0.48 & 0.67 & 0.35 & 0.55 & 0.38 & 0.5 \\
 & ChrF & 41.27 & 71.06 & 27.29 & 66.27 & 33.42 & 57.32 & 20.29 & 52.89 \\
 & ROUGE-1 & 0.66 & 0.74 & 0.12 & 0.68 & 0.57 & 0.6 & 0.09 & 0.54 \\
 & ROUGE-2 & 0.49 & 0.6 & 0.11 & 0.49 & 0.39 & 0.45 & 0.08 & 0.36 \\
\multirow{-8}{*}{\textbf{\begin{tabular}[c]{@{}l@{}}Mistral-v0.3-7B\\ - Ft.\end{tabular}}} & ROUGE-L & 0.57 & 0.72 & 0.12 & 0.65 & 0.49 & 0.58 & 0.09 & 0.51 \\ \hline
 & BLEU & 8.99 & 37.86 & 34.21 & 29.25 & 1.59 & 10.56 & 12.83 & 6.76 \\
 & BERTScore & 0.89 & 0.92 & 0.92 & 0.91 & 0.62 & 0.73 & 0.66 & 0.68 \\
 & TER & 80.02 & 51.46 & 57.09 & 59.85 & 94.82 & 85.99 & 93.26 & 95.29 \\
 & METEOR & 0.25 & 0.62 & 0.58 & 0.55 & 0.08 & 0.28 & 0.28 & 0.21 \\
 & ChrF & 24.94 & 61.82 & 29.28 & 56.03 & 11.44 & 34.01 & 11.71 & 28.79 \\
 & ROUGE-1 & 0.53 & 0.66 & 0.18 & 0.58 & 0.31 & 0.35 & 0.1 & 0.25 \\
 & ROUGE-2 & 0.26 & 0.47 & 0.16 & 0.35 & 0.09 & 0.17 & 0.08 & 0.09 \\
\multirow{-8}{*}{\textbf{\begin{tabular}[c]{@{}l@{}}Mistral-v0.3-7B\\ - 0 Shot\end{tabular}}} & ROUGE-L & 0.4 & 0.63 & 0.18 & 0.54 & 0.23 & 0.31 & 0.09 & 0.22 \\ \hline
 & BLEU & 12.38 & 41.24 & 29.67 & 35.08 & 2.37 & 11.39 & 13.14 & 7.52 \\
 & BERTScore & 0.91 & 0.93 & 0.89 & 0.92 & 0.64 & 0.77 & 0.69 & 0.75 \\
 & TER & 75.66 & 47.52 & 59.61 & 52.67 & 92.12 & 84.13 & 81.89 & 94.24 \\
 & METEOR & 0.31 & 0.66 & 0.52 & 0.61 & 0.1 & 0.29 & 0.3 & 0.22 \\
 & ChrF & 29.17 & 64.35 & 26.76 & 60.68 & 12.84 & 34.29 & 12.62 & 29.71 \\
 & ROUGE-1 & 0.58 & 0.69 & 0.15 & 0.64 & 0.35 & 0.36 & 0.1 & 0.26 \\
 & ROUGE-2 & 0.32 & 0.51 & 0.14 & 0.41 & 0.12 & 0.18 & 0.08 & 0.1 \\
\multirow{-8}{*}{\textbf{\begin{tabular}[c]{@{}l@{}}Mistral-v0.3-7B\\ - 1 Shot\end{tabular}}} & ROUGE-L & 0.45 & 0.66 & 0.15 & 0.6 & 0.27 & 0.32 & 0.1 & 0.24 \\ \hline
 & BLEU & 13.64 & 41.86 & 21.4 & 34.17 & 4.76 & 15.02 & 8.54 & 11.69 \\
 & BERTScore & 0.91 & 0.93 & 0.87 & 0.92 & 0.7 & 0.76 & 0.67 & 0.74 \\
 & TER & 74.2 & 46.61 & 64.85 & 53.98 & 86.73 & 76.58 & 80.51 & 83.67 \\
 & METEOR & 0.33 & 0.66 & 0.44 & 0.6 & 0.17 & 0.34 & 0.25 & 0.3 \\
 & ChrF & 30.12 & 64.65 & 21.72 & 59.53 & 18.25 & 38.97 & 10.1 & 35.59 \\
 & ROUGE-1 & 0.59 & 0.7 & 0.13 & 0.63 & 0.45 & 0.42 & 0.08 & 0.35 \\
 & ROUGE-2 & 0.34 & 0.51 & 0.12 & 0.4 & 0.19 & 0.22 & 0.06 & 0.15 \\
\multirow{-8}{*}{\textbf{\begin{tabular}[c]{@{}l@{}}Mistral-v0.3-7B\\ - 8 Shot\end{tabular}}} & ROUGE-L & 0.46 & 0.67 & 0.13 & 0.59 & 0.33 & 0.38 & 0.07 & 0.31 \\ \hline
 & BLEU & 14.22 & 42.83 & 21.35 & 35.08 & 4.9 & 15.64 & 9.83 & 12.0 \\
 & BERTScore & 0.91 & 0.93 & 0.87 & 0.92 & 0.71 & 0.77 & 0.69 & 0.75 \\
 & TER & 73.33 & 45.48 & 64.69 & 52.67 & 86.31 & 75.73 & 78.37 & 82.11 \\
 & METEOR & 0.34 & 0.67 & 0.44 & 0.61 & 0.18 & 0.35 & 0.27 & 0.3 \\
 & ChrF & 30.98 & 65.76 & 21.77 & 60.68 & 18.45 & 39.6 & 11.42 & 36.29 \\
 & ROUGE-1 & 0.6 & 0.71 & 0.13 & 0.64 & 0.46 & 0.43 & 0.09 & 0.36 \\
 & ROUGE-2 & 0.35 & 0.52 & 0.12 & 0.41 & 0.19 & 0.23 & 0.07 & 0.16 \\
\multirow{-8}{*}{\textbf{\begin{tabular}[c]{@{}l@{}}Mistral-v0.3-7B\\ - 16 Shot\end{tabular}}} & ROUGE-L & 0.47 & 0.68 & 0.13 & 0.6 & 0.33 & 0.38 & 0.09 & 0.32 \\ \hline
 & BLEU & 14.58 & 43.36 & 21.35 & 35.05 & 5.8 & 15.81 & 9.81 & 12.3 \\
 & BERTScore & 0.91 & 0.93 & 0.87 & 0.92 & 0.73 & 0.77 & 0.69 & 0.75 \\
 & TER & 73.14 & 45.49 & 64.51 & 53.02 & 84.79 & 75.53 & 78.33 & 82.1 \\
 & METEOR & 0.34 & 0.68 & 0.44 & 0.61 & 0.2 & 0.35 & 0.27 & 0.31 \\
 & ChrF & 31.29 & 66.06 & 21.77 & 60.7 & 20.24 & 39.73 & 11.41 & 36.63 \\
 & ROUGE-1 & 0.6 & 0.71 & 0.13 & 0.64 & 0.49 & 0.43 & 0.09 & 0.36 \\
 & ROUGE-2 & 0.36 & 0.53 & 0.11 & 0.41 & 0.22 & 0.23 & 0.07 & 0.16 \\
\multirow{-8}{*}{\textbf{\begin{tabular}[c]{@{}l@{}}Mistral-v0.3-7B\\ - 32 Shot\end{tabular}}} & ROUGE-L & 0.48 & 0.68 & 0.13 & 0.6 & 0.35 & 0.39 & 0.08 & 0.32 \\ \hline
\caption{\textbf{In-context Learning Results}. Full fine-tuning, zero-shot, few-shot on both ground-truth transcript and \gls{ASR} transcript in the cascaded setting. \textbf{English to X} results are reported in this table.\\
All cascaded models use Whisper$_{small-mono}$ as \gls{ASR} model (Whisper \gls{ASR} is fine-tuned monolingually - on each source language separately). Its \gls{WER} on test set is 29.6\%, 33.8\%, 31.3\%, 26.3\%, 45.7\% for Vietnamese, English, Chinese, German and French respectively.}
\label{tab:appx_nmt_fewshot-En-X}
\end{longtable}

%% file: tables/appx_nmt_fewshot-Vi-X.tex
\setlength{\tabcolsep}{4pt}
\begin{longtable}{l|l|cccc|cccc}
 &  & \multicolumn{4}{c|}{\textbf{Ground-truth}} & \multicolumn{4}{c}{\textbf{ASR}} \\ \cline{3-10} 
\multirow{-2}{*}{\textbf{Model}} & \multirow{-2}{*}{\textbf{Metrics}} & \textbf{vi-en} & \textbf{vi-fr} & \textbf{vi-zh} & \textbf{vi-de} & \textbf{vi-en} & \textbf{vi-fr} & \textbf{vi-zh} & \textbf{vi-de} \\ \hline
\endfirsthead
\multicolumn{10}{c}%
{{\bfseries Table \thetable\ continued from previous page}} \\
 &  & \multicolumn{4}{c|}{\textbf{Ground-truth}} & \multicolumn{4}{c}{\textbf{ASR}} \\ \cline{3-10} 
\multirow{-2}{*}{\textbf{Model}} & \multirow{-2}{*}{\textbf{Metrics}} & \textbf{vi-en} & \textbf{vi-fr} & \textbf{vi-zh} & \textbf{vi-de} & \textbf{vi-en} & \textbf{vi-fr} & \textbf{vi-zh} & \textbf{vi-de} \\ \hline
\endhead
\hline
\endfoot
\endlastfoot
 & BLEU & 23.16 & 15.57 & 16.09 & 11.61 & 14.55 & 10.29 & 11.56 & 7.71 \\
 & BERTScore & 0.92 & 0.79 & 0.74 & 0.77 & 0.78 & 0.75 & 0.73 & 0.73 \\
 & TER & 83.07 & 100.23 & 120.58 & 112.85 & 98.39 & 112.97 & 131.0 & 122.33 \\
 & METEOR & 0.57 & 0.45 & 0.5 & 0.39 & 0.43 & 0.35 & 0.4 & 0.31 \\
 & ChrF & 52.29 & 47.18 & 21.03 & 44.25 & 41.46 & 39.22 & 15.65 & 37.02 \\
 & ROUGE-1 & 0.56 & 0.45 & 0.03 & 0.4 & 0.44 & 0.36 & 0.02 & 0.32 \\
 & ROUGE-2 & 0.33 & 0.25 & 0.01 & 0.19 & 0.22 & 0.18 & 0.01 & 0.12 \\
\multirow{-8}{*}{\textbf{\begin{tabular}[c]{@{}l@{}}Llama-3.1-8B \\ - Ft.\end{tabular}}} & ROUGE-L & 0.51 & 0.4 & 0.03 & 0.36 & 0.39 & 0.31 & 0.02 & 0.28 \\ \hline
 & BLEU & 17.01 & 11.95 & 4.23 & 9.93 & 11.3 & 08.06 & 3.11 & 6.75 \\
 & BERTScore & 0.79 & 0.76 & 0.61 & 0.76 & 0.88 & 0.73 & 0.54 & 0.72 \\
 & TER & 81.7 & 89.32 & 180.31 & 92.37 & 95.71 & 98.96 & 182.42 & 102.77 \\
 & METEOR & 0.42 & 0.33 & 0.19 & 0.3 & 0.34 & 0.26 & 0.15 & 0.24 \\
 & ChrF & 43.0 & 38.83 & 6.21 & 37.28 & 35.53 & 33.2 & 4.79 & 32.0 \\
 & ROUGE-1 & 0.47 & 0.39 & 0.02 & 0.35 & 0.38 & 0.32 & 0.01 & 0.29 \\
 & ROUGE-2 & 0.24 & 0.19 & 0.01 & 0.14 & 0.17 & 0.14 & 0.0 & 0.1 \\
\multirow{-8}{*}{\textbf{\begin{tabular}[c]{@{}l@{}}Llama-3.1-8B\\ - 0 Shot\end{tabular}}} & ROUGE-L & 0.43 & 0.34 & 0.02 & 0.31 & 0.34 & 0.27 & 0.01 & 0.25 \\ \hline
 & BLEU & 24.75 & 16.02 & 18.84 & 11.93 & 14.08 & 9.94 & 12.24 & 7.62 \\
 & BERTScore & 0.83 & 0.79 & 0.76 & 0.77 & 0.89 & 0.75 & 0.7 & 0.73 \\
 & TER & 65.83 & 80.97 & 75.82 & 85.74 & 86.83 & 95.22 & 89.86 & 97.14 \\
 & METEOR & 0.53 & 0.38 & 0.43 & 0.32 & 0.37 & 0.3 & 0.34 & 0.25 \\
 & ChrF & 50.29 & 43.02 & 17.01 & 39.7 & 39.04 & 36.81 & 12.43 & 33.4 \\
 & ROUGE-1 & 0.56 & 0.43 & 0.03 & 0.38 & 0.42 & 0.36 & 0.02 & 0.3 \\
 & ROUGE-2 & 0.32 & 0.23 & 0.01 & 0.16 & 0.2 & 0.16 & 0.01 & 0.1 \\
\multirow{-8}{*}{\textbf{\begin{tabular}[c]{@{}l@{}}Llama-3.1-8B\\ - 1 Shot\end{tabular}}} & ROUGE-L & 0.51 & 0.38 & 0.03 & 0.33 & 0.38 & 0.31 & 0.02 & 0.26 \\ \hline
 & BLEU & 24.24 & 15.42 & 15.03 & 11.25 & 15.8 & 10.29 & 11.89 & 8.84 \\
 & BERTScore & 0.82 & 0.78 & 0.77 & 0.77 & 0.9 & 0.75 & 0.7 & 0.74 \\
 & TER & 66.93 & 84.03 & 89.52 & 86.95 & 84.82 & 94.29 & 93.2 & 96.51 \\
 & METEOR & 0.52 & 0.37 & 0.41 & 0.31 & 0.41 & 0.29 & 0.34 & 0.27 \\
 & ChrF & 49.83 & 42.08 & 15.47 & 39.18 & 40.61 & 35.1 & 12.52 & 35.19 \\
 & ROUGE-1 & 0.55 & 0.41 & 0.03 & 0.37 & 0.44 & 0.33 & 0.02 & 0.32 \\
 & ROUGE-2 & 0.31 & 0.22 & 0.01 & 0.15 & 0.21 & 0.15 & 0.01 & 0.12 \\
\multirow{-8}{*}{\textbf{\begin{tabular}[c]{@{}l@{}}Llama-3.1-8B\\ - 8 Shot\end{tabular}}} & ROUGE-L & 0.51 & 0.36 & 0.03 & 0.32 & 0.39 & 0.29 & 0.02 & 0.28 \\ \hline
 & BLEU & 24.75 & 16.02 & 18.84 & 11.93 & 16.94 & 10.28 & 11.24 & 8.4 \\
 & BERTScore & 0.83 & 0.79 & 0.78 & 0.77 & 0.9 & 0.75 & 0.69 & 0.74 \\
 & TER & 65.83 & 80.97 & 75.82 & 85.74 & 81.87 & 98.72 & 101.17 & 98.04 \\
 & METEOR & 0.53 & 0.38 & 0.43 & 0.32 & 0.41 & 0.31 & 0.34 & 0.28 \\
 & ChrF & 50.29 & 43.02 & 17.01 & 39.7 & 40.78 & 37.23 & 12.52 & 35.24 \\
 & ROUGE-1 & 0.56 & 0.43 & 0.03 & 0.38 & 0.44 & 0.35 & 0.02 & 0.31 \\
 & ROUGE-2 & 0.32 & 0.23 & 0.01 & 0.16 & 0.22 & 0.16 & 0.01 & 0.12 \\
\multirow{-8}{*}{\textbf{\begin{tabular}[c]{@{}l@{}}Llama-3.1-8B\\ - 16 Shot\end{tabular}}} & ROUGE-L & 0.51 & 0.38 & 0.03 & 0.33 & 0.4 & 0.3 & 0.02 & 0.27 \\ \hline
 & BLEU & 24.53 & 15.58 & 17.45 & 12.44 & 15.11 & 10.21 & 12.28 & 7.76 \\
 & BERTScore & 0.82 & 0.79 & 0.78 & 0.77 & 0.9 & 0.75 & 0.7 & 0.74 \\
 & TER & 68.49 & 84.25 & 78.04 & 86.52 & 85.42 & 97.14 & 88.95 & 99.26 \\
 & METEOR & 0.52 & 0.39 & 0.42 & 0.34 & 0.39 & 0.31 & 0.33 & 0.27 \\
 & ChrF & 49.62 & 43.73 & 16.15 & 40.81 & 39.4 & 36.88 & 12.3 & 34.24 \\
 & ROUGE-1 & 0.54 & 0.43 & 0.03 & 0.39 & 0.42 & 0.34 & 0.02 & 0.31 \\
 & ROUGE-2 & 0.31 & 0.23 & 0.01 & 0.17 & 0.2 & 0.16 & 0.01 & 0.11 \\
\multirow{-8}{*}{\textbf{\begin{tabular}[c]{@{}l@{}}Llama-3.1-8B\\ - 32 Shot\end{tabular}}} & ROUGE-L & 0.5 & 0.38 & 0.03 & 0.34 & 0.37 & 0.3 & 0.02 & 0.26 \\ \hline
 & BLEU & 26.21 & 19.25 & 29.06 & 14.44 & 13.97 & 11.66 & 20.27 & 8.75 \\
 & BERTScore & 0.93 & 0.81 & 0.81 & 0.79 & 0.78 & 0.76 & 0.78 & 0.75 \\
 & TER & 71.1 & 80.86 & 60.93 & 83.75 & 101.68 & 101.51 & 74.57 & 101.85 \\
 & METEOR & 0.57 & 0.47 & 0.55 & 0.39 & 0.43 & 0.36 & 0.43 & 0.3 \\
 & ChrF & 53.79 & 48.94 & 25.1 & 43.34 & 41.76 & 40.73 & 18.0 & 36.74 \\
 & ROUGE-1 & 0.58 & 0.49 & 0.04 & 0.42 & 0.43 & 0.39 & 0.03 & 0.34 \\
 & ROUGE-2 & 0.35 & 0.28 & 0.02 & 0.19 & 0.21 & 0.19 & 0.02 & 0.13 \\
\multirow{-8}{*}{\textbf{\begin{tabular}[c]{@{}l@{}}Qwen-2.5-7B\\ - Ft.\end{tabular}}} & ROUGE-L & 0.53 & 0.44 & 0.04 & 0.38 & 0.39 & 0.33 & 0.03 & 0.29 \\ \hline
 & BLEU & 20.89 & 15.21 & 26.23 & 9.73 & 12.41 & 9.94 & 16.57 & 6.59 \\
 & BERTScore & 0.79 & 0.78 & 0.81 & 0.75 & 0.88 & 0.74 & 0.73 & 0.72 \\
 & TER & 77.98 & 85.59 & 73.88 & 92.37 & 93.79 & 97.19 & 91.72 & 100.89 \\
 & METEOR & 0.45 & 0.38 & 0.54 & 0.29 & 0.34 & 0.3 & 0.41 & 0.24 \\
 & ChrF & 44.94 & 42.86 & 24.17 & 36.87 & 35.59 & 36.05 & 16.52 & 32.04 \\
 & ROUGE-1 & 0.47 & 0.42 & 0.04 & 0.33 & 0.36 & 0.33 & 0.02 & 0.28 \\
 & ROUGE-2 & 0.25 & 0.22 & 0.02 & 0.13 & 0.16 & 0.15 & 0.01 & 0.09 \\
\multirow{-8}{*}{\textbf{\begin{tabular}[c]{@{}l@{}}Qwen-2.5-7B\\ - 0 Shot\end{tabular}}} & ROUGE-L & 0.43 & 0.37 & 0.04 & 0.29 & 0.32 & 0.29 & 0.02 & 0.23 \\ \hline
 & BLEU & 22.31 & 18.44 & 27.25 & 11.33 & 14.34 & 10.5 & 16.78 & 6.42 \\
 & BERTScore & 0.82 & 0.8 & 0.81 & 0.77 & 0.9 & 0.76 & 0.73 & 0.73 \\
 & TER & 84.56 & 79.35 & 62.93 & 92.8 & 87.87 & 96.22 & 80.56 & 104.4 \\
 & METEOR & 0.55 & 0.44 & 0.53 & 0.35 & 0.4 & 0.33 & 0.41 & 0.26 \\
 & ChrF & 52.13 & 48.14 & 23.53 & 41.25 & 40.29 & 38.73 & 15.85 & 34.64 \\
 & ROUGE-1 & 0.54 & 0.48 & 0.03 & 0.38 & 0.42 & 0.36 & 0.02 & 0.29 \\
 & ROUGE-2 & 0.32 & 0.27 & 0.02 & 0.16 & 0.2 & 0.17 & 0.01 & 0.1 \\
\multirow{-8}{*}{\textbf{\begin{tabular}[c]{@{}l@{}}Qwen-2.5-7B\\ - 1 Shot\end{tabular}}} & ROUGE-L & 0.5 & 0.43 & 0.03 & 0.34 & \cellcolor[HTML]{FFFFFF}0.37 & \cellcolor[HTML]{FFFFFF}0.31 & \cellcolor[HTML]{FFFFFF}0.02 & \cellcolor[HTML]{FFFFFF}0.25 \\ \hline
 & BLEU & 20.36 & 15.47 & 22.98 & 11.41 & 15.24 & 11.77 & 18.31 & 7.7 \\
 & BERTScore & 0.81 & 0.79 & 0.8 & 0.77 & 0.9 & 0.76 & 0.74 & 0.74 \\
 & TER & 90.02 & 95.22 & 80.69 & 88.14 & 90.58 & 93.75 & 81.27 & 99.6 \\
 & METEOR & 0.54 & 0.43 & 0.53 & 0.34 & 0.42 & 0.34 & 0.42 & 0.28 \\
 & ChrF & 51.07 & 47.05 & 22.71 & 40.77 & 42.27 & 40.16 & 17.26 & 35.89 \\
 & ROUGE-1 & 0.52 & 0.45 & 0.03 & 0.38 & 0.44 & 0.37 & 0.02 & 0.31 \\
 & ROUGE-2 & 0.3 & 0.25 & 0.02 & 0.16 & 0.21 & 0.18 & 0.01 & 0.11 \\
\multirow{-8}{*}{\textbf{\begin{tabular}[c]{@{}l@{}}Qwen-2.5-7B\\ - 8 Shot\end{tabular}}} & ROUGE-L & 0.48 & 0.4 & 0.03 & 0.34 & 0.39 & 0.33 & 0.02 & 0.27 \\ \hline
 & BLEU & 22.31 & 18.44 & 27.25 & 11.33 & 14.14 & 11.85 & 19.11 & 7.89 \\
 & BERTScore & 0.82 & 0.81 & 0.82 & 0.77 & 0.9 & 0.77 & 0.75 & 0.74 \\
 & TER & 84.56 & 79.35 & 62.93 & 92.8 & 100.74 & 96.37 & 78.32 & 99.54 \\
 & METEOR & 0.55 & 0.44 & 0.53 & 0.35 & 0.43 & 0.35 & 0.43 & 0.28 \\
 & ChrF & 52.13 & 48.14 & 23.53 & 41.25 & 42.39 & 40.74 & 17.59 & 35.87 \\
 & ROUGE-1 & 0.54 & 0.48 & 0.03 & 0.38 & 0.43 & 0.38 & 0.02 & 0.31 \\
 & ROUGE-2 & 0.32 & 0.27 & 0.02 & 0.16 & 0.21 & 0.19 & 0.01 & 0.11 \\
\multirow{-8}{*}{\textbf{\begin{tabular}[c]{@{}l@{}}Qwen-2.5-7B\\ - 16 Shot\end{tabular}}} & ROUGE-L & 0.5 & 0.43 & 0.03 & 0.34 & 0.38 & 0.33 & 0.02 & 0.27 \\ \hline
 & BLEU & 24.52 & 18.13 & 26.61 & 11.97 & 1.91 & 3.72 & 9.15 & 2.24 \\
 & BERTScore & 0.83 & 0.81 & 0.82 & 0.78 & 0.84 & 0.67 & 0.65 & 0.63 \\
 & TER & 73.31 & 81.12 & 65.47 & 88.73 & 171.22 & 116.81 & 112.45 & 111.3 \\
 & METEOR & 0.56 & 0.45 & 0.53 & 0.35 & 0.16 & 0.16 & 0.25 & 0.11 \\
 & ChrF & 52.88 & 48.54 & 22.98 & 41.82 & 18.97 & 22.33 & 9.67 & 15.9 \\
 & ROUGE-1 & 0.56 & 0.48 & 0.04 & 0.39 & 0.15 & 0.18 & 0.03 & 0.12 \\
 & ROUGE-2 & 0.33 & 0.27 & 0.02 & 0.17 & 0.04 & 0.06 & 0.01 & 0.03 \\
\multirow{-8}{*}{\textbf{\begin{tabular}[c]{@{}l@{}}Qwen-2.5-7B\\ - 32 Shot\end{tabular}}} & ROUGE-L & 0.52 & 0.43 & 0.04 & 0.35 & 0.13 & 0.16 & 0.03 & 0.11 \\ \hline
 & BLEU & 24.56 & 16.0 & 25.04 & 13.38 & 15.86 & 10.92 & 17.92 & 09.03 \\
 & BERTScore & 0.92 & 0.79 & 0.78 & 0.78 & 0.78 & 0.75 & 0.77 & 0.75 \\
 & TER & 70.37 & 87.87 & 65.43 & 84.13 & 85.32 & 97.69 & 76.97 & 94.56 \\
 & METEOR & 0.53 & 0.42 & 0.49 & 0.37 & 0.42 & 0.33 & 0.39 & 0.3 \\
 & ChrF & 48.73 & 43.93 & 21.84 & 40.96 & 39.64 & 37.19 & 16.1 & 34.82 \\
 & ROUGE-1 & 0.54 & 0.44 & 0.03 & 0.4 & 0.43 & 0.36 & 0.02 & 0.33 \\
 & ROUGE-2 & 0.31 & 0.24 & 0.01 & 0.18 & 0.21 & 0.17 & 0.01 & 0.12 \\
\multirow{-8}{*}{\textbf{\begin{tabular}[c]{@{}l@{}}Mistral-v0.3-7B\\ - Ft.\end{tabular}}} & ROUGE-L & 0.5 & 0.39 & 0.03 & 0.36 & 0.39 & 0.32 & 0.02 & 0.29 \\ \hline
 & BLEU & 17.98 & 10.65 & 18.42 & 8.01 & 4.84 & 1.52 & 1.27 & 1.15 \\
 & BERTScore & 0.8 & 0.76 & 0.78 & 0.75 & 0.85 & 0.61 & 0.46 & 0.6 \\
 & TER & 77.89 & 92.55 & 73.8 & 94.34 & 108.45 & 137.43 & 136.79 & 126.4 \\
 & METEOR & 0.44 & 0.33 & 0.42 & 0.29 & 0.19 & 0.1 & 0.04 & 0.08 \\
 & ChrF & 42.7 & 39.02 & 16.81 & 35.76 & 22.36 & 17.13 & 1.45 & 16.1 \\
 & ROUGE-1 & 0.48 & 0.38 & 0.03 & 0.33 & 0.24 & 0.13 & 0.01 & 0.1 \\
 & ROUGE-2 & 0.24 & 0.18 & 0.02 & 0.12 & 0.08 & 0.03 & 0.0 & 0.02 \\
\multirow{-8}{*}{\textbf{\begin{tabular}[c]{@{}l@{}}Mistral-v0.3-7B\\ - 0 Shot\end{tabular}}} & ROUGE-L & 0.43 & 0.34 & 0.03 & 0.29 & 0.21 & 0.11 & 0.01 & 0.09 \\ \hline
 & BLEU & 21.04 & 11.98 & 17.79 & 9.32 & 4.19 & 1.22 & 2.22 & 1.2 \\
 & BERTScore & 0.8 & 0.78 & 0.78 & 0.75 & 0.87 & 0.61 & 0.54 & 0.61 \\
 & TER & 73.49 & 89.13 & 71.38 & 89.96 & 136.22 & 147.83 & 145.54 & 125.87 \\
 & METEOR & 0.49 & 0.35 & 0.42 & 0.3 & 0.22 & 0.1 & 0.11 & 0.09 \\
 & ChrF & 45.78 & 39.85 & 16.51 & 36.87 & 24.27 & 17.7 & 3.64 & 17.85 \\
 & ROUGE-1 & 0.51 & 0.39 & 0.03 & 0.35 & 0.24 & 0.12 & 0.01 & 0.12 \\
 & ROUGE-2 & 0.27 & 0.19 & 0.01 & 0.13 & 0.08 & 0.03 & 0.0 & 0.02 \\
\multirow{-8}{*}{\textbf{\begin{tabular}[c]{@{}l@{}}Mistral-v0.3-7B\\ - 1 Shot\end{tabular}}} & ROUGE-L & 0.46 & 0.35 & 0.03 & 0.31 & 0.21 & 0.11 & 0.01 & 0.1 \\ \hline
 & BLEU & 20.02 & 11.25 & 18.5 & 8.7 & 6.61 & 1.79 & 5.22 & 1.34 \\
 & BERTScore & 0.81 & 0.77 & 0.79 & 0.76 & 0.86 & 0.62 & 0.6 & 0.6 \\
 & TER & 74.58 & 93.42 & 69.95 & 92.04 & 106.49 & 148.88 & 100.6 & 151.77 \\
 & METEOR & 0.47 & 0.35 & 0.43 & 0.29 & 0.24 & 0.11 & 0.17 & 0.1 \\
 & ChrF & 44.81 & 39.49 & 17.04 & 36.62 & 25.31 & 17.76 & 5.89 & 16.9 \\
 & ROUGE-1 & 0.5 & 0.38 & 0.03 & 0.34 & 0.27 & 0.13 & 0.02 & 0.11 \\
 & ROUGE-2 & 0.26 & 0.18 & 0.01 & 0.12 & 0.09 & 0.04 & 0.01 & 0.02 \\
\multirow{-8}{*}{\textbf{\begin{tabular}[c]{@{}l@{}}Mistral-v0.3-7B\\ - 8 Shot\end{tabular}}} & ROUGE-L & 0.45 & 0.33 & 0.03 & 0.29 & 0.23 & 0.11 & 0.02 & 0.09 \\ \hline
 & BLEU & 21.04 & 12.42 & 18.7 & 9.32 & 7.45 & 2.91 & 6.78 & 02.07 \\
 & BERTScore & 0.81 & 0.78 & 0.79 & 0.76 & 0.87 & 0.66 & 0.62 & 0.64 \\
 & TER & 73.49 & 89.54 & 69.36 & 89.96 & 98.15 & 125.22 & 88.89 & 128.43 \\
 & METEOR & 0.49 & 0.35 & 0.44 & 0.3 & 0.24 & 0.14 & 0.21 & 0.12 \\
 & ChrF & 45.78 & 40.01 & 17.23 & 36.87 & 25.2 & 21.37 & 7.58 & 19.61 \\
 & ROUGE-1 & 0.51 & 0.39 & 0.03 & 0.35 & 0.27 & 0.17 & 0.02 & 0.14 \\
 & ROUGE-2 & 0.27 & 0.19 & 0.01 & 0.13 & 0.1 & 0.05 & 0.01 & 0.03 \\
\multirow{-8}{*}{\textbf{\begin{tabular}[c]{@{}l@{}}Mistral-v0.3-7B\\ - 16 Shot\end{tabular}}} & ROUGE-L & 0.46 & 0.35 & 0.03 & 0.31 & 0.24 & 0.15 & 0.02 & 0.13 \\ \hline
 & BLEU & 19.89 & 11.74 & 18.2 & 8.82 & 7.11 & 08.03 & 13.63 & 6.27 \\
 & BERTScore & 0.81 & 0.77 & 0.79 & 0.76 & 0.87 & 0.75 & 0.72 & 0.73 \\
 & TER & 76.59 & 92.94 & 69.24 & 92.13 & 99.96 & 103.54 & 77.74 & 100.09 \\
 & METEOR & 0.48 & 0.35 & 0.43 & 0.29 & 0.24 & 0.29 & 0.35 & 0.24 \\
 & ChrF & 45.47 & 39.8 & 16.86 & 36.31 & 25.33 & 34.67 & 13.22 & 32.19 \\
 & ROUGE-1 & 0.5 & 0.39 & 0.03 & 0.34 & 0.27 & 0.32 & 0.02 & 0.28 \\
 & ROUGE-2 & 0.26 & 0.18 & 0.01 & 0.12 & 0.1 & 0.13 & 0.01 & 0.09 \\
\multirow{-8}{*}{\textbf{\begin{tabular}[c]{@{}l@{}}Mistral-v0.3-7B\\ - 32 Shot\end{tabular}}} & ROUGE-L & 0.45 & 0.34 & 0.03 & 0.29 & 0.24 & 0.27 & 0.02 & 0.24 \\ \hline
\caption{\textbf{In-context Learning Results}. Full fine-tuning, zero-shot, few-shot on both ground-truth transcript and \gls{ASR} transcript in the cascaded setting. \textbf{Vietnamese to X} results are reported in this table.\\
All cascaded models use Whisper$_{small-mono}$ as \gls{ASR} model (Whisper \gls{ASR} is fine-tuned monolingually - on each source language separately). Its \gls{WER} on test set is 29.6\%, 33.8\%, 31.3\%, 26.3\%, 45.7\% for Vietnamese, English, Chinese, German and French respectively.}
\label{tab:appx_nmt_fewshot-Vi-X}
\end{longtable}

%% file: tables/appx_nmt_fewshot-Fr-X.tex
\setlength{\tabcolsep}{4pt}
\begin{longtable}{l|l|cccc|cccc}
 &  & \multicolumn{4}{c|}{\textbf{Ground-truth}} & \multicolumn{4}{c}{\textbf{ASR}} \\ \cline{3-10} 
\multirow{-2}{*}{\textbf{Model}} & \multirow{-2}{*}{\textbf{Metrics}} & \textbf{fr-en} & \textbf{fr-vi} & \textbf{fr-zh} & \textbf{fr-de} & \textbf{fr-en} & \textbf{fr-vi} & \textbf{fr-zh} & \textbf{fr-de} \\ \hline
\endfirsthead
\multicolumn{10}{c}%
{{\bfseries Table \thetable\ continued from previous page}} \\
 &  & \multicolumn{4}{c|}{\textbf{Ground-truth}} & \multicolumn{4}{c}{\textbf{ASR}} \\ \cline{3-10} 
\multirow{-2}{*}{\textbf{Model}} & \multirow{-2}{*}{\textbf{Metrics}} & \textbf{fr-en} & \textbf{fr-vi} & \textbf{fr-zh} & \textbf{fr-de} & \textbf{fr-en} & \textbf{fr-vi} & \textbf{fr-zh} & \textbf{fr-de} \\ \hline
\endhead
\hline
\endfoot
\endlastfoot
 & BLEU & 50.18 & 39.63 & 29.25 & 27.46 & 30.15 & 25.36 & 20.28 & 16.38 \\
 & BERTScore & 0.95 & 0.86 & 0.79 & 0.81 & 0.82 & 0.8 & 0.75 & 0.74 \\
 & TER & 42.23 & 54.71 & 68.66 & 82.2 & 65.8 & 71.69 & 80.06 & 99.6 \\
 & METEOR & 0.76 & 0.65 & 0.52 & 0.56 & 0.52 & 0.47 & 0.4 & 0.4 \\
 & ChrF & 69.44 & 55.16 & 25.3 & 56.35 & 49.71 & 41.24 & 17.84 & 44.43 \\
 & ROUGE-1 & 0.75 & 0.77 & 0.11 & 0.54 & 0.58 & 0.67 & 0.08 & 0.41 \\
 & ROUGE-2 & 0.58 & 0.56 & 0.07 & 0.34 & 0.39 & 0.44 & 0.05 & 0.22 \\
\multirow{-8}{*}{\textbf{\begin{tabular}[c]{@{}l@{}}Llama-3.1-8B \\ - Ft.\end{tabular}}} & ROUGE-L & 0.73 & 0.67 & 0.11 & 0.5 & 0.54 & 0.55 & 0.08 & 0.37 \\ \hline
 & BLEU & 38.9 & 26.76 & 6.82 & 26.87 & 22.72 & 15.96 & 07.05 & 15.73 \\
 & BERTScore & 0.86 & 0.82 & 0.62 & 0.82 & 0.91 & 0.78 & 0.59 & 0.77 \\
 & TER & 49.46 & 62.96 & 159.39 & 61.65 & 68.02 & 75.66 & 133.79 & 75.69 \\
 & METEOR & 0.66 & 0.52 & 0.22 & 0.53 & 0.46 & 0.37 & 0.2 & 0.36 \\
 & ChrF & 61.25 & 44.58 & 8.95 & 51.9 & 45.22 & 33.7 & 7.82 & 39.92 \\
 & ROUGE-1 & 0.69 & 0.69 & 0.07 & 0.57 & 0.54 & 0.61 & 0.07 & 0.44 \\
 & ROUGE-2 & 0.49 & 0.47 & 0.05 & 0.33 & 0.35 & 0.37 & 0.05 & 0.23 \\
\multirow{-8}{*}{\textbf{\begin{tabular}[c]{@{}l@{}}Llama-3.1-8B\\ - 0 Shot\end{tabular}}} & ROUGE-L & 0.66 & 0.59 & 0.07 & 0.52 & 0.5 & 0.48 & 0.06 & 0.4 \\ \hline
 & BLEU & 48.24 & 34.22 & 21.89 & 31.84 & 25.39 & 19.62 & 14.37 & 17.22 \\
 & BERTScore & 0.89 & 0.85 & 0.79 & 0.84 & 0.92 & 0.79 & 0.73 & 0.78 \\
 & TER & 41.44 & 55.64 & 68.08 & 57.83 & 65.29 & 71.64 & 76.66 & 73.8 \\
 & METEOR & 0.73 & 0.6 & 0.46 & 0.57 & 0.48 & 0.42 & 0.34 & 0.38 \\
 & ChrF & 67.71 & 51.17 & 19.56 & 55.77 & 47.3 & 37.8 & 14.16 & 41.72 \\
 & ROUGE-1 & 0.75 & 0.75 & 0.13 & 0.6 & 0.56 & 0.66 & 0.09 & 0.46 \\
 & ROUGE-2 & 0.56 & 0.53 & 0.1 & 0.37 & 0.36 & 0.41 & 0.07 & 0.24 \\
\multirow{-8}{*}{\textbf{\begin{tabular}[c]{@{}l@{}}Llama-3.1-8B\\ - 1 Shot\end{tabular}}} & ROUGE-L & 0.72 & 0.64 & 0.13 & 0.56 & 0.52 & 0.52 & 0.09 & 0.42 \\ \hline
 & BLEU & 48.25 & 34.08 & 20.97 & 32.13 & 28.69 & 20.92 & 14.97 & 19.07 \\
 & BERTScore & 0.89 & 0.86 & 0.79 & 0.85 & 0.91 & 0.8 & 0.73 & 0.78 \\
 & TER & 41.9 & 55.64 & 69.83 & 58.46 & 62.69 & 71.54 & 75.89 & 73.3 \\
 & METEOR & 0.73 & 0.6 & 0.44 & 0.57 & 0.51 & 0.42 & 0.34 & 0.4 \\
 & ChrF & 67.69 & 51.65 & 18.89 & 55.53 & 49.39 & 38.34 & 14.81 & 43.09 \\
 & ROUGE-1 & 0.74 & 0.76 & 0.12 & 0.6 & 0.58 & 0.65 & 0.08 & 0.47 \\
 & ROUGE-2 & 0.55 & 0.53 & 0.09 & 0.37 & 0.39 & 0.41 & 0.06 & 0.26 \\
\multirow{-8}{*}{\textbf{\begin{tabular}[c]{@{}l@{}}Llama-3.1-8B\\ - 8 Shot\end{tabular}}} & ROUGE-L & 0.72 & 0.65 & 0.12 & 0.56 & 0.54 & 0.52 & 0.08 & 0.43 \\ \hline
 & BLEU & 48.24 & 34.22 & 21.89 & 31.84 & 29.17 & 22.03 & 16.74 & 20.64 \\
 & BERTScore & 0.89 & 0.85 & 0.8 & 0.84 & 0.92 & 0.79 & 0.73 & 0.79 \\
 & TER & 41.44 & 55.64 & 68.08 & 57.83 & 62.62 & 70.32 & 73.95 & 73.17 \\
 & METEOR & 0.73 & 0.6 & 0.46 & 0.57 & 0.51 & 0.43 & 0.36 & 0.4 \\
 & ChrF & 67.71 & 51.17 & 19.56 & 55.77 & 49.83 & 38.95 & 16.22 & 43.47 \\
 & ROUGE-1 & 0.75 & 0.75 & 0.13 & 0.6 & 0.58 & 0.64 & 0.1 & 0.47 \\
 & ROUGE-2 & 0.56 & 0.53 & 0.1 & 0.37 & 0.4 & 0.41 & 0.07 & 0.27 \\
\multirow{-8}{*}{\textbf{\begin{tabular}[c]{@{}l@{}}Llama-3.1-8B\\ - 16 Shot\end{tabular}}} & ROUGE-L & 0.72 & 0.64 & 0.13 & 0.56 & 0.55 & 0.52 & 0.1 & 0.43 \\ \hline
 & BLEU & 49.7 & 34.83 & 22.38 & 31.31 & 28.66 & 21.09 & 14.24 & 18.48 \\
 & BERTScore & 0.89 & 0.85 & 0.8 & 0.84 & 0.91 & 0.79 & 0.73 & 0.78 \\
 & TER & 40.74 & 56.07 & 68.51 & 58.73 & 63.23 & 71.72 & 75.78 & 74.11 \\
 & METEOR & 0.74 & 0.6 & 0.46 & 0.57 & 0.51 & 0.42 & 0.34 & 0.38 \\
 & ChrF & 68.13 & 51.36 & 19.94 & 55.15 & 49.56 & 38.04 & 14.09 & 42.15 \\
 & ROUGE-1 & 0.75 & 0.75 & 0.13 & 0.6 & 0.58 & 0.64 & 0.09 & 0.46 \\
 & ROUGE-2 & 0.57 & 0.53 & 0.1 & 0.36 & 0.4 & 0.4 & 0.07 & 0.25 \\
\multirow{-8}{*}{\textbf{\begin{tabular}[c]{@{}l@{}}Llama-3.1-8B\\ - 32 Shot\end{tabular}}} & ROUGE-L & 0.72 & 0.64 & 0.13 & 0.55 & 0.55 & 0.51 & 0.09 & 0.42 \\ \hline
 & BLEU & 49.69 & 40.67 & 20.97 & 33.91 & 30.35 & 25.59 & 15.33 & 20.38 \\
 & BERTScore & 0.95 & 0.86 & 0.78 & 0.84 & 0.81 & 0.8 & 0.76 & 0.78 \\
 & TER & 42.17 & 52.57 & 67.39 & 59.28 & 69.6 & 71.42 & 73.88 & 76.51 \\
 & METEOR & 0.76 & 0.66 & 0.43 & 0.58 & 0.52 & 0.47 & 0.36 & 0.4 \\
 & ChrF & 70.03 & 56.12 & 20.86 & 56.82 & 49.88 & 41.85 & 15.7 & 43.63 \\
 & ROUGE-1 & 0.76 & 0.78 & 0.09 & 0.6 & 0.57 & 0.67 & 0.07 & 0.47 \\
 & ROUGE-2 & 0.58 & 0.58 & 0.06 & 0.37 & 0.39 & 0.45 & 0.05 & 0.26 \\
\multirow{-8}{*}{\textbf{\begin{tabular}[c]{@{}l@{}}Qwen-2.5-7B\\ - Ft.\end{tabular}}} & ROUGE-L & 0.73 & 0.68 & 0.08 & 0.56 & 0.54 & 0.55 & 0.07 & 0.43 \\ \hline
 & BLEU & 39.14 & 27.77 & 26.36 & 24.27 & 22.81 & 16.46 & 21.05 & 13.9 \\
 & BERTScore & 0.83 & 0.81 & 0.8 & 0.8 & 0.89 & 0.75 & 0.73 & 0.74 \\
 & TER & 59.08 & 65.38 & 80.39 & 69.54 & 76.68 & 80.86 & 84.82 & 81.83 \\
 & METEOR & 0.61 & 0.5 & 0.56 & 0.47 & 0.43 & 0.34 & 0.4 & 0.32 \\
 & ChrF & 57.98 & 44.54 & 26.81 & 48.68 & 43.3 & 32.15 & 18.48 & 36.85 \\
 & ROUGE-1 & 0.61 & 0.67 & 0.13 & 0.5 & 0.48 & 0.56 & 0.09 & 0.38 \\
 & ROUGE-2 & 0.43 & 0.45 & 0.1 & 0.27 & 0.29 & 0.33 & 0.07 & 0.18 \\
\multirow{-8}{*}{\textbf{\begin{tabular}[c]{@{}l@{}}Qwen-2.5-7B\\ - 0 Shot\end{tabular}}} & ROUGE-L & 0.58 & 0.56 & 0.13 & 0.46 & 0.44 & 0.44 & 0.09 & 0.34 \\ \hline
 & BLEU & 44.9 & 31.33 & 18.7 & 26.55 & 27.03 & 16.24 & 20.12 & 16.14 \\
 & BERTScore & 0.89 & 0.83 & 0.79 & 0.81 & 0.91 & 0.78 & 0.74 & 0.77 \\
 & TER & 51.04 & 60.82 & 68.76 & 64.25 & 65.99 & 74.57 & 74.12 & 76.47 \\
 & METEOR & 0.72 & 0.55 & 0.42 & 0.5 & 0.49 & 0.37 & 0.4 & 0.36 \\
 & ChrF & 66.38 & 47.26 & 18.54 & 49.7 & 48.73 & 33.57 & 18.39 & 40.39 \\
 & ROUGE-1 & 0.71 & 0.7 & 0.09 & 0.53 & 0.57 & 0.59 & 0.1 & 0.44 \\
 & ROUGE-2 & 0.53 & 0.49 & 0.07 & 0.31 & 0.37 & 0.37 & 0.08 & 0.22 \\
\multirow{-8}{*}{\textbf{\begin{tabular}[c]{@{}l@{}}Qwen-2.5-7B\\ - 1 Shot\end{tabular}}} & ROUGE-L & 0.68 & 0.6 & 0.09 & 0.49 & \cellcolor[HTML]{FFFFFF}0.53 & \cellcolor[HTML]{FFFFFF}0.47 & \cellcolor[HTML]{FFFFFF}0.1 & \cellcolor[HTML]{FFFFFF}0.39 \\ \hline
 & BLEU & 39.58 & 32.28 & 23.72 & 23.08 & 28.75 & 20.7 & 14.51 & 13.75 \\
 & BERTScore & 0.85 & 0.84 & 0.79 & 0.76 & 0.91 & 0.78 & 0.73 & 0.72 \\
 & TER & 62.65 & 58.75 & 72.23 & 71.32 & 76.87 & 74.53 & 75.67 & 83.31 \\
 & METEOR & 0.7 & 0.57 & 0.46 & 0.42 & 0.5 & 0.41 & 0.34 & 0.31 \\
 & ChrF & 64.76 & 48.86 & 21.33 & 43.77 & 48.97 & 37.33 & 14.89 & 35.02 \\
 & ROUGE-1 & 0.67 & 0.72 & 0.1 & 0.45 & 0.53 & 0.62 & 0.08 & 0.36 \\
 & ROUGE-2 & 0.49 & 0.51 & 0.08 & 0.25 & 0.35 & 0.39 & 0.06 & 0.18 \\
\multirow{-8}{*}{\textbf{\begin{tabular}[c]{@{}l@{}}Qwen-2.5-7B\\ - 8 Shot\end{tabular}}} & ROUGE-L & 0.64 & 0.62 & 0.1 & 0.41 & 0.5 & 0.5 & 0.07 & 0.33 \\ \hline
 & BLEU & 44.9 & 31.33 & 18.7 & 26.55 & 28.7 & 21.72 & 14.96 & 18.01 \\
 & BERTScore & 0.87 & 0.83 & 0.79 & 0.81 & 0.91 & 0.78 & 0.74 & 0.77 \\
 & TER & 51.04 & 60.82 & 68.76 & 64.25 & 76.32 & 75.03 & 75.23 & 76.62 \\
 & METEOR & 0.72 & 0.55 & 0.42 & 0.5 & 0.51 & 0.42 & 0.35 & 0.38 \\
 & ChrF & 66.38 & 47.26 & 18.54 & 49.7 & 49.28 & 38.09 & 15.52 & 41.47 \\
 & ROUGE-1 & 0.71 & 0.7 & 0.09 & 0.53 & 0.54 & 0.63 & 0.07 & 0.44 \\
 & ROUGE-2 & 0.53 & 0.49 & 0.07 & 0.31 & 0.36 & 0.4 & 0.05 & 0.24 \\
\multirow{-8}{*}{\textbf{\begin{tabular}[c]{@{}l@{}}Qwen-2.5-7B\\ - 16 Shot\end{tabular}}} & ROUGE-L & 0.68 & 0.6 & 0.09 & 0.49 & 0.51 & 0.51 & 0.07 & 0.4 \\ \hline
 & BLEU & 49.71 & 34.93 & 19.4 & 30.03 & 29.17 & 21.13 & 13.88 & 17.91 \\
 & BERTScore & 0.9 & 0.85 & 0.79 & 0.84 & 0.91 & 0.79 & 0.73 & 0.78 \\
 & TER & 41.23 & 55.79 & 68.27 & 60.61 & 66.49 & 74.12 & 74.79 & 76.11 \\
 & METEOR & 0.74 & 0.6 & 0.42 & 0.55 & 0.51 & 0.42 & 0.34 & 0.38 \\
 & ChrF & 68.28 & 51.18 & 19.2 & 54.6 & 49.55 & 38.05 & 14.66 & 41.88 \\
 & ROUGE-1 & 0.75 & 0.74 & 0.09 & 0.58 & 0.57 & 0.64 & 0.07 & 0.45 \\
 & ROUGE-2 & 0.56 & 0.53 & 0.08 & 0.34 & 0.38 & 0.4 & 0.06 & 0.24 \\
\multirow{-8}{*}{\textbf{\begin{tabular}[c]{@{}l@{}}Qwen-2.5-7B\\ - 32 Shot\end{tabular}}} & ROUGE-L & 0.72 & 0.64 & 0.09 & 0.53 & 0.54 & 0.51 & 0.07 & 0.41 \\ \hline
 & BLEU & 42.49 & 14.47 & 19.92 & 33.73 & 29.35 & 9.2 & 13.94 & 18.65 \\
 & BERTScore & 0.93 & 0.79 & 0.78 & 0.85 & 0.79 & 0.74 & 0.76 & 0.78 \\
 & TER & 56.11 & 74.72 & 67.02 & 57.73 & 74.28 & 82.83 & 74.11 & 74.55 \\
 & METEOR & 0.75 & 0.36 & 0.42 & 0.58 & 0.52 & 0.26 & 0.34 & 0.39 \\
 & ChrF & 68.36 & 30.93 & 20.91 & 56.06 & 49.85 & 23.69 & 15.16 & 41.66 \\
 & ROUGE-1 & 0.72 & 0.6 & 0.08 & 0.61 & 0.55 & 0.52 & 0.07 & 0.46 \\
 & ROUGE-2 & 0.55 & 0.37 & 0.05 & 0.38 & 0.38 & 0.27 & 0.05 & 0.25 \\
\multirow{-8}{*}{\textbf{\begin{tabular}[c]{@{}l@{}}Mistral-v0.3-7B\\ - Ft.\end{tabular}}} & ROUGE-L & 0.7 & 0.49 & 0.08 & 0.57 & 0.52 & 0.4 & 0.07 & 0.42 \\ \hline
 & BLEU & 42.47 & 4.86 & 22.46 & 21.79 & 25.09 & 2.45 & 14.33 & 12.68 \\
 & BERTScore & 0.87 & 0.7 & 0.78 & 0.81 & 0.91 & 0.65 & 0.7 & 0.75 \\
 & TER & 47.18 & 86.55 & 72.83 & 67.21 & 67.69 & 92.55 & 79.4 & 82.32 \\
 & METEOR & 0.7 & 0.2 & 0.45 & 0.48 & 0.48 & 0.12 & 0.33 & 0.32 \\
 & ChrF & 64.09 & 19.74 & 19.69 & 48.49 & 47.21 & 14.7 & 13.6 & 36.69 \\
 & ROUGE-1 & 0.71 & 0.47 & 0.12 & 0.52 & 0.55 & 0.39 & 0.08 & 0.39 \\
 & ROUGE-2 & 0.5 & 0.21 & 0.09 & 0.28 & 0.35 & 0.14 & 0.06 & 0.19 \\
\multirow{-8}{*}{\textbf{\begin{tabular}[c]{@{}l@{}}Mistral-v0.3-7B\\ - 0 Shot\end{tabular}}} & ROUGE-L & 0.68 & 0.35 & 0.12 & 0.48 & 0.51 & 0.29 & 0.08 & 0.35 \\ \hline
 & BLEU & 46.08 & 10.11 & 22.5 & 24.75 & 27.81 & 5.32 & 13.14 & 15.01 \\
 & BERTScore & 0.89 & 0.78 & 0.8 & 0.82 & 0.92 & 0.73 & 0.72 & 0.77 \\
 & TER & 44.67 & 78.64 & 67.06 & 64.05 & 63.69 & 85.9 & 75.34 & 76.91 \\
 & METEOR & 0.73 & 0.3 & 0.46 & 0.51 & 0.51 & 0.21 & 0.33 & 0.35 \\
 & ChrF & 66.58 & 26.79 & 20.3 & 50.42 & 49.02 & 20.47 & 13.9 & 39.43 \\
 & ROUGE-1 & 0.73 & 0.58 & 0.13 & 0.55 & 0.58 & 0.51 & 0.08 & 0.43 \\
 & ROUGE-2 & 0.53 & 0.31 & 0.09 & 0.3 & 0.39 & 0.23 & 0.06 & 0.21 \\
\multirow{-8}{*}{\textbf{\begin{tabular}[c]{@{}l@{}}Mistral-v0.3-7B\\ - 1 Shot\end{tabular}}} & ROUGE-L & 0.7 & 0.44 & 0.13 & 0.51 & 0.54 & 0.37 & 0.08 & 0.39 \\ \hline
 & BLEU & 48.65 & 9.48 & 15.76 & 27.24 & 30.24 & 8.36 & 12.72 & 17.61 \\
 & BERTScore & 0.89 & 0.77 & 0.78 & 0.84 & 0.92 & 0.76 & 0.73 & 0.78 \\
 & TER & 42.4 & 79.29 & 70.68 & 60.95 & 62.14 & 82.09 & 75.07 & 74.34 \\
 & METEOR & 0.74 & 0.3 & 0.39 & 0.54 & 0.52 & 0.26 & 0.33 & 0.38 \\
 & ChrF & 67.56 & 26.29 & 16.65 & 52.79 & 50.59 & 24.14 & 14.24 & 41.48 \\
 & ROUGE-1 & 0.74 & 0.57 & 0.1 & 0.58 & 0.58 & 0.55 & 0.07 & 0.46 \\
 & ROUGE-2 & 0.55 & 0.3 & 0.07 & 0.33 & 0.4 & 0.28 & 0.06 & 0.24 \\
\multirow{-8}{*}{\textbf{\begin{tabular}[c]{@{}l@{}}Mistral-v0.3-7B\\ - 8 Shot\end{tabular}}} & ROUGE-L & 0.71 & 0.43 & 0.1 & 0.54 & 0.55 & 0.41 & 0.07 & 0.42 \\ \hline
 & BLEU & 48.33 & 10.11 & 16.15 & 28.19 & 30.89 & 09.06 & 13.81 & 18.76 \\
 & BERTScore & 0.89 & 0.78 & 0.78 & 0.84 & 0.92 & 0.76 & 0.73 & 0.78 \\
 & TER & 41.91 & 78.64 & 70.06 & 60.43 & 61.73 & 81.41 & 74.05 & 72.97 \\
 & METEOR & 0.74 & 0.3 & 0.39 & 0.55 & 0.53 & 0.27 & 0.34 & 0.39 \\
 & ChrF & 67.39 & 26.79 & 17.05 & 53.07 & 50.88 & 24.54 & 15.33 & 42.08 \\
 & ROUGE-1 & 0.75 & 0.58 & 0.1 & 0.59 & 0.59 & 0.55 & 0.08 & 0.47 \\
 & ROUGE-2 & 0.55 & 0.31 & 0.07 & 0.34 & 0.41 & 0.28 & 0.05 & 0.26 \\
\multirow{-8}{*}{\textbf{\begin{tabular}[c]{@{}l@{}}Mistral-v0.3-7B\\ - 16 Shot\end{tabular}}} & ROUGE-L & 0.72 & 0.44 & 0.1 & 0.54 & 0.56 & 0.41 & 0.08 & 0.43 \\ \hline
 & BLEU & 49.35 & 10.57 & 16.67 & 28.83 & 29.19 & 8.12 & 12.39 & 17.13 \\
 & BERTScore & 0.89 & 0.78 & 0.79 & 0.84 & 0.92 & 0.76 & 0.73 & 0.78 \\
 & TER & 41.31 & 78.39 & 69.57 & 60.23 & 63.3 & 82.07 & 75.49 & 73.87 \\
 & METEOR & 0.74 & 0.31 & 0.4 & 0.55 & 0.51 & 0.26 & 0.33 & 0.38 \\
 & ChrF & 67.61 & 27.19 & 17.48 & 53.54 & 49.69 & 23.99 & 13.84 & 41.4 \\
 & ROUGE-1 & 0.75 & 0.58 & 0.1 & 0.59 & 0.58 & 0.54 & 0.07 & 0.46 \\
 & ROUGE-2 & 0.56 & 0.32 & 0.07 & 0.34 & 0.39 & 0.28 & 0.05 & 0.24 \\
\multirow{-8}{*}{\textbf{\begin{tabular}[c]{@{}l@{}}Mistral-v0.3-7B\\ - 32 Shot\end{tabular}}} & ROUGE-L & 0.72 & 0.45 & 0.1 & 0.55 & 0.55 & 0.4 & 0.07 & 0.42 \\ \hline
\caption{\textbf{In-context Learning Results}. Full fine-tuning, zero-shot, few-shot on both ground-truth transcript and \gls{ASR} transcript in the cascaded setting. \textbf{French to X} results are reported in this table.\\
All cascaded models use Whisper$_{small-mono}$ as \gls{ASR} model (Whisper \gls{ASR} is fine-tuned monolingually - on each source language separately). Its \gls{WER} on test set is 29.6\%, 33.8\%, 31.3\%, 26.3\%, 45.7\% for Vietnamese, English, Chinese, German and French respectively.}
\label{tab:appx_nmt_fewshot-Fr-X}
\end{longtable}

%% file: tables/appx_nmt_fewshot-De-X.tex
\setlength{\tabcolsep}{4pt}
\begin{longtable}{l|l|cccc|cccc}
 &  & \multicolumn{4}{c|}{\textbf{Ground-truth}} & \multicolumn{4}{c}{\textbf{ASR}} \\ \cline{3-10} 
\multirow{-2}{*}{\textbf{Model}} & \multirow{-2}{*}{\textbf{Metrics}} & \textbf{de-en} & \textbf{de-vi} & \textbf{de-fr} & \textbf{de-zh} & \textbf{de-en} & \textbf{de-vi} & \textbf{de-fr} & \textbf{de-zh} \\ \hline
\endfirsthead
\multicolumn{10}{c}%
{{\bfseries Table \thetable\ continued from previous page}} \\
 &  & \multicolumn{4}{c|}{\textbf{Ground-truth}} & \multicolumn{4}{c}{\textbf{ASR}} \\ \cline{3-10} 
\multirow{-2}{*}{\textbf{Model}} & \multirow{-2}{*}{\textbf{Metrics}} & \textbf{de-en} & \textbf{de-vi} & \textbf{de-fr} & \textbf{de-zh} & \textbf{de-en} & \textbf{de-vi} & \textbf{de-fr} & \textbf{de-zh} \\ \hline
\endhead
\hline
\endfoot
\endlastfoot
 & BLEU & 49.44 & 40.01 & 33.45 & 31.16 & 40.63 & 33.63 & 26.97 & 26.31 \\
 & BERTScore & 0.95 & 0.87 & 0.84 & 0.81 & 0.86 & 0.84 & 0.8 & 0.79 \\
 & TER & 44.99 & 54.45 & 72.36 & 62.08 & 56.53 & 63.07 & 79.21 & 68.19 \\
 & METEOR & 0.76 & 0.67 & 0.62 & 0.54 & 0.63 & 0.56 & 0.52 & 0.47 \\
 & ChrF & 69.74 & 57.57 & 61.36 & 27.18 & 58.95 & 49.74 & 54.06 & 22.87 \\
 & ROUGE-1 & 0.75 & 0.78 & 0.61 & 0.12 & 0.65 & 0.73 & 0.54 & 0.1 \\
 & ROUGE-2 & 0.57 & 0.58 & 0.42 & 0.09 & 0.48 & 0.51 & 0.35 & 0.07 \\
\multirow{-8}{*}{\textbf{\begin{tabular}[c]{@{}l@{}}Llama-3.1-8B \\ - Ft.\end{tabular}}} & ROUGE-L & 0.72 & 0.68 & 0.57 & 0.12 & 0.62 & 0.61 & 0.49 & 0.1 \\ \hline
 & BLEU & 40.09 & 25.98 & 29.35 & 10.77 & 30.35 & 20.74 & 22.36 & 10.55 \\
 & BERTScore & 0.86 & 0.82 & 0.83 & 0.67 & 0.92 & 0.8 & 0.8 & 0.64 \\
 & TER & 51.7 & 65.04 & 63.64 & 136.05 & 61.48 & 70.86 & 70.74 & 124.82 \\
 & METEOR & 0.66 & 0.5 & 0.54 & 0.32 & 0.54 & 0.43 & 0.44 & 0.28 \\
 & ChrF & 61.07 & 43.55 & 53.42 & 13.06 & 51.9 & 38.24 & 46.34 & 11.47 \\
 & ROUGE-1 & 0.68 & 0.66 & 0.58 & 0.09 & 0.59 & 0.63 & 0.51 & 0.08 \\
 & ROUGE-2 & 0.48 & 0.44 & 0.38 & 0.07 & 0.4 & 0.4 & 0.32 & 0.06 \\
\multirow{-8}{*}{\textbf{\begin{tabular}[c]{@{}l@{}}Llama-3.1-8B\\ - 0 Shot\end{tabular}}} & ROUGE-L & 0.64 & 0.54 & 0.54 & 0.09 & 0.56 & 0.5 & 0.47 & 0.08 \\ \hline
 & BLEU & 49.78 & 34.38 & 35.13 & 22.03 & 34.38 & 24.15 & 25.43 & 20.26 \\
 & BERTScore & 0.9 & 0.86 & 0.86 & 0.79 & 0.93 & 0.82 & 0.82 & 0.76 \\
 & TER & 41.64 & 56.94 & 58.72 & 68.59 & 56.38 & 64.68 & 67.46 & 70.79 \\
 & METEOR & 0.76 & 0.61 & 0.6 & 0.46 & 0.59 & 0.48 & 0.49 & 0.42 \\
 & ChrF & 68.35 & 52.11 & 58.58 & 19.58 & 55.43 & 42.51 & 49.63 & 18.39 \\
 & ROUGE-1 & 0.75 & 0.76 & 0.63 & 0.11 & 0.64 & 0.69 & 0.55 & 0.1 \\
 & ROUGE-2 & 0.56 & 0.53 & 0.42 & 0.08 & 0.44 & 0.45 & 0.35 & 0.08 \\
\multirow{-8}{*}{\textbf{\begin{tabular}[c]{@{}l@{}}Llama-3.1-8B\\ - 1 Shot\end{tabular}}} & ROUGE-L & 0.72 & 0.64 & 0.58 & 0.11 & 0.6 & 0.56 & 0.5 & 0.1 \\ \hline
 & BLEU & 48.83 & 32.76 & 35.32 & 22.4 & 36.04 & 26.67 & 27.05 & 18.32 \\
 & BERTScore & 0.9 & 0.86 & 0.86 & 0.8 & 0.93 & 0.83 & 0.82 & 0.75 \\
 & TER & 42.03 & 56.98 & 57.29 & 65.82 & 55.59 & 63.7 & 66.22 & 73.43 \\
 & METEOR & 0.75 & 0.6 & 0.61 & 0.47 & 0.61 & 0.51 & 0.5 & 0.4 \\
 & ChrF & 67.67 & 50.99 & 58.99 & 20.0 & 56.8 & 44.67 & 51.05 & 16.95 \\
 & ROUGE-1 & 0.75 & 0.76 & 0.63 & 0.12 & 0.64 & 0.7 & 0.56 & 0.1 \\
 & ROUGE-2 & 0.55 & 0.52 & 0.43 & 0.09 & 0.45 & 0.46 & 0.36 & 0.07 \\
\multirow{-8}{*}{\textbf{\begin{tabular}[c]{@{}l@{}}Llama-3.1-8B\\ - 8 Shot\end{tabular}}} & ROUGE-L & 0.71 & 0.63 & 0.59 & 0.12 & 0.61 & 0.58 & 0.51 & 0.1 \\ \hline
 & BLEU & 49.78 & 34.38 & 35.13 & 22.03 & 37.7 & 27.72 & 27.25 & 19.7 \\
 & BERTScore & 0.9 & 0.86 & 0.86 & 0.79 & 0.94 & 0.83 & 0.83 & 0.75 \\
 & TER & 41.64 & 56.94 & 58.72 & 68.59 & 53.54 & 64.49 & 65.72 & 71.5 \\
 & METEOR & 0.76 & 0.61 & 0.6 & 0.46 & 0.62 & 0.51 & 0.5 & 0.41 \\
 & ChrF & 68.35 & 52.11 & 58.58 & 19.58 & 57.92 & 45.3 & 50.88 & 18.11 \\
 & ROUGE-1 & 0.75 & 0.76 & 0.63 & 0.11 & 0.66 & 0.7 & 0.56 & 0.1 \\
 & ROUGE-2 & 0.56 & 0.53 & 0.42 & 0.08 & 0.46 & 0.47 & 0.36 & 0.07 \\
\multirow{-8}{*}{\textbf{\begin{tabular}[c]{@{}l@{}}Llama-3.1-8B\\ - 16 Shot\end{tabular}}} & ROUGE-L & 0.72 & 0.64 & 0.58 & 0.11 & 0.62 & 0.57 & 0.51 & 0.1 \\ \hline
 & BLEU & 49.07 & 35.32 & 34.95 & 23.09 & 36.84 & 27.09 & 26.84 & 19.25 \\
 & BERTScore & 0.9 & 0.86 & 0.86 & 0.8 & 0.93 & 0.83 & 0.83 & 0.76 \\
 & TER & 42.3 & 55.79 & 57.43 & 65.91 & 54.34 & 64.29 & 66.25 & 70.33 \\
 & METEOR & 0.75 & 0.62 & 0.6 & 0.47 & 0.61 & 0.51 & 0.5 & 0.41 \\
 & ChrF & 68.06 & 52.79 & 58.84 & 20.55 & 57.55 & 45.02 & 50.56 & 17.79 \\
 & ROUGE-1 & 0.75 & 0.77 & 0.64 & 0.12 & 0.66 & 0.7 & 0.56 & 0.11 \\
 & ROUGE-2 & 0.55 & 0.53 & 0.43 & 0.1 & 0.46 & 0.46 & 0.36 & 0.08 \\
\multirow{-8}{*}{\textbf{\begin{tabular}[c]{@{}l@{}}Llama-3.1-8B\\ - 32 Shot\end{tabular}}} & ROUGE-L & 0.72 & 0.65 & 0.59 & 0.12 & 0.62 & 0.57 & 0.51 & 0.11 \\ \hline
 & BLEU & 52.1 & 43.73 & 40.72 & 23.26 & 40.52 & 34.24 & 31.45 & 19.87 \\
 & BERTScore & 0.96 & 0.88 & 0.88 & 0.79 & 0.86 & 0.84 & 0.84 & 0.79 \\
 & TER & 39.94 & 48.63 & 51.44 & 63.95 & 55.76 & 59.16 & 61.65 & 68.06 \\
 & METEOR & 0.77 & 0.69 & 0.65 & 0.44 & 0.63 & 0.57 & 0.53 & 0.4 \\
 & ChrF & 70.13 & 59.36 & 62.75 & 23.28 & 59.05 & 50.58 & 54.36 & 20.15 \\
 & ROUGE-1 & 0.76 & 0.8 & 0.67 & 0.11 & 0.66 & 0.74 & 0.59 & 0.1 \\
 & ROUGE-2 & 0.58 & 0.6 & 0.47 & 0.09 & 0.47 & 0.53 & 0.4 & 0.08 \\
\multirow{-8}{*}{\textbf{\begin{tabular}[c]{@{}l@{}}Qwen-2.5-7B\\ - Ft.\end{tabular}}} & ROUGE-L & 0.73 & 0.7 & 0.63 & 0.11 & 0.62 & 0.62 & 0.54 & 0.1 \\ \hline
 & BLEU & 42.12 & 27.26 & 29.93 & 33.09 & 32.03 & 20.9 & 22.74 & 29.78 \\
 & BERTScore & 0.86 & 0.81 & 0.83 & 0.83 & 0.92 & 0.79 & 0.79 & 0.79 \\
 & TER & 54.87 & 65.49 & 63.02 & 63.12 & 65.25 & 71.74 & 70.78 & 67.85 \\
 & METEOR & 0.66 & 0.5 & 0.53 & 0.62 & 0.54 & 0.42 & 0.44 & 0.52 \\
 & ChrF & 61.0 & 43.92 & 53.64 & 31.26 & 51.78 & 37.93 & 46.53 & 25.63 \\
 & ROUGE-1 & 0.65 & 0.66 & 0.56 & 0.12 & 0.57 & 0.62 & 0.49 & 0.11 \\
 & ROUGE-2 & 0.46 & 0.44 & 0.36 & 0.1 & 0.38 & 0.39 & 0.3 & 0.09 \\
\multirow{-8}{*}{\textbf{\begin{tabular}[c]{@{}l@{}}Qwen-2.5-7B\\ - 0 Shot\end{tabular}}} & ROUGE-L & 0.62 & 0.55 & 0.52 & 0.12 & 0.53 & 0.5 & 0.45 & 0.11 \\ \hline
 & BLEU & 40.55 & 28.29 & 32.3 & 19.06 & 35.85 & 23.42 & 25.1 & 19.58 \\
 & BERTScore & 0.89 & 0.83 & 0.85 & 0.79 & 0.93 & 0.81 & 0.82 & 0.76 \\
 & TER & 57.2 & 65.45 & 62.21 & 68.52 & 57.82 & 66.64 & 67.69 & 72.04 \\
 & METEOR & 0.6 & 0.5 & 0.56 & 0.4 & 0.6 & 0.47 & 0.47 & 0.4 \\
 & ChrF & 58.35 & 44.68 & 56.84 & 19.01 & 56.18 & 41.7 & 49.54 & 18.35 \\
 & ROUGE-1 & 0.6 & 0.66 & 0.59 & 0.09 & 0.63 & 0.67 & 0.53 & 0.09 \\
 & ROUGE-2 & 0.42 & 0.44 & 0.38 & 0.07 & 0.43 & 0.44 & 0.33 & 0.07 \\
\multirow{-8}{*}{\textbf{\begin{tabular}[c]{@{}l@{}}Qwen-2.5-7B\\ - 1 Shot\end{tabular}}} & ROUGE-L & 0.57 & 0.55 & 0.55 & 0.09 & \cellcolor[HTML]{FFFFFF}0.6 & \cellcolor[HTML]{FFFFFF}0.55 & \cellcolor[HTML]{FFFFFF}0.49 & \cellcolor[HTML]{FFFFFF}0.09 \\ \hline
 & BLEU & 24.79 & 31.91 & 29.74 & 18.83 & 27.78 & 24.65 & 26.31 & 18.09 \\
 & BERTScore & 0.72 & 0.84 & 0.81 & 0.73 & 0.89 & 0.81 & 0.82 & 0.76 \\
 & TER & 75.48 & 61.27 & 63.77 & 94.17 & 75.66 & 67.64 & 65.23 & 71.02 \\
 & METEOR & 0.39 & 0.57 & 0.51 & 0.35 & 0.46 & 0.46 & 0.49 & 0.38 \\
 & ChrF & 38.36 & 49.02 & 52.61 & 16.91 & 45.07 & 41.71 & 50.37 & 17.82 \\
 & ROUGE-1 & 0.41 & 0.72 & 0.54 & 0.07 & 0.48 & 0.65 & 0.54 & 0.09 \\
 & ROUGE-2 & 0.27 & 0.49 & 0.35 & 0.06 & 0.31 & 0.43 & 0.35 & 0.07 \\
\multirow{-8}{*}{\textbf{\begin{tabular}[c]{@{}l@{}}Qwen-2.5-7B\\ - 8 Shot\end{tabular}}} & ROUGE-L & 0.38 & 0.6 & 0.5 & 0.07 & 0.45 & 0.53 & 0.5 & 0.09 \\ \hline
 & BLEU & 40.55 & 28.29 & 32.3 & 19.06 & 36.96 & 25.86 & 26.52 & 16.81 \\
 & BERTScore & 0.82 & 0.81 & 0.84 & 0.79 & 0.93 & 0.81 & 0.82 & 0.76 \\
 & TER & 57.2 & 65.45 & 62.21 & 68.52 & 60.11 & 68.28 & 65.34 & 70.93 \\
 & METEOR & 0.6 & 0.5 & 0.56 & 0.4 & 0.6 & 0.48 & 0.5 & 0.37 \\
 & ChrF & 58.35 & 44.68 & 56.84 & 19.01 & 56.41 & 42.7 & 51.05 & 17.27 \\
 & ROUGE-1 & 0.6 & 0.66 & 0.59 & 0.09 & 0.62 & 0.66 & 0.55 & 0.09 \\
 & ROUGE-2 & 0.42 & 0.44 & 0.38 & 0.07 & 0.43 & 0.43 & 0.35 & 0.07 \\
\multirow{-8}{*}{\textbf{\begin{tabular}[c]{@{}l@{}}Qwen-2.5-7B\\ - 16 Shot\end{tabular}}} & ROUGE-L & 0.57 & 0.55 & 0.55 & 0.09 & 0.59 & 0.54 & 0.51 & 0.09 \\ \hline
 & BLEU & 48.91 & 33.61 & 35.22 & 20.52 & 37.22 & 26.41 & 26.92 & 17.3 \\
 & BERTScore & 0.9 & 0.84 & 0.86 & 0.8 & 0.93 & 0.81 & 0.83 & 0.76 \\
 & TER & 42.63 & 59.35 & 55.55 & 66.94 & 55.91 & 67.06 & 64.63 & 70.33 \\
 & METEOR & 0.75 & 0.59 & 0.6 & 0.42 & 0.61 & 0.49 & 0.5 & 0.38 \\
 & ChrF & 67.98 & 50.16 & 59.21 & 20.28 & 57.36 & 43.45 & 51.4 & 17.6 \\
 & ROUGE-1 & 0.75 & 0.73 & 0.63 & 0.1 & 0.65 & 0.68 & 0.56 & 0.1 \\
 & ROUGE-2 & 0.55 & 0.51 & 0.42 & 0.08 & 0.45 & 0.45 & 0.36 & 0.07 \\
\multirow{-8}{*}{\textbf{\begin{tabular}[c]{@{}l@{}}Qwen-2.5-7B\\ - 32 Shot\end{tabular}}} & ROUGE-L & 0.71 & 0.61 & 0.59 & 0.1 & 0.61 & 0.55 & 0.51 & 0.09 \\ \hline
 & BLEU & 36.39 & 15.68 & 40.77 & 21.28 & 28.33 & 12.38 & 31.15 & 17.82 \\
 & BERTScore & 0.92 & 0.79 & 0.86 & 0.78 & 0.78 & 0.77 & 0.83 & 0.78 \\
 & TER & 80.09 & 73.73 & 53.76 & 66.07 & 91.13 & 77.89 & 63.4 & 70.11 \\
 & METEOR & 0.72 & 0.35 & 0.63 & 0.42 & 0.6 & 0.3 & 0.52 & 0.37 \\
 & ChrF & 66.77 & 32.48 & 61.34 & 22.37 & 57.3 & 28.44 & 52.73 & 18.9 \\
 & ROUGE-1 & 0.64 & 0.61 & 0.66 & 0.09 & 0.55 & 0.57 & 0.57 & 0.09 \\
 & ROUGE-2 & 0.49 & 0.38 & 0.47 & 0.07 & 0.4 & 0.33 & 0.39 & 0.07 \\
\multirow{-8}{*}{\textbf{\begin{tabular}[c]{@{}l@{}}Mistral-v0.3-7B\\ - Ft.\end{tabular}}} & ROUGE-L & 0.61 & 0.48 & 0.62 & 0.09 & 0.52 & 0.44 & 0.53 & 0.09 \\ \hline
 & BLEU & 44.26 & 6.71 & 28.82 & 27.14 & 34.54 & 4.98 & 21.44 & 21.22 \\
 & BERTScore & 0.89 & 0.73 & 0.84 & 0.81 & 0.93 & 0.71 & 0.8 & 0.75 \\
 & TER & 48.08 & 84.37 & 63.72 & 63.83 & 59.16 & 87.37 & 71.68 & 72.01 \\
 & METEOR & 0.72 & 0.22 & 0.54 & 0.52 & 0.59 & 0.18 & 0.44 & 0.43 \\
 & ChrF & 64.65 & 22.36 & 54.28 & 23.4 & 55.07 & 19.2 & 46.78 & 18.85 \\
 & ROUGE-1 & 0.71 & 0.52 & 0.59 & 0.12 & 0.62 & 0.48 & 0.51 & 0.1 \\
 & ROUGE-2 & 0.5 & 0.24 & 0.36 & 0.1 & 0.42 & 0.2 & 0.3 & 0.08 \\
\multirow{-8}{*}{\textbf{\begin{tabular}[c]{@{}l@{}}Mistral-v0.3-7B\\ - 0 Shot\end{tabular}}} & ROUGE-L & 0.67 & 0.37 & 0.54 & 0.12 & 0.58 & 0.34 & 0.46 & 0.1 \\ \hline
 & BLEU & 48.04 & 10.15 & 30.75 & 24.33 & 36.31 & 6.99 & 22.89 & 16.01 \\
 & BERTScore & 0.9 & 0.76 & 0.85 & 0.81 & 0.93 & 0.77 & 0.82 & 0.74 \\
 & TER & 44.93 & 79.06 & 60.96 & 64.11 & 56.18 & 83.25 & 68.21 & 72.27 \\
 & METEOR & 0.75 & 0.29 & 0.57 & 0.48 & 0.6 & 0.24 & 0.46 & 0.37 \\
 & ChrF & 67.21 & 26.69 & 55.61 & 21.94 & 56.33 & 23.1 & 48.16 & 16.11 \\
 & ROUGE-1 & 0.74 & 0.57 & 0.6 & 0.11 & 0.64 & 0.54 & 0.53 & 0.09 \\
 & ROUGE-2 & 0.54 & 0.3 & 0.38 & 0.09 & 0.44 & 0.26 & 0.32 & 0.07 \\
\multirow{-8}{*}{\textbf{\begin{tabular}[c]{@{}l@{}}Mistral-v0.3-7B\\ - 1 Shot\end{tabular}}} & ROUGE-L & 0.7 & 0.43 & 0.56 & 0.11 & 0.6 & 0.39 & 0.48 & 0.09 \\ \hline
 & BLEU & 47.79 & 9.31 & 31.72 & 15.2 & 37.63 & 8.39 & 24.35 & 13.54 \\
 & BERTScore & 0.9 & 0.77 & 0.85 & 0.78 & 0.93 & 0.76 & 0.82 & 0.75 \\
 & TER & 44.84 & 79.09 & 59.29 & 70.09 & 54.56 & 80.9 & 66.67 & 72.64 \\
 & METEOR & 0.75 & 0.29 & 0.57 & 0.37 & 0.62 & 0.27 & 0.47 & 0.34 \\
 & ChrF & 67.13 & 26.2 & 56.37 & 16.5 & 57.28 & 24.75 & 49.07 & 15.11 \\
 & ROUGE-1 & 0.74 & 0.57 & 0.61 & 0.09 & 0.65 & 0.56 & 0.54 & 0.07 \\
 & ROUGE-2 & 0.54 & 0.3 & 0.39 & 0.07 & 0.46 & 0.28 & 0.34 & 0.06 \\
\multirow{-8}{*}{\textbf{\begin{tabular}[c]{@{}l@{}}Mistral-v0.3-7B\\ - 8 Shot\end{tabular}}} & ROUGE-L & 0.7 & 0.43 & 0.57 & 0.09 & 0.62 & 0.41 & 0.5 & 0.07 \\ \hline
 & BLEU & 48.04 & 10.15 & 32.05 & 15.2 & 38.29 & 9.12 & 25.27 & 14.0 \\
 & BERTScore & 0.9 & 0.77 & 0.85 & 0.78 & 0.93 & 0.77 & 0.82 & 0.74 \\
 & TER & 44.93 & 79.06 & 59.36 & 70.52 & 55.25 & 80.37 & 66.71 & 72.83 \\
 & METEOR & 0.75 & 0.29 & 0.58 & 0.37 & 0.62 & 0.27 & 0.48 & 0.34 \\
 & ChrF & 67.21 & 26.69 & 56.27 & 16.48 & 57.55 & 25.3 & 49.37 & 15.38 \\
 & ROUGE-1 & 0.74 & 0.57 & 0.61 & 0.09 & 0.65 & 0.56 & 0.54 & 0.09 \\
 & ROUGE-2 & 0.54 & 0.3 & 0.4 & 0.07 & 0.46 & 0.29 & 0.34 & 0.07 \\
\multirow{-8}{*}{\textbf{\begin{tabular}[c]{@{}l@{}}Mistral-v0.3-7B\\ - 16 Shot\end{tabular}}} & ROUGE-L & 0.7 & 0.43 & 0.57 & 0.09 & 0.61 & 0.41 & 0.5 & 0.09 \\ \hline
 & BLEU & 48.21 & 10.66 & 31.71 & 16.08 & 37.85 & 8.81 & 24.2 & 13.87 \\
 & BERTScore & 0.9 & 0.78 & 0.85 & 0.78 & 0.93 & 0.76 & 0.82 & 0.75 \\
 & TER & 44.1 & 78.22 & 58.9 & 70.11 & 55.57 & 80.55 & 66.83 & 72.73 \\
 & METEOR & 0.75 & 0.3 & 0.57 & 0.38 & 0.62 & 0.27 & 0.47 & 0.34 \\
 & ChrF & 67.3 & 27.12 & 56.11 & 17.19 & 57.32 & 25.25 & 48.92 & 15.29 \\
 & ROUGE-1 & 0.74 & 0.57 & 0.61 & 0.08 & 0.65 & 0.56 & 0.54 & 0.07 \\
 & ROUGE-2 & 0.54 & 0.31 & 0.39 & 0.06 & 0.45 & 0.29 & 0.33 & 0.06 \\
\multirow{-8}{*}{\textbf{\begin{tabular}[c]{@{}l@{}}Mistral-v0.3-7B\\ - 32 Shot\end{tabular}}} & ROUGE-L & 0.71 & 0.43 & 0.56 & 0.08 & 0.62 & 0.41 & 0.5 & 0.07 \\ \hline
\caption{\textbf{In-context Learning Results}. Full fine-tuning, zero-shot, few-shot on both ground-truth transcript and \gls{ASR} transcript in the cascaded setting. \textbf{German to X} results are reported in this table.\\
All cascaded models use Whisper$_{small-mono}$ as \gls{ASR} model (Whisper \gls{ASR} is fine-tuned monolingually - on each source language separately). Its \gls{WER} on test set is 29.6\%, 33.8\%, 31.3\%, 26.3\%, 45.7\% for Vietnamese, English, Chinese, German and French respectively.}
\label{tab:appx_nmt_fewshot-De-X}
\end{longtable}

%% file: tables/appx_nmt_fewshot-Zh-X.tex
\setlength{\tabcolsep}{4pt}
\begin{longtable}{l|l|cccc|cccc}
 &  & \multicolumn{4}{c|}{\textbf{Ground-truth}} & \multicolumn{4}{c}{\textbf{ASR}} \\ \cline{3-10} 
\multirow{-2}{*}{\textbf{Model}} & \multirow{-2}{*}{\textbf{Metrics}} & \textbf{zh-en} & \textbf{zh-vi} & \textbf{zh-fr} & \textbf{zh-de} & \textbf{zh-en} & \textbf{zh-vi} & \textbf{zh-fr} & \textbf{zh-de} \\ \hline
\endfirsthead
\multicolumn{10}{c}%
{{\bfseries Table \thetable\ continued from previous page}} \\
 &  & \multicolumn{4}{c|}{\textbf{Ground-truth}} & \multicolumn{4}{c}{\textbf{ASR}} \\ \cline{3-10} 
\multirow{-2}{*}{\textbf{Model}} & \multirow{-2}{*}{\textbf{Metrics}} & \textbf{zh-en} & \textbf{zh-vi} & \textbf{zh-fr} & \textbf{zh-de} & \textbf{zh-en} & \textbf{zh-vi} & \textbf{zh-fr} & \textbf{zh-de} \\ \hline
\endhead
\hline
\endfoot
\endlastfoot
 & BLEU & 28.21 & 23.49 & 18.87 & 13.07 & 19.01 & 17.65 & 13.84 & 11.13 \\
 & BERTScore & 0.91 & 0.79 & 0.77 & 0.74 & 0.78 & 0.75 & 0.75 & 0.74 \\
 & TER & 84.4 & 96.2 & 99.18 & 107.48 & 101.13 & 104.68 & 106.77 & 109.71 \\
 & METEOR & 0.59 & 0.55 & 0.46 & 0.4 & 0.48 & 0.45 & 0.39 & 0.35 \\
 & ChrF & 47.42 & 41.33 & 41.69 & 32.09 & 40.96 & 34.68 & 36.97 & 32.37 \\
 & ROUGE-1 & 0.57 & 0.63 & 0.46 & 0.37 & 0.47 & 0.58 & 0.39 & 0.34 \\
 & ROUGE-2 & 0.37 & 0.44 & 0.28 & 0.19 & 0.27 & 0.36 & 0.21 & 0.16 \\
\multirow{-8}{*}{\textbf{\begin{tabular}[c]{@{}l@{}}Llama-3.1-8B \\ - Ft.\end{tabular}}} & ROUGE-L & 0.53 & 0.53 & 0.41 & 0.33 & 0.43 & 0.46 & 0.34 & 0.3 \\ \hline
 & BLEU & 16.21 & 16.01 & 11.95 & 7.41 & 12.29 & 12.17 & 8.78 & 06.02 \\
 & BERTScore & 0.79 & 0.86 & 0.76 & 0.75 & 0.89 & 0.77 & 0.74 & 0.73 \\
 & TER & 80.87 & 79.51 & 85.88 & 91.67 & 89.32 & 87.12 & 93.56 & 95.73 \\
 & METEOR & 0.46 & 0.44 & 0.35 & 0.31 & 0.37 & 0.35 & 0.29 & 0.26 \\
 & ChrF & 36.29 & 31.91 & 33.47 & 26.79 & 32.05 & 27.57 & 28.97 & 23.41 \\
 & ROUGE-1 & 0.47 & 0.61 & 0.4 & 0.35 & 0.39 & 0.55 & 0.33 & 0.28 \\
 & ROUGE-2 & 0.25 & 0.38 & 0.2 & 0.15 & 0.18 & 0.31 & 0.15 & 0.12 \\
\multirow{-8}{*}{\textbf{\begin{tabular}[c]{@{}l@{}}Llama-3.1-8B\\ - 0 Shot\end{tabular}}} & ROUGE-L & 0.43 & 0.49 & 0.35 & 0.31 & 0.35 & 0.43 & 0.28 & 0.25 \\ \hline
 & BLEU & 27.32 & 24.22 & 16.23 & 12.69 & 15.79 & 13.94 & 9.28 & 08.02 \\
 & BERTScore & 0.85 & 0.91 & 0.81 & 0.78 & 0.92 & 0.8 & 0.78 & 0.74 \\
 & TER & 65.78 & 70.48 & 79.85 & 83.63 & 84.27 & 84.98 & 90.0 & 92.65 \\
 & METEOR & 0.59 & 0.55 & 0.44 & 0.41 & 0.43 & 0.39 & 0.31 & 0.3 \\
 & ChrF & 48.7 & 41.34 & 40.15 & 35.9 & 36.21 & 29.88 & 30.57 & 25.37 \\
 & ROUGE-1 & 0.59 & 0.7 & 0.47 & 0.44 & 0.44 & 0.6 & 0.35 & 0.32 \\
 & ROUGE-2 & 0.36 & 0.47 & 0.27 & 0.21 & 0.22 & 0.34 & 0.17 & 0.15 \\
\multirow{-8}{*}{\textbf{\begin{tabular}[c]{@{}l@{}}Llama-3.1-8B\\ - 1 Shot\end{tabular}}} & ROUGE-L & 0.55 & 0.58 & 0.42 & 0.4 & 0.39 & 0.46 & 0.3 & 0.29 \\ \hline
 & BLEU & 27.76 & 24.74 & 15.35 & 11.82 & 21.16 & 19.56 & 14.14 & 12.08 \\
 & BERTScore & 0.85 & 0.91 & 0.8 & 0.8 & 0.92 & 0.8 & 0.78 & 0.79 \\
 & TER & 66.94 & 69.48 & 79.05 & 82.4 & 80.6 & 81.08 & 89.56 & 87.74 \\
 & METEOR & 0.59 & 0.56 & 0.43 & 0.4 & 0.48 & 0.44 & 0.36 & 0.36 \\
 & ChrF & 48.41 & 41.96 & 39.48 & 34.26 & 41.81 & 35.35 & 36.66 & 33.07 \\
 & ROUGE-1 & 0.59 & 0.71 & 0.46 & 0.44 & 0.49 & 0.64 & 0.4 & 0.4 \\
 & ROUGE-2 & 0.36 & 0.48 & 0.25 & 0.2 & 0.28 & 0.38 & 0.21 & 0.21 \\
\multirow{-8}{*}{\textbf{\begin{tabular}[c]{@{}l@{}}Llama-3.1-8B\\ - 8 Shot\end{tabular}}} & ROUGE-L & 0.55 & 0.59 & 0.42 & 0.39 & 0.44 & 0.5 & 0.34 & 0.36 \\ \hline
 & BLEU & 27.32 & 24.22 & 16.23 & 12.69 & 24.3 & 21.86 & 16.72 & 17.19 \\
 & BERTScore & 0.85 & 0.91 & 0.81 & 0.81 & 0.92 & 0.8 & 0.78 & 0.79 \\
 & TER & 65.78 & 70.48 & 79.85 & 83.63 & 78.01 & 79.45 & 86.03 & 85.58 \\
 & METEOR & 0.59 & 0.55 & 0.44 & 0.41 & 0.49 & 0.46 & 0.38 & 0.39 \\
 & ChrF & 48.7 & 41.34 & 40.15 & 35.9 & 44.39 & 37.37 & 38.22 & 37.05 \\
 & ROUGE-1 & 0.59 & 0.7 & 0.47 & 0.44 & 0.5 & 0.65 & 0.41 & 0.42 \\
 & ROUGE-2 & 0.36 & 0.47 & 0.27 & 0.21 & 0.29 & 0.4 & 0.22 & 0.23 \\
\multirow{-8}{*}{\textbf{\begin{tabular}[c]{@{}l@{}}Llama-3.1-8B\\ - 16 Shot\end{tabular}}} & ROUGE-L & 0.55 & 0.58 & 0.42 & 0.4 & 0.45 & 0.51 & 0.36 & 0.38 \\ \hline
 & BLEU & 28.69 & 24.39 & 16.43 & 12.34 & 19.18 & 19.54 & 12.37 & 10.51 \\
 & BERTScore & 0.85 & 0.91 & 0.81 & 0.81 & 0.92 & 0.81 & 0.78 & 0.78 \\
 & TER & 67.3 & 69.1 & 81.45 & 82.31 & 81.2 & 79.47 & 88.9 & 89.66 \\
 & METEOR & 0.6 & 0.55 & 0.45 & 0.42 & 0.47 & 0.46 & 0.36 & 0.37 \\
 & ChrF & 49.42 & 41.82 & 40.47 & 36.57 & 41.43 & 36.04 & 36.19 & 33.27 \\
 & ROUGE-1 & 0.6 & 0.71 & 0.47 & 0.45 & 0.48 & 0.65 & 0.4 & 0.4 \\
 & ROUGE-2 & 0.37 & 0.47 & 0.27 & 0.21 & 0.26 & 0.39 & 0.2 & 0.18 \\
\multirow{-8}{*}{\textbf{\begin{tabular}[c]{@{}l@{}}Llama-3.1-8B\\ - 32 Shot\end{tabular}}} & ROUGE-L & 0.55 & 0.59 & 0.42 & 0.41 & 0.43 & 0.51 & 0.35 & 0.34 \\ \hline
 & BLEU & 35.63 & 32.95 & 24.05 & 16.95 & 25.36 & 26.31 & 17.84 & 12.61 \\
 & BERTScore & 0.95 & 0.85 & 0.84 & 0.83 & 0.82 & 0.81 & 0.79 & 0.78 \\
 & TER & 60.98 & 67.86 & 73.13 & 77.31 & 79.48 & 80.4 & 84.11 & 91.04 \\
 & METEOR & 0.67 & 0.63 & 0.53 & 0.5 & 0.53 & 0.53 & 0.43 & 0.4 \\
 & ChrF & 55.3 & 47.73 & 47.09 & 39.65 & 45.87 & 41.37 & 40.88 & 34.86 \\
 & ROUGE-1 & 0.66 & 0.74 & 0.55 & 0.52 & 0.53 & 0.67 & 0.45 & 0.42 \\
 & ROUGE-2 & 0.45 & 0.53 & 0.35 & 0.28 & 0.32 & 0.45 & 0.26 & 0.2 \\
\multirow{-8}{*}{\textbf{\begin{tabular}[c]{@{}l@{}}Qwen-2.5-7B\\ - Ft.\end{tabular}}} & ROUGE-L & 0.62 & 0.63 & 0.5 & 0.48 & 0.48 & 0.55 & 0.4 & 0.37 \\ \hline
 & BLEU & 22.51 & 15.93 & 17.86 & 7.58 & 16.9 & 14.42 & 12.07 & 7.78 \\
 & BERTScore & 0.83 & 0.82 & 0.8 & 0.77 & 0.91 & 0.77 & 0.77 & 0.74 \\
 & TER & 74.01 & 76.17 & 78.49 & 87.24 & 87.43 & 85.97 & 88.63 & 95.0 \\
 & METEOR & 0.53 & 0.41 & 0.44 & 0.34 & 0.43 & 0.37 & 0.35 & 0.31 \\
 & ChrF & 44.85 & 32.32 & 40.84 & 29.61 & 37.97 & 29.62 & 33.97 & 28.17 \\
 & ROUGE-1 & 0.56 & 0.61 & 0.47 & 0.39 & 0.44 & 0.57 & 0.39 & 0.34 \\
 & ROUGE-2 & 0.32 & 0.39 & 0.27 & 0.16 & 0.23 & 0.34 & 0.2 & 0.14 \\
\multirow{-8}{*}{\textbf{\begin{tabular}[c]{@{}l@{}}Qwen-2.5-7B\\ - 0 Shot\end{tabular}}} & ROUGE-L & 0.51 & 0.5 & 0.42 & 0.34 & 0.4 & 0.45 & 0.34 & 0.3 \\ \hline
 & BLEU & 28.38 & 22.63 & 20.23 & 11.97 & 20.61 & 16.91 & 13.16 & 9.4 \\
 & BERTScore & 0.86 & 0.85 & 0.81 & 0.8 & 0.91 & 0.8 & 0.79 & 0.77 \\
 & TER & 67.96 & 71.83 & 76.08 & 82.62 & 83.98 & 79.69 & 87.86 & 93.12 \\
 & METEOR & 0.59 & 0.5 & 0.48 & 0.42 & 0.49 & 0.41 & 0.35 & 0.33 \\
 & ChrF & 47.86 & 37.41 & 43.55 & 34.08 & 42.1 & 33.0 & 35.74 & 30.35 \\
 & ROUGE-1 & 0.59 & 0.64 & 0.5 & 0.46 & 0.49 & 0.61 & 0.39 & 0.36 \\
 & ROUGE-2 & 0.38 & 0.44 & 0.31 & 0.22 & 0.27 & 0.38 & 0.2 & 0.16 \\
\multirow{-8}{*}{\textbf{\begin{tabular}[c]{@{}l@{}}Qwen-2.5-7B\\ - 1 Shot\end{tabular}}} & ROUGE-L & 0.55 & 0.54 & 0.46 & 0.42 & \cellcolor[HTML]{FFFFFF}0.44 & \cellcolor[HTML]{FFFFFF}0.48 & \cellcolor[HTML]{FFFFFF}0.34 & \cellcolor[HTML]{FFFFFF}0.32 \\ \hline
 & BLEU & 32.37 & 22.67 & 20.25 & 12.79 & 21.73 & 18.36 & 16.13 & 12.29 \\
 & BERTScore & 0.85 & 0.86 & 0.81 & 0.8 & 0.9 & 0.79 & 0.79 & 0.78 \\
 & TER & 64.09 & 70.76 & 75.17 & 81.02 & 80.02 & 79.28 & 85.2 & 89.03 \\
 & METEOR & 0.62 & 0.5 & 0.48 & 0.43 & 0.46 & 0.43 & 0.39 & 0.36 \\
 & ChrF & 51.54 & 38.91 & 43.96 & 35.4 & 40.54 & 34.22 & 38.5 & 33.67 \\
 & ROUGE-1 & 0.61 & 0.66 & 0.5 & 0.45 & 0.47 & 0.62 & 0.43 & 0.41 \\
 & ROUGE-2 & 0.4 & 0.45 & 0.3 & 0.22 & 0.28 & 0.39 & 0.24 & 0.2 \\
\multirow{-8}{*}{\textbf{\begin{tabular}[c]{@{}l@{}}Qwen-2.5-7B\\ - 8 Shot\end{tabular}}} & ROUGE-L & 0.58 & 0.56 & 0.46 & 0.41 & 0.42 & 0.49 & 0.38 & 0.37 \\ \hline
 & BLEU & 28.38 & 22.63 & 20.23 & 11.29 & 22.03 & 20.66 & 17.3 & 14.75 \\
 & BERTScore & 0.84 & 0.85 & 0.81 & 0.79 & 0.91 & 0.81 & 0.79 & 0.77 \\
 & TER & 67.96 & 71.83 & 76.08 & 83.85 & 81.52 & 76.13 & 82.84 & 87.24 \\
 & METEOR & 0.59 & 0.5 & 0.48 & 0.42 & 0.48 & 0.45 & 0.41 & 0.38 \\
 & ChrF & 47.86 & 37.41 & 43.55 & 33.81 & 40.79 & 36.53 & 39.15 & 33.33 \\
 & ROUGE-1 & 0.59 & 0.64 & 0.5 & 0.45 & 0.49 & 0.63 & 0.43 & 0.4 \\
 & ROUGE-2 & 0.38 & 0.44 & 0.31 & 0.22 & 0.29 & 0.41 & 0.25 & 0.22 \\
\multirow{-8}{*}{\textbf{\begin{tabular}[c]{@{}l@{}}Qwen-2.5-7B\\ - 16 Shot\end{tabular}}} & ROUGE-L & 0.55 & 0.54 & 0.46 & 0.41 & 0.44 & 0.51 & 0.38 & 0.37 \\ \hline
 & BLEU & 33.23 & 25.13 & 20.92 & 11.48 & 23.3 & 18.88 & 15.24 & 10.5 \\
 & BERTScore & 0.86 & 0.87 & 0.82 & 0.8 & 0.92 & 0.79 & 0.79 & 0.78 \\
 & TER & 64.02 & 69.67 & 75.67 & 84.04 & 77.07 & 83.31 & 85.2 & 90.95 \\
 & METEOR & 0.64 & 0.53 & 0.5 & 0.41 & 0.52 & 0.43 & 0.4 & 0.37 \\
 & ChrF & 53.52 & 41.0 & 44.95 & 34.77 & 44.66 & 33.94 & 38.97 & 33.36 \\
 & ROUGE-1 & 0.64 & 0.68 & 0.52 & 0.45 & 0.52 & 0.61 & 0.43 & 0.4 \\
 & ROUGE-2 & 0.42 & 0.47 & 0.31 & 0.21 & 0.3 & 0.38 & 0.24 & 0.19 \\
\multirow{-8}{*}{\textbf{\begin{tabular}[c]{@{}l@{}}Qwen-2.5-7B\\ - 32 Shot\end{tabular}}} & ROUGE-L & 0.59 & 0.57 & 0.47 & 0.4 & 0.47 & 0.49 & 0.38 & 0.35 \\ \hline
 & BLEU & 27.68 & 10.67 & 18.46 & 11.4 & 20.17 & 08.01 & 12.58 & 7.14 \\
 & BERTScore & 0.93 & 0.75 & 0.8 & 0.76 & 0.8 & 0.73 & 0.76 & 0.72 \\
 & TER & 72.65 & 82.73 & 81.15 & 90.07 & 83.87 & 85.69 & 87.85 & 98.87 \\
 & METEOR & 0.59 & 0.33 & 0.45 & 0.39 & 0.48 & 0.27 & 0.37 & 0.29 \\
 & ChrF & 47.23 & 24.79 & 39.83 & 31.55 & 40.58 & 21.71 & 34.68 & 25.42 \\
 & ROUGE-1 & 0.57 & 0.56 & 0.47 & 0.4 & 0.48 & 0.53 & 0.39 & 0.3 \\
 & ROUGE-2 & 0.36 & 0.32 & 0.28 & 0.2 & 0.26 & 0.27 & 0.2 & 0.13 \\
\multirow{-8}{*}{\textbf{\begin{tabular}[c]{@{}l@{}}Mistral-v0.3-7B\\ - Ft.\end{tabular}}} & ROUGE-L & 0.53 & 0.43 & 0.42 & 0.37 & 0.43 & 0.4 & 0.34 & 0.27 \\ \hline
 & BLEU & 22.94 & 16.76 & 17.51 & 8.66 & 12.4 & 1.22 & 3.65 & 01.05 \\
 & BERTScore & 0.83 & 0.83 & 0.8 & 0.78 & 0.89 & 0.6 & 0.64 & 0.59 \\
 & TER & 72.61 & 75.49 & 78.0 & 88.21 & 93.41 & 98.06 & 113.15 & 114.58 \\
 & METEOR & 0.54 & 0.43 & 0.44 & 0.36 & 0.37 & 0.07 & 0.16 & 0.09 \\
 & ChrF & 44.74 & 33.32 & 40.21 & 30.97 & 32.91 & 8.79 & 19.5 & 12.75 \\
 & ROUGE-1 & 0.55 & 0.62 & 0.47 & 0.4 & 0.39 & 0.28 & 0.18 & 0.1 \\
 & ROUGE-2 & 0.32 & 0.4 & 0.28 & 0.17 & 0.17 & 0.08 & 0.07 & 0.02 \\
\multirow{-8}{*}{\textbf{\begin{tabular}[c]{@{}l@{}}Mistral-v0.3-7B\\ - 0 Shot\end{tabular}}} & ROUGE-L & 0.5 & 0.51 & 0.42 & 0.35 & 0.34 & 0.21 & 0.16 & 0.09 \\ \hline
 & BLEU & 30.82 & 17.97 & 19.57 & 11.3 & 16.52 & 2.42 & 6.53 & 2.85 \\
 & BERTScore & 0.85 & 0.83 & 0.8 & 0.81 & 0.92 & 0.74 & 0.67 & 0.62 \\
 & TER & 65.66 & 73.89 & 76.46 & 83.38 & 90.42 & 95.99 & 105.51 & 112.79 \\
 & METEOR & 0.62 & 0.44 & 0.47 & 0.4 & 0.43 & 0.12 & 0.22 & 0.14 \\
 & ChrF & 51.57 & 34.53 & 43.3 & 33.91 & 37.51 & 12.0 & 24.06 & 16.69 \\
 & ROUGE-1 & 0.62 & 0.61 & 0.51 & 0.44 & 0.43 & 0.37 & 0.24 & 0.16 \\
 & ROUGE-2 & 0.4 & 0.41 & 0.29 & 0.21 & 0.21 & 0.13 & 0.11 & 0.05 \\
\multirow{-8}{*}{\textbf{\begin{tabular}[c]{@{}l@{}}Mistral-v0.3-7B\\ - 1 Shot\end{tabular}}} & ROUGE-L & 0.58 & 0.51 & 0.45 & 0.39 & 0.39 & 0.27 & 0.21 & 0.14 \\ \hline
 & BLEU & 32.47 & 22.55 & 20.32 & 12.8 & 19.65 & 6.8 & 11.95 & 9.42 \\
 & BERTScore & 0.85 & 0.85 & 0.81 & 0.8 & 0.92 & 0.73 & 0.76 & 0.75 \\
 & TER & 64.66 & 71.4 & 74.39 & 83.66 & 83.2 & 87.81 & 90.14 & 95.41 \\
 & METEOR & 0.62 & 0.48 & 0.49 & 0.43 & 0.47 & 0.24 & 0.34 & 0.32 \\
 & ChrF & 51.57 & 38.16 & 43.95 & 34.8 & 40.99 & 20.18 & 33.31 & 29.01 \\
 & ROUGE-1 & 0.61 & 0.64 & 0.5 & 0.45 & 0.48 & 0.52 & 0.37 & 0.34 \\
 & ROUGE-2 & 0.4 & 0.43 & 0.3 & 0.22 & 0.26 & 0.25 & 0.19 & 0.16 \\
\multirow{-8}{*}{\textbf{\begin{tabular}[c]{@{}l@{}}Mistral-v0.3-7B\\ - 8 Shot\end{tabular}}} & ROUGE-L & 0.58 & 0.54 & 0.46 & 0.41 & 0.43 & 0.38 & 0.32 & 0.3 \\ \hline
 & BLEU & 27.42 & 24.62 & 16.8 & 12.0 & 22.51 & 6.97 & 14.27 & 12.93 \\
 & BERTScore & 0.85 & 0.91 & 0.8 & 0.81 & 0.92 & 0.74 & 0.77 & 0.75 \\
 & TER & 66.91 & 70.38 & 79.92 & 82.87 & 81.31 & 88.06 & 88.7 & 93.21 \\
 & METEOR & 0.59 & 0.55 & 0.43 & 0.41 & 0.49 & 0.25 & 0.36 & 0.35 \\
 & ChrF & 49.02 & 41.34 & 39.91 & 35.45 & 42.73 & 20.12 & 34.96 & 29.34 \\
 & ROUGE-1 & 0.6 & 0.7 & 0.46 & 0.44 & 0.49 & 0.53 & 0.4 & 0.37 \\
 & ROUGE-2 & 0.36 & 0.47 & 0.26 & 0.21 & 0.28 & 0.25 & 0.22 & 0.21 \\
\multirow{-8}{*}{\textbf{\begin{tabular}[c]{@{}l@{}}Mistral-v0.3-7B\\ - 16 Shot\end{tabular}}} & ROUGE-L & 0.55 & 0.58 & 0.41 & 0.4 & 0.44 & 0.38 & 0.35 & 0.34 \\ \hline
 & BLEU & 27.67 & 8.27 & 15.7 & 9.56 & 19.19 & 6.72 & 12.05 & 7.59 \\
 & BERTScore & 0.84 & 0.88 & 0.8 & 0.78 & 0.92 & 0.75 & 0.77 & 0.76 \\
 & TER & 69.14 & 83.79 & 82.81 & 87.74 & 82.35 & 86.05 & 89.62 & 93.68 \\
 & METEOR & 0.6 & 0.31 & 0.43 & 0.38 & 0.48 & 0.27 & 0.35 & 0.31 \\
 & ChrF & 48.47 & 23.3 & 39.44 & 32.17 & 40.86 & 21.51 & 34.87 & 28.86 \\
 & ROUGE-1 & 0.59 & 0.58 & 0.44 & 0.41 & 0.48 & 0.55 & 0.38 & 0.35 \\
 & ROUGE-2 & 0.36 & 0.3 & 0.24 & 0.17 & 0.25 & 0.26 & 0.19 & 0.13 \\
\multirow{-8}{*}{\textbf{\begin{tabular}[c]{@{}l@{}}Mistral-v0.3-7B\\ - 32 Shot\end{tabular}}} & ROUGE-L & 0.54 & 0.43 & 0.4 & 0.36 & 0.43 & 0.4 & 0.32 & 0.3 \\ \hline
\caption{\textbf{In-context Learning Results}. Full fine-tuning, zero-shot, few-shot on both ground-truth transcript and \gls{ASR} transcript in the cascaded setting. \textbf{Chinese to X} results are reported in this table.\\
All cascaded models use Whisper$_{small-mono}$ as \gls{ASR} model (Whisper \gls{ASR} is fine-tuned monolingually - on each source language separately). Its \gls{WER} on test set is 29.6\%, 33.8\%, 31.3\%, 26.3\%, 45.7\% for Vietnamese, English, Chinese, German and French respectively.}
\label{tab:appx_nmt_fewshot-Zh-X}
\end{longtable}

%% file: tables/appx_nmt_gt_allMetrics-En-X.tex
\begin{longtable}{llcccc}
\hline
\multicolumn{1}{l|}{\textbf{MT}} & \multicolumn{1}{l|}{\textbf{Metrics}} & \textbf{en-vi} & \textbf{en-fr} & \textbf{en-zh} & \textbf{en-de} \\ \hline
\endfirsthead
\multicolumn{6}{c}%
{{\bfseries Table \thetable\ continued from previous page}} \\
\hline
\multicolumn{1}{l|}{\textbf{MT}} & \multicolumn{1}{l|}{\textbf{Metrics}} & \textbf{en-vi} & \textbf{en-fr} & \textbf{en-zh} & \textbf{en-de} \\ \hline
\endhead
\multicolumn{6}{c}{Decoder} \\ \hline
\multicolumn{1}{l|}{\multirow{8}{*}{\begin{tabular}[c]{@{}l@{}}Llama\\ -3.1-8B\end{tabular}}} & \multicolumn{1}{l|}{BLEU} & 53.44 & 48.24 & 37.50 & 40.49 \\
\multicolumn{1}{l|}{} & \multicolumn{1}{l|}{BERTSc} & 0.90 & 0.89 & 0.83 & 0.87 \\
\multicolumn{1}{l|}{} & \multicolumn{1}{l|}{TER} & 39.42 & 46.9 & 58.37 & 53.12 \\
\multicolumn{1}{l|}{} & \multicolumn{1}{l|}{METEOR} & 0.77 & 0.72 & 0.6 & 0.66 \\
\multicolumn{1}{l|}{} & \multicolumn{1}{l|}{ChrF} & 67.74 & 70.97 & 32.39 & 66.08 \\
\multicolumn{1}{l|}{} & \multicolumn{1}{l|}{ROUGE-1} & 0.83 & 0.73 & 0.15 & 0.67 \\
\multicolumn{1}{l|}{} & \multicolumn{1}{l|}{ROUGE-2} & 0.68 & 0.57 & 0.13 & 0.47 \\
\multicolumn{1}{l|}{} & \multicolumn{1}{l|}{ROUGE-L} & 0.76 & 0.7 & 0.15 & 0.63 \\ \hline
\multicolumn{1}{l|}{\multirow{8}{*}{\begin{tabular}[c]{@{}l@{}}Qwen\\ -2.5-7B\end{tabular}}} & \multicolumn{1}{l|}{BLEU} & 54.5 & 49.63 & 28.61 & 38.75 \\
\multicolumn{1}{l|}{} & \multicolumn{1}{l|}{BERTSc} & 0.9 & 0.9 & 0.81 & 0.87 \\
\multicolumn{1}{l|}{} & \multicolumn{1}{l|}{TER} & 38.21 & 42.42 & 59.1 & 51.55 \\
\multicolumn{1}{l|}{} & \multicolumn{1}{l|}{METEOR} & 0.77 & 0.72 & 0.5 & 0.64 \\
\multicolumn{1}{l|}{} & \multicolumn{1}{l|}{ChrF} & 67.82 & 70.5 & 27.81 & 63.39 \\
\multicolumn{1}{l|}{} & \multicolumn{1}{l|}{ROUGE-1} & 0.83 & 0.74 & 0.14 & 0.65 \\
\multicolumn{1}{l|}{} & \multicolumn{1}{l|}{ROUGE-2} & 0.69 & 0.57 & 0.13 & 0.43 \\
\multicolumn{1}{l|}{} & \multicolumn{1}{l|}{ROUGE-L} & 0.77 & 0.71 & 0.14 & 0.62 \\ \hline
\multicolumn{1}{l|}{\multirow{8}{*}{\begin{tabular}[c]{@{}l@{}}Mistral\\ -v0.3-7B\end{tabular}}} & \multicolumn{1}{l|}{BLEU} & 24.77 & 51.71 & 26.38 & 43.99 \\
\multicolumn{1}{l|}{} & \multicolumn{1}{l|}{BERTSc} & 0.82 & 0.89 & 0.81 & 0.88 \\
\multicolumn{1}{l|}{} & \multicolumn{1}{l|}{TER} & 63.33 & 43.0 & 60.43 & 48.54 \\
\multicolumn{1}{l|}{} & \multicolumn{1}{l|}{METEOR} & 0.45 & 0.73 & 0.48 & 0.67 \\
\multicolumn{1}{l|}{} & \multicolumn{1}{l|}{ChrF} & 41.27 & 71.06 & 27.29 & 66.27 \\
\multicolumn{1}{l|}{} & \multicolumn{1}{l|}{ROUGE-1} & 0.66 & 0.74 & 0.12 & 0.68 \\
\multicolumn{1}{l|}{} & \multicolumn{1}{l|}{ROUGE-2} & 0.49 & 0.6 & 0.11 & 0.49 \\
\multicolumn{1}{l|}{} & \multicolumn{1}{l|}{ROUGE-L} & 0.57 & 0.72 & 0.12 & 0.65 \\ \hline
\multicolumn{6}{c}{Encoder-decoder} \\ \hline
\multicolumn{1}{l|}{\multirow{8}{*}{\begin{tabular}[c]{@{}l@{}}mBart\\ -large-50\end{tabular}}} & \multicolumn{1}{l|}{BLEU} & 62.08 & 57.04 & 44.77 & 47.28 \\
\multicolumn{1}{l|}{} & \multicolumn{1}{l|}{BERTSc} & 0.92 & 0.92 & 0.86 & 0.89 \\
\multicolumn{1}{l|}{} & \multicolumn{1}{l|}{TER} & 29.87 & 34.26 & 41.08 & 42.76 \\
\multicolumn{1}{l|}{} & \multicolumn{1}{l|}{METEOR} & 0.81 & 0.77 & 0.68 & 0.7 \\
\multicolumn{1}{l|}{} & \multicolumn{1}{l|}{ChrF} & 72.75 & 75.3 & 38.7 & 68.74 \\
\multicolumn{1}{l|}{} & \multicolumn{1}{l|}{ROUGE-1} & 0.86 & 0.78 & 0.19 & 0.71 \\
\multicolumn{1}{l|}{} & \multicolumn{1}{l|}{ROUGE-2} & 0.74 & 0.64 & 0.17 & 0.52 \\
\multicolumn{1}{l|}{} & \multicolumn{1}{l|}{ROUGE-L} & 0.81 & 0.76 & 0.19 & 0.68 \\ \hline
\multicolumn{1}{l|}{\multirow{8}{*}{\begin{tabular}[c]{@{}l@{}}M2M100\\ -418M\end{tabular}}} & \multicolumn{1}{l|}{BLEU} & 62.31 & 57.49 & 46.38 & 49.36 \\
\multicolumn{1}{l|}{} & \multicolumn{1}{l|}{BERTSc} & 0.97 & 0.95 & 0.93 & 0.94 \\
\multicolumn{1}{l|}{} & \multicolumn{1}{l|}{TER} & 29.4 & 33.52 & 39.38 & 40.6 \\
\multicolumn{1}{l|}{} & \multicolumn{1}{l|}{METEOR} & 0.81 & 0.77 & 0.7 & 0.72 \\
\multicolumn{1}{l|}{} & \multicolumn{1}{l|}{ChrF} & 73.15 & 75.72 & 40.11 & 71.04 \\
\multicolumn{1}{l|}{} & \multicolumn{1}{l|}{ROUGE-1} & 0.86 & 0.79 & 0.2 & 0.73 \\
\multicolumn{1}{l|}{} & \multicolumn{1}{l|}{ROUGE-2} & 0.74 & 0.64 & 0.19 & 0.54 \\
\multicolumn{1}{l|}{} & \multicolumn{1}{l|}{ROUGE-L} & 0.81 & 0.76 & 0.2 & 0.7 \\ \hline
\multicolumn{1}{l|}{\multirow{8}{*}{Marian}} & \multicolumn{1}{l|}{BLEU} & 58.22 & 53.84 & 38.67 & 45.81 \\
\multicolumn{1}{l|}{} & \multicolumn{1}{l|}{BERTSc} & 0.91 & 0.91 & 0.85 & 0.89 \\
\multicolumn{1}{l|}{} & \multicolumn{1}{l|}{TER} & 32.56 & 36.53 & 45.9 & 43.77 \\
\multicolumn{1}{l|}{} & \multicolumn{1}{l|}{METEOR} & 0.79 & 0.75 & 0.64 & 0.69 \\
\multicolumn{1}{l|}{} & \multicolumn{1}{l|}{ChrF} & 70.19 & 73.68 & 33.27 & 68.72 \\
\multicolumn{1}{l|}{} & \multicolumn{1}{l|}{ROUGE-1} & 0.85 & 0.77 & 0.19 & 0.71 \\
\multicolumn{1}{l|}{} & \multicolumn{1}{l|}{ROUGE-2} & 0.71 & 0.61 & 0.18 & 0.5 \\
\multicolumn{1}{l|}{} & \multicolumn{1}{l|}{ROUGE-L} & 0.79 & 0.74 & 0.19 & 0.67 \\ \hline
\multicolumn{6}{c}{Commercial tool} \\ \hline
\multicolumn{1}{l|}{\multirow{8}{*}{\begin{tabular}[c]{@{}l@{}}Google\\ Translate\end{tabular}}} & \multicolumn{1}{l|}{BLEU} & 46.21 & 44.77 & 44.74 & 36.29 \\
\multicolumn{1}{l|}{} & \multicolumn{1}{l|}{BERTSc} & 0.91 & 0.91 & 0.9 & 0.89 \\
\multicolumn{1}{l|}{} & \multicolumn{1}{l|}{TER} & 48.02 & 50.96 & 52.47 & 59.78 \\
\multicolumn{1}{l|}{} & \multicolumn{1}{l|}{METEOR} & 0.67 & 0.64 & 0.63 & 0.58 \\
\multicolumn{1}{l|}{} & \multicolumn{1}{l|}{ChrF} & 59.7 & 64.44 & 39.17 & 59.39 \\
\multicolumn{1}{l|}{} & \multicolumn{1}{l|}{ROUGE-1} & 0.78 & 0.7 & 0.16 & 0.63 \\
\multicolumn{1}{l|}{} & \multicolumn{1}{l|}{ROUGE-2} & 0.64 & 0.56 & 0.14 & 0.44 \\
\multicolumn{1}{l|}{} & \multicolumn{1}{l|}{ROUGE-L} & 0.72 & 0.67 & 0.16 & 0.6
\\ \hline
\caption{\textbf{Full Results: Ground-truth Translation Baselines.} \textbf{English to X} results are reported in this table. Extension of Table \ref{tab:translation-groundtruth} in the main paper.}
\label{tab:appx_nmt_gt_allMetrics-En-X}
\end{longtable}

%% file: tables/appx_nmt_gt_allMetrics-Vi-X.tex
\begin{longtable}{llcccc}
\hline
\multicolumn{1}{l|}{\textbf{MT}} & \multicolumn{1}{l|}{\textbf{Metrics}} & \textbf{vi-en} & \textbf{vi-fr} & \textbf{vi-zh} & \textbf{vi-de} \\ \hline
\endfirsthead
\multicolumn{6}{c}%
{{\bfseries Table \thetable\ continued from previous page}} \\
\hline
\multicolumn{1}{l|}{\textbf{MT}} & \multicolumn{1}{l|}{\textbf{Metrics}} & \textbf{vi-en} & \textbf{vi-fr} & \textbf{vi-zh} & \textbf{vi-de} \\ \hline
\endhead
\multicolumn{6}{c}{Decoder} \\ \hline
\multicolumn{1}{l|}{\multirow{8}{*}{\begin{tabular}[c]{@{}l@{}}Llama\\ -3.1-8B\end{tabular}}} & \multicolumn{1}{l|}{BLEU} & 23.16 & 15.57 & 16.09 & 11.61 \\
\multicolumn{1}{l|}{} & \multicolumn{1}{l|}{BERTSc} & 0.92 & 0.79 & 0.74 & 0.77 \\
\multicolumn{1}{l|}{} & \multicolumn{1}{l|}{TER} & 83.07 & 100.23 & 120.58 & 112.85 \\
\multicolumn{1}{l|}{} & \multicolumn{1}{l|}{METEOR} & 0.57 & 0.45 & 0.5 & 0.39 \\
\multicolumn{1}{l|}{} & \multicolumn{1}{l|}{ChrF} & 52.29 & 47.18 & 21.03 & 44.25 \\
\multicolumn{1}{l|}{} & \multicolumn{1}{l|}{ROUGE-1} & 0.56 & 0.45 & 0.03 & 0.4 \\
\multicolumn{1}{l|}{} & \multicolumn{1}{l|}{ROUGE-2} & 0.33 & 0.25 & 0.01 & 0.19 \\
\multicolumn{1}{l|}{} & \multicolumn{1}{l|}{ROUGE-L} & 0.51 & 0.4 & 0.03 & 0.36 \\ \hline
\multicolumn{1}{l|}{\multirow{8}{*}{\begin{tabular}[c]{@{}l@{}}Qwen\\ -2.5-7B\end{tabular}}} & \multicolumn{1}{l|}{BLEU} & 26.21 & 19.25 & 29.06 & 14.44 \\
\multicolumn{1}{l|}{} & \multicolumn{1}{l|}{BERTSc} & 0.93 & 0.81 & 0.81 & 0.79 \\
\multicolumn{1}{l|}{} & \multicolumn{1}{l|}{TER} & 71.1 & 80.86 & 60.93 & 83.75 \\
\multicolumn{1}{l|}{} & \multicolumn{1}{l|}{METEOR} & 0.57 & 0.47 & 0.55 & 0.39 \\
\multicolumn{1}{l|}{} & \multicolumn{1}{l|}{ChrF} & 53.79 & 48.94 & 25.1 & 43.34 \\
\multicolumn{1}{l|}{} & \multicolumn{1}{l|}{ROUGE-1} & 0.58 & 0.49 & 0.04 & 0.42 \\
\multicolumn{1}{l|}{} & \multicolumn{1}{l|}{ROUGE-2} & 0.35 & 0.28 & 0.02 & 0.19 \\
\multicolumn{1}{l|}{} & \multicolumn{1}{l|}{ROUGE-L} & 0.53 & 0.44 & 0.04 & 0.38 \\ \hline
\multicolumn{1}{l|}{\multirow{8}{*}{\begin{tabular}[c]{@{}l@{}}Mistral\\ -v0.3-7B\end{tabular}}} & \multicolumn{1}{l|}{BLEU} & 24.56 & 16.0 & 25.04 & 13.38 \\
\multicolumn{1}{l|}{} & \multicolumn{1}{l|}{BERTSc} & 0.92 & 0.79 & 0.78 & 0.78 \\
\multicolumn{1}{l|}{} & \multicolumn{1}{l|}{TER} & 70.37 & 87.87 & 65.43 & 84.13 \\
\multicolumn{1}{l|}{} & \multicolumn{1}{l|}{METEOR} & 0.53 & 0.42 & 0.49 & 0.37 \\
\multicolumn{1}{l|}{} & \multicolumn{1}{l|}{ChrF} & 48.73 & 43.93 & 21.84 & 40.96 \\
\multicolumn{1}{l|}{} & \multicolumn{1}{l|}{ROUGE-1} & 0.54 & 0.44 & 0.03 & 0.4 \\
\multicolumn{1}{l|}{} & \multicolumn{1}{l|}{ROUGE-2} & 0.31 & 0.24 & 0.01 & 0.18 \\
\multicolumn{1}{l|}{} & \multicolumn{1}{l|}{ROUGE-L} & 0.5 & 0.39 & 0.03 & 0.36 \\ \hline
\multicolumn{6}{c}{Encoder-decoder} \\ \hline
\multicolumn{1}{l|}{\multirow{8}{*}{\begin{tabular}[c]{@{}l@{}}mBart\\ -large-50\end{tabular}}} & \multicolumn{1}{l|}{BLEU} & 13.34 & 17.9 & 22.97 & 9.85 \\
\multicolumn{1}{l|}{} & \multicolumn{1}{l|}{BERTSc} & 0.89 & 0.8 & 0.78 & 0.75 \\
\multicolumn{1}{l|}{} & \multicolumn{1}{l|}{TER} & 79.27 & 76.57 & 62.13 & 95.17 \\
\multicolumn{1}{l|}{} & \multicolumn{1}{l|}{METEOR} & 0.38 & 0.41 & 0.48 & 0.29 \\
\multicolumn{1}{l|}{} & \multicolumn{1}{l|}{ChrF} & 36.03 & 43.81 & 20.38 & 34.93 \\
\multicolumn{1}{l|}{} & \multicolumn{1}{l|}{ROUGE-1} & 0.42 & 0.45 & 0.03 & 0.32 \\
\multicolumn{1}{l|}{} & \multicolumn{1}{l|}{ROUGE-2} & 0.18 & 0.25 & 0.01 & 0.13 \\
\multicolumn{1}{l|}{} & \multicolumn{1}{l|}{ROUGE-L} & 0.38 & 0.41 & 0.03 & 0.29 \\ \hline
\multicolumn{1}{l|}{\multirow{8}{*}{\begin{tabular}[c]{@{}l@{}}M2M100\\ -418M\end{tabular}}} & \multicolumn{1}{l|}{BLEU} & 23.01 & 21.01 & 24.95 & 16.72 \\
\multicolumn{1}{l|}{} & \multicolumn{1}{l|}{BERTSc} & 0.82 & 0.81 & 0.8 & 0.79 \\
\multicolumn{1}{l|}{} & \multicolumn{1}{l|}{TER} & 68.26 & 73.61 & 60.67 & 77.59 \\
\multicolumn{1}{l|}{} & \multicolumn{1}{l|}{METEOR} & 0.51 & 0.45 & 0.51 & 0.4 \\
\multicolumn{1}{l|}{} & \multicolumn{1}{l|}{ChrF} & 48.0 & 48.05 & 21.95 & 44.52 \\
\multicolumn{1}{l|}{} & \multicolumn{1}{l|}{ROUGE-1} & 0.54 & 0.49 & 0.04 & 0.45 \\
\multicolumn{1}{l|}{} & \multicolumn{1}{l|}{ROUGE-2} & 0.29 & 0.29 & 0.02 & 0.21 \\
\multicolumn{1}{l|}{} & \multicolumn{1}{l|}{ROUGE-L} & 0.49 & 0.45 & 0.04 & 0.4 \\ \hline
\multicolumn{1}{l|}{\multirow{8}{*}{Marian}} & \multicolumn{1}{l|}{BLEU} & 17.63 & 15.97 & 15.56 & 12.84 \\
\multicolumn{1}{l|}{} & \multicolumn{1}{l|}{BERTSc} & 0.8 & 0.79 & 0.78 & 0.77 \\
\multicolumn{1}{l|}{} & \multicolumn{1}{l|}{TER} & 75.09 & 79.31 & 71.43 & 82.01 \\
\multicolumn{1}{l|}{} & \multicolumn{1}{l|}{METEOR} & 0.45 & 0.39 & 0.4 & 0.35 \\
\multicolumn{1}{l|}{} & \multicolumn{1}{l|}{ChrF} & 42.61 & 42.83 & 14.79 & 39.82 \\
\multicolumn{1}{l|}{} & \multicolumn{1}{l|}{ROUGE-1} & 0.48 & 0.44 & 0.03 & 0.4 \\
\multicolumn{1}{l|}{} & \multicolumn{1}{l|}{ROUGE-2} & 0.23 & 0.23 & 0.01 & 0.17 \\
\multicolumn{1}{l|}{} & \multicolumn{1}{l|}{ROUGE-L} & 0.44 & 0.4 & 0.03 & 0.35 \\ \hline
\multicolumn{6}{c}{Commercial tool} \\ \hline
\multicolumn{1}{l|}{\multirow{8}{*}{\begin{tabular}[c]{@{}l@{}}Google\\ Translate\end{tabular}}} & \multicolumn{1}{l|}{BLEU} & 18.79 & 16.42 & 21.63 & 12.54 \\
\multicolumn{1}{l|}{} & \multicolumn{1}{l|}{BERTSc} & 0.84 & 0.83 & 0.83 & 0.81 \\
\multicolumn{1}{l|}{} & \multicolumn{1}{l|}{TER} & 75.72 & 82.76 & 71.96 & 87.12 \\
\multicolumn{1}{l|}{} & \multicolumn{1}{l|}{METEOR} & 0.43 & 0.37 & 0.44 & 0.32 \\
\multicolumn{1}{l|}{} & \multicolumn{1}{l|}{ChrF} & 43.91 & 42.82 & 19.03 & 40.21 \\
\multicolumn{1}{l|}{} & \multicolumn{1}{l|}{ROUGE-1} & 0.49 & 0.43 & 0.02 & 0.38 \\
\multicolumn{1}{l|}{} & \multicolumn{1}{l|}{ROUGE-2} & 0.26 & 0.24 & 0.01 & 0.17 \\
\multicolumn{1}{l|}{} & \multicolumn{1}{l|}{ROUGE-L} & 0.45 & 0.38 & 0.02 & 0.33
\\ \hline
\caption{\textbf{Full Results: Ground-truth Translation Baselines.} \textbf{Vietnamese to X} results are reported in this table. Extension of Table \ref{tab:translation-groundtruth} in the main paper.}
\label{tab:appx_nmt_gt_allMetrics-Vi-X}
\end{longtable}

%% file: tables/appx_nmt_gt_allMetrics-Fr-X.tex
\begin{longtable}{llcccc}
\hline
\multicolumn{1}{l|}{\textbf{MT}} & \multicolumn{1}{l|}{\textbf{Metrics}} & \textbf{fr-en} & \textbf{fr-vi} & \textbf{fr-zh} & \textbf{fr-de} \\ \hline
\endfirsthead
\multicolumn{6}{c}%
{{\bfseries Table \thetable\ continued from previous page}} \\
\hline
\multicolumn{1}{l|}{\textbf{MT}} & \multicolumn{1}{l|}{\textbf{Metrics}} & \textbf{fr-en} & \textbf{fr-vi} & \textbf{fr-zh} & \textbf{fr-de} \\ \hline
\endhead
\multicolumn{6}{c}{Decoder} \\ \hline
\multicolumn{1}{l|}{\multirow{8}{*}{\begin{tabular}[c]{@{}l@{}}Llama\\ -3.1-8B\end{tabular}}} & \multicolumn{1}{l|}{BLEU} & 50.18 & 39.63 & 29.25 & 27.46 \\
\multicolumn{1}{l|}{} & \multicolumn{1}{l|}{BERTSc} & 0.95 & 0.86 & 0.79 & 0.81 \\
\multicolumn{1}{l|}{} & \multicolumn{1}{l|}{TER} & 42.23 & 54.71 & 68.66 & 82.2 \\
\multicolumn{1}{l|}{} & \multicolumn{1}{l|}{METEOR} & 0.76 & 0.65 & 0.52 & 0.56 \\
\multicolumn{1}{l|}{} & \multicolumn{1}{l|}{ChrF} & 69.44 & 55.16 & 25.3 & 56.35 \\
\multicolumn{1}{l|}{} & \multicolumn{1}{l|}{ROUGE-1} & 0.75 & 0.77 & 0.11 & 0.54 \\
\multicolumn{1}{l|}{} & \multicolumn{1}{l|}{ROUGE-2} & 0.58 & 0.56 & 0.07 & 0.34 \\
\multicolumn{1}{l|}{} & \multicolumn{1}{l|}{ROUGE-L} & 0.73 & 0.67 & 0.11 & 0.5 \\ \hline
\multicolumn{1}{l|}{\multirow{8}{*}{\begin{tabular}[c]{@{}l@{}}Qwen\\ -2.5-7B\end{tabular}}} & \multicolumn{1}{l|}{BLEU} & 49.69 & 40.67 & 20.97 & 33.91 \\
\multicolumn{1}{l|}{} & \multicolumn{1}{l|}{BERTSc} & 0.95 & 0.86 & 0.78 & 0.84 \\
\multicolumn{1}{l|}{} & \multicolumn{1}{l|}{TER} & 42.17 & 52.57 & 67.39 & 59.28 \\
\multicolumn{1}{l|}{} & \multicolumn{1}{l|}{METEOR} & 0.76 & 0.66 & 0.43 & 0.58 \\
\multicolumn{1}{l|}{} & \multicolumn{1}{l|}{ChrF} & 70.03 & 56.12 & 20.86 & 56.82 \\
\multicolumn{1}{l|}{} & \multicolumn{1}{l|}{ROUGE-1} & 0.76 & 0.78 & 0.09 & 0.6 \\
\multicolumn{1}{l|}{} & \multicolumn{1}{l|}{ROUGE-2} & 0.58 & 0.58 & 0.06 & 0.37 \\
\multicolumn{1}{l|}{} & \multicolumn{1}{l|}{ROUGE-L} & 0.73 & 0.68 & 0.08 & 0.56 \\ \hline
\multicolumn{1}{l|}{\multirow{8}{*}{\begin{tabular}[c]{@{}l@{}}Mistral\\ -v0.3-7B\end{tabular}}} & \multicolumn{1}{l|}{BLEU} & 42.49 & 14.47 & 19.92 & 33.73 \\
\multicolumn{1}{l|}{} & \multicolumn{1}{l|}{BERTSc} & 0.93 & 0.79 & 0.78 & 0.85 \\
\multicolumn{1}{l|}{} & \multicolumn{1}{l|}{TER} & 56.11 & 74.72 & 67.02 & 57.73 \\
\multicolumn{1}{l|}{} & \multicolumn{1}{l|}{METEOR} & 0.75 & 0.36 & 0.42 & 0.58 \\
\multicolumn{1}{l|}{} & \multicolumn{1}{l|}{ChrF} & 68.36 & 30.93 & 20.91 & 56.06 \\
\multicolumn{1}{l|}{} & \multicolumn{1}{l|}{ROUGE-1} & 0.72 & 0.6 & 0.08 & 0.61 \\
\multicolumn{1}{l|}{} & \multicolumn{1}{l|}{ROUGE-2} & 0.55 & 0.37 & 0.05 & 0.38 \\
\multicolumn{1}{l|}{} & \multicolumn{1}{l|}{ROUGE-L} & 0.7 & 0.49 & 0.08 & 0.57 \\ \hline
\multicolumn{6}{c}{Encoder-decoder} \\ \hline
\multicolumn{1}{l|}{\multirow{8}{*}{\begin{tabular}[c]{@{}l@{}}mBart\\ -large-50\end{tabular}}} & \multicolumn{1}{l|}{BLEU} & 39.79 & 37.26 & 24.63 & 29.03 \\
\multicolumn{1}{l|}{} & \multicolumn{1}{l|}{BERTSc} & 0.93 & 0.86 & 0.77 & 0.83 \\
\multicolumn{1}{l|}{} & \multicolumn{1}{l|}{TER} & 51.84 & 54.81 & 63.66 & 64.46 \\
\multicolumn{1}{l|}{} & \multicolumn{1}{l|}{METEOR} & 0.67 & 0.62 & 0.49 & 0.53 \\
\multicolumn{1}{l|}{} & \multicolumn{1}{l|}{ChrF} & 59.96 & 53.25 & 21.41 & 51.13 \\
\multicolumn{1}{l|}{} & \multicolumn{1}{l|}{ROUGE-1} & 0.67 & 0.76 & 0.13 & 0.55 \\
\multicolumn{1}{l|}{} & \multicolumn{1}{l|}{ROUGE-2} & 0.45 & 0.53 & 0.1 & 0.32 \\
\multicolumn{1}{l|}{} & \multicolumn{1}{l|}{ROUGE-L} & 0.63 & 0.65 & 0.13 & 0.5 \\ \hline
\multicolumn{1}{l|}{\multirow{8}{*}{\begin{tabular}[c]{@{}l@{}}M2M100\\ -418M\end{tabular}}} & \multicolumn{1}{l|}{BLEU} & 43.73 & 35.04 & 29.41 & 34.72 \\
\multicolumn{1}{l|}{} & \multicolumn{1}{l|}{BERTSc} & 0.88 & 0.82 & 0.82 & 0.83 \\
\multicolumn{1}{l|}{} & \multicolumn{1}{l|}{TER} & 45.48 & 57.02 & 53.94 & 55.57 \\
\multicolumn{1}{l|}{} & \multicolumn{1}{l|}{METEOR} & 0.71 & 0.57 & 0.56 & 0.59 \\
\multicolumn{1}{l|}{} & \multicolumn{1}{l|}{ChrF} & 63.79 & 49.8 & 25.65 & 57.69 \\
\multicolumn{1}{l|}{} & \multicolumn{1}{l|}{ROUGE-1} & 0.71 & 0.67 & 0.15 & 0.6 \\
\multicolumn{1}{l|}{} & \multicolumn{1}{l|}{ROUGE-2} & 0.49 & 0.49 & 0.11 & 0.37 \\
\multicolumn{1}{l|}{} & \multicolumn{1}{l|}{ROUGE-L} & 0.68 & 0.59 & 0.15 & 0.56 \\ \hline
\multicolumn{1}{l|}{\multirow{8}{*}{Marian}} & \multicolumn{1}{l|}{BLEU} & 39.97 & 33.41 & 17.13 & 32.62 \\
\multicolumn{1}{l|}{} & \multicolumn{1}{l|}{BERTSc} & 0.87 & 0.86 & 0.78 & 0.85 \\
\multicolumn{1}{l|}{} & \multicolumn{1}{l|}{TER} & 48.76 & 57.15 & 67.55 & 56.39 \\
\multicolumn{1}{l|}{} & \multicolumn{1}{l|}{METEOR} & 0.68 & 0.62 & 0.41 & 0.59 \\
\multicolumn{1}{l|}{} & \multicolumn{1}{l|}{ChrF} & 60.97 & 52.06 & 16.05 & 56.73 \\
\multicolumn{1}{l|}{} & \multicolumn{1}{l|}{ROUGE-1} & 0.69 & 0.76 & 0.14 & 0.61 \\
\multicolumn{1}{l|}{} & \multicolumn{1}{l|}{ROUGE-2} & 0.46 & 0.53 & 0.1 & 0.36 \\
\multicolumn{1}{l|}{} & \multicolumn{1}{l|}{ROUGE-L} & 0.65 & 0.65 & 0.14 & 0.57 \\ \hline
\multicolumn{6}{c}{Commercial tool} \\ \hline
\multicolumn{1}{l|}{\multirow{8}{*}{\begin{tabular}[c]{@{}l@{}}Google\\ Translate\end{tabular}}} & \multicolumn{1}{l|}{BLEU} & 27.82 & 24.18 & 24.49 & 22.38 \\
\multicolumn{1}{l|}{} & \multicolumn{1}{l|}{BERTSc} & 0.88 & 0.86 & 0.85 & 0.86 \\
\multicolumn{1}{l|}{} & \multicolumn{1}{l|}{TER} & 63.0 & 67.34 & 66.27 & 70.46 \\
\multicolumn{1}{l|}{} & \multicolumn{1}{l|}{METEOR} & 0.51 & 0.46 & 0.44 & 0.43 \\
\multicolumn{1}{l|}{} & \multicolumn{1}{l|}{ChrF} & 50.04 & 42.11 & 22.47 & 46.46 \\
\multicolumn{1}{l|}{} & \multicolumn{1}{l|}{ROUGE-1} & 0.59 & 0.68 & 0.1 & 0.5 \\
\multicolumn{1}{l|}{} & \multicolumn{1}{l|}{ROUGE-2} & 0.41 & 0.46 & 0.07 & 0.3 \\
\multicolumn{1}{l|}{} & \multicolumn{1}{l|}{ROUGE-L} & 0.56 & 0.56 & 0.1 & 0.47
\\ \hline
\caption{\textbf{Full Results: Ground-truth Translation Baselines.} \textbf{French to X} results are reported in this table. Extension of Table \ref{tab:translation-groundtruth} in the main paper.}
\label{tab:appx_nmt_gt_allMetrics-Fr-X}
\end{longtable}

%% file: tables/appx_nmt_gt_allMetrics-De-X.tex
\begin{longtable}{llcccc}
\hline
\multicolumn{1}{l|}{\textbf{MT}} & \multicolumn{1}{l|}{\textbf{Metrics}} & \textbf{de-en} & \textbf{de-vi} & \textbf{de-fr} & \textbf{de-zh} \\ \hline
\endfirsthead
\multicolumn{6}{c}%
{{\bfseries Table \thetable\ continued from previous page}} \\
\hline
\multicolumn{1}{l|}{\textbf{MT}} & \multicolumn{1}{l|}{\textbf{Metrics}} & \textbf{de-en} & \textbf{de-vi} & \textbf{de-fr} & \textbf{de-zh} \\ \hline
\endhead
\multicolumn{6}{c}{Decoder} \\ \hline
\multicolumn{1}{l|}{\multirow{8}{*}{\begin{tabular}[c]{@{}l@{}}Llama\\ -3.1-8B\end{tabular}}} & \multicolumn{1}{l|}{BLEU} & 49.44 & 40.01 & 33.45 & 31.16 \\
\multicolumn{1}{l|}{} & \multicolumn{1}{l|}{BERTSc} & 0.95 & 0.87 & 0.84 & 0.81 \\
\multicolumn{1}{l|}{} & \multicolumn{1}{l|}{TER} & 44.99 & 54.45 & 72.36 & 62.08 \\
\multicolumn{1}{l|}{} & \multicolumn{1}{l|}{METEOR} & 0.76 & 0.67 & 0.62 & 0.54 \\
\multicolumn{1}{l|}{} & \multicolumn{1}{l|}{ChrF} & 69.74 & 57.57 & 61.36 & 27.18 \\
\multicolumn{1}{l|}{} & \multicolumn{1}{l|}{ROUGE-1} & 0.75 & 0.78 & 0.61 & 0.12 \\
\multicolumn{1}{l|}{} & \multicolumn{1}{l|}{ROUGE-2} & 0.57 & 0.58 & 0.42 & 0.09 \\
\multicolumn{1}{l|}{} & \multicolumn{1}{l|}{ROUGE-L} & 0.72 & 0.68 & 0.57 & 0.12 \\ \hline
\multicolumn{1}{l|}{\multirow{8}{*}{\begin{tabular}[c]{@{}l@{}}Qwen\\ -2.5-7B\end{tabular}}} & \multicolumn{1}{l|}{BLEU} & 52.1 & 43.73 & 40.72 & 23.26 \\
\multicolumn{1}{l|}{} & \multicolumn{1}{l|}{BERTSc} & 0.96 & 0.88 & 0.88 & 0.79 \\
\multicolumn{1}{l|}{} & \multicolumn{1}{l|}{TER} & 39.94 & 48.63 & 51.44 & 63.95 \\
\multicolumn{1}{l|}{} & \multicolumn{1}{l|}{METEOR} & 0.77 & 0.69 & 0.65 & 0.44 \\
\multicolumn{1}{l|}{} & \multicolumn{1}{l|}{ChrF} & 70.13 & 59.36 & 62.75 & 23.28 \\
\multicolumn{1}{l|}{} & \multicolumn{1}{l|}{ROUGE-1} & 0.76 & 0.8 & 0.67 & 0.11 \\
\multicolumn{1}{l|}{} & \multicolumn{1}{l|}{ROUGE-2} & 0.58 & 0.6 & 0.47 & 0.09 \\
\multicolumn{1}{l|}{} & \multicolumn{1}{l|}{ROUGE-L} & 0.73 & 0.7 & 0.63 & 0.11 \\ \hline
\multicolumn{1}{l|}{\multirow{8}{*}{\begin{tabular}[c]{@{}l@{}}Mistral\\ -v0.3-7B\end{tabular}}} & \multicolumn{1}{l|}{BLEU} & 36.39 & 15.68 & 40.77 & 21.28 \\
\multicolumn{1}{l|}{} & \multicolumn{1}{l|}{BERTSc} & 0.92 & 0.79 & 0.86 & 0.78 \\
\multicolumn{1}{l|}{} & \multicolumn{1}{l|}{TER} & 80.09 & 73.73 & 53.76 & 66.07 \\
\multicolumn{1}{l|}{} & \multicolumn{1}{l|}{METEOR} & 0.72 & 0.35 & 0.63 & 0.42 \\
\multicolumn{1}{l|}{} & \multicolumn{1}{l|}{ChrF} & 66.77 & 32.48 & 61.34 & 22.37 \\
\multicolumn{1}{l|}{} & \multicolumn{1}{l|}{ROUGE-1} & 0.64 & 0.61 & 0.66 & 0.09 \\
\multicolumn{1}{l|}{} & \multicolumn{1}{l|}{ROUGE-2} & 0.49 & 0.38 & 0.47 & 0.07 \\
\multicolumn{1}{l|}{} & \multicolumn{1}{l|}{ROUGE-L} & 0.61 & 0.48 & 0.62 & 0.09 \\ \hline
\multicolumn{6}{c}{Encoder-decoder} \\ \hline
\multicolumn{1}{l|}{\multirow{8}{*}{\begin{tabular}[c]{@{}l@{}}mBart\\ -large-50\end{tabular}}} & \multicolumn{1}{l|}{BLEU} & 41.45 & 41.12 & 40.48 & 30.43 \\
\multicolumn{1}{l|}{} & \multicolumn{1}{l|}{BERTSc} & 0.94 & 0.87 & 0.87 & 0.8 \\
\multicolumn{1}{l|}{} & \multicolumn{1}{l|}{TER} & 48.34 & 51.45 & 50.82 & 56.79 \\
\multicolumn{1}{l|}{} & \multicolumn{1}{l|}{METEOR} & 0.68 & 0.65 & 0.63 & 0.54 \\
\multicolumn{1}{l|}{} & \multicolumn{1}{l|}{ChrF} & 60.02 & 55.8 & 60.48 & 26.35 \\
\multicolumn{1}{l|}{} & \multicolumn{1}{l|}{ROUGE-1} & 0.68 & 0.77 & 0.65 & 0.13 \\
\multicolumn{1}{l|}{} & \multicolumn{1}{l|}{ROUGE-2} & 0.47 & 0.56 & 0.46 & 0.11 \\
\multicolumn{1}{l|}{} & \multicolumn{1}{l|}{ROUGE-L} & 0.65 & 0.67 & 0.61 & 0.13 \\ \hline
\multicolumn{1}{l|}{\multirow{8}{*}{\begin{tabular}[c]{@{}l@{}}M2M100\\ -418M\end{tabular}}} & \multicolumn{1}{l|}{BLEU} & 44.76 & 43.83 & 43.53 & 30.42 \\
\multicolumn{1}{l|}{} & \multicolumn{1}{l|}{BERTSc} & 0.83 & 0.85 & 0.88 & 0.75 \\
\multicolumn{1}{l|}{} & \multicolumn{1}{l|}{TER} & 46.56 & 48.72 & 47.31 & 54.14 \\
\multicolumn{1}{l|}{} & \multicolumn{1}{l|}{METEOR} & 0.68 & 0.67 & 0.67 & 0.53 \\
\multicolumn{1}{l|}{} & \multicolumn{1}{l|}{ChrF} & 62.33 & 58.19 & 64.91 & 27.56 \\
\multicolumn{1}{l|}{} & \multicolumn{1}{l|}{ROUGE-1} & 0.67 & 0.77 & 0.69 & 0.13 \\
\multicolumn{1}{l|}{} & \multicolumn{1}{l|}{ROUGE-2} & 0.48 & 0.58 & 0.5 & 0.11 \\
\multicolumn{1}{l|}{} & \multicolumn{1}{l|}{ROUGE-L} & 0.64 & 0.67 & 0.65 & 0.13 \\ \hline
\multicolumn{1}{l|}{\multirow{8}{*}{Marian}} & \multicolumn{1}{l|}{BLEU} & 42.74 & 38.26 & 39.59 & 18.11 \\
\multicolumn{1}{l|}{} & \multicolumn{1}{l|}{BERTSc} & 0.88 & 0.87 & 0.87 & 0.78 \\
\multicolumn{1}{l|}{} & \multicolumn{1}{l|}{TER} & 47.25 & 52.21 & 50.87 & 67.04 \\
\multicolumn{1}{l|}{} & \multicolumn{1}{l|}{METEOR} & 0.71 & 0.64 & 0.64 & 0.42 \\
\multicolumn{1}{l|}{} & \multicolumn{1}{l|}{ChrF} & 62.6 & 54.26 & 61.55 & 16.79 \\
\multicolumn{1}{l|}{} & \multicolumn{1}{l|}{ROUGE-1} & 0.7 & 0.78 & 0.66 & 0.13 \\
\multicolumn{1}{l|}{} & \multicolumn{1}{l|}{ROUGE-2} & 0.49 & 0.55 & 0.46 & 0.12 \\
\multicolumn{1}{l|}{} & \multicolumn{1}{l|}{ROUGE-L} & 0.67 & 0.66 & 0.62 & 0.13 \\ \hline
\multicolumn{6}{c}{Commercial tool} \\ \hline
\multicolumn{1}{l|}{\multirow{8}{*}{\begin{tabular}[c]{@{}l@{}}Google\\ Translate\end{tabular}}} & \multicolumn{1}{l|}{BLEU} & 40.74 & 32.69 & 33.15 & 31.89 \\
\multicolumn{1}{l|}{} & \multicolumn{1}{l|}{BERTSc} & 0.9 & 0.88 & 0.88 & 0.86 \\
\multicolumn{1}{l|}{} & \multicolumn{1}{l|}{TER} & 50.5 & 57.38 & 58.0 & 57.64 \\
\multicolumn{1}{l|}{} & \multicolumn{1}{l|}{METEOR} & 0.63 & 0.56 & 0.55 & 0.54 \\
\multicolumn{1}{l|}{} & \multicolumn{1}{l|}{ChrF} & 60.31 & 50.13 & 56.07 & 28.01 \\
\multicolumn{1}{l|}{} & \multicolumn{1}{l|}{ROUGE-1} & 0.68 & 0.74 & 0.61 & 0.12 \\
\multicolumn{1}{l|}{} & \multicolumn{1}{l|}{ROUGE-2} & 0.51 & 0.53 & 0.43 & 0.09 \\
\multicolumn{1}{l|}{} & \multicolumn{1}{l|}{ROUGE-L} & 0.65 & 0.63 & 0.57 & 0.12
\\ \hline
\caption{\textbf{Full Results: Ground-truth Translation Baselines.} \textbf{German to X} results are reported in this table. Extension of Table \ref{tab:translation-groundtruth} in the main paper.}
\label{tab:appx_nmt_gt_allMetrics-De-X}
\end{longtable}

%% file: tables/appx_nmt_gt_allMetrics-Zh-X.tex
\begin{longtable}{llcccc}
\hline
\multicolumn{1}{l|}{\textbf{MT}} & \multicolumn{1}{l|}{\textbf{Metrics}} & \textbf{zh-en} & \textbf{zh-vi} & \textbf{zh-fr} & \textbf{zh-de} \\ \hline
\endfirsthead
\multicolumn{6}{c}%
{{\bfseries Table \thetable\ continued from previous page}} \\
\hline
\multicolumn{1}{l|}{\textbf{MT}} & \multicolumn{1}{l|}{\textbf{Metrics}} & \textbf{zh-en} & \textbf{zh-vi} & \textbf{zh-fr} & \textbf{zh-de} \\ \hline
\endhead
\multicolumn{6}{c}{Decoder} \\ \hline
\multicolumn{1}{l|}{\multirow{8}{*}{\begin{tabular}[c]{@{}l@{}}Llama\\ -3.1-8B\end{tabular}}} & \multicolumn{1}{l|}{BLEU} & 28.21 & 23.49 & 18.87 & 13.07 \\
\multicolumn{1}{l|}{} & \multicolumn{1}{l|}{BERTSc} & 0.91 & 0.79 & 0.77 & 0.74 \\
\multicolumn{1}{l|}{} & \multicolumn{1}{l|}{TER} & 84.4 & 96.2 & 99.18 & 107.48 \\
\multicolumn{1}{l|}{} & \multicolumn{1}{l|}{METEOR} & 0.59 & 0.55 & 0.46 & 0.4 \\
\multicolumn{1}{l|}{} & \multicolumn{1}{l|}{ChrF} & 47.42 & 41.33 & 41.69 & 32.09 \\
\multicolumn{1}{l|}{} & \multicolumn{1}{l|}{ROUGE-1} & 0.57 & 0.63 & 0.46 & 0.37 \\
\multicolumn{1}{l|}{} & \multicolumn{1}{l|}{ROUGE-2} & 0.37 & 0.44 & 0.28 & 0.19 \\
\multicolumn{1}{l|}{} & \multicolumn{1}{l|}{ROUGE-L} & 0.53 & 0.53 & 0.41 & 0.33 \\ \hline
\multicolumn{1}{l|}{\multirow{8}{*}{\begin{tabular}[c]{@{}l@{}}Qwen\\ -2.5-7B\end{tabular}}} & \multicolumn{1}{l|}{BLEU} & 35.63 & 32.95 & 24.05 & 16.95 \\
\multicolumn{1}{l|}{} & \multicolumn{1}{l|}{BERTSc} & 0.95 & 0.85 & 0.84 & 0.83 \\
\multicolumn{1}{l|}{} & \multicolumn{1}{l|}{TER} & 60.98 & 67.86 & 73.13 & 77.31 \\
\multicolumn{1}{l|}{} & \multicolumn{1}{l|}{METEOR} & 0.67 & 0.63 & 0.53 & 0.5 \\
\multicolumn{1}{l|}{} & \multicolumn{1}{l|}{ChrF} & 55.3 & 47.73 & 47.09 & 39.65 \\
\multicolumn{1}{l|}{} & \multicolumn{1}{l|}{ROUGE-1} & 0.66 & 0.74 & 0.55 & 0.52 \\
\multicolumn{1}{l|}{} & \multicolumn{1}{l|}{ROUGE-2} & 0.45 & 0.53 & 0.35 & 0.28 \\
\multicolumn{1}{l|}{} & \multicolumn{1}{l|}{ROUGE-L} & 0.62 & 0.63 & 0.5 & 0.48 \\ \hline
\multicolumn{1}{l|}{\multirow{8}{*}{\begin{tabular}[c]{@{}l@{}}Mistral\\ -v0.3-7B\end{tabular}}} & \multicolumn{1}{l|}{BLEU} & 27.68 & 10.67 & 18.46 & 11.4 \\
\multicolumn{1}{l|}{} & \multicolumn{1}{l|}{BERTSc} & 0.93 & 0.75 & 0.8 & 0.76 \\
\multicolumn{1}{l|}{} & \multicolumn{1}{l|}{TER} & 72.65 & 82.73 & 81.15 & 90.07 \\
\multicolumn{1}{l|}{} & \multicolumn{1}{l|}{METEOR} & 0.59 & 0.33 & 0.45 & 0.39 \\
\multicolumn{1}{l|}{} & \multicolumn{1}{l|}{ChrF} & 47.23 & 24.79 & 39.83 & 31.55 \\
\multicolumn{1}{l|}{} & \multicolumn{1}{l|}{ROUGE-1} & 0.57 & 0.56 & 0.47 & 0.4 \\
\multicolumn{1}{l|}{} & \multicolumn{1}{l|}{ROUGE-2} & 0.36 & 0.32 & 0.28 & 0.2 \\
\multicolumn{1}{l|}{} & \multicolumn{1}{l|}{ROUGE-L} & 0.53 & 0.43 & 0.42 & 0.37 \\ \hline
\multicolumn{6}{c}{Encoder-decoder} \\ \hline
\multicolumn{1}{l|}{\multirow{8}{*}{\begin{tabular}[c]{@{}l@{}}mBart\\ -large-50\end{tabular}}} & \multicolumn{1}{l|}{BLEU} & 15.03 & 22.28 & 15.7 & 10.67 \\
\multicolumn{1}{l|}{} & \multicolumn{1}{l|}{BERTSc} & 0.9 & 0.82 & 0.79 & 0.77 \\
\multicolumn{1}{l|}{} & \multicolumn{1}{l|}{TER} & 76.4 & 71.86 & 79.51 & 85.8 \\
\multicolumn{1}{l|}{} & \multicolumn{1}{l|}{METEOR} & 0.42 & 0.49 & 0.39 & 0.36 \\
\multicolumn{1}{l|}{} & \multicolumn{1}{l|}{ChrF} & 36.18 & 39.38 & 38.64 & 33.11 \\
\multicolumn{1}{l|}{} & \multicolumn{1}{l|}{ROUGE-1} & 0.46 & 0.68 & 0.44 & 0.4 \\
\multicolumn{1}{l|}{} & \multicolumn{1}{l|}{ROUGE-2} & 0.23 & 0.44 & 0.23 & 0.18 \\
\multicolumn{1}{l|}{} & \multicolumn{1}{l|}{ROUGE-L} & 0.42 & 0.55 & 0.39 & 0.34 \\ \hline
\multicolumn{1}{l|}{\multirow{8}{*}{\begin{tabular}[c]{@{}l@{}}M2M100\\ -418M\end{tabular}}} & \multicolumn{1}{l|}{BLEU} & 21.65 & 27.69 & 21.88 & 15.17 \\
\multicolumn{1}{l|}{} & \multicolumn{1}{l|}{BERTSc} & 0.76 & 0.85 & 0.82 & 0.82 \\
\multicolumn{1}{l|}{} & \multicolumn{1}{l|}{TER} & 72.58 & 66.12 & 72.93 & 78.47 \\
\multicolumn{1}{l|}{} & \multicolumn{1}{l|}{METEOR} & 0.52 & 0.55 & 0.5 & 0.46 \\
\multicolumn{1}{l|}{} & \multicolumn{1}{l|}{ChrF} & 43.67 & 43.39 & 46.57 & 39.77 \\
\multicolumn{1}{l|}{} & \multicolumn{1}{l|}{ROUGE-1} & 0.51 & 0.67 & 0.52 & 0.49 \\
\multicolumn{1}{l|}{} & \multicolumn{1}{l|}{ROUGE-2} & 0.28 & 0.47 & 0.32 & 0.25 \\
\multicolumn{1}{l|}{} & \multicolumn{1}{l|}{ROUGE-L} & 0.47 & 0.57 & 0.47 & 0.44 \\ \hline
\multicolumn{1}{l|}{\multirow{8}{*}{Marian}} & \multicolumn{1}{l|}{BLEU} & 11.44 & 16.14 & 11.33 & 6.24 \\
\multicolumn{1}{l|}{} & \multicolumn{1}{l|}{BERTSc} & 0.78 & 0.79 & 0.77 & 0.75 \\
\multicolumn{1}{l|}{} & \multicolumn{1}{l|}{TER} & 86.03 & 83.81 & 89.53 & 104.78 \\
\multicolumn{1}{l|}{} & \multicolumn{1}{l|}{METEOR} & 0.38 & 0.42 & 0.33 & 0.3 \\
\multicolumn{1}{l|}{} & \multicolumn{1}{l|}{ChrF} & 32.3 & 33.04 & 33.64 & 29.48 \\
\multicolumn{1}{l|}{} & \multicolumn{1}{l|}{ROUGE-1} & 0.39 & 0.63 & 0.37 & 0.33 \\
\multicolumn{1}{l|}{} & \multicolumn{1}{l|}{ROUGE-2} & 0.16 & 0.35 & 0.17 & 0.11 \\
\multicolumn{1}{l|}{} & \multicolumn{1}{l|}{ROUGE-L} & 0.35 & 0.48 & 0.32 & 0.28 \\ \hline
\multicolumn{6}{c}{Commercial tool} \\ \hline
\multicolumn{1}{l|}{\multirow{8}{*}{\begin{tabular}[c]{@{}l@{}}Google\\ Translate\end{tabular}}} & \multicolumn{1}{l|}{BLEU} & 27.74 & 30.7 & 20.71 & 19.11 \\
\multicolumn{1}{l|}{} & \multicolumn{1}{l|}{BERTSc} & 0.88 & 0.87 & 0.85 & 0.77 \\
\multicolumn{1}{l|}{} & \multicolumn{1}{l|}{TER} & 74.94 & 72.9 & 81.21 & 83.94 \\
\multicolumn{1}{l|}{} & \multicolumn{1}{l|}{METEOR} & 0.52 & 0.52 & 0.43 & 0.45 \\
\multicolumn{1}{l|}{} & \multicolumn{1}{l|}{ChrF} & 51.33 & 49.89 & 49.91 & 53.27 \\
\multicolumn{1}{l|}{} & \multicolumn{1}{l|}{ROUGE-1} & 0.55 & 0.69 & 0.48 & 0.48 \\
\multicolumn{1}{l|}{} & \multicolumn{1}{l|}{ROUGE-2} & 0.34 & 0.46 & 0.29 & 0.26 \\
\multicolumn{1}{l|}{} & \multicolumn{1}{l|}{ROUGE-L} & 0.51 & 0.56 & 0.43 & 0.43
\\ \hline
\caption{\textbf{Full Results: Ground-truth Translation Baselines.} \textbf{Chinese to X} results are reported in this table. Extension of Table \ref{tab:translation-groundtruth} in the main paper.}
\label{tab:appx_nmt_gt_allMetrics-Zh-X}
\end{longtable}

%% file: tables/appx_nmt_asr_allMetrics-En-X.tex
\begin{longtable}{llcccc}
\hline
\multicolumn{1}{l|}{\textbf{MT}} & \multicolumn{1}{l|}{\textbf{Metrics}} & \textbf{en-vi} & \textbf{en-fr} & \textbf{en-zh} & \textbf{en-de} \\ \hline
\endfirsthead
\multicolumn{6}{c}%
{{\bfseries Table \thetable\ continued from previous page}} \\
\hline
\multicolumn{1}{l|}{\textbf{MT}} & \multicolumn{1}{l|}{\textbf{Metrics}} & \textbf{en-vi} & \textbf{en-fr} & \textbf{en-zh} & \textbf{en-de} \\ \hline
\endhead
\hline
\endfoot
\endlastfoot
\multicolumn{6}{c}{Decoder} \\ \hline
\multicolumn{1}{l|}{\multirow{8}{*}{\begin{tabular}[c]{@{}l@{}}Llama\\ -3.1-8B\end{tabular}}} & \multicolumn{1}{l|}{BLEU} & 43.32 & 37.92 & 30.78 & 31.36 \\
\multicolumn{1}{l|}{} & \multicolumn{1}{l|}{BERTSc} & 0.78 & 0.76 & 0.73 & 0.74 \\
\multicolumn{1}{l|}{} & \multicolumn{1}{l|}{TER} & 53.71 & 60.93 & 66.54 & 67.05 \\
\multicolumn{1}{l|}{} & \multicolumn{1}{l|}{METEOR} & 0.63 & 0.59 & 0.51 & 0.53 \\
\multicolumn{1}{l|}{} & \multicolumn{1}{l|}{ChrF} & 57.23 & 60.63 & 26.34 & 56.58 \\
\multicolumn{1}{l|}{} & \multicolumn{1}{l|}{ROUGE-1} & 0.76 & 0.63 & 0.13 & 0.57 \\
\multicolumn{1}{l|}{} & \multicolumn{1}{l|}{ROUGE-2} & 0.58 & 0.47 & 0.11 & 0.37 \\
\multicolumn{1}{l|}{} & \multicolumn{1}{l|}{ROUGE-L} & 0.67 & 0.59 & 0.13 & 0.53 \\ \hline
\multicolumn{1}{l|}{\multirow{8}{*}{\begin{tabular}[c]{@{}l@{}}Qwen\\ -2.5-7B\end{tabular}}} & \multicolumn{1}{l|}{BLEU} & 43.37 & 37.34 & 23.46 & 28.5 \\
\multicolumn{1}{l|}{} & \multicolumn{1}{l|}{BERTSc} & 0.77 & 0.76 & 0.74 & 0.74 \\
\multicolumn{1}{l|}{} & \multicolumn{1}{l|}{TER} & 53.52 & 57.76 & 64.75 & 66.7 \\
\multicolumn{1}{l|}{} & \multicolumn{1}{l|}{METEOR} & 0.63 & 0.58 & 0.44 & 0.51 \\
\multicolumn{1}{l|}{} & \multicolumn{1}{l|}{ChrF} & 57.34 & 60.21 & 23.19 & 54.19 \\
\multicolumn{1}{l|}{} & \multicolumn{1}{l|}{ROUGE-1} & 0.76 & 0.63 & 0.12 & 0.55 \\
\multicolumn{1}{l|}{} & \multicolumn{1}{l|}{ROUGE-2} & 0.59 & 0.47 & 0.1 & 0.34 \\
\multicolumn{1}{l|}{} & \multicolumn{1}{l|}{ROUGE-L} & 0.67 & 0.6 & 0.12 & 0.51 \\ \hline
\multicolumn{1}{l|}{\multirow{8}{*}{\begin{tabular}[c]{@{}l@{}}Mistral\\ -v0.3-7B\end{tabular}}} & \multicolumn{1}{l|}{BLEU} & 17.72 & 36.58 & 20.27 & 29.9 \\
\multicolumn{1}{l|}{} & \multicolumn{1}{l|}{BERTSc} & 0.68 & 0.74 & 0.69 & 0.72 \\
\multicolumn{1}{l|}{} & \multicolumn{1}{l|}{TER} & 71.27 & 59.78 & 70.94 & 65.81 \\
\multicolumn{1}{l|}{} & \multicolumn{1}{l|}{METEOR} & 0.35 & 0.55 & 0.38 & 0.5 \\
\multicolumn{1}{l|}{} & \multicolumn{1}{l|}{ChrF} & 33.42 & 57.32 & 20.29 & 52.89 \\
\multicolumn{1}{l|}{} & \multicolumn{1}{l|}{ROUGE-1} & 0.57 & 0.6 & 0.09 & 0.54 \\
\multicolumn{1}{l|}{} & \multicolumn{1}{l|}{ROUGE-2} & 0.39 & 0.45 & 0.08 & 0.36 \\
\multicolumn{1}{l|}{} & \multicolumn{1}{l|}{ROUGE-L} & 0.49 & 0.58 & 0.09 & 0.51 \\ \hline
\multicolumn{6}{c}{Encoder-decoder} \\ \hline
\multicolumn{1}{l|}{\multirow{8}{*}{\begin{tabular}[c]{@{}l@{}}mBart\\ -large-50\end{tabular}}} & \multicolumn{1}{l|}{BLEU} & 48.0 & 43.2 & 35.7 & 35.07 \\
\multicolumn{1}{l|}{} & \multicolumn{1}{l|}{BERTSc} & 0.87 & 0.86 & 0.81 & 0.84 \\
\multicolumn{1}{l|}{} & \multicolumn{1}{l|}{TER} & 46.21 & 51.32 & 54.44 & 59.01 \\
\multicolumn{1}{l|}{} & \multicolumn{1}{l|}{METEOR} & 0.66 & 0.62 & 0.57 & 0.56 \\
\multicolumn{1}{l|}{} & \multicolumn{1}{l|}{ChrF} & 61.1 & 64.07 & 31.25 & 58.51 \\
\multicolumn{1}{l|}{} & \multicolumn{1}{l|}{ROUGE-1} & 0.78 & 0.67 & 0.15 & 0.61 \\
\multicolumn{1}{l|}{} & \multicolumn{1}{l|}{ROUGE-2} & 0.63 & 0.52 & 0.13 & 0.41 \\
\multicolumn{1}{l|}{} & \multicolumn{1}{l|}{ROUGE-L} & 0.71 & 0.64 & 0.15 & 0.57 \\ \hline
\multicolumn{1}{l|}{\multirow{8}{*}{\begin{tabular}[c]{@{}l@{}}M2M100\\ -418M\end{tabular}}} & \multicolumn{1}{l|}{BLEU} & 48.21 & 43.16 & 36.94 & 36.55 \\
\multicolumn{1}{l|}{} & \multicolumn{1}{l|}{BERTSc} & 0.95 & 0.92 & 0.92 & 0.92 \\
\multicolumn{1}{l|}{} & \multicolumn{1}{l|}{TER} & 46.94 & 51.58 & 53.79 & 57.62 \\
\multicolumn{1}{l|}{} & \multicolumn{1}{l|}{METEOR} & 0.67 & 0.63 & 0.58 & 0.57 \\
\multicolumn{1}{l|}{} & \multicolumn{1}{l|}{ChrF} & 61.29 & 64.22 & 32.22 & 60.13 \\
\multicolumn{1}{l|}{} & \multicolumn{1}{l|}{ROUGE-1} & 0.78 & 0.67 & 0.16 & 0.62 \\
\multicolumn{1}{l|}{} & \multicolumn{1}{l|}{ROUGE-2} & 0.63 & 0.52 & 0.14 & 0.43 \\
\multicolumn{1}{l|}{} & \multicolumn{1}{l|}{ROUGE-L} & 0.71 & 0.65 & 0.16 & 0.58 \\ \hline
\multicolumn{1}{l|}{\multirow{8}{*}{Marian}} & \multicolumn{1}{l|}{BLEU} & 45.07 & 40.54 & 31.17 & 33.9 \\
\multicolumn{1}{l|}{} & \multicolumn{1}{l|}{BERTSc} & 0.87 & 0.86 & 0.82 & 0.84 \\
\multicolumn{1}{l|}{} & \multicolumn{1}{l|}{TER} & 49.1 & 53.41 & 57.5 & 59.75 \\
\multicolumn{1}{l|}{} & \multicolumn{1}{l|}{METEOR} & 0.65 & 0.61 & 0.53 & 0.56 \\
\multicolumn{1}{l|}{} & \multicolumn{1}{l|}{ChrF} & 58.93 & 62.68 & 27.33 & 58.41 \\
\multicolumn{1}{l|}{} & \multicolumn{1}{l|}{ROUGE-1} & 0.77 & 0.66 & 0.15 & 0.6 \\
\multicolumn{1}{l|}{} & \multicolumn{1}{l|}{ROUGE-2} & 0.61 & 0.5 & 0.13 & 0.4 \\
\multicolumn{1}{l|}{} & \multicolumn{1}{l|}{ROUGE-L} & 0.69 & 0.63 & 0.15 & 0.57
\\ \hline
\caption{\textbf{Extra Results: Cascaded \gls{ST} Baselines.} \textbf{English to X} results are reported in this table. Supplement of Table \ref{tab:asr-translation} in the main paper.\\
All cascaded models use Whisper$_{small-mono}$ as \gls{ASR} model (Whisper \gls{ASR} is fine-tuned monolingually - on each source language separately). Its \gls{WER} on test set is 29.6\%, 33.8\%, 31.3\%, 26.3\%, 45.7\% for Vietnamese, English, Chinese, German and French respectively.}
\label{tab:appx_nmt_asr_allMetrics-En-X}
\end{longtable}

%% file: tables/appx_nmt_asr_allMetrics-Vi-X.tex
\begin{longtable}{llcccc}
\hline
\multicolumn{1}{l|}{\textbf{MT}} & \multicolumn{1}{l|}{\textbf{Metrics}} & \textbf{vi-en} & \textbf{vi-fr} & \textbf{vi-zh} & \textbf{vi-de} \\ \hline
\endfirsthead
\multicolumn{6}{c}%
{{\bfseries Table \thetable\ continued from previous page}} \\
\hline
\multicolumn{1}{l|}{\textbf{MT}} & \multicolumn{1}{l|}{\textbf{Metrics}} & \textbf{vi-en} & \textbf{vi-fr} & \textbf{vi-zh} & \textbf{vi-de} \\ \hline
\endhead
\hline
\endfoot
\endlastfoot
\multicolumn{6}{c}{Decoder} \\ \hline
\multicolumn{1}{l|}{\multirow{8}{*}{\begin{tabular}[c]{@{}l@{}}Llama\\ -3.1-8B\end{tabular}}} & \multicolumn{1}{l|}{BLEU} & 14.55 & 10.29 & 11.56 & 7.71 \\
\multicolumn{1}{l|}{} & \multicolumn{1}{l|}{BERTSc} & 0.78 & 0.75 & 0.73 & 0.73 \\
\multicolumn{1}{l|}{} & \multicolumn{1}{l|}{TER} & 98.39 & 112.97 & 131.0 & 122.33 \\
\multicolumn{1}{l|}{} & \multicolumn{1}{l|}{METEOR} & 0.43 & 0.35 & 0.4 & 0.31 \\
\multicolumn{1}{l|}{} & \multicolumn{1}{l|}{ChrF} & 41.46 & 39.22 & 15.65 & 37.02 \\
\multicolumn{1}{l|}{} & \multicolumn{1}{l|}{ROUGE-1} & 0.44 & 0.36 & 0.02 & 0.32 \\
\multicolumn{1}{l|}{} & \multicolumn{1}{l|}{ROUGE-2} & 0.22 & 0.18 & 0.01 & 0.12 \\
\multicolumn{1}{l|}{} & \multicolumn{1}{l|}{ROUGE-L} & 0.39 & 0.31 & 0.02 & 0.28 \\ \hline
\multicolumn{1}{l|}{\multirow{8}{*}{\begin{tabular}[c]{@{}l@{}}Qwen\\ -2.5-7B\end{tabular}}} & \multicolumn{1}{l|}{BLEU} & 13.97 & 11.66 & 20.27 & 8.75 \\
\multicolumn{1}{l|}{} & \multicolumn{1}{l|}{BERTSc} & 0.78 & 0.76 & 0.78 & 0.75 \\
\multicolumn{1}{l|}{} & \multicolumn{1}{l|}{TER} & 101.68 & 101.51 & 74.57 & 101.85 \\
\multicolumn{1}{l|}{} & \multicolumn{1}{l|}{METEOR} & 0.43 & 0.36 & 0.43 & 0.3 \\
\multicolumn{1}{l|}{} & \multicolumn{1}{l|}{ChrF} & 41.76 & 40.73 & 18.0 & 36.74 \\
\multicolumn{1}{l|}{} & \multicolumn{1}{l|}{ROUGE-1} & 0.43 & 0.39 & 0.03 & 0.34 \\
\multicolumn{1}{l|}{} & \multicolumn{1}{l|}{ROUGE-2} & 0.21 & 0.19 & 0.02 & 0.13 \\
\multicolumn{1}{l|}{} & \multicolumn{1}{l|}{ROUGE-L} & 0.39 & 0.33 & 0.03 & 0.29 \\ \hline
\multicolumn{1}{l|}{\multirow{8}{*}{\begin{tabular}[c]{@{}l@{}}Mistral\\ -v0.3-7B\end{tabular}}} & \multicolumn{1}{l|}{BLEU} & 15.86 & 10.92 & 17.92 & 09.03 \\
\multicolumn{1}{l|}{} & \multicolumn{1}{l|}{BERTSc} & 0.78 & 0.75 & 0.77 & 0.75 \\
\multicolumn{1}{l|}{} & \multicolumn{1}{l|}{TER} & 85.32 & 97.69 & 76.97 & 94.56 \\
\multicolumn{1}{l|}{} & \multicolumn{1}{l|}{METEOR} & 0.42 & 0.33 & 0.39 & 0.3 \\
\multicolumn{1}{l|}{} & \multicolumn{1}{l|}{ChrF} & 39.64 & 37.19 & 16.1 & 34.82 \\
\multicolumn{1}{l|}{} & \multicolumn{1}{l|}{ROUGE-1} & 0.43 & 0.36 & 0.02 & 0.33 \\
\multicolumn{1}{l|}{} & \multicolumn{1}{l|}{ROUGE-2} & 0.21 & 0.17 & 0.01 & 0.12 \\
\multicolumn{1}{l|}{} & \multicolumn{1}{l|}{ROUGE-L} & 0.39 & 0.32 & 0.02 & 0.29 \\ \hline
\multicolumn{6}{c}{Encoder-decoder} \\ \hline
\multicolumn{1}{l|}{\multirow{8}{*}{\begin{tabular}[c]{@{}l@{}}mBart\\ -large-50\end{tabular}}} & \multicolumn{1}{l|}{BLEU} & 10.17 & 12.8 & 16.77 & 7.23 \\
\multicolumn{1}{l|}{} & \multicolumn{1}{l|}{BERTSc} & 0.88 & 0.76 & 0.73 & 0.72 \\
\multicolumn{1}{l|}{} & \multicolumn{1}{l|}{TER} & 86.47 & 87.93 & 74.4 & 104.87 \\
\multicolumn{1}{l|}{} & \multicolumn{1}{l|}{METEOR} & 0.32 & 0.33 & 0.39 & 0.25 \\
\multicolumn{1}{l|}{} & \multicolumn{1}{l|}{ChrF} & 31.78 & 37.76 & 15.29 & 31.98 \\
\multicolumn{1}{l|}{} & \multicolumn{1}{l|}{ROUGE-1} & 0.36 & 0.37 & 0.02 & 0.28 \\
\multicolumn{1}{l|}{} & \multicolumn{1}{l|}{ROUGE-2} & 0.14 & 0.18 & 0.01 & 0.1 \\
\multicolumn{1}{l|}{} & \multicolumn{1}{l|}{ROUGE-L} & 0.32 & 0.33 & 0.02 & 0.24 \\ \hline
\multicolumn{1}{l|}{\multirow{8}{*}{\begin{tabular}[c]{@{}l@{}}M2M100\\ -418M\end{tabular}}} & \multicolumn{1}{l|}{BLEU} & 15.64 & 13.95 & 16.99 & 11.1 \\
\multicolumn{1}{l|}{} & \multicolumn{1}{l|}{BERTSc} & 0.78 & 0.77 & 0.74 & 0.75 \\
\multicolumn{1}{l|}{} & \multicolumn{1}{l|}{TER} & 82.03 & 86.73 & 75.18 & 89.49 \\
\multicolumn{1}{l|}{} & \multicolumn{1}{l|}{METEOR} & 0.41 & 0.36 & 0.39 & 0.32 \\
\multicolumn{1}{l|}{} & \multicolumn{1}{l|}{ChrF} & 39.68 & 40.25 & 15.51 & 37.54 \\
\multicolumn{1}{l|}{} & \multicolumn{1}{l|}{ROUGE-1} & 0.43 & 0.4 & 0.03 & 0.36 \\
\multicolumn{1}{l|}{} & \multicolumn{1}{l|}{ROUGE-2} & 0.2 & 0.2 & 0.01 & 0.14 \\
\multicolumn{1}{l|}{} & \multicolumn{1}{l|}{ROUGE-L} & 0.39 & 0.35 & 0.03 & 0.31 \\ \hline
\multicolumn{1}{l|}{\multirow{8}{*}{Marian}} & \multicolumn{1}{l|}{BLEU} & 12.95 & 11.23 & 12.09 & 09.08 \\
\multicolumn{1}{l|}{} & \multicolumn{1}{l|}{BERTSc} & 0.77 & 0.76 & 0.75 & 0.74 \\
\multicolumn{1}{l|}{} & \multicolumn{1}{l|}{TER} & 85.96 & 89.57 & 79.67 & 91.54 \\
\multicolumn{1}{l|}{} & \multicolumn{1}{l|}{METEOR} & 0.37 & 0.32 & 0.33 & 0.28 \\
\multicolumn{1}{l|}{} & \multicolumn{1}{l|}{ChrF} & 36.61 & 37.06 & 11.93 & 34.59 \\
\multicolumn{1}{l|}{} & \multicolumn{1}{l|}{ROUGE-1} & 0.4 & 0.36 & 0.02 & 0.33 \\
\multicolumn{1}{l|}{} & \multicolumn{1}{l|}{ROUGE-2} & 0.17 & 0.17 & 0.01 & 0.12 \\
\multicolumn{1}{l|}{} & \multicolumn{1}{l|}{ROUGE-L} & 0.36 & 0.32 & 0.02 & 0.28 \\ \hline
\caption{\textbf{Extra Results: Cascaded \gls{ST} Baselines.} \textbf{Vietnamese to X} results are reported in this table. Supplement of Table \ref{tab:asr-translation} in the main paper.\\
All cascaded models use Whisper$_{small-mono}$ as \gls{ASR} model (Whisper \gls{ASR} is fine-tuned monolingually - on each source language separately). Its \gls{WER} on test set is 29.6\%, 33.8\%, 31.3\%, 26.3\%, 45.7\% for Vietnamese, English, Chinese, German and French respectively.}
\label{tab:appx_nmt_asr_allMetrics-Vi-X}
\end{longtable}

%% file: tables/appx_nmt_asr_allMetrics-Fr-X.tex
\begin{longtable}{llcccc}
\hline
\multicolumn{1}{l|}{\textbf{MT}} & \multicolumn{1}{l|}{\textbf{Metrics}} & \textbf{fr-en} & \textbf{fr-vi} & \textbf{fr-zh} & \textbf{fr-de} \\ \hline
\endfirsthead
\multicolumn{6}{c}%
{{\bfseries Table \thetable\ continued from previous page}} \\
\hline
\multicolumn{1}{l|}{\textbf{MT}} & \multicolumn{1}{l|}{\textbf{Metrics}} & \textbf{fr-en} & \textbf{fr-vi} & \textbf{fr-zh} & \textbf{fr-de} \\ \hline
\endhead
\hline
\endfoot
\endlastfoot
\multicolumn{6}{c}{Decoder} \\ \hline
\multicolumn{1}{l|}{\multirow{8}{*}{\begin{tabular}[c]{@{}l@{}}Llama\\ -3.1-8B\end{tabular}}} & \multicolumn{1}{l|}{BLEU} & 30.15 & 25.36 & 20.28 & 16.38 \\
\multicolumn{1}{l|}{} & \multicolumn{1}{l|}{BERTSc} & 0.82 & 0.8 & 0.75 & 0.74 \\
\multicolumn{1}{l|}{} & \multicolumn{1}{l|}{TER} & 65.8 & 71.69 & 80.06 & 99.6 \\
\multicolumn{1}{l|}{} & \multicolumn{1}{l|}{METEOR} & 0.52 & 0.47 & 0.4 & 0.4 \\
\multicolumn{1}{l|}{} & \multicolumn{1}{l|}{ChrF} & 49.71 & 41.24 & 17.84 & 44.43 \\
\multicolumn{1}{l|}{} & \multicolumn{1}{l|}{ROUGE-1} & 0.58 & 0.67 & 0.08 & 0.41 \\
\multicolumn{1}{l|}{} & \multicolumn{1}{l|}{ROUGE-2} & 0.39 & 0.44 & 0.05 & 0.22 \\
\multicolumn{1}{l|}{} & \multicolumn{1}{l|}{ROUGE-L} & 0.54 & 0.55 & 0.08 & 0.37 \\ \hline
\multicolumn{1}{l|}{\multirow{8}{*}{\begin{tabular}[c]{@{}l@{}}Qwen\\ -2.5-7B\end{tabular}}} & \multicolumn{1}{l|}{BLEU} & 30.35 & 25.59 & 15.33 & 20.38 \\
\multicolumn{1}{l|}{} & \multicolumn{1}{l|}{BERTSc} & 0.81 & 0.8 & 0.76 & 0.78 \\
\multicolumn{1}{l|}{} & \multicolumn{1}{l|}{TER} & 69.6 & 71.42 & 73.88 & 76.51 \\
\multicolumn{1}{l|}{} & \multicolumn{1}{l|}{METEOR} & 0.52 & 0.47 & 0.36 & 0.4 \\
\multicolumn{1}{l|}{} & \multicolumn{1}{l|}{ChrF} & 49.88 & 41.85 & 15.7 & 43.63 \\
\multicolumn{1}{l|}{} & \multicolumn{1}{l|}{ROUGE-1} & 0.57 & 0.67 & 0.07 & 0.47 \\
\multicolumn{1}{l|}{} & \multicolumn{1}{l|}{ROUGE-2} & 0.39 & 0.45 & 0.05 & 0.26 \\
\multicolumn{1}{l|}{} & \multicolumn{1}{l|}{ROUGE-L} & 0.54 & 0.55 & 0.07 & 0.43 \\ \hline
\multicolumn{1}{l|}{\multirow{8}{*}{\begin{tabular}[c]{@{}l@{}}Mistral\\ -v0.3-7B\end{tabular}}} & \multicolumn{1}{l|}{BLEU} & 29.35 & 9.2 & 13.94 & 18.65 \\
\multicolumn{1}{l|}{} & \multicolumn{1}{l|}{BERTSc} & 0.79 & 0.74 & 0.76 & 0.78 \\
\multicolumn{1}{l|}{} & \multicolumn{1}{l|}{TER} & 74.28 & 82.83 & 74.11 & 74.55 \\
\multicolumn{1}{l|}{} & \multicolumn{1}{l|}{METEOR} & 0.52 & 0.26 & 0.34 & 0.39 \\
\multicolumn{1}{l|}{} & \multicolumn{1}{l|}{ChrF} & 49.85 & 23.69 & 15.16 & 41.66 \\
\multicolumn{1}{l|}{} & \multicolumn{1}{l|}{ROUGE-1} & 0.55 & 0.52 & 0.07 & 0.46 \\
\multicolumn{1}{l|}{} & \multicolumn{1}{l|}{ROUGE-2} & 0.38 & 0.27 & 0.05 & 0.25 \\
\multicolumn{1}{l|}{} & \multicolumn{1}{l|}{ROUGE-L} & 0.52 & 0.4 & 0.07 & 0.42 \\ \hline
\multicolumn{6}{c}{Encoder-decoder} \\ \hline
\multicolumn{1}{l|}{\multirow{8}{*}{\begin{tabular}[c]{@{}l@{}}mBart\\ -large-50\end{tabular}}} & \multicolumn{1}{l|}{BLEU} & 23.82 & 22.86 & 16.46 & 17.39 \\
\multicolumn{1}{l|}{} & \multicolumn{1}{l|}{BERTSc} & 0.9 & 0.8 & 0.72 & 0.77 \\
\multicolumn{1}{l|}{} & \multicolumn{1}{l|}{TER} & 71.43 & 71.72 & 74.54 & 77.56 \\
\multicolumn{1}{l|}{} & \multicolumn{1}{l|}{METEOR} & 0.47 & 0.44 & 0.36 & 0.38 \\
\multicolumn{1}{l|}{} & \multicolumn{1}{l|}{ChrF} & 44.7 & 39.14 & 15.58 & 40.21 \\
\multicolumn{1}{l|}{} & \multicolumn{1}{l|}{ROUGE-1} & 0.52 & 0.66 & 0.08 & 0.44 \\
\multicolumn{1}{l|}{} & \multicolumn{1}{l|}{ROUGE-2} & 0.32 & 0.41 & 0.05 & 0.22 \\
\multicolumn{1}{l|}{} & \multicolumn{1}{l|}{ROUGE-L} & 0.48 & 0.52 & 0.08 & 0.39 \\ \hline
\multicolumn{1}{l|}{\multirow{8}{*}{\begin{tabular}[c]{@{}l@{}}M2M100\\ -418M\end{tabular}}} & \multicolumn{1}{l|}{BLEU} & 25.65 & 21.88 & 18.44 & 19.98 \\
\multicolumn{1}{l|}{} & \multicolumn{1}{l|}{BERTSc} & 0.81 & 0.76 & 0.75 & 0.76 \\
\multicolumn{1}{l|}{} & \multicolumn{1}{l|}{TER} & 66.39 & 72.12 & 69.65 & 72.77 \\
\multicolumn{1}{l|}{} & \multicolumn{1}{l|}{METEOR} & 0.49 & 0.41 & 0.39 & 0.4 \\
\multicolumn{1}{l|}{} & \multicolumn{1}{l|}{ChrF} & 47.0 & 37.81 & 17.68 & 43.01 \\
\multicolumn{1}{l|}{} & \multicolumn{1}{l|}{ROUGE-1} & 0.55 & 0.58 & 0.1 & 0.47 \\
\multicolumn{1}{l|}{} & \multicolumn{1}{l|}{ROUGE-2} & 0.35 & 0.39 & 0.07 & 0.26 \\
\multicolumn{1}{l|}{} & \multicolumn{1}{l|}{ROUGE-L} & 0.51 & 0.48 & 0.1 & 0.43 \\ \hline
\multicolumn{1}{l|}{\multirow{8}{*}{Marian}} & \multicolumn{1}{l|}{BLEU} & 24.03 & 22.2 & 11.27 & 19.14 \\
\multicolumn{1}{l|}{} & \multicolumn{1}{l|}{BERTSc} & 0.81 & 0.8 & 0.74 & 0.79 \\
\multicolumn{1}{l|}{} & \multicolumn{1}{l|}{TER} & 67.61 & 73.99 & 78.35 & 73.19 \\
\multicolumn{1}{l|}{} & \multicolumn{1}{l|}{METEOR} & 0.47 & 0.44 & 0.31 & 0.4 \\
\multicolumn{1}{l|}{} & \multicolumn{1}{l|}{ChrF} & 45.53 & 38.42 & 11.78 & 42.62 \\
\multicolumn{1}{l|}{} & \multicolumn{1}{l|}{ROUGE-1} & 0.54 & 0.66 & 0.09 & 0.47 \\
\multicolumn{1}{l|}{} & \multicolumn{1}{l|}{ROUGE-2} & 0.33 & 0.41 & 0.06 & 0.25 \\
\multicolumn{1}{l|}{} & \multicolumn{1}{l|}{ROUGE-L} & 0.5 & 0.53 & 0.09 & 0.43 \\ \hline
\caption{\textbf{Extra Results: Cascaded \gls{ST} Baselines.} \textbf{French to X} results are reported in this table. Supplement of Table \ref{tab:asr-translation} in the main paper.\\
All cascaded models use Whisper$_{small-mono}$ as \gls{ASR} model (Whisper \gls{ASR} is fine-tuned monolingually - on each source language separately). Its \gls{WER} on test set is 29.6\%, 33.8\%, 31.3\%, 26.3\%, 45.7\% for Vietnamese, English, Chinese, German and French respectively.}
\label{tab:appx_nmt_asr_allMetrics-Fr-X}
\end{longtable}

%% file: tables/appx_nmt_asr_allMetrics-De-X.tex
\begin{longtable}{llcccc}
\hline
\multicolumn{1}{l|}{\textbf{MT}} & \multicolumn{1}{l|}{\textbf{Metrics}} & \textbf{de-en} & \textbf{de-vi} & \textbf{de-fr} & \textbf{de-zh} \\ \hline
\endfirsthead
\multicolumn{6}{c}%
{{\bfseries Table \thetable\ continued from previous page}} \\
\hline
\multicolumn{1}{l|}{\textbf{MT}} & \multicolumn{1}{l|}{\textbf{Metrics}} & \textbf{de-en} & \textbf{de-vi} & \textbf{de-fr} & \textbf{de-zh} \\ \hline
\endhead
\hline
\endfoot
\endlastfoot
\multicolumn{6}{c}{Decoder} \\ \hline
\multicolumn{1}{l|}{\multirow{8}{*}{\begin{tabular}[c]{@{}l@{}}Llama\\ -3.1-8B\end{tabular}}} & \multicolumn{1}{l|}{BLEU} & 40.63 & 33.63 & 26.97 & 26.31 \\
\multicolumn{1}{l|}{} & \multicolumn{1}{l|}{BERTSc} & 0.86 & 0.84 & 0.8 & 0.79 \\
\multicolumn{1}{l|}{} & \multicolumn{1}{l|}{TER} & 56.53 & 63.07 & 79.21 & 68.19 \\
\multicolumn{1}{l|}{} & \multicolumn{1}{l|}{METEOR} & 0.63 & 0.56 & 0.52 & 0.47 \\
\multicolumn{1}{l|}{} & \multicolumn{1}{l|}{ChrF} & 58.95 & 49.74 & 54.06 & 22.87 \\
\multicolumn{1}{l|}{} & \multicolumn{1}{l|}{ROUGE-1} & 0.65 & 0.73 & 0.54 & 0.1 \\
\multicolumn{1}{l|}{} & \multicolumn{1}{l|}{ROUGE-2} & 0.48 & 0.51 & 0.35 & 0.07 \\
\multicolumn{1}{l|}{} & \multicolumn{1}{l|}{ROUGE-L} & 0.62 & 0.61 & 0.49 & 0.1 \\ \hline
\multicolumn{1}{l|}{\multirow{8}{*}{\begin{tabular}[c]{@{}l@{}}Qwen\\ -2.5-7B\end{tabular}}} & \multicolumn{1}{l|}{BLEU} & 40.52 & 34.24 & 31.45 & 19.87 \\
\multicolumn{1}{l|}{} & \multicolumn{1}{l|}{BERTSc} & 0.86 & 0.84 & 0.84 & 0.79 \\
\multicolumn{1}{l|}{} & \multicolumn{1}{l|}{TER} & 55.76 & 59.16 & 61.65 & 68.06 \\
\multicolumn{1}{l|}{} & \multicolumn{1}{l|}{METEOR} & 0.63 & 0.57 & 0.53 & 0.4 \\
\multicolumn{1}{l|}{} & \multicolumn{1}{l|}{ChrF} & 59.05 & 50.58 & 54.36 & 20.15 \\
\multicolumn{1}{l|}{} & \multicolumn{1}{l|}{ROUGE-1} & 0.66 & 0.74 & 0.59 & 0.1 \\
\multicolumn{1}{l|}{} & \multicolumn{1}{l|}{ROUGE-2} & 0.47 & 0.53 & 0.4 & 0.08 \\
\multicolumn{1}{l|}{} & \multicolumn{1}{l|}{ROUGE-L} & 0.62 & 0.62 & 0.54 & 0.1 \\ \hline
\multicolumn{1}{l|}{\multirow{8}{*}{\begin{tabular}[c]{@{}l@{}}Mistral\\ -v0.3-7B\end{tabular}}} & \multicolumn{1}{l|}{BLEU} & 28.33 & 12.38 & 31.15 & 17.82 \\
\multicolumn{1}{l|}{} & \multicolumn{1}{l|}{BERTSc} & 0.78 & 0.77 & 0.83 & 0.78 \\
\multicolumn{1}{l|}{} & \multicolumn{1}{l|}{TER} & 91.13 & 77.89 & 63.4 & 70.11 \\
\multicolumn{1}{l|}{} & \multicolumn{1}{l|}{METEOR} & 0.6 & 0.3 & 0.52 & 0.37 \\
\multicolumn{1}{l|}{} & \multicolumn{1}{l|}{ChrF} & 57.3 & 28.44 & 52.73 & 18.9 \\
\multicolumn{1}{l|}{} & \multicolumn{1}{l|}{ROUGE-1} & 0.55 & 0.57 & 0.57 & 0.09 \\
\multicolumn{1}{l|}{} & \multicolumn{1}{l|}{ROUGE-2} & 0.4 & 0.33 & 0.39 & 0.07 \\
\multicolumn{1}{l|}{} & \multicolumn{1}{l|}{ROUGE-L} & 0.52 & 0.44 & 0.53 & 0.09 \\ \hline
\multicolumn{6}{c}{Encoder-decoder} \\ \hline
\multicolumn{1}{l|}{\multirow{8}{*}{\begin{tabular}[c]{@{}l@{}}mBart\\ -large-50\end{tabular}}} & \multicolumn{1}{l|}{BLEU} & 31.95 & 32.62 & 31.96 & 25.07 \\
\multicolumn{1}{l|}{} & \multicolumn{1}{l|}{BERTSc} & 0.92 & 0.84 & 0.84 & 0.77 \\
\multicolumn{1}{l|}{} & \multicolumn{1}{l|}{TER} & 58.6 & 60.62 & 60.52 & 63.61 \\
\multicolumn{1}{l|}{} & \multicolumn{1}{l|}{METEOR} & 0.56 & 0.54 & 0.52 & 0.46 \\
\multicolumn{1}{l|}{} & \multicolumn{1}{l|}{ChrF} & 51.5 & 48.21 & 52.74 & 22.5 \\
\multicolumn{1}{l|}{} & \multicolumn{1}{l|}{ROUGE-1} & 0.6 & 0.72 & 0.58 & 0.11 \\
\multicolumn{1}{l|}{} & \multicolumn{1}{l|}{ROUGE-2} & 0.39 & 0.5 & 0.39 & 0.08 \\
\multicolumn{1}{l|}{} & \multicolumn{1}{l|}{ROUGE-L} & 0.57 & 0.6 & 0.54 & 0.11 \\ \hline
\multicolumn{1}{l|}{\multirow{8}{*}{\begin{tabular}[c]{@{}l@{}}M2M100\\ -418M\end{tabular}}} & \multicolumn{1}{l|}{BLEU} & 33.66 & 34.7 & 34.67 & 24.31 \\
\multicolumn{1}{l|}{} & \multicolumn{1}{l|}{BERTSc} & 0.77 & 0.81 & 0.85 & 0.72 \\
\multicolumn{1}{l|}{} & \multicolumn{1}{l|}{TER} & 58.65 & 58.99 & 58.23 & 62.56 \\
\multicolumn{1}{l|}{} & \multicolumn{1}{l|}{METEOR} & 0.55 & 0.56 & 0.56 & 0.45 \\
\multicolumn{1}{l|}{} & \multicolumn{1}{l|}{ChrF} & 52.4 & 49.93 & 56.08 & 23.1 \\
\multicolumn{1}{l|}{} & \multicolumn{1}{l|}{ROUGE-1} & 0.58 & 0.71 & 0.61 & 0.1 \\
\multicolumn{1}{l|}{} & \multicolumn{1}{l|}{ROUGE-2} & 0.4 & 0.51 & 0.43 & 0.09 \\
\multicolumn{1}{l|}{} & \multicolumn{1}{l|}{ROUGE-L} & 0.55 & 0.6 & 0.57 & 0.1 \\ \hline
\multicolumn{1}{l|}{\multirow{8}{*}{Marian}} & \multicolumn{1}{l|}{BLEU} & 34.09 & 29.72 & 30.48 & 14.79 \\
\multicolumn{1}{l|}{} & \multicolumn{1}{l|}{BERTSc} & 0.85 & 0.83 & 0.84 & 0.76 \\
\multicolumn{1}{l|}{} & \multicolumn{1}{l|}{TER} & 56.82 & 61.64 & 61.09 & 71.92 \\
\multicolumn{1}{l|}{} & \multicolumn{1}{l|}{METEOR} & 0.58 & 0.53 & 0.52 & 0.36 \\
\multicolumn{1}{l|}{} & \multicolumn{1}{l|}{ChrF} & 53.91 & 46.44 & 53.0 & 14.45 \\
\multicolumn{1}{l|}{} & \multicolumn{1}{l|}{ROUGE-1} & 0.62 & 0.72 & 0.58 & 0.12 \\
\multicolumn{1}{l|}{} & \multicolumn{1}{l|}{ROUGE-2} & 0.41 & 0.48 & 0.39 & 0.09 \\
\multicolumn{1}{l|}{} & \multicolumn{1}{l|}{ROUGE-L} & 0.59 & 0.59 & 0.54 & 0.12 \\ \hline
\caption{\textbf{Extra Results: Cascaded \gls{ST} Baselines.} \textbf{German to X} results are reported in this table. Supplement of Table \ref{tab:asr-translation} in the main paper.\\
All cascaded models use Whisper$_{small-mono}$ as \gls{ASR} model (Whisper \gls{ASR} is fine-tuned monolingually - on each source language separately). Its \gls{WER} on test set is 29.6\%, 33.8\%, 31.3\%, 26.3\%, 45.7\% for Vietnamese, English, Chinese, German and French respectively.}
\label{tab:appx_nmt_asr_allMetrics-De-X}
\end{longtable}

%% file: tables/appx_nmt_asr_allMetrics-Zh-X.tex
\begin{longtable}{llcccc}
\hline
\multicolumn{1}{l|}{\textbf{MT}} & \multicolumn{1}{l|}{\textbf{Metrics}} & \textbf{zh-en} & \textbf{zh-vi} & \textbf{zh-fr} & \textbf{zh-de} \\ \hline
\endfirsthead
\multicolumn{6}{c}%
{{\bfseries Table \thetable\ continued from previous page}} \\
\hline
\multicolumn{1}{l|}{\textbf{MT}} & \multicolumn{1}{l|}{\textbf{Metrics}} & \textbf{zh-en} & \textbf{zh-vi} & \textbf{zh-fr} & \textbf{zh-de} \\ \hline
\endhead
\hline
\endfoot
\endlastfoot
\multicolumn{6}{c}{Decoder} \\ \hline
\multicolumn{1}{l|}{\multirow{8}{*}{\begin{tabular}[c]{@{}l@{}}Llama\\ -3.1-8B\end{tabular}}} & \multicolumn{1}{l|}{BLEU} & 19.01 & 17.65 & 13.84 & 11.13 \\
\multicolumn{1}{l|}{} & \multicolumn{1}{l|}{BERTSc} & 0.78 & 0.75 & 0.75 & 0.74 \\
\multicolumn{1}{l|}{} & \multicolumn{1}{l|}{TER} & 101.13 & 104.68 & 106.77 & 109.71 \\
\multicolumn{1}{l|}{} & \multicolumn{1}{l|}{METEOR} & 0.48 & 0.45 & 0.39 & 0.35 \\
\multicolumn{1}{l|}{} & \multicolumn{1}{l|}{ChrF} & 40.96 & 34.68 & 36.97 & 32.37 \\
\multicolumn{1}{l|}{} & \multicolumn{1}{l|}{ROUGE-1} & 0.47 & 0.58 & 0.39 & 0.34 \\
\multicolumn{1}{l|}{} & \multicolumn{1}{l|}{ROUGE-2} & 0.27 & 0.36 & 0.21 & 0.16 \\
\multicolumn{1}{l|}{} & \multicolumn{1}{l|}{ROUGE-L} & 0.43 & 0.46 & 0.34 & 0.3 \\ \hline
\multicolumn{1}{l|}{\multirow{8}{*}{\begin{tabular}[c]{@{}l@{}}Qwen\\ -2.5-7B\end{tabular}}} & \multicolumn{1}{l|}{BLEU} & 25.36 & 26.31 & 17.84 & 12.61 \\
\multicolumn{1}{l|}{} & \multicolumn{1}{l|}{BERTSc} & 0.82 & 0.81 & 0.79 & 0.78 \\
\multicolumn{1}{l|}{} & \multicolumn{1}{l|}{TER} & 79.48 & 80.4 & 84.11 & 91.04 \\
\multicolumn{1}{l|}{} & \multicolumn{1}{l|}{METEOR} & 0.53 & 0.53 & 0.43 & 0.4 \\
\multicolumn{1}{l|}{} & \multicolumn{1}{l|}{ChrF} & 45.87 & 41.37 & 40.88 & 34.86 \\
\multicolumn{1}{l|}{} & \multicolumn{1}{l|}{ROUGE-1} & 0.53 & 0.67 & 0.45 & 0.42 \\
\multicolumn{1}{l|}{} & \multicolumn{1}{l|}{ROUGE-2} & 0.32 & 0.45 & 0.26 & 0.2 \\
\multicolumn{1}{l|}{} & \multicolumn{1}{l|}{ROUGE-L} & 0.48 & 0.55 & 0.4 & 0.37 \\ \hline
\multicolumn{1}{l|}{\multirow{8}{*}{\begin{tabular}[c]{@{}l@{}}Mistral\\ -v0.3-7B\end{tabular}}} & \multicolumn{1}{l|}{BLEU} & 20.17 & 08.01 & 12.58 & 7.14 \\
\multicolumn{1}{l|}{} & \multicolumn{1}{l|}{BERTSc} & 0.8 & 0.73 & 0.76 & 0.72 \\
\multicolumn{1}{l|}{} & \multicolumn{1}{l|}{TER} & 83.87 & 85.69 & 87.85 & 98.87 \\
\multicolumn{1}{l|}{} & \multicolumn{1}{l|}{METEOR} & 0.48 & 0.27 & 0.37 & 0.29 \\
\multicolumn{1}{l|}{} & \multicolumn{1}{l|}{ChrF} & 40.58 & 21.71 & 34.68 & 25.42 \\
\multicolumn{1}{l|}{} & \multicolumn{1}{l|}{ROUGE-1} & 0.48 & 0.53 & 0.39 & 0.3 \\
\multicolumn{1}{l|}{} & \multicolumn{1}{l|}{ROUGE-2} & 0.26 & 0.27 & 0.2 & 0.13 \\
\multicolumn{1}{l|}{} & \multicolumn{1}{l|}{ROUGE-L} & 0.43 & 0.4 & 0.34 & 0.27 \\ \hline
\multicolumn{6}{c}{Encoder-decoder} \\ \hline
\multicolumn{1}{l|}{\multirow{8}{*}{\begin{tabular}[c]{@{}l@{}}mBart\\ -large-50\end{tabular}}} & \multicolumn{1}{l|}{BLEU} & 11.88 & 18.04 & 12.3 & 9.64 \\
\multicolumn{1}{l|}{} & \multicolumn{1}{l|}{BERTSc} & 0.89 & 0.79 & 0.76 & 0.75 \\
\multicolumn{1}{l|}{} & \multicolumn{1}{l|}{TER} & 83.71 & 80.4 & 89.41 & 96.86 \\
\multicolumn{1}{l|}{} & \multicolumn{1}{l|}{METEOR} & 0.36 & 0.42 & 0.34 & 0.32 \\
\multicolumn{1}{l|}{} & \multicolumn{1}{l|}{ChrF} & 32.19 & 34.81 & 34.69 & 30.11 \\
\multicolumn{1}{l|}{} & \multicolumn{1}{l|}{ROUGE-1} & 0.4 & 0.64 & 0.38 & 0.35 \\
\multicolumn{1}{l|}{} & \multicolumn{1}{l|}{ROUGE-2} & 0.18 & 0.38 & 0.19 & 0.15 \\
\multicolumn{1}{l|}{} & \multicolumn{1}{l|}{ROUGE-L} & 0.35 & 0.49 & 0.33 & 0.3 \\ \hline
\multicolumn{1}{l|}{\multirow{8}{*}{\begin{tabular}[c]{@{}l@{}}M2M100\\ -418M\end{tabular}}} & \multicolumn{1}{l|}{BLEU} & 16.65 & 21.83 & 16.94 & 13.06 \\
\multicolumn{1}{l|}{} & \multicolumn{1}{l|}{BERTSc} & 0.76 & 0.83 & 0.79 & 0.78 \\
\multicolumn{1}{l|}{} & \multicolumn{1}{l|}{TER} & 85.02 & 78.88 & 84.12 & 90.04 \\
\multicolumn{1}{l|}{} & \multicolumn{1}{l|}{METEOR} & 0.44 & 0.46 & 0.41 & 0.4 \\
\multicolumn{1}{l|}{} & \multicolumn{1}{l|}{ChrF} & 39.11 & 37.38 & 41.01 & 37.33 \\
\multicolumn{1}{l|}{} & \multicolumn{1}{l|}{ROUGE-1} & 0.44 & 0.62 & 0.45 & 0.42 \\
\multicolumn{1}{l|}{} & \multicolumn{1}{l|}{ROUGE-2} & 0.22 & 0.4 & 0.25 & 0.2 \\
\multicolumn{1}{l|}{} & \multicolumn{1}{l|}{ROUGE-L} & 0.4 & 0.49 & 0.39 & 0.37 \\ \hline
\multicolumn{1}{l|}{\multirow{8}{*}{Marian}} & \multicolumn{1}{l|}{BLEU} & 8.5 & 13.37 & 8.39 & 5.73 \\
\multicolumn{1}{l|}{} & \multicolumn{1}{l|}{BERTSc} & 0.75 & 0.77 & 0.74 & 0.73 \\
\multicolumn{1}{l|}{} & \multicolumn{1}{l|}{TER} & 93.56 & 86.57 & 97.14 & 109.11 \\
\multicolumn{1}{l|}{} & \multicolumn{1}{l|}{METEOR} & 0.32 & 0.36 & 0.28 & 0.27 \\
\multicolumn{1}{l|}{} & \multicolumn{1}{l|}{ChrF} & 28.26 & 30.13 & 30.09 & 27.23 \\
\multicolumn{1}{l|}{} & \multicolumn{1}{l|}{ROUGE-1} & 0.34 & 0.61 & 0.33 & 0.3 \\
\multicolumn{1}{l|}{} & \multicolumn{1}{l|}{ROUGE-2} & 0.12 & 0.31 & 0.14 & 0.11 \\
\multicolumn{1}{l|}{} & \multicolumn{1}{l|}{ROUGE-L} & 0.3 & 0.44 & 0.27 & 0.25 \\ \hline
\caption{\textbf{Extra Results: Cascaded \gls{ST} Baselines.} \textbf{Chinese to X} results are reported in this table. Supplement of Table \ref{tab:asr-translation} in the main paper.\\
All cascaded models use Whisper$_{small-mono}$ as \gls{ASR} model (Whisper \gls{ASR} is fine-tuned monolingually - on each source language separately). Its \gls{WER} on test set is 29.6\%, 33.8\%, 31.3\%, 26.3\%, 45.7\% for Vietnamese, English, Chinese, German and French respectively.}
\label{tab:appx_nmt_asr_allMetrics-Zh-X}
\end{longtable}